%% file: main.tex
\documentclass[manuscript,screen]{acmart}  
\AtBeginDocument{%
  \providecommand\BibTeX{{%
    \normalfont B\kern-0.5em{\scshape i\kern-0.25em b}\kern-0.8em\TeX}}}

\setcopyright{acmlicensed}
\copyrightyear{2018}
\acmYear{2018}
\acmDOI{XXXXXXX.XXXXXXX}

\acmJournal{JACM}
\acmVolume{37}
\acmNumber{4}
\acmArticle{111}
\acmMonth{8}

\usepackage{tabularx}
\usepackage{multirow}
\usepackage{geometry}
\usepackage{pifont}   
\usepackage{array}
\usepackage{longtable}

\usepackage{booktabs}

\newcolumntype{L}[1]{>{\raggedright\arraybackslash}m{#1}}
\usepackage{subcaption}  
\usepackage{graphicx}    
\usepackage{xcolor}
\usepackage{wrapfig}
\usepackage{pgf}
\usepackage{tikz} 
\usepackage[edges]{forest}
\usepackage{longtable}
\usepackage{enumitem}
\usepackage[most]{tcolorbox} 
\usepackage{setspace}
\usepackage{wrapfig}
\usepackage[section]{placeins}

\definecolor{customred}{RGB}{207, 55, 49}
\definecolor{customorange}{RGB}{255, 128, 0}
\definecolor{customgreen}{RGB}{0, 204, 0}
\definecolor{customyellow}{RGB}{251, 239, 214}

\newtcolorbox{takeaway}[2][]{%
    colback=blue!5!white, 
    colframe=blue!75!black, 
    fonttitle=\bfseries, 
    title=#2, #1 
}

\newtcolorbox{takeaway2}[1][]{%
    colback=blue!5!white, 
    colframe=blue!75!black, 
    sharp corners, 
    boxrule=1.2pt, 
    #1 
}

\newcolumntype{C}[1]{>{\centering\arraybackslash}p{#1}}

\begin{document}

\title[A Survey of Small Language Models]{A Comprehensive Survey of Small Language Models in the Era of Large Language Models: Techniques, Enhancements, Applications, Collaboration with LLMs, and Trustworthiness}
\author{Fali Wang}
\email{fqw5095@psu.edu}

\author{Zhiwei Zhang}

\author{Xianren Zhang}

\author{Zongyu Wu}
\affiliation{%
  \institution{The Pennsylvania State University}
  \city{University Park}
  \country{USA}}

\author{TzuHao Mo}
\affiliation{
  \institution{University of Pennsylvania}
  \city{Philadelphia}
  \country{USA}}

\author{Qiuhao Lu}

\author{Wanjing Wang}

\author{Rui Li}
\affiliation{
  \institution{UTHealth Houston}
  \city{Houston}
  \country{USA}}

\author{Junjie Xu}
\affiliation{%
  \institution{The Pennsylvania State University}
  \city{University Park}
  \country{USA}}

\author{Xianfeng Tang}

\author{Qi He}
\affiliation{
  \institution{Amazon}
  \city{Palo Alto}
  \country{USA}}

\author{Yao Ma}
\affiliation{%
  \institution{Rensselaer Polytechnic Institute}
  \city{Troy}
  \country{USA}}

\author{Ming Huang}
\affiliation{
  \institution{UTHealth Houston}
  \city{Houston}
  \country{USA}}

\author{Suhang Wang}
\authornote{Corresponding author.} 
\affiliation{%
  \institution{The Pennsylvania State University}
  \city{University Park}
  \country{USA}}
\email{szw494@psu.edu}

\renewcommand{\shortauthors}{Fali Wang, et al.}

\begin{abstract}
Large language models (LLMs) have demonstrated emergent abilities in text generation, question answering, and reasoning, facilitating various tasks and domains. 
Despite their proficiency in various tasks, LLMs like PaLM 540B and Llama-3.1 405B face limitations due to large parameter sizes and computational demands, often requiring cloud API use which raises privacy concerns, limits real-time applications on edge devices, and increases fine-tuning costs.
Additionally, LLMs often underperform in specialized domains such as healthcare and law due to insufficient domain-specific knowledge, necessitating specialized models. 
Therefore, Small Language Models (SLMs) are increasingly favored for their low inference latency, cost-effectiveness, efficient development, and easy customization and adaptability. These models are particularly well-suited for resource-limited environments and domain knowledge acquisition, addressing LLMs' challenges and proving ideal for applications that require localized data handling for privacy, minimal inference latency for efficiency, and domain knowledge acquisition through lightweight fine-tuning.
The rising demand for SLMs has spurred extensive research and development. However, a comprehensive survey investigating issues related to the definition, acquisition, application, enhancement, and reliability of SLM remains lacking, prompting us to conduct a detailed survey on these topics.
The definition of SLMs varies widely, thus to standardize, we propose defining SLMs by their capability to perform specialized tasks and suitability for resource-constrained settings, setting boundaries based on the minimal size for emergent abilities and the maximum size sustainable under resource constraints.
For other aspects, we provide a taxonomy of relevant models/methods and develop general frameworks for each category to enhance and utilize SLMs effectively.
We have compiled the collected SLM models and related methods on GitHub: \url{https://github.com/FairyFali/SLMs-Survey}.
\end{abstract}



\received{20 February 2007}
\received[revised]{12 March 2009}
\received[accepted]{5 June 2009}

\maketitle

\input{math}

\input{sections/1.intro}

\input{sections/3.construction_of_slm}

\input{sections/4.enhancement}

\input{sections/5.application}

\input{sections/6.model}

\input{sections/7.slm4llm}

\input{sections/8.synergy}
\input{sections/9.security}

\input{sections/10.challenges}

\section{Conclusion}
This paper provides a comprehensive survey of Small Language Models (SLMs) with up to 7 billion parameters. Initially, we address the need to clearly define SLMs due to existing ambiguities in their characterization. We then present the foundational concepts essential for constructing SLMs. The survey progresses to explore enhancement techniques, including knowledge distillation and quantization, as well as strategies for adapting Large Language Models (LLMs) to SLM contexts. We survey representative SLMs, both general-domain and domain-specific, discussing their preferred datasets and architectural decisions. We also assess their applications across various tasks and deployment strategies on devices. Further, we investigate their role in augmenting the capabilities of LLMs, serving as proxies for fine-tuning and facilitating two types of synergies: cloud-local and task-centric. Additionally, we discuss the critical aspect of their trustworthiness. The paper concludes with key insights aimed at guiding future research on small language models.

\bibliographystyle{ACM-Reference-Format}
\bibliography{reference}

\appendix

\end{document}

%% file: math.tex


\newcommand{\figleft}{{\em (Left)}}
\newcommand{\figcenter}{{\em (Center)}}
\newcommand{\figright}{{\em (Right)}}
\newcommand{\figtop}{{\em (Top)}}
\newcommand{\figbottom}{{\em (Bottom)}}
\newcommand{\captiona}{{\em (a)}}
\newcommand{\captionb}{{\em (b)}}
\newcommand{\captionc}{{\em (c)}}
\newcommand{\captiond}{{\em (d)}}

\newcommand{\newterm}[1]{{\bf #1}}

\def\figref#1{figure~\ref{#1}}
\def\Figref#1{Figure~\ref{#1}}
\def\twofigref#1#2{figures \ref{#1} and \ref{#2}}
\def\quadfigref#1#2#3#4{figures \ref{#1}, \ref{#2}, \ref{#3} and \ref{#4}}
\def\secref#1{section~\ref{#1}}
\def\Secref#1{Section~\ref{#1}}
\def\twosecrefs#1#2{sections \ref{#1} and \ref{#2}}
\def\secrefs#1#2#3{sections \ref{#1}, \ref{#2} and \ref{#3}}
\def\eqref#1{equation~\ref{#1}}
\def\Eqref#1{Equation~\ref{#1}}
\def\plaineqref#1{\ref{#1}}
\def\chapref#1{chapter~\ref{#1}}
\def\Chapref#1{Chapter~\ref{#1}}
\def\rangechapref#1#2{chapters\ref{#1}--\ref{#2}}
\def\algref#1{algorithm~\ref{#1}}
\def\Algref#1{Algorithm~\ref{#1}}
\def\twoalgref#1#2{algorithms \ref{#1} and \ref{#2}}
\def\Twoalgref#1#2{Algorithms \ref{#1} and \ref{#2}}
\def\partref#1{part~\ref{#1}}
\def\Partref#1{Part~\ref{#1}}
\def\twopartref#1#2{parts \ref{#1} and \ref{#2}}

\def\ceil#1{\lceil #1 \rceil}
\def\floor#1{\lfloor #1 \rfloor}
\def\1{\bm{1}}
\newcommand{\train}{\mathcal{D}}
\newcommand{\valid}{\mathcal{D_{\mathrm{valid}}}}
\newcommand{\test}{\mathcal{D_{\mathrm{test}}}}

\def\eps{{\epsilon}}

\def\reta{{\textnormal{$\eta$}}}
\def\ra{{\textnormal{a}}}
\def\rb{{\textnormal{b}}}
\def\rc{{\textnormal{c}}}
\def\rd{{\textnormal{d}}}
\def\re{{\textnormal{e}}}
\def\rf{{\textnormal{f}}}
\def\rg{{\textnormal{g}}}
\def\rh{{\textnormal{h}}}
\def\ri{{\textnormal{i}}}
\def\rj{{\textnormal{j}}}
\def\rk{{\textnormal{k}}}
\def\rl{{\textnormal{l}}}
\def\rn{{\textnormal{n}}}
\def\ro{{\textnormal{o}}}
\def\rp{{\textnormal{p}}}
\def\rq{{\textnormal{q}}}
\def\rr{{\textnormal{r}}}
\def\rs{{\textnormal{s}}}
\def\rt{{\textnormal{t}}}
\def\ru{{\textnormal{u}}}
\def\rv{{\textnormal{v}}}
\def\rw{{\textnormal{w}}}
\def\rx{{\textnormal{x}}}
\def\ry{{\textnormal{y}}}
\def\rz{{\textnormal{z}}}

\def\rvepsilon{{\mathbf{\epsilon}}}
\def\rvtheta{{\mathbf{\theta}}}
\def\rva{{\mathbf{a}}}
\def\rvb{{\mathbf{b}}}
\def\rvc{{\mathbf{c}}}
\def\rvd{{\mathbf{d}}}
\def\rve{{\mathbf{e}}}
\def\rvf{{\mathbf{f}}}
\def\rvg{{\mathbf{g}}}
\def\rvh{{\mathbf{h}}}
\def\rvu{{\mathbf{i}}}
\def\rvj{{\mathbf{j}}}
\def\rvk{{\mathbf{k}}}
\def\rvl{{\mathbf{l}}}
\def\rvm{{\mathbf{m}}}
\def\rvn{{\mathbf{n}}}
\def\rvo{{\mathbf{o}}}
\def\rvp{{\mathbf{p}}}
\def\rvq{{\mathbf{q}}}
\def\rvr{{\mathbf{r}}}
\def\rvs{{\mathbf{s}}}
\def\rvt{{\mathbf{t}}}
\def\rvu{{\mathbf{u}}}
\def\rvv{{\mathbf{v}}}
\def\rvw{{\mathbf{w}}}
\def\rvx{{\mathbf{x}}}
\def\rvy{{\mathbf{y}}}
\def\rvz{{\mathbf{z}}}

\def\erva{{\textnormal{a}}}
\def\ervb{{\textnormal{b}}}
\def\ervc{{\textnormal{c}}}
\def\ervd{{\textnormal{d}}}
\def\erve{{\textnormal{e}}}
\def\ervf{{\textnormal{f}}}
\def\ervg{{\textnormal{g}}}
\def\ervh{{\textnormal{h}}}
\def\ervi{{\textnormal{i}}}
\def\ervj{{\textnormal{j}}}
\def\ervk{{\textnormal{k}}}
\def\ervl{{\textnormal{l}}}
\def\ervm{{\textnormal{m}}}
\def\ervn{{\textnormal{n}}}
\def\ervo{{\textnormal{o}}}
\def\ervp{{\textnormal{p}}}
\def\ervq{{\textnormal{q}}}
\def\ervr{{\textnormal{r}}}
\def\ervs{{\textnormal{s}}}
\def\ervt{{\textnormal{t}}}
\def\ervu{{\textnormal{u}}}
\def\ervv{{\textnormal{v}}}
\def\ervw{{\textnormal{w}}}
\def\ervx{{\textnormal{x}}}
\def\ervy{{\textnormal{y}}}
\def\ervz{{\textnormal{z}}}

\def\rmA{{\mathbf{A}}}
\def\rmB{{\mathbf{B}}}
\def\rmC{{\mathbf{C}}}
\def\rmD{{\mathbf{D}}}
\def\rmE{{\mathbf{E}}}
\def\rmF{{\mathbf{F}}}
\def\rmG{{\mathbf{G}}}
\def\rmH{{\mathbf{H}}}
\def\rmI{{\mathbf{I}}}
\def\rmJ{{\mathbf{J}}}
\def\rmK{{\mathbf{K}}}
\def\rmL{{\mathbf{L}}}
\def\rmM{{\mathbf{M}}}
\def\rmN{{\mathbf{N}}}
\def\rmO{{\mathbf{O}}}
\def\rmP{{\mathbf{P}}}
\def\rmQ{{\mathbf{Q}}}
\def\rmR{{\mathbf{R}}}
\def\rmS{{\mathbf{S}}}
\def\rmT{{\mathbf{T}}}
\def\rmU{{\mathbf{U}}}
\def\rmV{{\mathbf{V}}}
\def\rmW{{\mathbf{W}}}
\def\rmX{{\mathbf{X}}}
\def\rmY{{\mathbf{Y}}}
\def\rmZ{{\mathbf{Z}}}

\def\ermA{{\textnormal{A}}}
\def\ermB{{\textnormal{B}}}
\def\ermC{{\textnormal{C}}}
\def\ermD{{\textnormal{D}}}
\def\ermE{{\textnormal{E}}}
\def\ermF{{\textnormal{F}}}
\def\ermG{{\textnormal{G}}}
\def\ermH{{\textnormal{H}}}
\def\ermI{{\textnormal{I}}}
\def\ermJ{{\textnormal{J}}}
\def\ermK{{\textnormal{K}}}
\def\ermL{{\textnormal{L}}}
\def\ermM{{\textnormal{M}}}
\def\ermN{{\textnormal{N}}}
\def\ermO{{\textnormal{O}}}
\def\ermP{{\textnormal{P}}}
\def\ermQ{{\textnormal{Q}}}
\def\ermR{{\textnormal{R}}}
\def\ermS{{\textnormal{S}}}
\def\ermT{{\textnormal{T}}}
\def\ermU{{\textnormal{U}}}
\def\ermV{{\textnormal{V}}}
\def\ermW{{\textnormal{W}}}
\def\ermX{{\textnormal{X}}}
\def\ermY{{\textnormal{Y}}}
\def\ermZ{{\textnormal{Z}}}

\def\vzero{{\bm{0}}}
\def\vone{{\bm{1}}}
\def\vmu{{\bm{\mu}}}
\def\vtheta{{\bm{\theta}}}
\def\va{{\bm{a}}}
\def\vb{{\bm{b}}}
\def\vc{{\bm{c}}}
\def\vd{{\bm{d}}}
\def\ve{{\bm{e}}}
\def\vf{{\bm{f}}}
\def\vg{{\bm{g}}}
\def\vh{{\bm{h}}}
\def\vi{{\bm{i}}}
\def\vj{{\bm{j}}}
\def\vk{{\bm{k}}}
\def\vl{{\bm{l}}}
\def\vm{{\bm{m}}}
\def\vn{{\bm{n}}}
\def\vo{{\bm{o}}}
\def\vp{{\bm{p}}}
\def\vq{{\bm{q}}}
\def\vr{{\bm{r}}}
\def\vs{{\bm{s}}}
\def\vt{{\bm{t}}}
\def\vu{{\bm{u}}}
\def\vv{{\bm{v}}}
\def\vw{{\bm{w}}}
\def\vx{{\bm{x}}}
\def\vy{{\bm{y}}}
\def\vz{{\bm{z}}}

\def\evalpha{{\alpha}}
\def\evbeta{{\beta}}
\def\evepsilon{{\epsilon}}
\def\evlambda{{\lambda}}
\def\evomega{{\omega}}
\def\evmu{{\mu}}
\def\evpsi{{\psi}}
\def\evsigma{{\sigma}}
\def\evtheta{{\theta}}
\def\eva{{a}}
\def\evb{{b}}
\def\evc{{c}}
\def\evd{{d}}
\def\eve{{e}}
\def\evf{{f}}
\def\evg{{g}}
\def\evh{{h}}
\def\evi{{i}}
\def\evj{{j}}
\def\evk{{k}}
\def\evl{{l}}
\def\evm{{m}}
\def\evn{{n}}
\def\evo{{o}}
\def\evp{{p}}
\def\evq{{q}}
\def\evr{{r}}
\def\evs{{s}}
\def\evt{{t}}
\def\evu{{u}}
\def\evv{{v}}
\def\evw{{w}}
\def\evx{{x}}
\def\evy{{y}}
\def\evz{{z}}

\def\mA{{\bm{A}}}
\def\mB{{\bm{B}}}
\def\mC{{\bm{C}}}
\def\mD{{\bm{D}}}
\def\mE{{\bm{E}}}
\def\mF{{\bm{F}}}
\def\mG{{\bm{G}}}
\def\mH{{\bm{H}}}
\def\mI{{\bm{I}}}
\def\mJ{{\bm{J}}}
\def\mK{{\bm{K}}}
\def\mL{{\bm{L}}}
\def\mM{{\bm{M}}}
\def\mN{{\bm{N}}}
\def\mO{{\bm{O}}}
\def\mP{{\bm{P}}}
\def\mQ{{\bm{Q}}}
\def\mR{{\bm{R}}}
\def\mS{{\bm{S}}}
\def\mT{{\bm{T}}}
\def\mU{{\bm{U}}}
\def\mV{{\bm{V}}}
\def\mW{{\bm{W}}}
\def\mX{{\bm{X}}}
\def\mY{{\bm{Y}}}
\def\mZ{{\bm{Z}}}
\def\mBeta{{\bm{\beta}}}
\def\mPhi{{\bm{\Phi}}}
\def\mLambda{{\bm{\Lambda}}}
\def\mSigma{{\bm{\Sigma}}}

\newcommand{\tens}[1]{\bm{\mathsfit{#1}}}
\def\tA{{\tens{A}}}
\def\tB{{\tens{B}}}
\def\tC{{\tens{C}}}
\def\tD{{\tens{D}}}
\def\tE{{\tens{E}}}
\def\tF{{\tens{F}}}
\def\tG{{\tens{G}}}
\def\tH{{\tens{H}}}
\def\tI{{\tens{I}}}
\def\tJ{{\tens{J}}}
\def\tK{{\tens{K}}}
\def\tL{{\tens{L}}}
\def\tM{{\tens{M}}}
\def\tN{{\tens{N}}}
\def\tO{{\tens{O}}}
\def\tP{{\tens{P}}}
\def\tQ{{\tens{Q}}}
\def\tR{{\tens{R}}}
\def\tS{{\tens{S}}}
\def\tT{{\tens{T}}}
\def\tU{{\tens{U}}}
\def\tV{{\tens{V}}}
\def\tW{{\tens{W}}}
\def\tX{{\tens{X}}}
\def\tY{{\tens{Y}}}
\def\tZ{{\tens{Z}}}

\def\gA{{\mathcal{A}}}
\def\gB{{\mathcal{B}}}
\def\gC{{\mathcal{C}}}
\def\gD{{\mathcal{D}}}
\def\gE{{\mathcal{E}}}
\def\gF{{\mathcal{F}}}
\def\gG{{\mathcal{G}}}
\def\gH{{\mathcal{H}}}
\def\gI{{\mathcal{I}}}
\def\gJ{{\mathcal{J}}}
\def\gK{{\mathcal{K}}}
\def\gL{{\mathcal{L}}}
\def\gM{{\mathcal{M}}}
\def\gN{{\mathcal{N}}}
\def\gO{{\mathcal{O}}}
\def\gP{{\mathcal{P}}}
\def\gQ{{\mathcal{Q}}}
\def\gR{{\mathcal{R}}}
\def\gS{{\mathcal{S}}}
\def\gT{{\mathcal{T}}}
\def\gU{{\mathcal{U}}}
\def\gV{{\mathcal{V}}}
\def\gW{{\mathcal{W}}}
\def\gX{{\mathcal{X}}}
\def\gY{{\mathcal{Y}}}
\def\gZ{{\mathcal{Z}}}

\def\sA{{\mathbb{A}}}
\def\sB{{\mathbb{B}}}
\def\sC{{\mathbb{C}}}
\def\sD{{\mathbb{D}}}
\def\sF{{\mathbb{F}}}
\def\sG{{\mathbb{G}}}
\def\sH{{\mathbb{H}}}
\def\sI{{\mathbb{I}}}
\def\sJ{{\mathbb{J}}}
\def\sK{{\mathbb{K}}}
\def\sL{{\mathbb{L}}}
\def\sM{{\mathbb{M}}}
\def\sN{{\mathbb{N}}}
\def\sO{{\mathbb{O}}}
\def\sP{{\mathbb{P}}}
\def\sQ{{\mathbb{Q}}}
\def\sR{{\mathbb{R}}}
\def\sS{{\mathbb{S}}}
\def\sT{{\mathbb{T}}}
\def\sU{{\mathbb{U}}}
\def\sV{{\mathbb{V}}}
\def\sW{{\mathbb{W}}}
\def\sX{{\mathbb{X}}}
\def\sY{{\mathbb{Y}}}
\def\sZ{{\mathbb{Z}}}

\def\emLambda{{\Lambda}}
\def\emA{{A}}
\def\emB{{B}}
\def\emC{{C}}
\def\emD{{D}}
\def\emE{{E}}
\def\emF{{F}}
\def\emG{{G}}
\def\emH{{H}}
\def\emI{{I}}
\def\emJ{{J}}
\def\emK{{K}}
\def\emL{{L}}
\def\emM{{M}}
\def\emN{{N}}
\def\emO{{O}}
\def\emP{{P}}
\def\emQ{{Q}}
\def\emR{{R}}
\def\emS{{S}}
\def\emT{{T}}
\def\emU{{U}}
\def\emV{{V}}
\def\emW{{W}}
\def\emX{{X}}
\def\emY{{Y}}
\def\emZ{{Z}}
\def\emSigma{{\Sigma}}

\newcommand{\etens}[1]{\mathsfit{#1}}
\def\etLambda{{\etens{\Lambda}}}
\def\etA{{\etens{A}}}
\def\etB{{\etens{B}}}
\def\etC{{\etens{C}}}
\def\etD{{\etens{D}}}
\def\etE{{\etens{E}}}
\def\etF{{\etens{F}}}
\def\etG{{\etens{G}}}
\def\etH{{\etens{H}}}
\def\etI{{\etens{I}}}
\def\etJ{{\etens{J}}}
\def\etK{{\etens{K}}}
\def\etL{{\etens{L}}}
\def\etM{{\etens{M}}}
\def\etN{{\etens{N}}}
\def\etO{{\etens{O}}}
\def\etP{{\etens{P}}}
\def\etQ{{\etens{Q}}}
\def\etR{{\etens{R}}}
\def\etS{{\etens{S}}}
\def\etT{{\etens{T}}}
\def\etU{{\etens{U}}}
\def\etV{{\etens{V}}}
\def\etW{{\etens{W}}}
\def\etX{{\etens{X}}}
\def\etY{{\etens{Y}}}
\def\etZ{{\etens{Z}}}

\newcommand{\pdata}{p_{\rm{data}}}
\newcommand{\ptrain}{\hat{p}_{\rm{data}}}
\newcommand{\Ptrain}{\hat{P}_{\rm{data}}}
\newcommand{\pmodel}{p_{\rm{model}}}
\newcommand{\Pmodel}{P_{\rm{model}}}
\newcommand{\ptildemodel}{\tilde{p}_{\rm{model}}}
\newcommand{\pencode}{p_{\rm{encoder}}}
\newcommand{\pdecode}{p_{\rm{decoder}}}
\newcommand{\precons}{p_{\rm{reconstruct}}}

\newcommand{\E}{\mathbb{E}}
\newcommand{\Ls}{\mathcal{L}}
\newcommand{\R}{\mathbb{R}}
\newcommand{\emp}{\tilde{p}}
\newcommand{\lr}{\alpha}
\newcommand{\reg}{\lambda}
\newcommand{\rect}{\mathrm{rectifier}}
\newcommand{\softmax}{\mathrm{softmax}}
\newcommand{\sigmoid}{\sigma}
\newcommand{\softplus}{\zeta}
\newcommand{\KL}{D_{\mathrm{KL}}}
\newcommand{\Var}{\mathrm{Var}}
\newcommand{\standarderror}{\mathrm{SE}}
\newcommand{\Cov}{\mathrm{Cov}}
\newcommand{\normlzero}{L^0}
\newcommand{\normlone}{L^1}
\newcommand{\normltwo}{L^2}
\newcommand{\normlp}{L^p}
\newcommand{\normmax}{L^\infty}

\newcommand{\parents}{Pa} 

\let\ab\allowbreak

%% file: sections/1.intro.tex
\section{Introduction}
\label{intro}

\begin{figure}[!htbp]
\centering
\begin{spacing}{1.}
\resizebox{0.95\linewidth}{!}{
\begin{forest}
	for tree={
		draw,
		shape=rectangle,
		rounded corners,
		top color=white,
		grow'=0,
		l sep'=1.2em,
		reversed=true,
		anchor=west,
		child anchor=west,
	},
	forked edges,
	root/.style={
            draw=customred,
		rotate=90, shading angle=90, bottom color=white!40,
		anchor=north, font=\normalsize, inner sep=0.5em},
	level1/.style={
            draw=customred,
            text width=2cm,
		bottom color=white!30, font=\normalsize, inner sep=0.3em,
		s sep=0.2em},
	level2/.style={
            draw=customred,
            text width=4cm,
		bottom color=white!40, font=\small, inner sep=0.25em, 
		s sep=0.1em},
	level3/.style={
            draw=customred,
            text width=4cm,  
		bottom color=white!40, font=\small, inner sep=0.2em, 
		l sep'=0.5em},
	level4/.style={
            draw=customred,
            text width=6.7cm,
            top color=customyellow!100, 
		bottom color=customyellow!100, font=\small, inner sep=0.2em,
		l sep'=0.5em},
	where n=0{root}{},
	where level=1{level1}{},
	where level=2{level2}{},
	where level=3{level3}{},
	where level>=4{level4}{},
	[Small Language Models
	    [Introduction (\S \ref{intro})
		]
		[Concepts in Building LMs (\S \ref{construction})
                [Architecture (\S \ref{architecture})]
                [Training techniques (\S \ref{training_techniques})]
                [Obtain SLMs from LLMs (\S \ref{compression})
                    [Pruning (\S \ref{pruning})
                        [Unstructured Pruning~\cite{frantar2023sparsegpt,sun2024a,zhang2023loraprune,zhang2024plug,li2023sparse,shao2024one,das2023beyond}; Structured Pruning~\cite{men2024shortgpt,ma2023llm,li2024nuteprune,yang2024laco,zhaoapt,li2024lorap,an2024fluctuation,ashkboos2024slicegpt,chen2023lorashear,xia2024sheared,guo2023compresso}]
                    ]
                    [Knowledge Distillation (\S \ref{knowledge_distillation})
                        [White-Box KD \cite{zhang2023towards,gu2024minillm,agarwal2024policy,ko2024distillm,jha2024justchopembarrassinglysimple,kim2024token,padmanabhan2024propagating}; Black-Box KD \cite{peng2023instruction,chen2023mcc,wang2023scott}]
                    ]
                    [Quantization (\S \ref{quantization})
                        [SqueezeLLM \cite{kim2023squeezellm}; JSQ \cite{guo2024compressing}; FrameQuant \cite{adepu2024framequant}; OneBit \cite{xu2024onebit}; BiLLM \cite{huang2024billm}; LQER \cite{zhang2024lqer}; I-LLM \cite{hu2024llm}; PV-Tuning \cite{malinovskii2024pv}; BitNet \cite{wang2023bitnet}; BitNet b1.58 \cite{ma2024era}; PEQA \cite{kim2024memory}; QLoRA \cite{dettmers2024qlora}]
                    ]
                ]
		]
		[Enhancement of SLMs (\S \ref{enhancement})
                [Training Methods for SLMs from Scratch (\S \ref{Training4SLMFromScratch})
                    [MobiLlama~\cite{thawakar2024mobillama}; MobileLLMs \cite{liu2024mobilellm}; MindLLMs~\cite{yang2023mindllm}; \citet{tang2024rethinking}, for tree={text width=11cm, top color=customyellow!100, bottom color=customyellow!100} ]
                ]
			[Supervised Fine-Tuning (\S \ref{SFT})
                    [Alpaca~\cite{alpaca}; UltraChat \cite{ding2023enhancing}; WizardLM \cite{xu2023wizardlm}; SlimOrca \cite{SlimOrca}; ShareGPT \cite{wang2024openchat}; Capybara \cite{daniele2023amplify-instruct}; Deita \cite{liu2023makes}; MetaMathQA \cite{yu2024metamath}; MobileBERT \cite{sun2020mobilebert}; StableLM~\cite{bellagente2024stable, StableLM-3B-4E1T}; RLHF \cite{ouyang2022training}; DPO \cite{rafailov2024direct}, for tree={text width=11cm, top color=customyellow!100, bottom color=customyellow!100}]
                ]
			[Data quality in Knowledge Distillation (\S \ref{data_quality4enhance})
                    [TinyStory \cite{eldan2023tinystories}; Self-Amplify \cite{bhan2024self}; AS-ES Learning \cite{xi2024learning}; \citet{huang2023large}; \citet{bhan2024self}; \citet{tian2024toward}, for tree={text width=11cm, top color=customyellow!100, bottom color=customyellow!100}]
                ]
			[Distillation techniques for enhancing SLM (\S \ref{distillation4enhance})
                    [GKD~\cite{agarwal2024policy}; DistiLLM~\cite{ko2024distillm}; Adapt-and-Distill~\cite{yao2021adapt}; Bit-level Inference Scaling Laws~\cite{dettmers2023case}, for tree={text width=11cm, top color=customyellow!100, bottom color=customyellow!100}]
                ]
			[Performance improvement through quantization (\S \ref{quantization4enhance})
                    [Bit-level Inference Scaling Laws~\cite{dettmers2023case}; BiLLM~\cite{huang2024billm}; LLM.int8()~\cite{dettmers2022gptint}; PB-LLM~\cite{shang2023pbllmpartiallybinarizedlarge}; OneBit~\cite{xu2024onebit}; BitNet~\cite{wang2023bitnet}; LLM-QAT~\cite{liu2023llm}, for tree={text width=11cm, top color=customyellow!100, bottom color=customyellow!100}]
                ]
			[Techniques in LLMs contributing SLMs (\S \ref{techniques_in_llms_for_slms})
                    [RAG for SLMs \cite{yazan2024impact, liu2024can}; MoE for SLMs~\cite{kim2023mixture,li2024examining}, for tree={text width=11cm, top color=customyellow!100, bottom color=customyellow!100}]
                ]
		]
		[Applications of SLMs (\S \ref{application})
			[Task-specific SLM applications (\S \ref{task_specific_application})
				[SLM applications in QA (\S \ref{slm_application_qa})
                        [Phi-series \cite{gunasekar2023textbooksneed, abdin2024phi}; Orca 2 \cite{mitra2023orca}; BioGPT-Large \cite{guo2023improvingsmalllanguagemodels}; Stable Beluga 7B \cite{StableBelugaModels}; \citet{phogat2024finetuningsmallerlanguagemodels}; \citet{hartill2023teachingsmallerlanguagemodels}; \citet{gichamba2024colbertretrievalensembleresponse}; \citet{jeong2023testtimeselfadaptivesmalllanguage}; Rationale Ranking \cite{hartill2023answeringunseenquestionssmaller}]
                    ]
				[SLM applications in Coding (\S \ref{slm_application_coding})
                        [DeepSeek-Coder \cite{guo2024deepseek}; Phi-1~\cite{gunasekar2023textbooksneed}; Phi-3.5-mini \cite{abdin2024phi}; CodeGemma \cite{team2024codegemma}; CodeLlama \cite{roziere2023code}]
                    ]
				[SLM applications in Recommender Systems (\S \ref{slm_application_recommender})
                        [PromptRec \cite{wu2024could}; SLIM \cite{wang2024can}; BiLLP \cite{shi2024large}; RecLoRA \cite{zhu2024lifelong}; \citet{wu2021empowering}]
                    ]
				[SLM applications in Web Search (\S \ref{slm_application_web})
                        [Content encoder \cite{changpre, humeaupoly, lu2020twinbert}; H-ERNIE \cite{chu2022h}; \citet{zou2022pre}; \citet{peng2023soft}; CoCondenser \cite{gao2022unsupervised}; Implicit Interaction ($I^3$) \cite{dong2023i3}; InPars \cite{bonifacio2022inpars}; rewrite-retrieve-read \cite{ma2023query}]
                    ]
				[SLM applications in Mobile-device (\S \ref{slm_application_mobile})
                        [Octopus \cite{chen2024octopusondevicelanguagemodel}; MobileAgent~\cite{ding2024mobileagentenhancingmobilecontrol}; \citet{carreira2023revolutionizingmobileinteractionenabling}; AutoDroid~\cite{wen2024autodroidllmpoweredtaskautomation}; \citet{qin2023enabling}; \citet{zhu2023ondeviceagenttextrewriting}]
                    ]
                ]
			[Techniques during SLM Deployment (\S \ref{slm_deployment_mobile})
				[Memory efficiency (\S \ref{memory_efficiency})
                        [MoE-I$^2$ \cite{yang-etal-2024-moe}; MobileAIBench \cite{murthy2024mobileaibenchbenchmarkingllmslmms}; MobileLLM \cite{liu2024mobilellm}; EdgeMoE \cite{yi2023edgemoefastondeviceinference}; GEAR \cite{kang2024gear}; HETLORA \cite{cho2024heterogeneouslorafederatedfinetuning}; MobiLlama \cite{thawakar2024mobillama}]
                    ]
				[Computing efficiency (\S \ref{runtime_efficiency})
                        [COST-EFF \cite{shen-etal-2022-cost}; MobileLLM \cite{liu2024mobilellm}; MobiLlama \cite{thawakar2024mobillama}; Merino \cite{zhao2024merinoentropydrivendesigngenerative}; EdgeMoE \cite{yi2023edgemoefastondeviceinference}; \citet{nawrot2024dynamic}; LLMCadLLM-Cad\cite{xu2023llmcadfastscalableondevice}; LinguaLinked \cite{zhao2023lingualinkeddistributedlargelanguage}]
                    ]
			]
		]
            [Models (\S \ref{model})
                [Generic-domain SLMs (\S \ref{generic_slms})
                    [PhoneLM \cite{yi2024phonelm}; Llama 3.2; Qwen \cite{bai2023qwentechnicalreport, yang2024qwen2}; Gemma \cite{team2024gemma, team2024gemma2}; StableLM \cite{bellagente2024stable, StableLM-3B-4E1T}; TinyLlama \cite{zhang2024tinyllamaopensourcesmalllanguage}; OLMo \cite{groeneveld2024olmo}; H2O-Danube3 \cite{pfeiffer2024h2o}; Fox-1 \cite{fox1}; MiniCPM \cite{hu2024minicpm}; Phi \cite{gunasekar2023textbooksneed,li2023textbooksneediiphi15,javaheripi2023phi,abdin2024phi,abdin2024phi}; BLOOM and BLOOMZ~\cite{le2023bloom}; Galactica~\cite{taylor2022galactica}; OPT \cite{zhang2022opt}; XGLM~\cite{lin-etal-2022-shot}; Megatron-gpt2 \cite{shoeybi2019megatron}; MINITRON \cite{muralidharan2024compact}; LaMini-LM \cite{kim2016sequence}; FlanT5 \cite{chung2024scaling} ... , for tree={text width=11cm, top color=customyellow!100, bottom color=customyellow!100}]
                ]
                [Specific-domain SLMs (\S \ref{specific_slms})
                    [Hippocrates \cite{acikgoz2024hippocrates}; BioMedLM \cite{bolton2024biomedlm}; MentaLLaMA \cite{yang2024mentallama}; ChemLLM \cite{zhang2024chemllm}; SciGLM \cite{zhang2024sciglm}; AstroLLaMA \cite{nguyen2023astrollama}; MindLLM \cite{yang2023mindllm}, for tree={text width=11cm, top color=customyellow!100, bottom color=customyellow!100}]
                ]
            ]
		[SLMs for LLMs (\S \ref{slm4llm})
			[SLM for reliable LLM generation (\S \ref{slm4reliable})
                    [POLAR \cite{zhao2023automatic}; SAPLMA \cite{azaria2023internal}; SuperICL \cite{xu2023small}; SuperContext \cite{yang2024supervised}; Self-RAG \cite{asai2024selfrag}; SKR \cite{wang2023self}; SlimPLM \cite{tan2024small}; CRAG \cite{yan2024corrective}; \citet{feng2023knowledge}; \citet{liu2024ra}; HaluAgent \cite{cheng2024small}; Toolformer \cite{schick2024toolformer}, for tree={text width=11cm, top color=customyellow!100, bottom color=customyellow!100}]
                ]
                [SLM for extracting LLM prompts (\S \ref{slm4extract})
                    [Prompt Stealing Attacks \cite{sha2024prompt}; Output2prompt \cite{zhang2024extracting}; Model Purifying \cite{li2024purifying}; \cite{zhang2024effectivepromptextractionlanguage} , for tree={text width=11cm, top color=customyellow!100, bottom color=customyellow!100}]
                ]
			[SLM for fine-tuning LLMs (\S \ref{slm4finetune})
                    [Weak-to-Strong Search \cite{zhou2024weaktostrong}; Emulated Fine-tuning \cite{mitchell2024an}; CROSSLM \cite{deng2023mutual}; \citet{swayamdipta2020dataset}; \citet{mekala2024smaller}; Proxy-tuning \cite{liu2024tuning}, for tree={text width=11cm, top color=customyellow!100, bottom color=customyellow!100}]
                ]
			[SLM for LLM applications (\S \ref{slm4application})
                    [SLCoLM \cite{tang2024small}; HEF \cite{yang2024enhancing}; BLADE \cite{li2024bladeenhancingblackboxlarge}; Contrastive Decoding \cite{li2023contrastive}; ``Train-Guide-Predict'' \cite{tang2024small}; \citet{sennrich2023mitigating}, for tree={text width=11cm, top color=customyellow!100, bottom color=customyellow!100}]
                ]
			[SLM for LLM safety (\S \ref{slm4safety})
                    [Llama Guard \cite{inan2023llama}; \citet{kwon-etal-2024-slm}, for tree={text width=11cm, top color=customyellow!100, bottom color=customyellow!100}]
                ]
			[SLM for LLM evaluation (\S \ref{slm4evaluation})
                    [SLIDE \cite{zhao2024slide}; Semantic uncertainty \cite{kuhn2023semantic}; Selfcheckgpt \cite{manakul-etal-2023-selfcheckgpt}; Proxylm \cite{anugraha2024proxylm}; Factscore \cite{huang2023look}, for tree={text width=11cm, top color=customyellow!100, bottom color=customyellow!100}]
                ]
		]
		[Synergy between SLMs and LLMs (\S \ref{synergy})
			[Cloud-edge synergy (\S \ref{cloud_edge_synergy})
                    [CoGenesis \cite{zhang2024cogenesis}; \citet{hao2024hybrid}; \citet{xu2024large}; LLM-to-SLM \cite{bergner2024think}; Synergy of Thoughts \cite{shang2024defint}; CROSSLM \cite{deng2023mutual}, for tree={text width=11cm, top color=customyellow!100, bottom color=customyellow!100}]
                ]
			[Task-centric synergy (\S \ref{task_centric_synergy})
                    [$\alpha$-UMi \cite{shen-etal-2024-small}; SynCID \cite{liang2024synergizing}; Filter-then-rerank Framework \cite{ma2023large}; LLMCad \cite{xu2023llmcadfastscalableondevice}, for tree={text width=11cm, top color=customyellow!100, bottom color=customyellow!100}]
                ]
            ]
		[Trustworthiness in SLMs (\S \ref{security}) 
                [HELM~\cite{liang2023holistic}; Do-Not-Answer~\cite{wang-etal-2024-answer}; PromptRobust~\cite{zhu2023promptbench}; GLUE-X~\cite{yang-etal-2023-glue}; HaluEval~\cite{li-etal-2023-halueval}; PrivLM-Bench~\cite{li-etal-2024-privlm}; FFT~\cite{cui2023fft}; ROBBIE~\cite{esiobu2023robbie}; TrustLLM~\cite{sun2024trustllm}; RAmBLA~\cite{bolton2024rambla}; JailbreakBench~\cite{chao2024jailbreakbench}; \citet{xie2024online}; OR-Bench~\cite{cui2024or}; SORRY-Bench~\cite{xie2024sorry}; BeHonest~\cite{chern2024behonest}; \citet{hong2024decoding}; \citet{nakka2024device}; RUPBench~\cite{wang2024rupbench}, for tree={text width=15.53cm, top color=customyellow!100, bottom color=customyellow!100}]
            ]
		[Future Directions (\S \ref{challenge})
		]
	]
\end{forest}
}
\end{spacing}
\vskip -1em
\caption{Overview of Small Language Models.}
\label{fig:overview}
\end{figure}

The evolution of neural language models (LMs) from BERT's \cite{devlin2019bert} pre-training and fine-tuning paradigm to T5's \cite{raffel2020exploring} pre-training plus prompting approach, and finally to GPT-3's \cite{NEURIPS2020_1457c0d6} pre-training plus in-context learning, has greatly enhanced natural language processing (NLP). These advancements have broadened NLP's application across various fields, including language understanding \cite{wang2018glue}, programming \cite{nam2024using, team2024codegemma}, recommendation systems \cite{wang2024can}, information retrieval \cite{spatharioti2023comparing, changpre, humeaupoly, lu2020twinbert}, mobile-device control \cite{ding2024mobileagentenhancingmobilecontrol}, scientific discovery \cite{shojaee2024llm, zhang2024chemllm}, medical question answering \cite{wang2023augmenting, bolton2024biomedlm}, and legal question answering \cite{almeida2024exploring}.
In particular, the recent emergence of proprietary commercial models including ChatGPT, Bard, and Claude, and open-sourced models such as Llama~\cite{touvron2023llama,touvron2023llama2,dubey2024llama} has led to rapid growth in the development of large language models (LLMs). 
Even though neural networks consistently improve on various tasks with longer training times, larger datasets, and increased model sizes—a phenomenon known as a neural scaling law \cite{kaplan2020scaling}, these models unpredictably exhibit a sudden acquisition of versatile abilities, termed "\textit{emergent ability}," once they reach a critical scale threshold, thereby supporting the "larger is better" trend. This ability is not present in small-scale models.
For instance, the latest Llama-3.1 model with 405 billion parameters performs better in dialogue, logical reasoning, and programming compared to the smaller 7B counterpart \cite{dubey2024llama}.

Despite their prowess in complex tasks, LLMs' huge parameters and computational needs impose significant limitations, hindering their adoption in many real-world applications. For example, the LLaMa 3.1 model with 405 billion parameters \cite{dubey2024llama}, trained on 16K H100 GPUs for 54 days, requires about 202.5 GB of GPU memory using int4 precision and has large inference latency. These issues present several challenges in specific contexts: (1) LLMs are generally hosted in the cloud and used via cloud-based APIs due to the large GPU memory and computational cost. Users need to upload their data to query LLMs,  raising data leakage and privacy concerns, especially in high-stake scenarios such as healthcare, finance, and e-commerce;
(2) Driven by personal agents, on-device deployment is a critical requirement. Several factors, including cloud costs, latency, and privacy concerns, hinder the on-device processing of cloud-based LLMs, and direct deployment is impractical due to their high parameter and cache requirements, which often exceed the capabilities of devices such as mobile phones;
(3) Their large parameter count can cause inference delays from seconds to minutes, unsuitable for real-time applications. For instance, Llama 2 7B takes approximately 84 seconds to process 100 tokens on benchmarks including HellaSwag, TruthfulQA, MMLU, and Arc\_C when run on a smartphone equipped with a Snapdragon 685 processor~\cite{thawakar2024mobillama};
(4) To boost performance in specialized domains such as healthcare and law, where generic LLMs underperform, LLMs are often fine-tuned. However, this process is computationally expensive due to their large size.
(5) Though general-purpose LLMs are powerful, many real-world applications require only specific abilities and domain knowledge, deploying general-purpose LLMs would be a waste of resources and such LLMs often cannot match the performance of models tailored for specific tasks \cite{phogat2024finetuningsmallerlanguagemodels,guo2023improvingsmalllanguagemodels,jeong2023testtimeselfadaptivesmalllanguage,wang2024can,chen2024octopusondevicelanguagemodel}. 

\begin{figure}[t]
    \centering
    \includegraphics[width=0.86\linewidth]{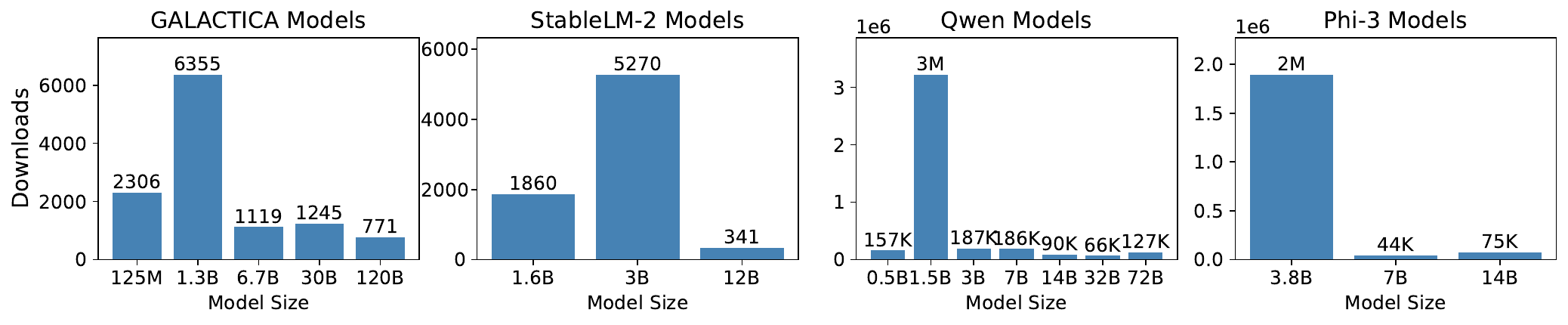}
    \vskip -1.5em
    \caption{Download Statistics Last Month in Huggingface for LLMs with Various Model Sizes, obtained on October 7, 2024.}
    \vskip -1em
\label{fig:slm_downloads_last_month}
\end{figure}

\begin{figure}[!t]
    \centering
    \includegraphics[width=0.8\linewidth]{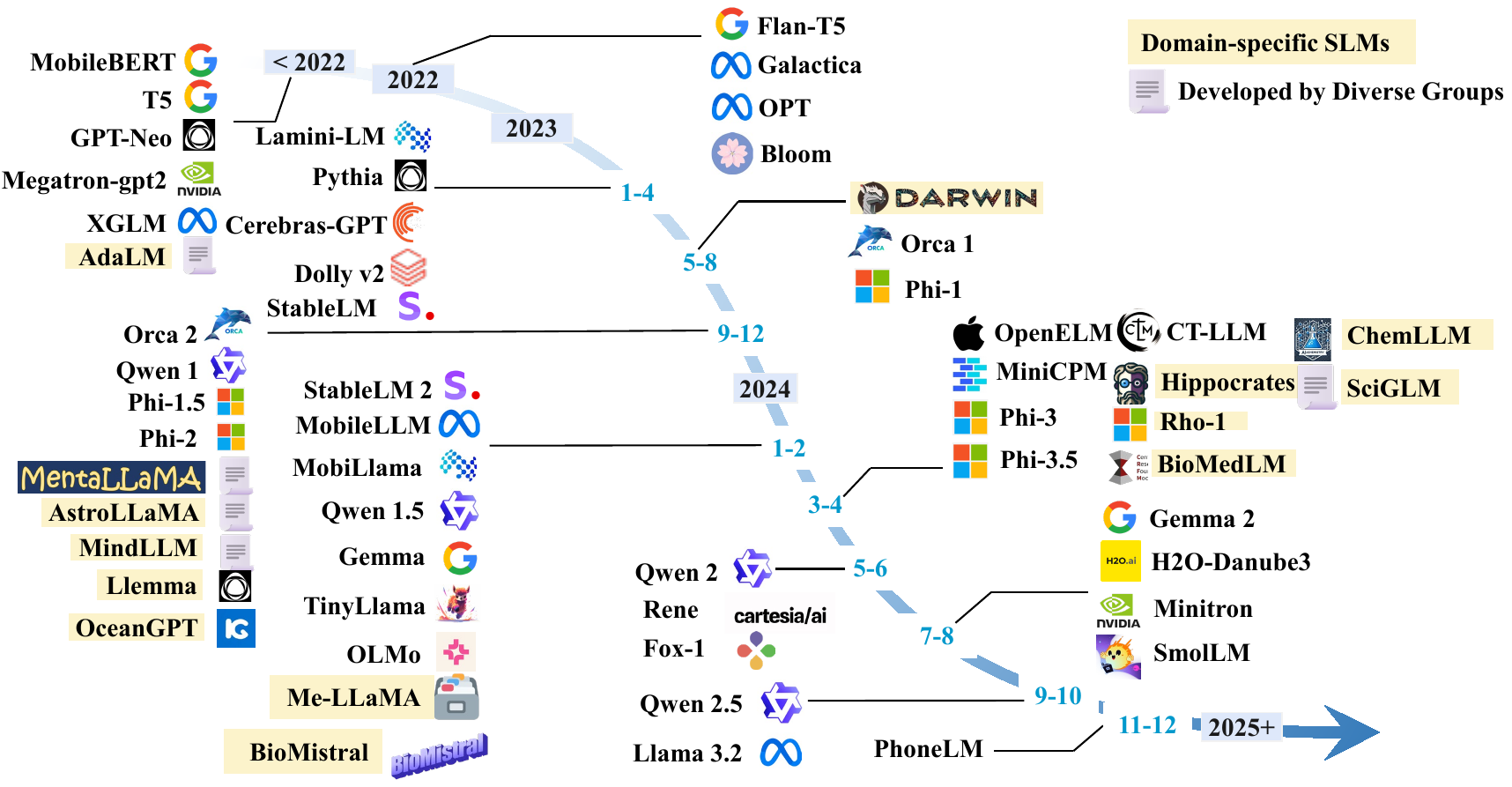}
    \vskip -1.5em
    \caption{A timeline of existing small language models.}
    \vskip -1em
    \label{fig:timeline}
\end{figure}

Recently, small language models (SLMs) have shown great potential in alleviating these issues while achieving performance comparable to LLMs for domain-specific problems \cite{yang2024qwen2, team2024gemma2, bellagente2024stable, pfeiffer2024h2o, zhang2024tinyllamaopensourcesmalllanguage, groeneveld2024olmo, hu2024minicpm, abdin2024phi, thawakar2024mobillama, liu2024mobilellm}. 
Owing to fewer parameters, SLMs excel in efficiency, cost, flexibility, and customization. They provide significant computational savings in pre-training and inference with reduced memory and storage needs, which is vital for applications requiring efficient resource use. These small models are especially effective in resource-limited settings, performing well on low-power devices such as edge devices. Besides, SLMs improve on-device processing by enhancing privacy, security, response times, and personalization. This supports advanced personal assistants and cloud-independent applications, boosting energy efficiency and reducing carbon emissions. For example, the Llama 3.2 models (1B \& 3B) demonstrate that local processing enables immediate execution of prompts and responses \cite{llama3.2}. This approach protects privacy by keeping sensitive data such as patient health information (PHI), business data, personal messages, and calendar details local, enhancing confidentiality. It also allows for precise control over which queries are processed on-device versus those requiring cloud-based models. Therefore, small language models are gaining increasing attention as alternatives to LLMs, as indicated in Figure \ref{fig:slm_downloads_last_month}, which shows that SLMs are downloaded more frequently than larger models in the Hugging Face community, and Figure \ref{fig:timeline}, which illustrates the growing popularity of SLM releases over time.

Typically, LMs that exhibit emergent abilities are classified as LLMs. However, the categorization of SLMs remains unclear. Studies vary in their contexts: some define SLMs as models with fewer than one billion parameters \cite{liu2024mobilellm}, while others consider the term ``small language model'' relative to the larger counterparts \cite{wang2024can, tang2024small, lee2024can}, with no consensus on a unified definition in the current landscape of LLMs. Research suggests SLMs for mobile devices, typically possessing around 6GB of memory, consist of sub-billion parameter models \cite{liu2024mobilellm}, whereas others classify models with up to 10 billion parameters as small, noting their lack of emergent abilities \cite{fu2023specializing}. Given their use in resource-constrained environments and for specific tasks, we propose a generalized definition: \textit{Given specific tasks and resource constraints, we define SLMs as falling within a range where the lower bound is the minimum size at which the model exhibits emergent abilities for a specialized task, and the upper bound is the largest size manageable within limited resource conditions.} This definition integrates various perspectives and addresses factors related to mobile computing and capability thresholds.


Due to the growing demand for SLMs, extensive literature has emerged on various aspects of SLMs. For example, several resource-efficient techniques \cite{xu2024survey} and training methods optimized for SLMs, such as quantization-aware training \cite{xu2024onebit,wang2023bitnet,liu2023llm} and selective architectural component choices \cite{qu2024survey, thawakar2024mobillama, liu2024mobilellm}, aim to enhance performance in specific applications \cite{phogat2024finetuningsmallerlanguagemodels, roziere2023code, wu2024could, bonifacio2022inpars, chen2024octopusondevicelanguagemodel}. These methods have led to the development of numerous open-source, general-purpose, and domain-specific SLMs \cite{yang2024qwen2, team2024gemma2, bellagente2024stable, acikgoz2024hippocrates, bolton2024biomedlm, zhang2024sciglm}. Beyond their inherent capabilities, SLMs can enhance LLMs by serving as modules or effective proxies \cite{zhao2023automatic, xu2023understanding, wu2024divide, sha2024prompt, mitchell2024an, yang2024enhancing}. Furthermore, the complementary advantages of SLMs and LLMs can be leveraged collectively to better complete tasks \cite{zhang2024cogenesis, deng2023mutual, shen-etal-2024-small, liang2024synergizing, ma2023large}. Despite the commendable performance of SLMs, it is crucial not to overlook their credibility issues, such as the risks of adversarial attacks, producing hallucinations, and privacy breaches \cite{dominguez2023questioning, hong2024decoding, egashira2024exploiting, kumar2024increased,nakka2024device,perez-etal-2023-discovering,mo2023trustworthy,wang2024rupbench,kumar2024increased, WangCPXKZXXDSTA23,yuan2024s, liu-etal-2023-maximum, wang2021macrobert,wang-etal-2024-unlocking, zhang2024does}. However, currently, there is no comprehensive survey thoroughly exploring these works on SLMs in the era of LLMs. Therefore, this paper offers the first comprehensive survey that analyzes various aspects of SLMs in the LLM era and their future directions. The overview structure of our paper is shown in Figure \ref{fig:overview}. To summarize, our major contributions are: 
\begin{itemize}[leftmargin=*]
\setlength{\itemsep}{0pt}
\setlength{\parsep}{0pt}
\setlength{\parskip}{0pt}
    \item In Section \ref{enhancement}, we examine various techniques for improving the performance of SLMs, including training from scratch, fine-tuning, knowledge distillation, quantization, and leveraging LLM-enhancing technologies to optimize SLMs.
    \item In Section \ref{application}, we discuss the tasks that SLMs can enhance and the deployment strategies that enable models to fit within the resource constraints of edge devices while maintaining acceptable inference speed.
    \item In Section \ref{model}, we collect SLMs with fewer than 7 billion parameters across both general-purpose and domain-specific applications, reviewing common architectural choices, training techniques, and datasets, and providing a comparative summary of performance across different model sizes. Recent SLMs are listed.
    \item In Section \ref{slm4llm}, we explore how SLMs can address key challenges faced by LLMs, such as high inference latency, labor-intensive fine-tuning, susceptibility to knowledge noise, and risks of copyright infringement. 
    \item In Section \ref{synergy}, we survey two kinds of synergies between LLMs and SLMs: one involves cloud-based LLMs and local SLMs, while the other leverages the unique advantages of both to more effectively solve tasks. 
    \item In Section \ref{security}, we investigate the trustworthiness issues of SLMs, including hallucination and privacy concerns, by providing a taxonomic summary of current evaluation methods.
\end{itemize}

Concurrently with our survey, \citet{lu2024small} evaluate open-source SLMs, focusing on their architectures, datasets, algorithms, and on-device performance metrics such as inference latency and memory usage. \citet{van2024survey} delve into optimization strategies for SLMs, including model compression, pruning, and quantization. \citet{chen2024role} investigate how SLMs enhance LLMs and vice versa. In contrast, our survey offers a more comprehensive review with the following differences: (1) we present a detailed taxonomy of recent advancements in SLMs in the era of LLMs; (2) we define SLMs based on emergent capabilities and device specifications, which refines previous unclear definitions related to LLMs; (3) we discuss SLM applications, especially in on-device tasks and deployment, topics previously unexplored; (4) we examine domain-specific SLMs previously overlooked; and (5) we additionally consider the synergy between SLMs and LLMs.

%% file: sections/3.construction_of_slm.tex
\section{Foundational Concepts in Building Language Models}  
\label{construction}
This section will introduce foundational concepts and background knowledge for language models, including the concepts of architecture and the training process, as well as methods for obtaining SLMs from LLMs. The advanced training strategy to improve SLM performance will be introduced in Section \ref{enhancement}.

\subsection{Architecture of SLMs}
\label{architecture}

\begin{wrapfigure}[17]{r}{0.3\textwidth}
    \centering
    \vskip -1.5em
    \includegraphics[width=0.3\textwidth]{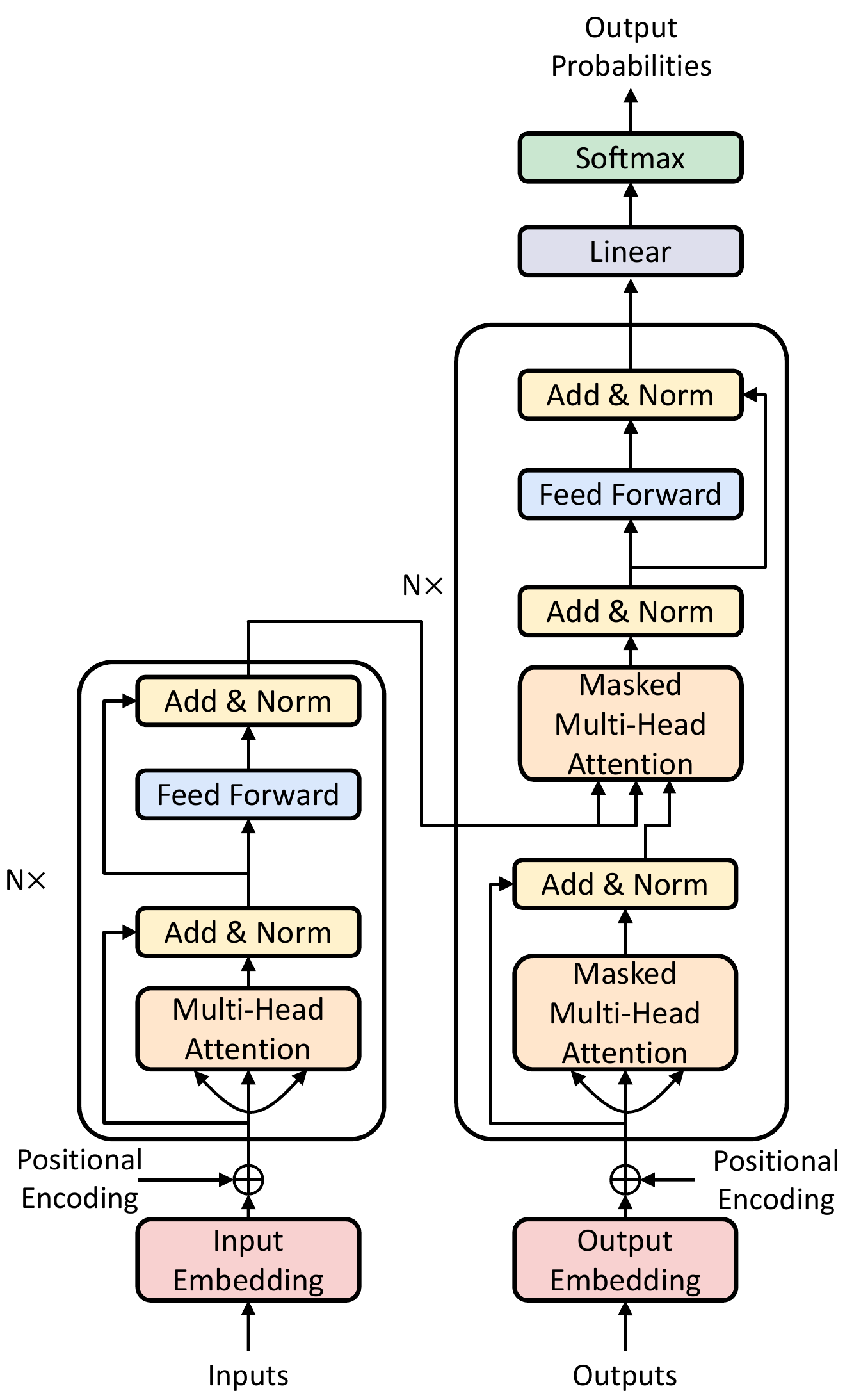}
    \vskip -1em
    \caption{Transformer architecture  \cite{vaswani2017attention}.}
    \label{fig:transformer}
\end{wrapfigure}

SLMs commonly employ the Transformer architecture \cite{vaswani2017attention} (see Figure \ref{fig:transformer}), which utilizes self-attention mechanisms to manage long-range text dependencies, essential for maintaining performance with constrained resources. However, due to the attention mechanism, Transformers have large inference cost. Hence, to alleviate the issue, several subquadratic-time architectures such as Mamba~\cite{gu2023mamba}, Hymba~\cite{dong2024hymba}, and xLSTM~\cite{beck2024xlstm} are proposed. Next, we will give details of Transformer due to its popularity and briefly introduce newly emerged models.

\subsubsection{Transformer}
The Transformer's self-attention mechanism \cite{vaswani2017attention} allows language models to efficiently capture contextual information across longer sequences, even with limited resources. The Transformer generally adopts an encoder-decoder structure featuring self-attention mechanisms, feedforward networks, positional embeddings, and layer normalization. The Transformer architecture design tailored for SLMs is detailed in Section \ref{model}; this section will provide only foundational concepts.

\textbf{Self-Attention Mechanism} enables the model to evaluate the importance of tokens relative to each other. The self-attention mechanism is written as
\begin{equation}
\centering
    \text{Attention}(\rmQ, \rmK, \rmV) = \text{softmax}\left(\frac{\rmQ \rmK^\top}{\sqrt{d_k}}\right)\rmV \nonumber
\end{equation}
where \( \rmQ \), \( \rmK \), and \( \rmV \) are query, key, and value matrices, scaled by \( \sqrt{d_k} \) for stability where $d_k$ is the dimension of key matrices. The dot product \( \rmQ \rmK^\top\) reflects the similarity between the query and key vectors. 

\textbf{Multi-Head Attention (MHA) \cite{vaswani2017attention}}  is the first method that uses multiple heads to capture diverse information.
MHA allows the model to attend to different parts of the input sequence using multiple attention heads as 
\begin{equation}
    \text{MultiHead}(\rmQ, \rmK, \rmV) = \text{Concat}(\text{head}_1, \text{head}_2, \dots, \text{head}_h)\rmW^O, ~\text{with}~ \text{head}_i = \text{Attention}(\rmQ \rmW_i^Q, \rmK \rmW_i^K, \rmV \rmW_i^V)
\end{equation}
Each head in the Multi-Head Attention mechanism operates independently, allowing the model to capture diverse aspects of the data. The outputs are combined using learned projection matrices \( \rmW_i^Q \), \( \rmW_i^K \), and \( \rmW_i^V \), concatenated, and passed through the output projection matrix \( \rmW^O \). 

Building on this foundation, several modifications have been introduced to further optimize self-attention mechanisms for specific challenges such as memory efficiency and computational speed. To address the KV-cache bottleneck in MHA, \textbf{Multi-Query Attention (MQA) \cite{shazeer2019fast}} proposes that all attention heads share the same set of keys and values, which reduces the memory and computational overhead associated with storing and managing multiple key-value pairs. \textbf{Grouped Query Attention (GQA) \cite{ainslie2023gqatraininggeneralizedmultiquery}} serves as a middle ground between MHA and MQA. It introduces subgroups of query heads (fewer than the total number of attention heads), where each subgroup shares a single key and value head. Unlike MQA and GQA, which reduce the number of key and value heads, \textbf{Multi-Head Latent Attention (MLA) \cite{liu2024deepseek}}  compresses the keys and values into a joint latent vector. This compression allows for efficient handling of key-value pairs while maintaining high performance, significantly reducing the KV-cache and improving inference efficiency. \textbf{Flash Attention}~\cite{dao2022flashattention,daoflashattention} accelerates the self-attention mechanism by minimizing the memory overhead typical of standard attention calculations. This optimization allows SLMs to process longer sequences more efficiently, enhancing their functionality under strict hardware constraints.

\textbf{Feedforward Network (FFN)} comprises two linear transformations separated by a non-linearity, typically modeled as \( \text{FFN}(\mathbf{x}) = \sigma(\mathbf{x}\rmW_1 + b_1)\rmW_2 + b_2 \). where \( \rmW_1 \) and \( \rmW_2 \) are the weight matrices, and \( b_1 \) and \( b_2 \) are bias terms. $\sigma$ is the activation function, which introduces non-linearity, allowing models to learn complex patterns. Generally, ReLU is used as the activation function. In addition to ReLU, activation functions such as GeLU and SiLU are also used in SLMs to improve performance. We give the details here:
(i) \textbf{ReLU (Rectified Linear Unit) \cite{relu}} is defined as $\sigma(x) = \max(0, x)$, which is commonly used for its simplicity and effectiveness.
(ii) \textbf{GELU (Gaussian Error Linear Unit) \cite{gelu}} is defined as 
$\text{GELU}(x) = x \cdot \Phi(x) = x \cdot \frac{1}{2} \left[1 + \text{erf}\left(\frac{x}{\sqrt{2}}\right)\right]$, where \( \Phi(x) \) is the standard Gaussian CDF and \textit{erf} is the error function. It is smoother than ReLU and widely used in models such as BERT~\cite{devlin2019bert} and GPT~\cite{radford2019language} for better gradient flow control. 
Since calculating the Gaussian error function for each neuron is computationally expensive and time-consuming, there are approximations using tanh and sigmoid functions, corresponding to $\text{GELU}_{\text{tanh}}$ and $\text{SiLU}$:
(iii) \textbf{GELU with tanh} is defined as $\text{GELU}_{\text{tanh}}(x) = 0.5 \cdot x \cdot \left(1 + \tanh\left(\sqrt{\frac{2}{\pi}} \cdot (x + 0.044715 \cdot x^3)\right)\right)$. This approximation uses the Tanh function to simplify computations.
(iv) \textbf{SiLU (Sigmoid Linear Unit) \cite{silu}} is calculated as $\text{SiLU}(x) = x \cdot \text{sigmoid}(x) = x \cdot \frac{1}{1 + e^{-x}}$. It effectively combines the sigmoid function with its input, enhancing modeling capabilities.
(v) \textbf{SwiGLU (Swish-Gated Linear Units)}~\cite{swiglu} integrates the Swish activation function with Gated Linear Units, defined as $\text{SwiGLU}(x) = \text{Swish}(x \cdot W + b) \odot (x \cdot V + c)$ where $W, V$ are the weight matrix and $b,c$ are the bias terms. The Swish function is expressed as $\text{Swish}(x) = x \cdot \text{sigmoid}(x)$. This combination enhances expressiveness and computational efficiency, making it a preferred choice in advanced models such as the Qwen series \cite{yang2024qwen2}.



\textbf{Positional Embeddings} in Transformer models \cite{vaswani2017attention} are essential for capturing token order, providing context about relative positions within a sequence. Traditional positional embeddings in the Transformer architecture utilize a sinusoidal function, defined as:
$
    PE(pos, 2i) = \sin\left(\frac{pos}{10000^{2i/d_{\text{model}}}}\right), \quad PE(pos, 2i+1) = \cos\left(\frac{pos}{10000^{2i/d_{\text{model}}}}\right)
$
where \( pos \) represents the position within the sequence, \( i \) is the dimension index, and \( d_{\text{model}} \) is the dimensionality of the model. 
To improve the model's capacity for understanding the relative positions of tokens within a sequence, \textbf{Rotary Positional Embedding (RoPE) \cite{su2024roformer}} introduces a rotational matrix to the embeddings. RoPE significantly enhances the positional encoding by maintaining the relative distances through rotational transformations, thus optimizing the model’s interpretative ability regarding sequence dynamics.

\textbf{Layer Normalization~\cite{lei2016layer}} stabilizes the training process by normalizing layer outputs, accelerating convergence. Two types of layer normalization are commonly used~\cite{lei2016layer}:
(i) \textbf{Non-Parametric Layer Norm} normalizes inputs using the mean and variance calculated across the layer's dimensions without learnable parameters as 
$
\text{LN}(x) = \frac{x - \mu}{\sigma}
$
where \( \mu \) is the mean and \( \sigma \) is the standard deviation of the inputs. Its simplicity makes it ideal for SLMs. (ii) \textbf{Parametric Layer Norm} includes learnable parameters \(\gamma\) and \(\beta\) for adaptive scaling and bias, enhancing model flexibility:
$
\text{PLN}(x) = \gamma \left(\frac{x - \mu}{\sigma}\right) + \beta
$
Additionally, \textbf{RMS Norm (Root Mean Square Layer Normalization)~\cite{zhang2019root}} simplifies the calculation by using the root mean square of inputs, reducing computational demands:
$
\text{RMSNorm}(x) = \gamma \frac{x}{\sqrt{\frac{1}{N}\sum_{i=1}^{N}x_i^2 + \epsilon}} + \beta
$
where \( N \) is the number of inputs, \( x_i \) is the \( i \)-th input, and \( \epsilon \) is a small constant to prevent division by zero.

\begin{wrapfigure}[10]{r}{0.4\textwidth}
    \centering
    \vskip -0.6em
    \includegraphics[width=\linewidth]{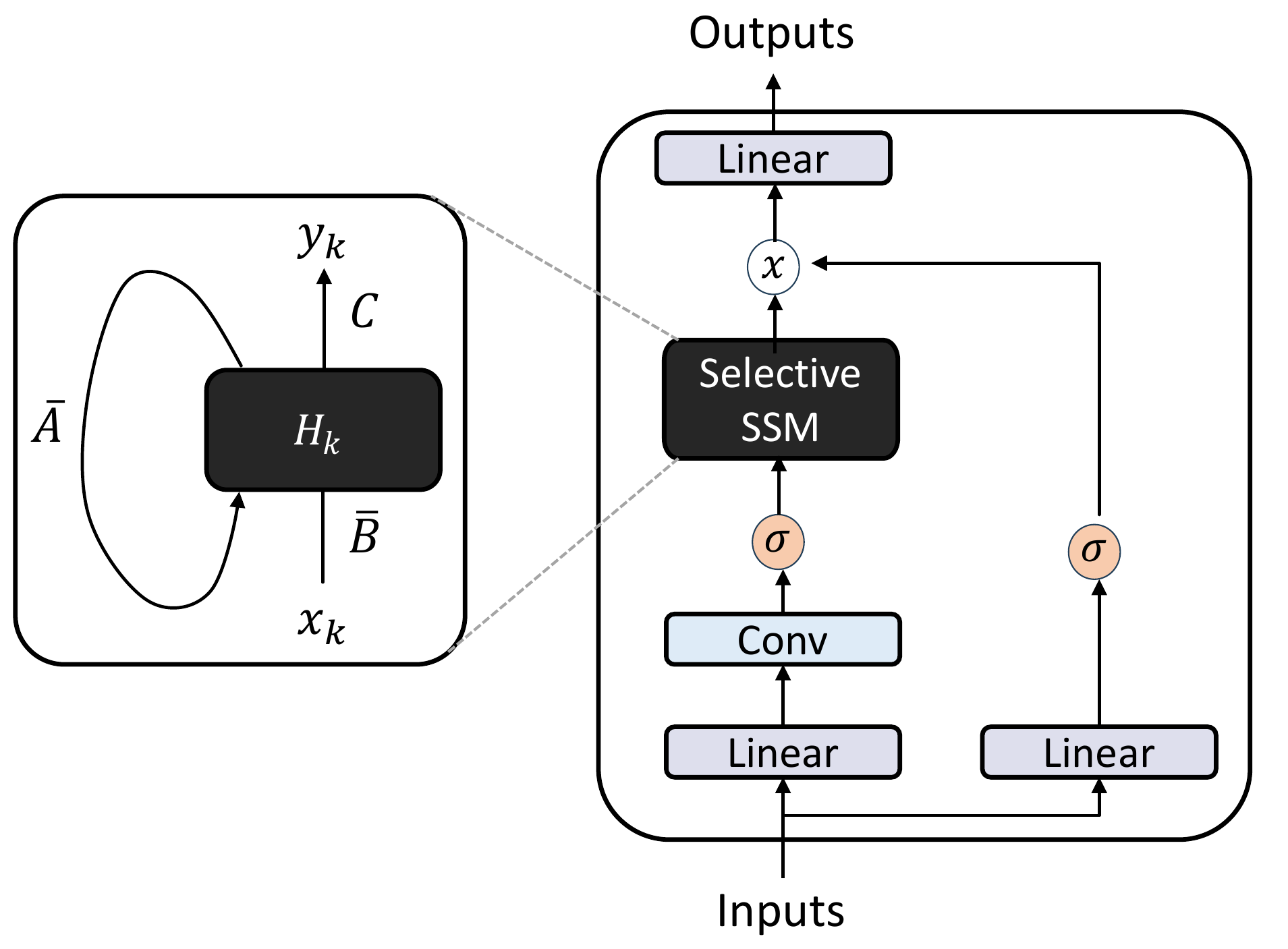}
    \vskip -1em
    \caption{Mamba 1 architecture \cite{gu2023mamba}. } 
    \label{fig:mamba}
\end{wrapfigure}
\subsubsection{Mamba}
The attention mechanism in Transformer suffers from a drawback: it requires recalculating attention scores with every previous token for each new token generated during inference, leading to quadratic time complexity. This increases the inference cost as sequence lengths grow. In contrast, Mamba \cite{gu2023mamba,dao2024transformers}, based on state space models (SSMs) \cite{kalman1960new} which are a superclass of recurrent neural networks (RNNs), rely only on the last hidden state for generating the next token, enabling faster inference speeds, as shown in Figure \ref{fig:mamba}. 
To address the Linear Time Invariant nature of traditional SSMs, which hinders their ability to focus on or ignore specific inputs, Mamba improves SSMs with a dynamic selection mechanism. This mechanism selectively filters out irrelevant information while retaining essential data, tailored to the content of the input. Leveraging this selective SSM foundation, Mamba adeptly captures complex global relationships within sequence data.
Due to its focus on the immediate previous hidden state, as opposed to Transformer which requires access to all previous hidden states, Mamba achieves a higher utilization rate of model parameters. This makes it more suitable for SLMs.
However, we identify two drawbacks of Mamba: (i) its focus on selectively capturing global information may compromise performance on tasks that require nuanced understanding, such as detailed sentiment analysis or complex entity recognition; and (ii) to balance inference speed, Mamba's recurrent structure primarily encodes static global information, which limits its effectiveness in handling multi-round tasks within a single query, such as interactive dialogue systems or iterative problem-solving scenarios.

In language modeling, Mamba 1 \cite{gu2023mamba} is pre-trained on the Pile dataset \cite{gao2020pile} using the training recipe from \cite{NEURIPS2020_1457c0d6} and ranges from 125M to 1.3B parameters. It outperforms comparable models such as Pythia \cite{biderman2023pythiasuiteanalyzinglarge} and RWKV \cite{peng-etal-2023-rwkv} in various tasks; for instance, Mamba-1.4B achieves a 32.8\% accuracy on the Arc-Challenge \cite{clark2018think} dataset, surpassing Pythia-1.4B's 28.5\% and RWKV-1.5B's 29.4\%. Mamba 2 \cite{dao2024transformers} develops a theoretical framework linking SSMs with attention mechanisms through structured semi-separable matrices, enhancing the selective SSM to achieve 2-8x faster speeds while competing with Transformer models. Training and configuration for Mamba 2 align with Mamba 1. Additionally, Mamba-series models are applied widely across different fields \cite{zuo2024falcon, qu2024survey, jiang2023mistral, qu2024ssd4rec}. Other follow-up Mamba-based language models such as Falcon Mamba 7B \cite{zuo2024falcon} and Mistral 7B \cite{jiang2023mistral} also demonstrate the efficiency of the architecture for NLP tasks. Falcon Mamba 7B scales Mamba’s long-sequence processing capabilities to extensive language data, reducing memory overhead and making it ideal for long-form NLP tasks. Similarly, Mistral 7B incorporates Mamba’s efficient sequence handling ability to improve processing speed and computational efficiency in long-context NLP tasks, showcasing Mamba’s scalability and practicality for large-scale language modeling.

\begin{wrapfigure}[8]{r}{0.6\textwidth}
    \centering
    \vskip -0.5em
    \includegraphics[width=0.6\textwidth]{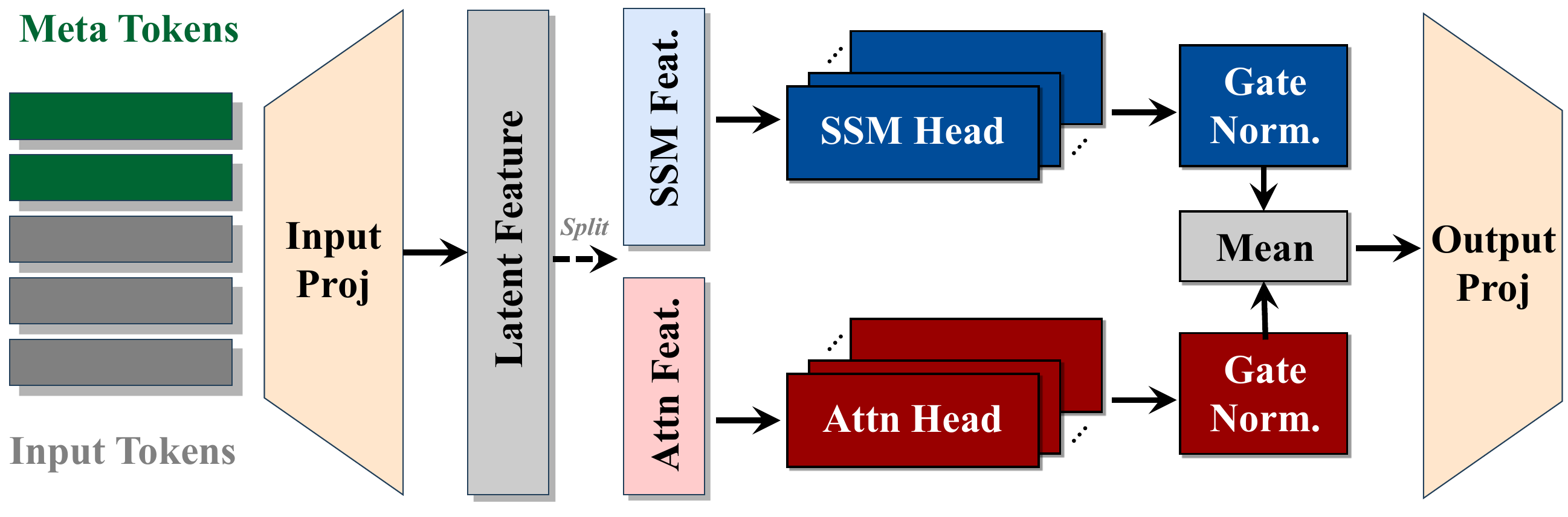}
    \vskip -1em
    \caption{Hymba \cite{dong2024hymba} architecture.}
    \label{fig:hymba}
\end{wrapfigure}
\subsubsection{Hymba}
Attention heads in the Transformer facilitate high-resolution recall, while SSM heads in Mamba efficiently summarize context. To balance performance and efficiency for SLMs, Hymba \cite{dong2024hymba} integrates both attention and SSM heads within the same layer, allowing for parallel and complementary processing of inputs, as depicted in Figure \ref{fig:hymba}.
This hybrid-head approach enables each layer to simultaneously leverage the high-resolution recall of attention heads and the contextual summarization of SSMs, increasing the model’s flexibility and expressiveness in managing diverse information flows and memory
access patterns. 

Hymba has been developed in models of varying sizes—125M, 350M, and 1.5B, trained on a combination of the DCLM-Baseline-1.0 \cite{li2024datacomp}, SmolLM-Corpus \cite{benallal2024smollmcorpus}, and a proprietary high-quality dataset, with token counts of 1 trillion, 250 billion, and 50 billion, respectively. The models incorporate the Warmup-Stable-Decay (WSD) learning rate scheduler \cite{hu2024minicpm} and the data annealing technique \cite{dubey2024llama} to ensure stable pretraining, conducted on 128 NVIDIA A100 GPUs. The 1.5B base model was post-trained using full finetuning (FFT), followed by direct preference optimization (DPO) \cite{rafailov2024direct} to develop the Hymba-1.5B-Instruct model. In commonsense reasoning tasks, the Hymba 1.5B model surpasses Llama-3.2-3B \cite{llama3.2} by achieving 1.32\% higher average accuracy, requiring an 11.67$\times$ smaller cache size, and delivering a 3.49$\times$ increase in processing speed.

\subsubsection{xLSTM} 
Long Short-Term Memory (LSTM) \cite{hochreiter1996lstm} shares a conceptual similarity with Mamba that achieves success in language modeling through the introduction of time-dependent weights. This similarity raises an intriguing question: how effective would LSTMs be at language modeling if scaled to billions of parameters, incorporating advanced techniques from modern LLMs while addressing known LSTM limitations? Inspired by this question, \citeauthor{beck2024xlstm}~\cite{beck2024xlstm} propose xLSTM architecture, which performs favorably compared to state-of-the-art Transformers and SSMs in empirical evaluations.
To address the limitations of LSTM, xLSTM designs \textit{exponential gates} to enhance effectiveness with long sequences, expand memory cells from scalars to matrices to increase storage capacity, and remove memory mixing to enable parallel processing.

To test the language modeling capabilities of xLSTM scaled to billions of parameters, it is trained on a large dataset comprising 300 billion tokens from SlimPajama \cite{cerebras2023slimpajama} across various model sizes (125M, 350M, 760M, 1.3B). The performance of pre-trained xLSTM is compared against RWKV-4 \cite{peng-etal-2023-rwkv}, Llama \cite{inan2023llama}, and Mamba \cite{gu2023mamba} across 571 text domains of the PALOMA benchmark \cite{magnusson2023paloma} and various downstream tasks. Across all model sizes and the majority of tasks, xLSTM consistently outperforms the others, suggesting that larger xLSTM models could become formidable competitors to existing Large Language Models that utilize Transformer technology.


\subsection{Training SLMs from Scratch}
\label{training_techniques}
Training SLMs from scratch entails several critical steps: 
(i) Pre-training, focused on acquiring general features and knowledge from the corpus; 
(ii) Fine-tuning, targeted at boosting the model's abilities and performance for specific tasks; 
(iii) Decoding strategies, which involve the methods used for iteratively selecting the next token during generation.



\subsubsection{Pre-training}
Typically, the pre-training paradigm for language models is divided into encoder-based and decoder-based approaches. Encoder-based models, such as BERT~\cite{devlin2019bert}, utilize Masked Language Modeling (MLM) tasks where the goal is to predict masked tokens within a sentence. This is achieved by maximizing:
\[
P(\text{masked token} \mid \text{context}) = \text{softmax}(\mathbf{W} \cdot \mathbf{h}_{\text{mask}} + b),
\]
where $\text{masked token}$ is the original token that has been masked, $\text{context}$ represents the other unmasked tokens in the sentence, $\mathbf{W}$ and $b$ are trainable parameters of a linear output layer, $\mathbf{h}_{\text{mask}}$ is the output from the transformer encoder for the masked position, and $\text{softmax}$ is the activation function that converts logits to probabilities over the vocabulary. This process enhances the model's language encoding capabilities. Decoder-based models, such as GPT~\cite{radford2019language}, employ Next Token Prediction (NTP) tasks, aiming to model the distribution of the next token by maximizing:
\[
P(\text{next token} \mid \text{context}) = \text{softmax}(\mathbf{W} \cdot \mathbf{h}_{\text{last}} + b),
\]
where $\text{next token}$ is the token that the model aims to predict, $\text{context}$ represents the sequence of tokens preceding the token to be predicted, and $\mathbf{h}_{\text{last}}$ is the output from the transformer encoder for the last token in the context.
Effective data preprocessing, crucial for optimizing the performance of SLMs trained from scratch, involves meticulous data cleaning and strategic tokenization.

\textbf{Data Cleaning} involves techniques such as filtering, deduplication, and noise reduction, which improve data quality and help the model generalize better. Filtering noisy or irrelevant data, addressing outliers, and handling imbalances in the dataset ensure that the training data is both representative and efficient. Deduplication, in particular, helps prevent overfitting by removing repeated instances, making the model more robust with efficient parameter usage. 

\textbf{Tokenization} plays a vital role in handling diverse vocabularies without increasing model size. Advanced methods such as Byte-Pair Encoding (BPE)~\cite{gage1994new} and WordPiece~\cite{song2021fast} break text into subwords \cite{devlin2019bert}, allowing the model to manage rare and compound words efficiently. These strategies ensure that SLMs maintain a balance between vocabulary coverage and model compactness, crucial for improving generalization while minimizing computational demands.

\subsubsection{Fine-Tuning} 
After the initial training, SLMs are fine-tuned on specific tasks using task-specific data and loss functions. Parameter-efficient fine-tuning methods \cite{hu2021lora, li2021prefix, houlsby2019parameter, gu-etal-2024-light}, such as Low-Rank Adaptation (LoRA), prefix-tuning, and adapter modules, are particularly effective for SLMs. \textbf{Low-Rank Adaptation (LoRA)~\cite{hu2021lora}} modifies Transformer weights by introducing trainable low-rank matrices \( \mathbf{A} \) and \( \mathbf{B} \) for efficient fine-tuning, avoiding significant alterations to pre-trained weights. The update is represented as:
$
    \Delta \mathbf{W} = \mathbf{A} \mathbf{B}^\top
$
The fine-tuned weight matrix used in Transformer operations then becomes:
$
    \mathbf{W}_{\text{ft}} = \mathbf{W} + \alpha \Delta \mathbf{W}
$
where \( \alpha \) is a scaling factor adjusting the adaptation's impact, allowing fine-tuning on a smaller set of parameters while retaining the model's foundational capabilities.
\textbf{Prefix-Tuning~\cite{li2021prefix}} prepends learnable prefixes to the input sequence, guiding the model's attention without altering core model parameters. It is especially useful for generative tasks. 
\textbf{Adapter Modules~\cite{houlsby2019parameter}} are small, trainable layers inserted into the pre-trained model. These layers are fine-tuned on task-specific data, allowing the base model to remain fixed while the adapters learn the necessary adjustments. The typical structure of an adapter module includes a down-projection, a non-linearity, and an up-projection:
$
\text{Adapter}(\mathbf{h}) = \mathbf{h} + \mathbf{W}_{\text{up}} \cdot \sigma(\mathbf{W}_{\text{down}} \cdot \mathbf{h} + \mathbf{b}_{\text{down}}) + \mathbf{b}_{\text{up}}
$
where \(\mathbf{h}\) is the input hidden state, \(\mathbf{W}_{\text{down}}\) and \(\mathbf{W}_{\text{up}}\) are the projection matrices, \(\mathbf{b}_{\text{down}}\) and \(\mathbf{b}_{\text{up}}\) are the bias terms, and \(\sigma\) is a non-linear activation function. 


\subsubsection{Decoding Strategies}
After pre-training or fine-tuning, employing an effective decoding strategy is crucial for generating output from language models. Decoding, the process of text generation from SLMs, involves iteratively selecting the next word. A fundamental method is the greedy search, which predicts the most likely token at each step. This is formally modeled as:
$x_i = \arg \max_x P(x \mid x_{<i})$,
where \(x_i\) is the token with the highest probability at the \(i\)-th step, conditioned on the preceding context \(x_{<i}\).
Other decoding strategies, such as beam search or top-k sampling, are crucial for generating high-quality outputs. Beam search balances exploration and exploitation by considering multiple possible sequences simultaneously, while top-k sampling introduces diversity and creativity in text generation. These strategies collectively ensure that SLMs are efficient and capable of delivering high performance across various natural language processing tasks.

\subsection{Obtain SLMs from LLMs} 
\label{compression}



Obtaining an SLM from an LLM is crucial for deploying in resource-constrained environments. Instead of training from scratch, leveraging an LLM allows for knowledge transfer, enabling SLMs to retain much of the LLM's linguistic and domain knowledge with reduced training time and data. To obtain SLMs from LLMs, three primary techniques are used: pruning, knowledge distillation, and quantization. Pruning removes less critical parameters, reducing model size while aiming to maintain performance.  Knowledge distillation transfers knowledge from a large teacher model to a smaller student model, preserving much of the original model's understanding. Quantization decreases parameter precision, significantly lowering memory and computation needs with minimal impact on accuracy. These methods balance size reduction, efficiency, and performance retention.



\begin{wrapfigure}[9]{r}{0.35\textwidth}
    \centering
    \vskip -2.4em
    \includegraphics[width=\linewidth]{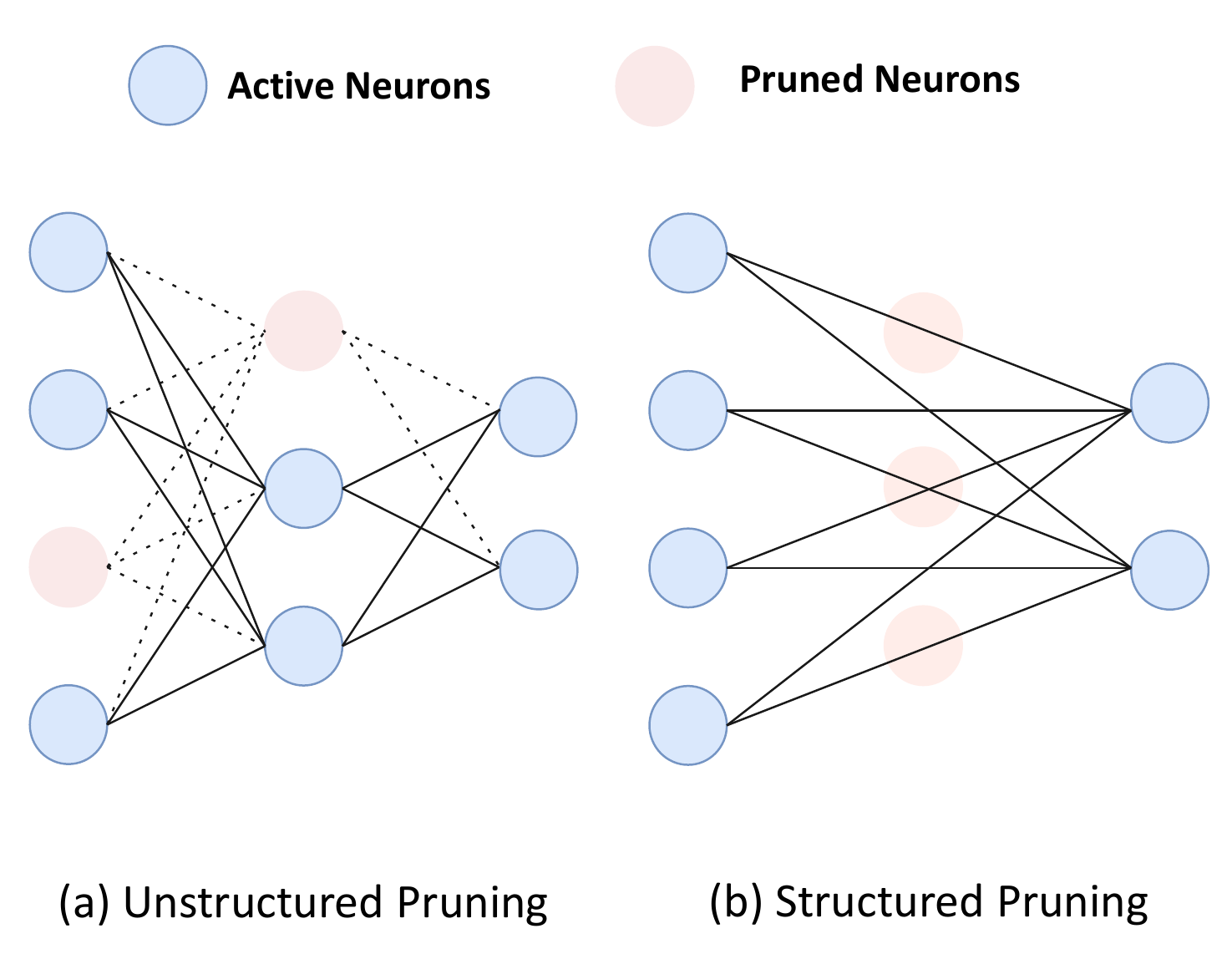}
    \vskip -1em
    \caption{Unstructured and structured pruning.}
    \label{fig:pruning}
\end{wrapfigure}
\subsubsection{Pruning}
\label{pruning}
Pruning is a technique used to reduce a model's size and computational requirements (e.g., LLMs) without significantly sacrificing its performance \cite{han2015learning}. This process involves identifying and removing less important or redundant parameters and components from the model. The primary goal of LLM pruning is to make the model more efficient, faster, and suitable for deployment in resource-constrained environments. Typically, pruning can be categorized into two main types: \textit{unstructured pruning} and \textit{structured pruning}. An illustration of unstructured pruning and structured pruning is shown in Figure~\ref{fig:pruning}.

\textbf{Unstructured Pruning}~\cite{frantar2023sparsegpt,sun2024a,zhang2023loraprune,zhang2024plug,li2023sparse,shao2024one,das2023beyond} prunes an LLM by removing weights individually without considering its internal structure. The least significant parameters are pruned according to specific criteria (e.g. magnitude or impact on the output). This method can achieve significant compression while maintaining performance. However, it can also lead to irregular memory access patterns and reduced hardware efficiency because the pruned model lacks a regular structure. SparseGPT \cite{frantar2023sparsegpt} is a representative unstructured pruning method that can reduce large-scale GPT models like OPT-175B \cite{zhang2022opt} and BLOOM-176B \cite{le2023bloom} to up to 60\% sparsity using a novel sparse regression solver. Wanda \cite{sun2024a} combines weight magnitudes with input activations to efficiently identify and discard less impactful parameters. It operates in a single forward pass, rapidly achieving high sparsity without retraining. It is also worth noting that recent studies specifically address the compatibility issues between pruning and Low-rank Adaptation (LoRA) \cite{hu2021lora}, such as LoRAPrune \cite{zhang2023loraprune}. 

\textbf{Structured Pruning}~\cite{men2024shortgpt,ma2023llm,li2024nuteprune,yang2024laco,zhaoapt,li2024lorap,an2024fluctuation,ashkboos2024slicegpt,chen2023lorashear,xia2024sheared,guo2023compresso,shen-etal-2024-pruning,shen-etal-2022-cost,gao2024displlm}, which prunes an LLM by targeting entire structural components—such as neurons, channels, or layers—rather. This approach allows for a direct reduction in dimensionality, thus efficiently reducing model complexity and memory usage. Although structured pruning may lead to higher accuracy degradation than unstructured pruning, it simplifies implementation without requiring specialized hardware. ShortGPT \cite{men2024shortgpt} proposes the Block Influence (BI) metric, which measures the significance of each layer based on its transformation of hidden states. Essentially, a transformer block's influence is measured by how much it alters the hidden states. By calculating BI scores, ShortGPT determines which layers contribute minimally to the overall performance and removes these low-importance layers. This simple yet effective layer removal strategy significantly reduces the model's parameters and computational requirements. LLM Pruner \cite{ma2023llm} offers a method to efficiently prune LLMs without access to the original training dataset. It employs a three-step compression pipeline: Discovery (identifying interdependent structures), Estimation (evaluating the performance impact of each group), and Recovery (post-training to address performance loss). NutePrune~\cite{li2024nuteprune} enhances structured pruning with a Numerous-teacher method, employing variable sparsity masks and LoRA modules to guide the pruning process. This approach effectively reduces model size and complexity. COST-EFF \cite{shen-etal-2022-cost} introduces a slenderized backbone—a form of structured pruning—and a multi-exit model that employs task-specific calibration through knowledge distillation. This slenderization reduces the model's spatial footprint, while the multi-exit strategy effectively balances utility and runtime costs. To enhance the flexibility of structural pruning, DISP-LLM \cite{gao2024displlm} breaks the structural dependencies in regular methods by allowing different layers to have different subsets of features along the embedding dimension.

\begin{figure}[t]
    \centering
    \includegraphics[width=0.65\linewidth]{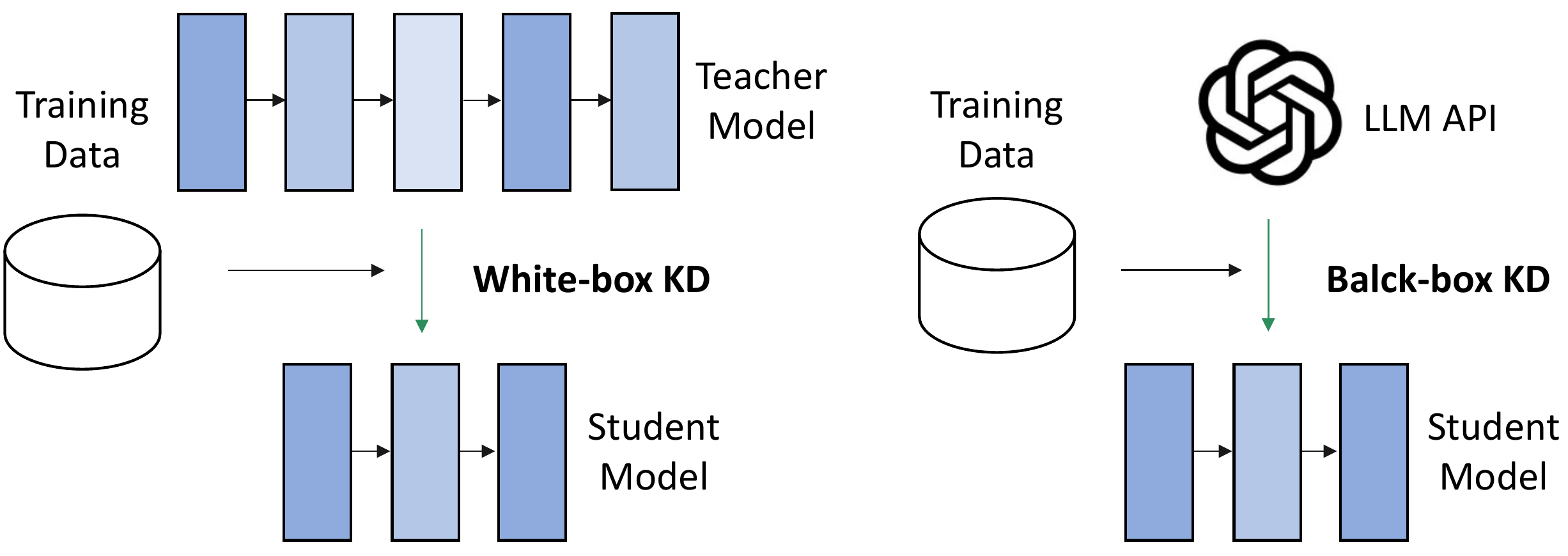}
    \vskip -1em
    \caption{Illustration of white-box and black-box knowledge distillation \cite{gominsu2024bible}.}
    \label{fig:SLM_kd}
\end{figure}

\subsubsection{Knowledge Distillation}
\label{knowledge_distillation}


Knowledge distillation (KD) compresses a larger teacher model into a smaller student model by training the student to mimic the teacher's outputs~\cite{hinton2015distilling}. This enables the student to retain much of the teacher’s capabilities with fewer parameters, making it ideal for scaling down LLMs for resource-limited environments while maintaining performance. KD can be categorized into \textit{white-box} and \textit{black-box} approaches~\cite{wang2024model,zhu2023survey,yang2024survey} as shown in Figure \ref{fig:SLM_kd}. In \textbf{White-Box KD}, the student has access to the teacher's internal states or output distributions \cite{zhang2023towards,gu2024minillm,agarwal2024policy,ko2024distillm,jha2024justchopembarrassinglysimple,kim2024token,padmanabhan2024propagating, wu-etal-2024-weight}. Generalized Knowledge Distillation (GKD) \cite{ko2024distillm} introduces skew KL divergence to stabilize gradients and enhance performance, using an adaptive off-policy approach to minimize noisy feedback and improve efficiency. \textbf{Black-Box KD} relies only on teacher outputs without having access to model internals~\cite{peng2023instruction,chen2023mcc,wang2023scott, lee-etal-2024-mentor}. Methods like Distilling Step-by-Step~\cite{hsieh2023distilling} use teacher-generated rationales to train smaller models, improving performance with fewer examples. LaMini-LM~\cite{wu-etal-2024-lamini} creates a diverse instruction dataset with GPT-3.5 Turbo responses, enabling robust performance in smaller models.

\begin{figure}[!b]
    \centering
    \includegraphics[width=0.67\linewidth]{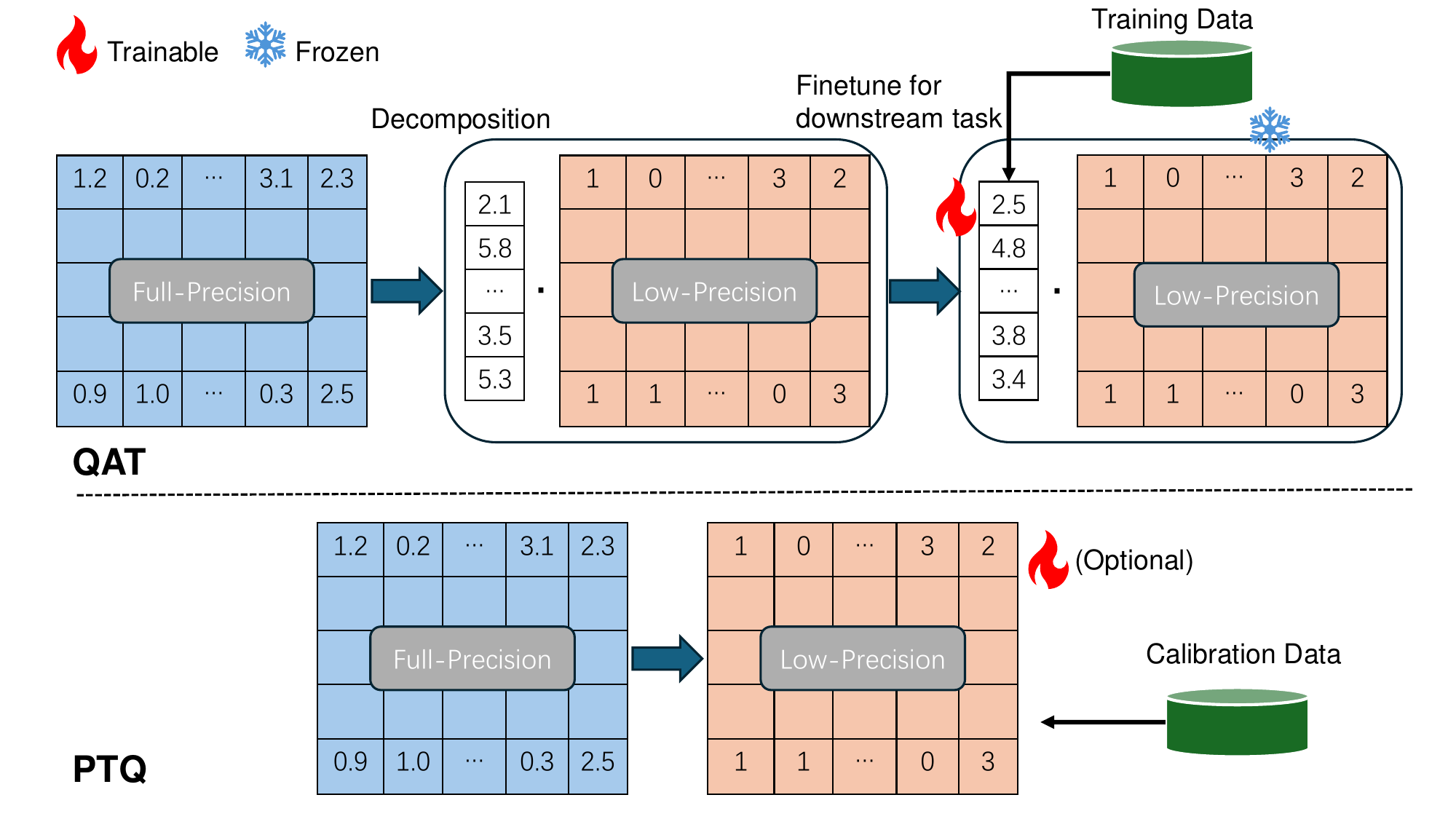}
    \vskip -1em
    \caption{Illustration of quantization-aware training (QAT) and post-training quantization (PTQ).}
    \label{fig:quantization}
\end{figure}

\subsubsection{Quantization}
\label{quantization}

Quantization reduces the storage and computational demands of LLMs by converting floating-point representations into lower-precision formats, significantly cutting both storage requirements and computational complexity. Existing methods fall into two categories: \textit{Post-Training Quantization} (PTQ) and \textit{Quantization-Aware Training} (QAT). 
Figure \ref{fig:quantization} illustrates the two quantization methods. \textbf{Post-Training Quantization}, applied after training, simplifies model compression without altering the architecture or requiring retraining, though it may result in precision loss. Consider a group or block of weights $\mathbf{w}$; the linear operation can be expressed as $y = \mathbf{w}\mathbf{x}$, while the quantized version is given by $y = Q(\mathbf{w})\mathbf{x}$. Generally, the quantization function $Q$ is defined as \cite{lin2024awq}:
$
Q(\mathbf{w}) = \Delta \cdot \text{Round}\left(\frac{\mathbf{w}}{\Delta}\right), \quad \Delta = \frac{\text{max}(|\mathbf{w}|)}{2^{N-1}},
$
where $N$ is the number of quantization bits, and $\Delta$ is the quantization scale factor determined by the absolute maximum value of $\mathbf{w}$. \textbf{Quantization-Aware Training (QAT)} enhances LLM efficiency by directly incorporating quantization into the training process, often resulting in higher accuracy than PTQ. During QAT, the forward pass utilizes quantized weights \( Q(\mathbf{W}) \) and activations \( Q(\mathbf{X}) \), while retaining full-precision values during the backward pass and for updating gradients to ensure stable learning dynamics.
The comparisons of the post-training quantization methods are summarized in Table \ref{tab:quantization}, detailing precision, addressed problems, and technical contributions of each method.

\begin{table}[!t]
\small
\centering
\caption{Representative quantization methods.}
\label{tab:quantization}
\vskip -1em
\resizebox{.99\textwidth}{!}{
\begin{tabular}{c|c|c|p{5cm}|c}
\hline
\textbf{Methods} & \textbf{Bit} & \textbf{Type} & \textbf{Technical Contribution} & \textbf{Problems} \\ \hline
SqueezeLLM \cite{kim2023squeezellm} & 3-bit & PTQ & Sensitivity-based non-uniform quantization, dense and sparse decomposition &  ultra-low bit
quantization \\ 
JSQ \cite{guo2024compressing} & Flexible & PTQ & Joint Sparsification and Quantization & better compression-accuracy trade-offs.  \\
FrameQuant \cite{adepu2024framequant} & Fractional bit  & PTQ &  Fractional bit widths  & better compression-accuracy trade-offs. \\
OneBit \cite{xu2024onebit} & 1-bit & PTQ & Quantization-aware knowledge distillation & 1-bit quantization \\
BiLLM \cite{huang2024billm} & 1-bit & PTQ & Crucial Weights Selection, Block-based error compensation & 1-bit quantization \\
LQER \cite{zhang2024lqer} & Flexible & PTQ & Quantization Error Minimization & better compression-accuracy trade-offs \\
I-LLM \cite{hu2024llm} & Flexible &PTQ & Fully-Smooth Block-Reconstruction, Dynamic Integer-only MatMul and Integer-only Non-linear Operators  & Integer-only Quantization \\
PV-Tuning \cite{malinovskii2024pv} & 1-bit/2-bit & PTQ & PV algorithm & better compression-accuracy trade-offs. \\ 
BitNet \cite{wang2023bitnet} & 1-bit & QAT & 1-bit Transformer Architecture & 1-bit quantization\\
BitNet b1.58 \cite{ma2024era} &  \{-1, 0, 1\} & QAT & Ternary Parameters & 1-bit quantization \\
PEQA \cite{kim2024memory} & Flexible & QAT &  Quantization Scales Optimization & Parameter-Efficient Finetuning \\
QLoRA \cite{dettmers2024qlora} & NF4  & QAT & 4-bit NormalFloat and Double Quantization & Parameter-Efficient Finetuning \\
\hline
\end{tabular}
}
\vskip -2em
\end{table}

\subsubsection{Low-Rank Techniques} 
\label{low_rank_techniques}
Low-rank techniques compress LLMs by approximating a high-dimensional weight matrix with two lower-dimensional matrices, reducing computational and memory requirements. A matrix \(\mathbf{W} \in \mathbb{R}^{m \times n}\) is approximated as \(\mathbf{W} \approx \mathbf{A} \times \mathbf{B}\), where \(\mathbf{A} \in \mathbb{R}^{m \times r}\) and \(\mathbf{B} \in \mathbb{R}^{r \times n}\), with \(r\) much smaller than \(m\) or \(n\), reducing the number of parameters. Building on this concept, \citet{ji2024feature} propose a low-rank method tailored for LLMs, leveraging the observation that while LLMs have high-rank weights, their feature interactions tend to exhibit low-rank properties. The method estimates feature distributions using pooled covariance matrices and allocates distinct compression ratios to layers based on their sensitivity to low-rank compression. A Bayesian optimization strategy, using a Gaussian process as the surrogate model, optimizes the allocation of low-rank dimensions, ensuring the model maintains performance while achieving significant compression. Transitioning from model compression to fine-tuning, \citet{cho2024heterogeneouslorafederatedfinetuning} tackles system and data heterogeneity with the HETLORA method, which uses heterogeneous low-rank approximations to accommodate the diverse capabilities of clients and data complexities. By combining local rank self-pruning with sparsity-weighted aggregation, it balances high and low-rank LoRA modules, improving convergence speed and performance compared to uniform approaches. LLM-Neo \cite{yang2024llm} combines knowledge distillation with low-rank adaptation (LoRA) to improve the efficiency of transferring knowledge from a teacher LLM to a compact student model.

%% file: sections/4.enhancement.tex
\section{Advanced Enhancement Strategies for Small Language Models} %
\label{enhancement}
With the foundational concepts introduced in Section~\ref{construction}, this section explores various advanced techniques that enhance the performance of SLMs, including innovative training methods for training SLMs from scratch, supervised fine-tuning (SFT) to align SLMs to adhere to instructions, advanced knowledge distillation and quantization techniques, and techniques frequently used in LLMs such as mixture-of-experts to enhance SLM for specific applications. A summary of enhancement techniques is also summarized in Table \ref{tab:enhancement_summary}.

\begin{table}[t]
\caption{Advanced enhancement methods for SLM.} 
\label{tab:enhancement_summary}
\vskip -1em
\small
\begin{tabular}{p{2cm}|l|p{9cm}}
\hline
Topic                                                                                & Method                                 & Main Contribution                                                                                                                                                                        \\ \hline
\multirow{3}{*}{\begin{tabular}[c]{@{}l@{}}Training \\ from \\ Scratch\end{tabular}} & MindLLM~\cite{yang2023mindllm}         & Bilingual models with advanced features.                                                                                                            \\
                                                                                     & MobiLlama~\cite{thawakar2024mobillama} & On-device SLM with dual objectives for efficiency and capability.                                                                                       \\
                                                                                     & MobileLLM~\cite{liu2024mobilellm}      & Optimizes LLM deployment on mobile with advanced architecture.                                                                                       \\ \hline
\multirow{4}{*}{\begin{tabular}[c]{@{}l@{}}Supervised \\ Fine-tuning \end{tabular}}  & MobileBERT \cite{sun2020mobilebert} & Compact BERT for efficient fine-tuning.  \\   
                                                                                    & Alpaca 7B~\cite{alpaca}              & Uses ChatGPT-generated tasks to tune Llama 7B. \\
                                                                                     &  RLHF \cite{ouyang2022training}           & Trains using human-preferred data and reinforcement learning.   \\
                                                                                     &  DPO \cite{rafailov2024direct}           & Dynamically adjusts log probabilities to prevent model degradation. \\ 
\hline
\multirow{3}{*}{\begin{tabular}[c]{@{}l@{}}Data \\ Quality \\ in KD\end{tabular}}    & TinyStory~\cite{eldan2023tinystories}  & Enhances narrative coherence in child-friendly datasets.                                                                           \\
                                                                                     & AS-ES~\cite{xi2024learning}            & Improves CoT by categorizing reasoning steps.                                                                            \\
                                                                                     & Self-Amplify~\cite{bhan2024self}       & Automates CoT data annotation for small models.                    
                                                                                     \\ \hline
\multirow{3}{*}{\begin{tabular}[c]{@{}l@{}}Distillation \\ for SLM\end{tabular}}     & GKD~\cite{agarwal2024policy}           & Aligns training and inference distributions using on-policy sequences.                                                                                    \\
                                                                                     & DistiLLM~\cite{ko2024distillm}         & Uses skew KL divergence and adaptive off-policy for output utilization.                                                          \\
                                                                                     & Adapt-and-Distill~\cite{yao2021adapt}  & Domain adapts both teacher and student models before distillation.                                                                                       \\ \hline
\multirow{7}{*}{Quantization}                                                        & SmoothQuant~\cite{xiao2023smoothquant} & Balances quantization difficulty using per-channel scaling.                                                       \\
                                                                                     & BiLLM~\cite{huang2024billm}            & Applies Hessian-based metrics for binary residual approximation. \\
                                                                                     & LLM-QAT~\cite{liu2023llm}              & Uses data-free knowledge distillation and logit distillation for fine-tuning.                                                                   \\
                                                                                     & PB-LLM~\cite{shang2023pbllmpartiallybinarizedlarge}              & Binarizes non-salient weights while preserving others in higher precision.                                                    \\
                                                                                     & OneBit~\cite{xu2024onebit}             & Achieves near 1-bit quantization with minimal performance loss.                                                     \\
                                                                                     & BitNet~\cite{wang2023bitnet}           & Introduces 1-bit Transformer architecture with BitLinear layers.              \\
                                                                                     & BitNet b1.58~\cite{ma2024era}          & Implements a ternary weight system in enhanced BitNet.                   \\ \hline
\multirow{3}{*}{\begin{tabular}[c]{@{}l@{}}LLM techniques \\for SLM\end{tabular}}                              & \citet{ma2023large}           & Combines filtering and re-ranking to improve Information Extraction tasks.         \\
& MoQE~\cite{kim2023mixture} & Applies quantization to expert weights to outperform dense models. \\
& SLM-RAG~\cite{liu2024can} & Shows that SLMs with RAG can match LLM performance. \\
\hline
\end{tabular}
\end{table}

\subsection{Innovative Training Methods for Small Language Models from Scratch}
\label{Training4SLMFromScratch}
\begin{figure}[!t]
    \centering
    \includegraphics[width=0.9\linewidth]{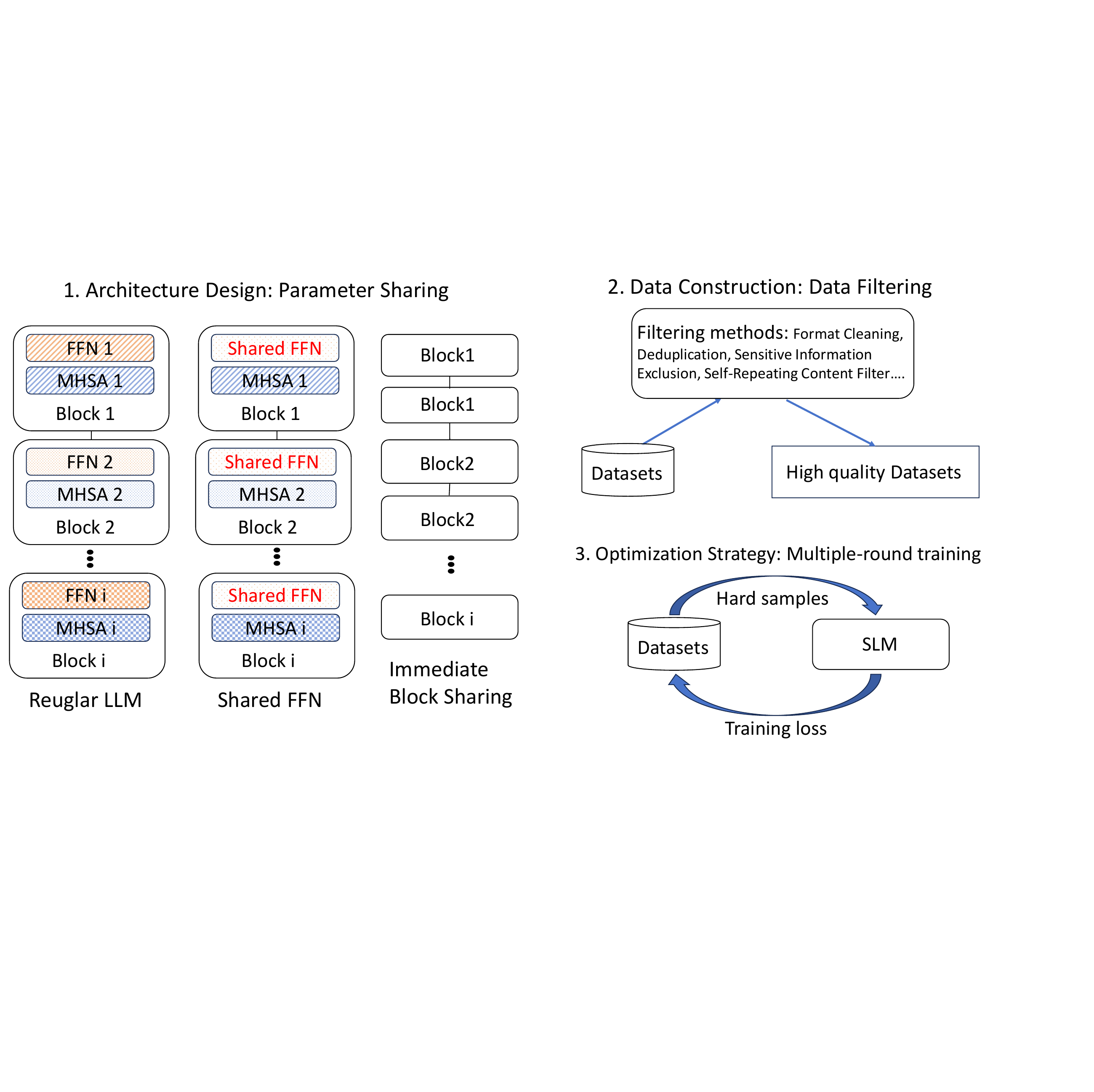}
    \vskip -1.5em
    \caption{Innovative Training Methods for Small Language Models from Scratch}
    \label{fig:train_from_scratch}
    \vskip -2em
\end{figure}

In scenarios with limited resources, we aim to train small language models to provide efficient, cost-effective solutions tailored for specific domains, while still maintaining competitive performance with larger models. Training small language models (SLMs) from scratch involves unique strategies that diverge significantly from those used for large language models (LLMs). This section synthesizes cutting-edge techniques tailored to optimize the inherent capabilities of SLMs, underscoring their potential to match or surpass larger counterparts in efficiency and effectiveness. As shown in  Figure \ref{fig:train_from_scratch}, the methods for training SLMs from scratch can be categorized into three primary categories: \textit{Architecture Design}, \textit{Data Construction}, and \textit{Optimization Strategy}. Next, we introduce each category in detail.


\textbf{Architecture Design for SLMs} When designing SLM architectures, parameter-sharing techniques are employed to minimize space usage and reduce the model's size. As shown in the first part of Figure \ref{fig:train_from_scratch}, parameter sharing is achieved by two approaches: (\textbf{i}) a single Feed-Forward Network (FFN) module is shared by every transformer layer. As shown in Figure~\ref{fig:train_from_scratch} (1) middle, \textbf{FFN layer sharing/reusing} can maintain a smaller size while still benefiting from the depth and complexity gained through repeated processing of input data. This technique is firstly applied in MobiLlama~\cite{thawakar2024mobillama} which surpasses the performance of existing SLMs of comparable size.
(\textbf{ii}) Entire transformer blocks are shared. As shown in Figure~\ref{fig:train_from_scratch} (1) right, \textbf{Transformer Block-wise Sharing} is another parameter-sharing approach that maintains depth and complexity. There are different transformer block-wise sharing strategies such as repeating the transformer blocks all over again or repeating the immediate transformer block. This technique is applied in MobileLLMs \cite{liu2024mobilellm} which has 125M and 350M parameters. MobileLLMs demonstrate performance improvements of 2.7\% and 4.3\%, respectively, compared to previous models with equivalent parameters. Moreover, they exhibit accuracy comparable to LLaMa-2-7B on API call tasks, highlighting the capabilities of smaller models in mobile environments.

\textbf{Data Construction} For SLMs, the emphasis on data quality surpasses that of quantity and diversity~\cite{yang2023mindllm}. Experiments demonstrate that using a quality filtering approach to remove low-quality data can lead to improved performance in SLMs~\cite{yang2023mindllm}. Unlike large models, which can handle diverse and large datasets, SLMs benefit more from cleaner, high-quality data probably due to their limited capacity against noise. Generally, data processing has several steps: 
(i) Remove HTML, CSS, JS, and non-text elements for clean text; (ii) Filter low text-to-content ratio web pages; (iii) Deduplicate using SimHash \cite{dasgupta2011fast,sadowski2007simhash}; (iv) Exclude sensitive/offensive content with heuristics and token replacements; (v) Remove self-repeating phrases of advertisements to enhance dataset informativeness \cite{chen2024data,yang2023mindllm}.
These steps collectively ensure that training data has high-quality, informative texts. SLMs also significantly benefit from these techniques. For example, MindLLMs~\cite{yang2023mindllm}, which are bilingual lightweight language models (available in 1.3B and 3B versions), adopt these data processing techniques and achieve improved capability acquisition.


\textbf{Training Strategy for SLMs} For LLMs, due to the large model size and data volume, LLMs are usually trained with one round. For SLMs, multiple-round training can be applied \cite{tang2024rethinking}. Considering some examples are hard to fit, hard examples can be trained with a high probability \cite{tang2024rethinking}. For each round of training, the data sampling probability is updated according to the overall loss of that sample. Experiments results show that two rounds of training and a 50\% sampling rate are a good trade-off between performance and training efficiency. \citet{tang2024rethinking} show that a deep and thin neural architecture and multiple-round training can enhance the performance of the trained Pangu 1.5B pro model. This model outperforms the conventionally trained Pangu 1.5B and a series of other comparable large language models with similar model sizes on multiple benchmark datasets, achieving an average performance increase of 8.87\%.
\subsection{Supervised Fine-Tuning (SFT) for Enhancing SLM performance}
\label{SFT}
\begin{figure}[!t]
    \centering
    \includegraphics[width=0.8\linewidth]{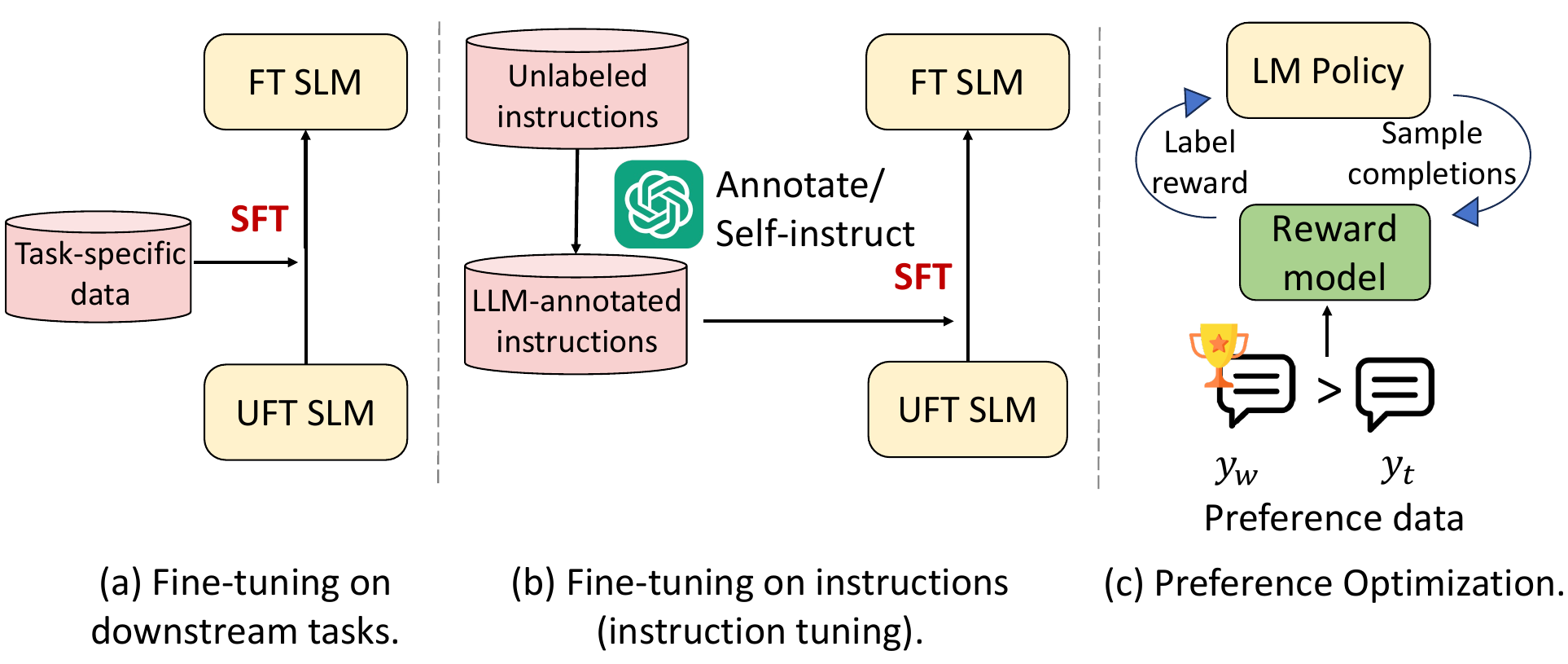}
    \vskip -1em
    \caption{Fine-tuning for Enhancing SLMs}
    \label{fig:fine_tuning}
\end{figure}
Supervised Fine-Tuning (SFT) employs a training methodology similar to pre-training but is specifically tailored to align models to adhere to the instructions encapsulated within various instructional datasets. This approach is designed to refine the model’s responsiveness and appropriateness to given contexts as dictated by the training data. For example, various models, such as Alpaca~\cite{alpaca}, UltraChat \cite{ding2023enhancing}, WizardLM \cite{xu2023wizardlm}, SlimOrca \cite{SlimOrca}, ShareGPT \cite{wang2024openchat}, Capybara \cite{daniele2023amplify-instruct}, Deita \cite{liu2023makes}, and MetaMathQA \cite{yu2024metamath}, incorporates a suite of conversational datasets to enhance their capabilities in context-aware dialogue and instruction adherence. Usually, as shown in Figure~\ref{fig:fine_tuning}, existing SFT methods can be categorized into three categories:
\begin{itemize}[leftmargin=*]
\setlength{\itemsep}{0pt}
\setlength{\parsep}{0pt}
\setlength{\parskip}{0pt}
\item \textbf{(i) Classical fine-tuning with downstream data}~\cite{devlin2019bert,radford2019language} trains SLMs on task-specific annotated data, transferring general language representations to specific tasks such as sentiment analysis. In the LLM era, this approach remains effective, such as enhancing LLMs by calibrating responses or assigning risk scores with smaller models such as BERT \cite{zhao2023automatic}, or optimizing for mobile devices with MobileBERT \cite{sun2020mobilebert}. 
\item \textbf{(ii) Instruction tuning} with LLM-generated data \cite{alpaca, ding2023enhancing, SlimOrca} or human-generated questions with LLM annotations \cite{wang2024openchat} aims to align generative models with specific instructions, enhancing their instruction-following and reasoning capabilities. For example, \textbf{Alpaca 7B}~\cite{alpaca} uses 52k ChatGPT-generated instruction-following examples from 175 self-instructed seed tasks to tune Llama 7B~\cite{touvron2023llama}. Meanwhile, StableLM~\cite{bellagente2024stable, StableLM-3B-4E1T} is trained on the Restruct-v1 dataset, which includes summarization, question-answering, and sentiment analysis tasks, using instruction data from \cite{longpre2023data}. 
\item \textbf{(iii) Preference optimization with human feedback}
\cite{rafailov2024direct, wang2024openchat, ouyang2022training} aims to better align language models with human preferences. Reinforcement Learning from Human Feedback (\textbf{RLHF}) \cite{ouyang2022training} gathers human-preferred data, trains a reward model, and fine-tunes the LM using reinforcement learning. Direct Preference Optimization (\textbf{DPO}) \cite{rafailov2024direct} provides a simpler alternative to RLHF. Unlike RLHF, DPO avoids explicit reward modeling and reinforcement learning techniques. Instead, it adjusts the log probabilities of preferred versus non-preferred responses using a dynamic weighting mechanism, preventing model degradation issues typical of methods relying on probability ratios. For instance, Llama 3.2 1B \& 3B apply SFT and DPO in post-training to enhance alignment with instructions and human preferences.
\end{itemize}


\begin{figure}[!t]
    \centering
    \includegraphics[width=0.9\linewidth]{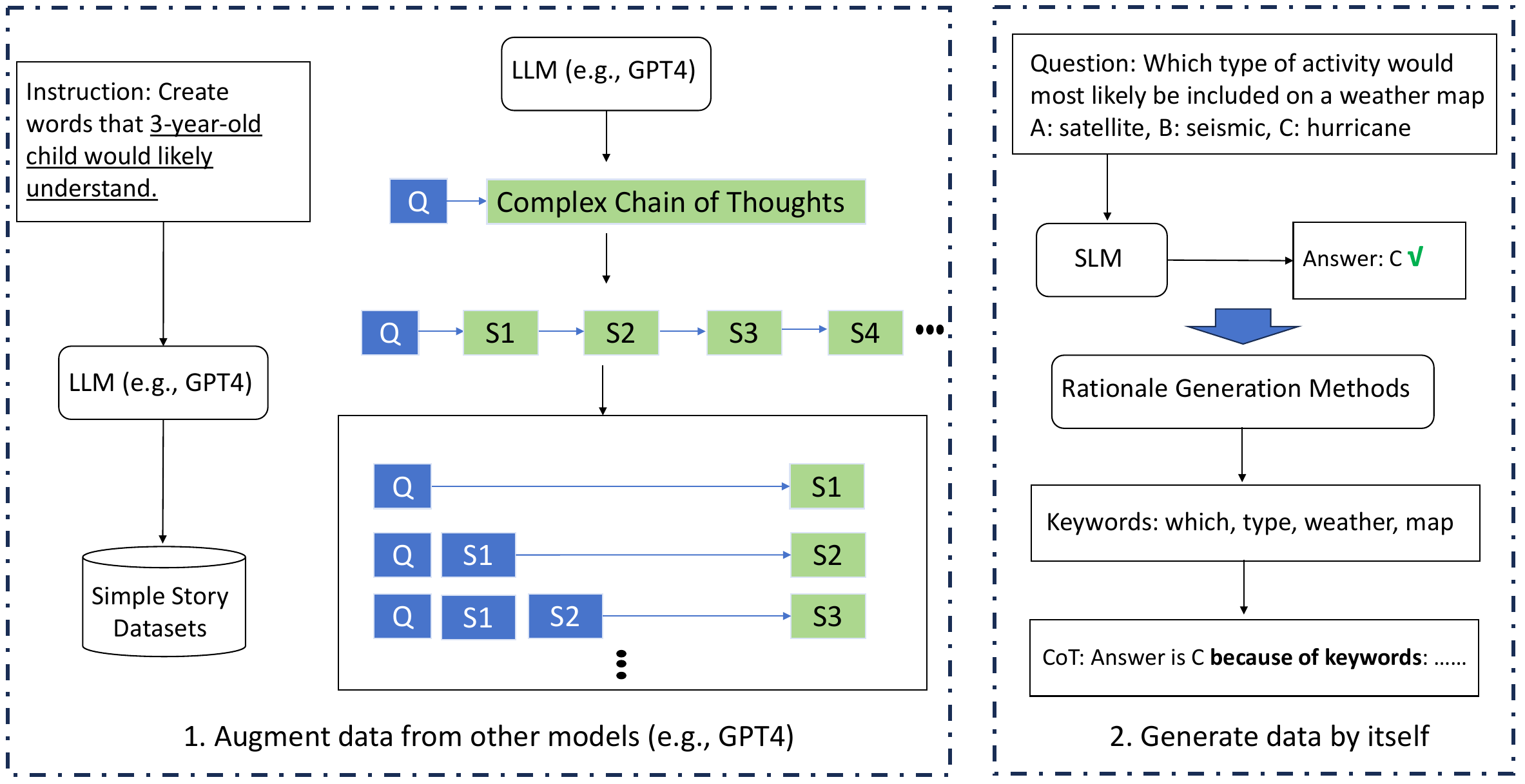}
    \vskip -1em
    \caption{Data Quality in Knowledge Distillation (KD)}
    \label{fig:data_quality}
\end{figure}

\subsection{Data Quality in Knowledge Distillation (KD)}
\label{data_quality4enhance}

Transitioning from the discussion on training SLMs from scratch, this section delves into the critical role of data quality in Knowledge Distillation (KD). The motivation here is to highlight how high-quality data generated from LLMs can significantly enhance the learning efficiency and performance of SLMs. The central idea is that meticulously crafted datasets when used in KD, enable SLMs to more effectively mimic the advanced capabilities of their larger counterparts. As shown in Figure \ref{fig:data_quality}, the data can come either from (1) other strong LLMs (e.g., GPT-4~\cite{achiam2023gpt}) which are much larger and more powerful than the target SLM, or (2) the target SLM itself.




\textbf{Augment Data from LLMs.} LLM-generated data could be categorized as \textit{pre-training data} and \textit{fine-tuning data}. 
Firstly, due to the limitations of model size, studies have shown that training SLMs requires simple and comprehensible data~\cite{eldan2023tinystories, xi2024learning, lee2024can,li2023mixed}. As shown in Figure \ref{fig:data_quality} (1) left, 
\textbf{TinyStory}~\cite{eldan2023tinystories} shows that small models (tens of millions of parameters) can generate coherent stories for 3-4-year-olds. GPT-3.5 or GPT-4~\cite{achiam2023gpt} prompts create simple stories from three keywords chosen from a 1,500-word vocabulary, which are then used to train SLMs for similar outputs. This approach shows that simple and comprehensible data can help smaller models exhibit behaviors similar to those of larger language models, such as obeying scaling laws and achieving enhanced performance. 
On the other hand, many efforts to enhance the Chain-of-Thought (CoT) capabilities of small models involve using LLMs to generate high-quality fine-tuning CoT data. 
As shown in Figure \ref{fig:data_quality} (1) right, these data train small models end-to-end to mimic CoT reasoning~\cite{xi2024learning, ma2023sci}. \textbf{AS-ES Learning}~\cite{xi2024learning} highlights that small models struggle with complex reasoning, even when provided detailed steps, as these require nuanced extraction and abstraction. Therefore, the study introduces a paradigm splitting reasoning into extractive segments (context reminders) and abstractive segments (inferred insights).


\textbf{Augment Data from Itself.} Besides distilling data from other LLMs, language models can also train on their own outputs~\cite{huang2023large,bhan2024self,tian2024toward}. Since voting strategies can improve the performance of LLMs, reasoning paths that lead to the majority answer can be further utilized to fine-tune LLMs~\cite{huang2023large}. Similarly, SLMs can generate their training data with the aid of existing rationale generation methods. \textbf{Self-Amplify}~\cite{bhan2024self} notes that human annotation of Chain-of-Thought (CoT) data is very time-consuming; thus, automated rationale generation methods have been proposed. These methods involve three main steps: (1) Selection of samples \( (x, y) \) that the model predicts correctly as few-shot examples; (2) Rationale generation, where rationales are produced using post hoc explanation methods; (3) Prompt design for SLMs, where the final prompt is crafted based on the previously generated rationales.

\begin{figure}[!t]
    \centering
    \includegraphics[width=0.9\linewidth]{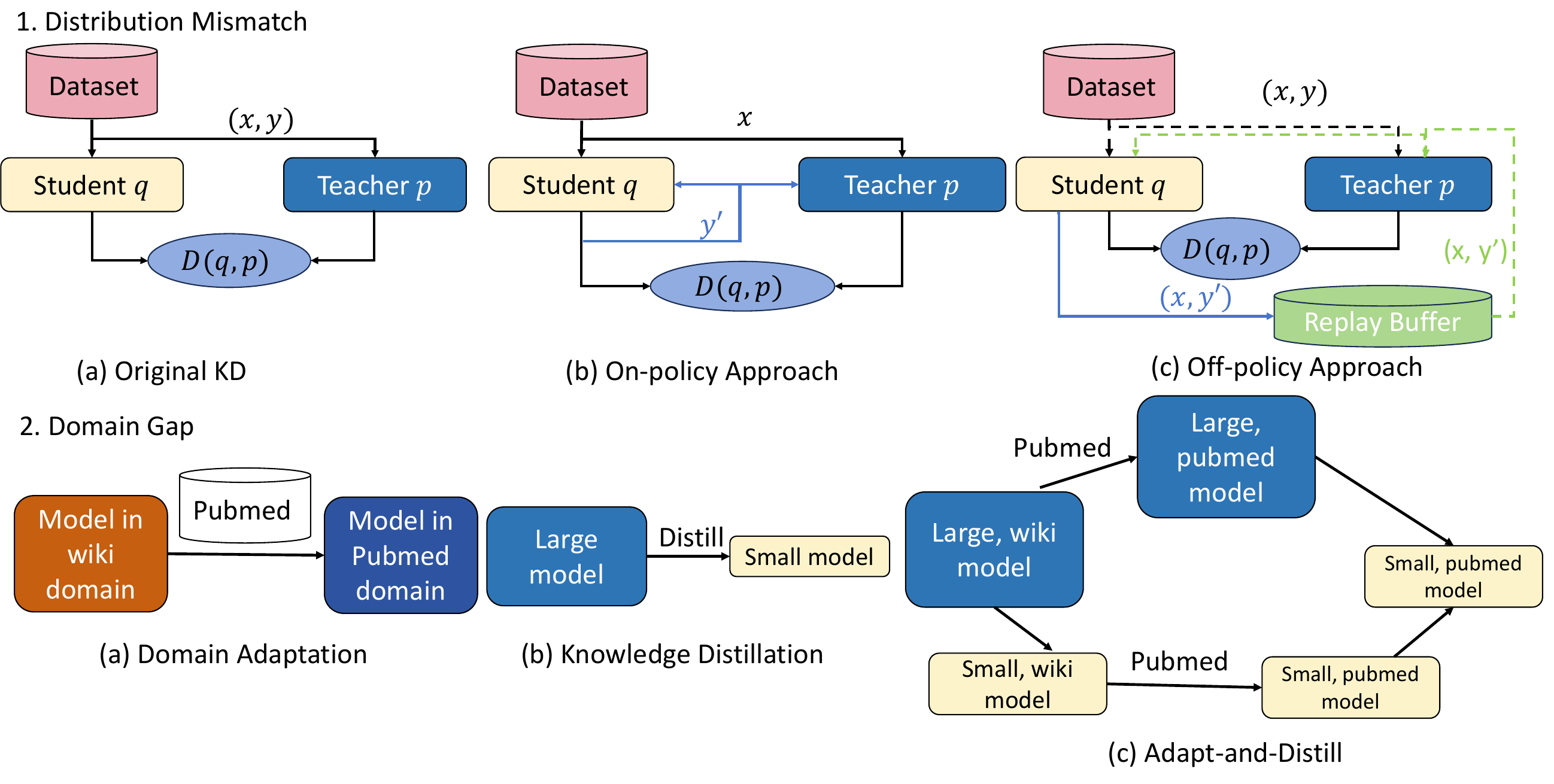}
    \vskip -2em
    \caption{Distillation Techniques for Enhancing SLM Performance. On-policy means learning only use data from the current student (policy), while off-policy permits the use of previously gathered data.}  
    \vskip -0em
    \label{fig:distill}
\end{figure}

\subsection{Distillation Techniques for Enhancing SLM Performance}
\label{distillation4enhance}


Following the discussion on data quality in KD, this section reviews specialized KD training strategies designed to enhance the performance of SLMs. The motivation is to address the unique challenges and constraints involved in distilling knowledge from LLMs to SLMs, ensuring that the smaller models can maximize their performance gains. As shown in Figure \ref{fig:distill}, two main gaps between LLMs and SLMs lead to challenges in distillation: \textit{distribution mismatch} and \textit{domain gap}. \textit{Distribution mismatch}~\cite{ko2024distillm,agarwal2024policy} occurs when the distribution of output sequences during training does not align with the distribution of sequences that SLMs produce during inference, leading to suboptimal performance of the student model. The \textit{domain gap}~\cite{yao2021adapt} arises when there is a discrepancy between the domains or tasks on which the LLMs and SLMs are trained and applied. This gap can cause significant degradation in the performance of the student model if not properly addressed during the distillation process. To address these issues, specialized strategies involve first aligning the teacher and student models with the target domain before proceeding with knowledge distillation. To explore these challenges further, we now delve into the details of these two branches of methods.



\textbf{Distribution Mismatch} In original knowledge distillation, illustrated in Figure \ref{fig:distill} Distribution Mismatch (a), the teacher and student are provided with the same input sequences $x$ and output labels $y$, producing probability distributions for the next token ($q$ and $p$). The loss is calculated as the difference between these two distributions, $D(q,p)$. However, a key challenge arises due to distribution mismatch: the output sequences during training ($y$) differ in distribution from those the SLMs produce during inference ($y^{\prime}$). To address this challenge, various techniques have been proposed. As shown in Figure \ref{fig:distill} Distribution Mismatch (b), one approach trains the student model using on-policy sequences—sequences generated by the student itself—guided by the teacher model's feedback. Specifically, both the student and teacher take the same input ($x$) and the student-generated output ($y^{\prime}$), producing probability distributions for the next token ($q$ and $p$, respectively). The loss is calculated as the difference between these two distributions, $D(q,p)$. This approach helps the student model reduce the distribution gap between training and inference by learning from the teacher's feedback on its own generated sequences. Generalized Knowledge Distillation (\textbf{GKD})~\cite{agarwal2024policy} is the first work using this technique and improves distillation outcomes. However, a drawback of this technique is that it requires the student to constantly produce new training sequences, which can be computationally expensive. To improve efficiency, as shown in Figure \ref{fig:distill} Distribution Mismatch (c), an adaptive off-policy approach can be used to efficiently manage student-generated outputs by storing them in a replay buffer, thereby reducing computational costs. \textbf{DistiLLM}~\cite{ko2024distillm} employs this off-policy approach and improves the efficiency of KD.

\textbf{Domain Gap} When training an SLM in a specific domain that differs from the domain of the LLMs, the gap between the two domains becomes problematic. As illustrated in Figure \ref{fig:distill} Domain Gap (a), domain adaptation fine-tunes a language model, initially trained on a general corpus, using a specialized dataset such as PubMed to enhance performance in that specific domain. As illustrated in Figure \ref{fig:distill} Domain Gap (b), Knowledge distillation transfers knowledge from the larger model to the smaller one. However, because the teacher model may not produce high-quality outputs on specialized datasets, domain adaptation is needed prior to knowledge distillation. As illustrated in Figure \ref{fig:distill} Domain Gap (c), \textbf{Adapt-and-Distill}~\cite{yao2021adapt} tackles the domain gap by distilling general large models into smaller ones. This paper introduces AdaLM and demonstrates that the ``Adapt-and-Distill'' strategy—first involving domain adaptation of both the large teacher model and the small student model, followed by distillation—is the most effective compared to three other strategies: training directly from scratch, distillation followed by adaptation, and adapting the teacher model before distillation into a general small student model. These innovative techniques are crucial for enhancing the capabilities of SLMs, making them more efficient and effective for various applications. However, adapting both the teacher (LLMs) and the student (SLMs) models to the target domain can be time-consuming. Future research could focus on efficiently solving the domain gap problem.

\begin{takeaway2}
\textbf{Insights}: Here are some insights from distillation techniques:
\begin{itemize}[leftmargin=*]
\setlength{\itemsep}{0pt}
\setlength{\parskip}{0pt}
\setlength{\parsep}{0pt}
    \item Sampling SLM outputs during the training process is the main approach to resolving distribution mismatch.
    \item Techniques like Adapt-and-Distill address the domain gap by first adapting both the teacher (LLMs) and the student (SLMs) models to the target domain before proceeding with distillation.
\end{itemize}
\end{takeaway2}



\subsection{Performance Improvement through Quantization}
\label{quantization4enhance}
As mentioned in Section \ref{construction}, quantization is one of the most effective methods for adapting LLMs to SLMs. However, compression to smaller sizes often compromises performance. To address the performance drop associated with quantization, various methods have been proposed. This section examines how these quantization methods specifically enhance the performance of SLMs. While the general introduction to compression methods is discussed in the compression section, the focus here is on detailing those approaches that boost the efficiency and effectiveness of SLMs. As shown in Figure \ref{fig:quantization}, we categorize these quantization methods into two main approaches: Post-Training Quantization (PTQ), where quantization is conducted on a well-trained fixed model, and Quantization-Aware Training (QAT), where quantization is integrated into the training process. This section introduces advanced techniques in PTQ and QAT respectively.

\textbf{Post-Training Quantization (PTQ)} primarily includes weight quantization and activation quantization. Weight quantization aims to quantize model parameters while preserving performance. \textbf{GPTQ}~\cite{frantar2023gptq} compresses LLMs to 4-bit or 2-bit by quantizing weights layer-by-layer to minimize layer-wise quantization errors. \textbf{PB-LLM}~\cite{shang2023pbllmpartiallybinarizedlarge}, applicable to both PTQ and QAT, retains the most salient weights while binarizing the rest based on magnitudes. \textbf{BiLLM}~\cite{huang2024billm}, another PTQ method, uses a Hessian-based metric to identify salient and non-salient weights. Salient weights undergo binary residual approximation to minimize loss, while non-salient weights are divided into sparse and concentrated groups for separate binarization, reducing quantization errors. Activation quantization faces challenges with outliers that can stretch the quantization range, causing most values to cluster at few bits and introducing significant errors. To address this, \textbf{LLM.int8()}~\cite{dettmers2022gptint} isolates outlier features for 16-bit processing and handles the rest in 8-bit. \textbf{SmoothQuant}~\cite{xiao2023smoothquant} circumvents per-channel quantization issues by employing a "smoothing" technique that shifts the quantization challenge from activations to weights through a per-channel scaling transformation. This balance between activating and weight quantization allows effective 8-bit quantization (W8A8), preserving accuracy while significantly reducing memory and computational costs. SmoothQuant thus enhances the efficiency of SLMs in resource-constrained environments.

\textbf{Quantization-Aware Training (QAT)} differs from PTQ in that it includes a training phase after the model has been quantized. When models are quantized to extremes, such as 2-bit or 1-bit, performance typically drops significantly, but further training can help the model retain its capabilities. For instance, to mitigate performance degradation from binarization,\textbf{ PB-LLM}~\cite{shang2023pbllmpartiallybinarizedlarge} selectively binarizes only non-salient weights, preserving the most salient ones at higher precision. This method effectively reduces the model size without significantly impacting performance. Salient weights are chosen based on their magnitude, ensuring that the most influential weights maintain higher precision to preserve the model's reasoning capabilities. The paper explores both post-training quantization (PTQ) and quantization-aware training (QAT) to fine-tune and recover the performance of partially binarized models, achieving a balance between compression and accuracy. \textbf{OneBit}~\cite{xu2024onebit} and \textbf{BitNet}~\cite{wang2023bitnet} address the severe performance degradation associated with 1-bit quantization by decomposing floating-point matrices and employing mixed-precision strategies. Specifically, OneBit introduces Sign-Value-Independent Decomposition (SVID), which decomposes a floating-point matrix into a 1-bit matrix and two floating-point vectors. This method allows LLMs to be quantized to a 1-bit level while minimizing performance loss. By retaining critical information with the floating-point vectors, OneBit effectively balances extreme compression with maintaining model accuracy. \textbf{BitNet b1.58}~\cite{ma2024era} improves on the original BitNet by introducing a ternary matrix weight system {-1, 0, 1}, resulting in a 1.58-bit model. BitNet b1.58 matches the performance of full-precision models starting from a 3 billion parameter size while further reducing memory and latency costs. \textbf{LLM-QAT}~\cite{liu2023llm} employs data-free knowledge distillation, where the pre-trained model itself generates data for fine-tuning the quantized model (student) using logit distillation from the full-precision model (teacher). This method incorporates quantization of weights, activations, and key-value cache, achieving accurate 4-bit quantization for weights and key-value caches, and 6-bit for activations, demonstrating substantial improvements over existing post-training quantization methods.


\begin{takeaway2}
\textbf{Insights}: Insights drawn from quantization strategies include:
\begin{itemize}[leftmargin=*]
\setlength{\itemsep}{0pt}
\setlength{\parskip}{0pt}
\setlength{\parsep}{0pt}
    \item Post-Training Quantization techniques primarily focus on quantizing model weights, where selecting salient weights is crucial. Beyond weight quantization, handling outliers in activation signals is a significant challenge in quantizing activations.
    \item Quantization-Aware Training methods show that low-bit quantization (e.g., 1-bit models) requires additional tuning to maintain performance. Knowledge can be distilled from the model before quantization to the quantized model.
\end{itemize}
\end{takeaway2}

\subsection{Techniques in LLMs Contributing to SLMs}
\label{techniques_in_llms_for_slms}
This subsection explores the potential of advanced techniques such as RAG and MoE, which enhance LLM performance, to also maintain or boost SLM performance within constrained computational budgets. However, effectively integrating these techniques into SLMs, which inherently possess limited capabilities, remains an unresolved challenge.

\textbf{Retrieval Augmented Generation (RAG)} enhances the capabilities of language models in knowledge-intensive tasks by incorporating a retrieval mechanism. This approach allows models to access relevant contextual information from a data repository in response to user queries. By integrating this retrieved data, RAG-equipped models better understand specific topics, enabling more informed and accurate outputs. 
For SLMs, a significant concern is whether they possess the capacity for long-context reasoning. 
A recent study \cite{liu2024can} compares SLMs at the 7B level with RAG to larger models such as GPT-3.5 and GPT-4, suggesting that SLMs equipped with RAG can sometimes perform comparably or even better than LLMs. These findings indicate that RAG for SLMs is effective and represents a promising direction for future research.


\textbf{Mixture-of-Experts (MoE) \cite{cai2024survey}} has emerged as an effective method for substantially scaling up model capacity with minimal computation overhead in LLMs. 
The MoE framework is founded on a straightforward yet potent concept: distinct components of a model, referred to as ``experts'', specialize in different tasks or data facets. In this paradigm, only the relevant experts are activated for a specific input, which manages computational costs while leveraging a vast pool of specialized knowledge. This scalable and adaptable approach enables increased model capacity without proportionally escalating computational demands. We argue that MoE is particularly suitable for SLM architectures \cite{jiang2024mixtral} as it minimizes both computational load and memory overhead. However, research on MoE for SLMs remains sparse. Future studies could investigate how large LLM MoE architectures can be effectively compressed into small ones or how to develop an SLM with MoE tailored for specific devices from scratch.

%% file: sections/5.application.tex
\section{Applications of Small Language Models}
\label{application}
In this section, we delve into the applications of small language models (SLMs) across various NLP tasks and their deployment strategies. Due to benefits such as enhanced privacy, faster inference, and lower memory requirements, many NLP applications are now leveraging SLMs over LLMs. Additionally, deploying SLMs often involves considerations of memory and runtime efficiency, which are crucial for optimizing resource use on budget-constrained edge devices, particularly mobile phones. Then, we will discuss task-specific applications of SLMs and their deployment methods on mobile and edge devices.

\subsection{Task-specific SLM Applications}
\label{task_specific_application}
This subsection explores the diverse NLP tasks to which SLMs can contribute. Question-answering and coding represent generative tasks, while recommender systems and web search (though not strictly within the NLP domain) typically leverage the encoding capabilities of SLMs. Additionally, the application of SLMs on mobile devices is particularly well-suited due to constraints in memory and computing resources. The representative works are systematically organized in Table \ref{tab:tasks}.

\begin{table}[!t]
\caption{Task-specific SLM Applications}
\label{tab:tasks}
\vskip -1em
\small
\centering
\begin{tabularx}{\textwidth}{>{\hsize=0.22\hsize}X|>{\hsize=0.5\hsize}X|>{\hsize=1.2\hsize}X}
\hline
\textbf{Aspect} & \textbf{Representative work} & \textbf{Key point} \\
\hline

\multirow{7}{=}{\textbf{SLM in QA}} 


& Alpaca \cite{alpaca} & Tune Llama 7B \cite{touvron2023llama} using 52k ChatGPT-generated examples. \\
& Stable Beluga 7B \cite{StableBelugaModels} &  Employ explanation tuning to Llama-2 7B \cite{touvron2023llama2} on an Orca-style dataset. \\ 
& Fine-tuned BioGPT \cite{guo2023improvingsmalllanguagemodels} & Fine-tuning BioGPT (1.6B) \cite{luo2022biogpt} on PubMedQA. \\

& Financial SLMs \cite{phogat2024finetuningsmallerlanguagemodels} & Transfer financial knowledge from GPT-4 \cite{achiam2023gpt} to multiple SLMs. \\

& ColBERT \cite{gichamba2024colbertretrievalensembleresponse} & Fetch retrieval documents for SLMs to answer domain-specific questions. \\
& Rationale Ranking \cite{hartill2023answeringunseenquestionssmaller} & For unseen questions, combine retrieval with LLM-generated rationales. \\
& T-SAS \cite{jeong2023testtimeselfadaptivesmalllanguage} & Enhance SLMs adaptability with self-generated pseudo labels. \\




\hline

\multirow{4}{=}{\textbf{SLM in Coding}} 
& Phi-3.5-mini \cite{abdin2024phi} & New addition to the Phi-3 series and focus on high-quality data. \\ 
& TinyLlama \cite{zhang2024tinyllamaopensourcesmalllanguage} & 1.1B Transformer model is trained on 3T corpus. \\
& CodeLlama \cite{roziere2023code} & A derivative of Llama 2 fine-tuned on domain-specific datasets. \\
& CodeGemma \cite{team2024codegemma} & Fine-tuning Gemma to enhancing coding capabilities. \\
\hline

\multirow{5}{=}{\textbf{SLM in Recommendation}}& PromptRec \cite{wu2024could} & Training on prompt templates \\
& SLIM \cite{wang2024can} & Step-by-step Knowledge Distillation \\
& BiLLP \cite{shi2024large} & LLaMa-2-7B as planner and reflector \\
& ONCE \cite{liu2024once}  & LLaMa-2-7B as Content Encoder \\
& RecLoRA \cite{zhu2024lifelong} & Personalized low-rank adaptation \\
\hline

\multirow{3}{=}{\textbf{SLM in Web Search}}
& Content encoder \cite{changpre, humeaupoly, lu2020twinbert} & Encode concatenated queries and documents. \\
& Ranker \cite{chu2022h, nogueira2019passage} & Re-rank retrieved documents using a specially SLM. \\
& Rewriter \cite{ma2023query} & Bridge the gap between queries and needed knowledge by rewriting inputs. \\
\hline

\multirow{6}{=}{\textbf{SLM in Mobile-device}} & Octopus \cite{chen2024octopusondevicelanguagemodel} & Calling software APIs via learning in documents \\ 
& MobileAgent \cite{ding2024mobileagentenhancingmobilecontrol} & Standard Operating Procedure (SOP) \\
& $\alpha$-UMI \cite{shen-etal-2024-small} & SLMs serve as Multi-agents in tool uses. \\
& Mobile Interaction \cite{carreira2023revolutionizingmobileinteractionenabling} & Text-to-action control and tests on 6GB and 4GB Android devices \\
& AutoDroid \cite{wen2024autodroidllmpoweredtaskautomation} & Interaction based on GUI and APP knowledge injection \\
& M4 \cite{yuan2024mobile} & a foundation model handling all mobile AI tasks. \\
& Agent for Text Rewriting \cite{zhu2023ondeviceagenttextrewriting} & Data Knowledge Distillation from LLMs \\
\hline

\end{tabularx}
\end{table}

\subsubsection{SLM Applications in Question-Answering} 
\label{slm_application_qa}
Question-answering (QA) is a fundamental task in the NLP field, demanding language models to exhibit abilities in understanding language, reasoning, common sense, and recalling specialized knowledge. Typically, larger language models yield better QA performance. However, the substantial size of these models introduces challenges such as immense computational requirements, privacy concerns when using proprietary LLMs, and difficulties in customization. These issues lead researchers and developers to favor SLMs in scenarios that demand efficiency, privacy, and customization. Therefore, we explore methods to enhance the capabilities of SLMs in QA across three key areas: (i) Instruction Tuning of Generic SLMs for QA, (ii) Instruction Tuning of Domain-Specific SLMs for QA, and (iii) Enhancing SLMs for Out-of-Domain Questions.
\textbf{Instruction Tuning Generic SLMs for QA.}
Despite the Phi series' high question-answering capability, its training cost with over 3.4T tokens on 512 H100 GPUs for 10 days \cite{abdin2024phi} is prohibitive for many researchers and developers. Instruction tuning \cite{wei2022finetuned} offers a cost-effective alternative, enhancing small models by fine-tuning on large model outputs. Alpaca 7B \cite{alpaca} tunes Llama 7B \cite{touvron2023llama} with 52k ChatGPT-generated examples from 175 seed tasks. This behavior cloning mimics teacher models effectively but struggles in reasoning-intensive QA tasks where accuracy is key, not style \cite{chia2024instructeval}. To counter it, explanation tuning \cite{StableBelugaModels} enhances Llama-2 7B \cite{touvron2023llama2} using explanatory LLM answers to improve reasoning. However, its effectiveness varies with system instructions, and those effective for larger models like GPT-4 may not suit smaller ones. SLMs also struggle to identify optimal system instructions for different tasks. Therefore, Orca 2~\cite{mitra2023orca} addresses this by promoting cautious reasoning, deciding which solution strategy to choose for a given task among direct answer generation, or ``Slow Thinking'' strategies (step-by-step, guess and check or explain-then-answer, etc.) and erasing specific system instructions during training. This involves (1) solution strategy is guided by the performance of Orca 1~\cite{mukherjee2023orca}, (2) writing task-specific system instructions corresponding to the chosen strategy to obtain teacher responses for each task, and (3) at training time, employing Prompt Erasing to replace student's system instructions with generic ones vacated of details of how to approach the task, encouraging students learn not just task solutions but also deeper reasoning abilities.

\textbf{Instruction Tuning Domain SLMs for QA.} 
Beyond instruction tuning for generic SLMs, tuning domain-specific SLMs is also crucial, as they provide specialized assistance where generic SLMs may underperform. Instruction-tuning generic SLMs can derive domain SLMs. We summarize some representatives in several domains. 
(1) In finance, \citet{phogat2024finetuningsmallerlanguagemodels} transfer financial QA abilities from teacher LLMs such as GPT-4~\cite{achiam2023gpt} to specialized SLMs such as Phi-3-Mini~\cite{abdin2024phi}, using datasets such as FinQA \cite{chen2021finqa}, ConvFinQA \cite{chen2022convfinqa}, and TATQA \cite{zhu2021tat}. They train SLMs with Python programs created by the teacher model, which detail steps for financial reasoning, including concept comprehension, formula identification, entity extraction, and calculations. During inference, SLMs generate Python code that an external interpreter executes.
(2) In the medical field, \citet{guo2023improvingsmalllanguagemodels} enhance student SLMs, including domain-specific BioGPT (1.6B) ~\cite{luo2022biogpt} and general Llama 7B~\cite{touvron2023llama}, by fine-tuning on enriched PubMedQA \cite{jin2019pubmedqa} data. This enhancement is achieved by generating new samples or rewriting existing ones using teacher LLMs, which include the highly knowledgeable GPT-4 and the relatively weaker ChatGPT. The best SLM, with under 1.6 billion parameters, achieves 75.4\% accuracy, surpassing GPT-4’s 74.4\% in few-shot settings on the PubmedQA test sets. It demonstrates that LLMs effectively refine and diversify question-answer pairs, leading to enhanced performance in a significantly smaller model after fine-tuning. 
We report the detailed results of comparisons of instruction-tuned domain-specific language models for QA and larger language models on FinQA~\cite{chen2021finqa} and PubMedQA~\cite{jin2019pubmedqa}, as shown in Table~\ref{table:instructionqa}.
\begin{table}[tb]
    \centering
    \vskip -0.8em
    \caption{Comparison of instruction-tuned domain SLMs for QA and LLMs on FinQA~\cite{chen2021finqa} and PubMedQA~\cite{jin2019pubmedqa}. }
    \label{table:instructionqa}
    \vskip -0.8em
    \small
    \begin{tabular}{c|c|c|c|c|c}
    \toprule
        \textbf{Model} & \textbf{Size} & \textbf{Instruction tuned?} & \textbf{Task Name} & \textbf{Shot Type} & \textbf{Accuracy (\%)} \\ \hline
        GPT-4~\cite{achiam2023gpt} & - & $\times$ & FinQA & Zero-shot & 77.5  \\ 
        Phi-3-Mini~\cite{abdin2024phi} & 2.7B & $\checkmark$ & FinQA & Zero-shot & 77.6  \\\hline
        Meditron-70B \cite{chen2023meditron} & 70B & $\times $ & PubMedQA & Zero-shot & 81.6 \\
        RankRAG-llama3-70B \cite{yu2024rankrag} & 70B & $\times $ & PubMedQA & Zero-shot & 79.8 \\
        Flan-PaLM \cite{singhal2023large} & 540B & $\times $ & PubMedQA & Few-shot & 79.0 \\
        GAL 120B \cite{taylor2022galactica} & 120B & $\times $ & PubMedQA & Zero-shot & 77.6 \\
        Flan-PaLM \cite{singhal2023large} & 62B & $\times $ & PubMedQA & Few-shot & 77.2 \\
        BioGPT \cite{luo2022biogpt} & 345M & $\checkmark$& PubMedQA & Zero-shot & 78.2 \\
        BioGPT-Large \cite{luo2022biogpt} & 1.5B &$\checkmark$ & PubMedQA & Zero-shot & 81.0 \\
        \bottomrule
    \end{tabular}
    \vskip -0.8em
\end{table}




\textbf{Enhancing SLMs for Out-of-Domain Questions.} 
One of the major advantages of LLMs is their strong comprehension and logical reasoning abilities, which SLMs often struggle to match due to their limited parameters, especially when handling unseen or out-of-domain questions. Various methods have been developed to address this limitation, including Retrieval-Augmented Generation (RAG) and self-adaptive techniques.

\begin{enumerate}[leftmargin=*]
\setlength{\itemsep}{0pt}
\setlength{\parskip}{0pt}
\setlength{\parsep}{0pt}
\item \textbf{Retrieval-Augmented Generation (RAG):}
\emph{Incorporating External Knowledge for Domain-Specific QA.}
RAG addresses OOD questions by integrating external knowledge during inference, allowing models to access information beyond their pre-trained parameters. By retrieving relevant documents in real time, RAG enables small language models to provide accurate answers on specialized topics.
In the telecommunications domain, \textbf{\citet{gichamba2024colbertretrievalensembleresponse}} use ColBERT as a dense retrieval system to fetch documents from technical datasets. By encoding queries and documents separately, ColBERT computes relevance scores, helping small models like Phi-2 and Falcon-7B retrieve precise technical information to answer complex telecom-related queries.
\textbf{Rationale Ranking} \cite{hartill2023answeringunseenquestionssmaller} addresses answering unseen questions using smaller language models by integrating external explanatory contexts from retrieval systems with reasoning rationales from LLMs. This method involves ranking both the retrieved explanatory contexts and LLM-generated rationales using a scoring module, which then combines them to form a cohesive context. Consequently, this integrated approach enhances the SLMs' performance on unseen questions.

\item \textbf{Self-Adaptive Techniques:} \emph{Enhancing Model Adaptability with Self-Generated Pseudo Labels.}
Fine-tuning, while effective in adapting domain knowledge, can be impractical in realistic scenarios where labeled datasets are scarce. To overcome this, self-adaptive techniques employ self-generated pseudo labels to activate specific aspects of the target tasks, thereby enhancing model adaptability~\cite{veksler2023test,shu2022test}. Test-time Self-Adaptive Small LMs (\textbf{T-SAS}) \cite{jeong2023testtimeselfadaptivesmalllanguage} first stochastically generates multiple answers for an unlabeled question. The most plausible answer is then selected via majority voting to enhance pseudo-label accuracy, serving as a pseudo-label for training during test-time. 

\end{enumerate}

\textbf{Comparison between LLMs and SLMs for QA.} 
When comparing LLMs such as GPT-4~\cite{achiam2023gpt} or BLOOM-175B~\cite{le2023bloom} with fine-tuned SLMs in QA tasks, the benefits of SLMs are clear. LLMs, while versatile across multiple domains due to extensive pre-training, are computationally demanding, making them less ideal for resource-limited settings. SLMs, however, when fine-tuned for specific domains, often match or exceed the performance of larger models within those specialties. The trade-off is between large-scale models' generalization and small-scale model's specialization: LLMs handle diverse domains but may need additional techniques such as knowledge injection for domain-specific queries. In contrast, domain-specific SLMs, though less flexible, provide higher accuracy and more relevant responses, making them ideal for edge deployments where computational resources are scarce but domain precision is crucial.

\subsubsection{SLM Applications in Coding}
\label{slm_application_coding}
The adoption of SLMs for coding offers an alternative to LLMs due to their lower computational needs and potential for domain-specific tuning. Despite LLMs' proficiency in code generation and programming support, SLMs are advantageous for their faster inference, reduced operational costs, and suitability for real-time environments where rapid responses are crucial. Representative works are discussed next. The Phi series~\cite{javaheripi2023phi, abdin2024phi, li2023textbooksneediiphi15} showcase SLMs' evolution in coding tasks. For instance, Phi-1~\cite{gunasekar2023textbooksneed}, a Transformer with 1.3B parameters, specializes in basic Python coding and achieves notable scores in benchmarks such as HumanEval \cite{gunasekar2023textbooksneed}, which includes 164 programming problems. Subsequent models, Phi-1.5 and Phi-2, have enhanced these capabilities, while Phi-3 demonstrated SLMs' potential to rival larger models \cite{abdin2024phi}. The latest model, Phi-3.5-mini, with 3.8B parameters, excels in long context tasks using advanced fine-tuning and optimization techniques, performing comparably to larger models such as Llama-3.1-8B-instruct \cite{dubey2024llama} and surpassing smaller ones like Gemma-2 \cite{team2024gemma2}.

Another avenue of development is the fine-tuning of general-purpose SLMs for coding tasks \cite{roziere2023code, team2024codegemma, guo2024deepseek, bai2023qwentechnicalreport, lozhkov2024starcoder}. For instance, CodeLlama models \cite{roziere2023code}, derivatives of Llama 2 \cite{touvron2023llama2}, undergo a rigorous fine-tuning process on domain-specific datasets, enhancing their proficiency in specific programming languages such as Python. They are trained to handle tasks such as syntax error detection, code suggestion, and infilling, where they learn to predict and complete missing parts of the code. This specialized fine-tuning improves their ability to interpret and execute detailed programming instructions, making them highly effective in real-time code editing environments \cite{roziere2023code}. CodeGemma models \cite{team2024codegemma}, stemming from Google DeepMind’s Gemma framework, also exhibit a focused approach to enhancing coding capabilities through fine-tuning. These models are specifically engineered for high-performance code generation and infilling, underpinned by extensive training on a vast corpus of over 500 billion to 1 trillion tokens, predominantly consisting of code. This comprehensive dataset enables CodeGemma models to excel in mathematical reasoning and complex problem-solving within code contexts, setting new benchmarks in latency-sensitive applications such as real-time IDE support and automated code reviews \cite{team2024codegemma}.

\begin{wraptable}[16]{r}{5.3cm}
\centering
\vskip -0.0em
\caption{Performance comparison between SLMs and LLMs in coding benchmarks. All models listed are chat or instruct versions, and performance are sourced from respective research papers or technical reports \cite{roziere2023code, Trufinescu2024,dubey2024llama,guo2024deepseek,team2024codegemma}.}
\vskip -1em
\scriptsize
\label{tab:slm_llm_performance_coding}
\begin{tabular}{lccc}
\toprule
\textbf{Model} & \textbf{Size} & \textbf{HumanEval} & \textbf{MBPP}\\
\midrule
DeepSeek-Coder~\cite{guo2024deepseek} & 1.3B & 65.2& 49.4\\
CodeGemma \cite{team2024codegemma} & 2B & 37.8 & 49.2\\
Gemma 2 \cite{team2024gemma2} & 2B & 17.7 & 40.2\\
Phi-3.5-mini \cite{Trufinescu2024} & 3.8B & 62.8 & 69.6\\
\midrule
DeepSeek-Coder~\cite{guo2024deepseek} & 6.7B & 78.6& 65.4\\
CodeGemma \cite{team2024codegemma} & 7B & 60.4 & 55.2\\
Llama 3.1~\cite{dubey2024llama}  & 8B & 66.5  & 69.4\\
Gemma 2 \cite{team2024gemma2} & 9B & 61.0 & 69.3\\
GPT-3.5 Turbo & - & 68.0 & 71.2\\
\midrule
DeepSeek-Coder~\cite{guo2024deepseek} & 33B & 79.3& 70.0\\
Llama 3.1~\cite{dubey2024llama}  & 70B & 80.5 & 75.4\\
Llama 3.1~\cite{dubey2024llama}  & 405B & 89.0 & 78.8\\
GPT-4o~\citet{gpt4o}     & - &90.2 &81.4\\
Claude 3.5 Sonnet \cite{anthropic2024claude} & - &92.0 &76.6\\
\bottomrule
\end{tabular}
\end{wraptable}
\textbf{Comparison between SLMs and LLMs on Coding}. 
Table~\ref{tab:slm_llm_performance_coding} provides a comparative analysis of SLMs and LLMs on coding benchmarks HumanEval \cite{chen2021evaluating} and MBPP \cite{austin2021program}. Insights include: 
(\textbf{i}) Small SLMs (1.3B - 3.8B Parameters) like Phi-3.5-mini \cite{Trufinescu2024} achieve high scores, demonstrating the efficacy of small models. Mid-sized SLMs (6.7B - 9B Parameters), such as DeepSeek-Coder 6.7B~\cite{guo2024deepseek} and Llama 3.1 8B~\cite{dubey2024llama}, show improved performance, indicating that larger model sizes and enhanced training contribute to better accuracy. Large models (33B and above) like Llama 3.1 405B~\cite{dubey2024llama}, GPT-4o~\cite{gpt4o}, and Claude 3.5 Sonnet \cite{anthropic2024claude} excel, supporting the idea that bigger models generalize better across diverse coding tasks; 
(\textbf{ii}) There's a notable trade-off between computational efficiency and performance, with larger models requiring more resources, impacting their practical deployment in constrained environments; 
(\textbf{iii}) Specialized training and fine-tuning, as used in models like DeepSeek-Coder~\cite{guo2024deepseek}, are crucial for excelling in coding tasks, though such models may not handle complex requests as effectively, highlighting the versatility of general SLMs for broader applications.

\subsubsection{SLM Applications in Recommender Systems} 
\label{slm_application_recommender}

Recommender systems are essential in various online services, helping to manage information overload and meet users' personal needs. SLMs enhance recommendation systems by (1) addressing the cold start problem; (2) reducing popularity bias; (3) improving long-term planning; (4) serving as personalized recommenders; and (5) acting as content encoders. These applications show the versatility and effectiveness of SLMs in boosting performance and personalization in recommendation. Next, we introduce the details. 

\textbf{SLM for System Cold Start Problem.} 
Traditional recommendation systems, which utilize historical user-item interactions such as clicks, purchases, and ratings to learn representations and match items to users, fail in scenarios lacking any user-item interactions, known as the cold-start recommendation problem, often occurring in start-up businesses \cite{rashid2008learning}. Although LLMs address this with in-context learning, their slow and costly inference restricts real-time use. 
Thus, \textbf{PromptRec} \cite{wu2024could} explores using SLMs as in-context recommenders for recommendation system cold-start problems. However, SLMs often struggle without emergent context-learning abilities. To overcome this, SLMs are enhanced by pre-training on relevant corpora, using a improved C4 corpus subset \cite{raffel2020exploring}, and by developing training prompts for different domains, enhancing cold-start performance. Results show that enhanced SLMs like BERT-mini \cite{devlin2019bert}, with 11.3M parameters, achieve BERT-large's performance in cold-start scenarios, with only 17\% of BERT-large's inference time. Similarly, many studies have addressed the cold-start problem by leveraging BERT \cite{zhang2019cold, noorian2024bert, zhuang2021bert, heidari2022attention}. For example, \textbf{ADLRS} \cite{heidari2022attention} employs BERT to convert web-crawled item profiles into vectors that highlight key aspects, aiding recommender systems in acquiring essential initial information.

\textbf{SLM for Mitigating Popularity Bias.}
Popularity bias in recommender systems, marked by discrepancies between item popularity in training datasets and the real world, often stems from using closed-loop datasets with limited information. Recent LLMs leverage their broad open-world knowledge to better reason about user-item interactions \cite{lin2024rella, liu2024once}, reducing this bias by providing recommenders with more extensive item details. Using the chain-of-thought (CoT) prompting, LLMs decompose complex tasks into intermediate reasoning steps, enhancing understanding of user behavior and interests. However, LLMs' high resource demands limit their practical use. To overcome this, the Step-by-step Knowledge Distillation Framework for Recommendation (\textbf{SLIM}) \cite{wang2024can} distills LLM reasoning capabilities into SLMs, keeping just 4\% of the original parameters, transitioning from ChatGPT to Llama 7B~\cite{touvron2023llama}. SLIM uses detailed LLM templates to extract reasoning steps and streamlined templates for fine-tuning, enabling SLMs to improve recommender systems by better reasoning on richer item information.

\textbf{SLM for Long-term Planning.} 
Traditional recommender systems focus on optimizing immediate user responses, often maximizing short-term gains but overlooking long-term engagement. This can trap users in echo chambers and filter bubbles \cite{10.1145/3534678.3539073, gao2023cirs}. To tackle this, integrating planning capabilities into recommendations to balance immediate and long-term outcomes is vital. LMs, with their extensive knowledge and reasoning abilities, are expected to enhance planning capabilities. \textbf{BiLLP} \cite{shi2024large} adopts a hierarchical learning approach with macro and micro-learning phases. In macro-learning, a Planner and a Reflector, both as SLM instances like Llama-2-7B \cite{touvron2023llama2}, operate; the Planner forms long-term plans using high-level experiences, while the Reflector updates plans based on past actions. Micro-learning uses an SLM-based Actor-Critic mechanism for personalized planning, with the Actor implementing plans and the Critic assessing actions for long-term benefits. The use of SLMs for long-term planning, similar to their use in cold-start scenarios, remains underexplored and merits further research.

\begin{wrapfigure}[12]{r}{0.4\textwidth}
    \centering
    \vskip -2em
    \includegraphics[width=\linewidth]{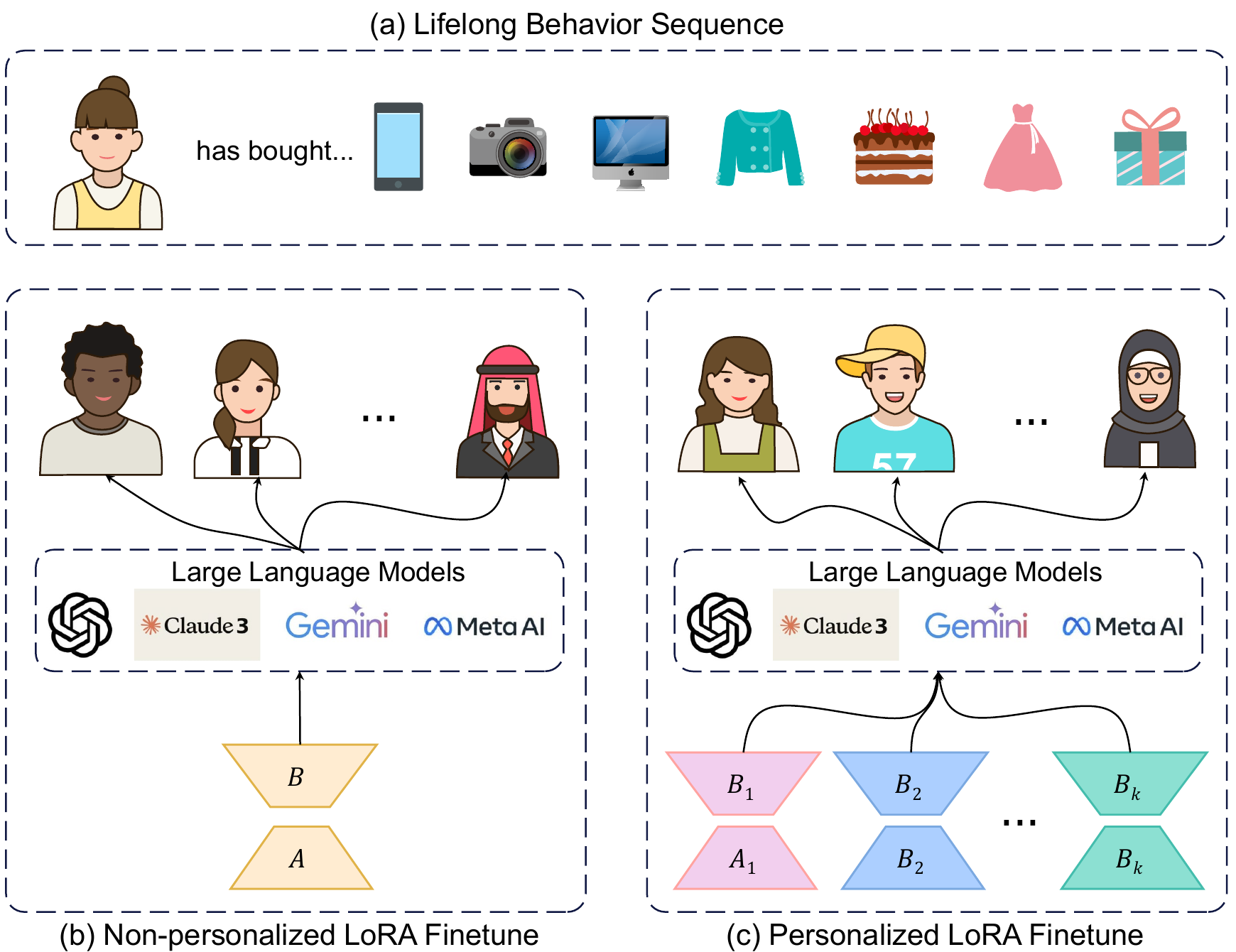}
    \vskip -1em
    \caption{The illustration of lifelong behavior sequence and personalized low-rank adaption (LoRA) for recommendation \cite{zhu2024lifelong}. }
    \label{fig:lifelong_recom}
\end{wrapfigure}
\textbf{SLMs as a Personalized Recommender.} Generative language model-based recommender systems require integrating user knowledge, typically achieved through fine-tuning. Fine-tuning techniques like LoRA \cite{hu2021lora} can incorporate extensive knowledge across all users by training an external module with a small number of parameters $\mathbf{A}$ and $\mathbf{B}$, but this approach often overlooks individual user preferences. To address this, \textbf{RecLoRA} \cite{zhu2024lifelong} utilizes Vicuna-7B \cite{vicuna2023} to integrate personalized knowledge into SLMs/LLMs tailored for recommendation tasks, as illustrated in Figure \ref{fig:lifelong_recom}. Specifically, RecLoRA maintains a set of parallel, independent LoRA weights $(\mathbf{A}_i, \mathbf{B}_i)$, allowing for the customization of language model parameters to match individual user preferences more effectively. 
\textbf{SLM as a Content Encoder.} 
Language models, particularly when deep, provide an effective starting point for fine-tuning on downstream tasks. In news recommendation systems, the representational capability of a model significantly impacts performance. Consequently, many news recommender systems now employ language models fine-tuned on specific datasets as text encoders. 
For example, \textbf{\citet{wu2021empowering}} conducts pioneering work using a pre-trained language model to enhance large-scale news recommender systems by substituting traditional news encoders with a BERT model \cite{devlin2019bert}. 
However, BERT may struggle to capture content as it is pre-trained on limited data. Therefore,
\textbf{ONCE} \cite{liu2024once} propose using Llama-2-7B~\cite{touvron2023llama2} as an encoder to overcome the limitations of BERT in content-based recommendations. Additionally, the study explores the synergistic use of LLMs in recommendation systems, finding that SLMs optimized with LoRA~\cite{hu2021lora} outperform the recommendation results of systems assisted by generic LLMs such as ChatGPT.

\subsubsection{SLM Applications in Web Search} 
\label{slm_application_web}
\begin{figure}[!t]
    \centering
    \includegraphics[width=0.8\linewidth]{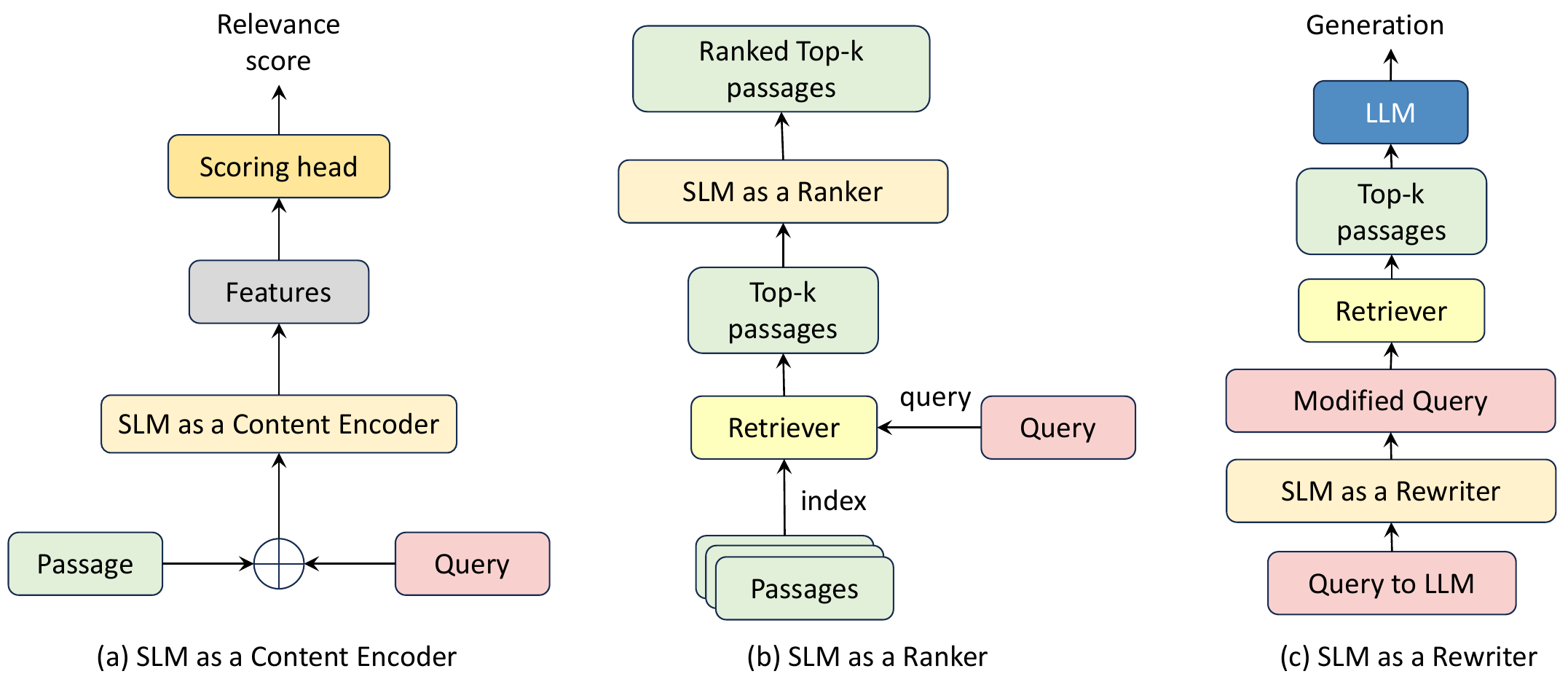}
    \vskip -1em
    \caption{Roles of SLM in Web Search.}
    \label{fig:slm_for_web_search}
    \vskip -2em
\end{figure}

Web search systems, involving retrieval and ranking, face challenges due to the diverse web documents and search queries. Traditional keyword-matching methods often fall short because of phrasing variations and the long-tail distribution of queries and content, complicating accurate semantic inference. Effective integration of retrieval and ranking models is also crucial. Language models, serving as content encoders, help overcome semantic challenges through their language understanding from pre-training \cite{chu2022h, dong2023i3, wang2023improving}. Joint training of retrieval and ranking models addresses integration, with SLMs ranking retrieved documents and acting as re-rankers. Additionally, SLMs serve as rewriters in scenarios requiring enhanced query understanding. Thus, in web search, SLMs fulfill three roles: (1) \textit{content encoder}, (2) \textit{ranker}, and (3) \textit{rewriter}, as depicted in Figure \ref{fig:slm_for_web_search}. Next, we give details.

\textbf{SLM as a Content Encoder.} 
Text embeddings are vector representations that encode semantic information, widely used in retrieval; SLM-based dense retrieval utilizes pre-trained deep language understanding to effectively tackle semantic challenges.
\textbf{H-ERNIE} \cite{chu2022h} employs a hierarchical model that encodes queries and documents at multiple granularity—character, word, and phrase—to improve specificity and relevance in web search results by aggregating finer details into coarser layers, addressing issues like ambiguous queries. \textbf{Implicit Interaction ($I^3$)} \cite{dong2023i3} uses BERT~\cite{devlin2019bert} as a content encoder, generating implicit pseudo-queries from passages to enable high online efficiency with offline caching of passage vectors. 
However, ERNIE and BERT-style models overlook advancements in SLMs such as context length extension \cite{roziere2023code}. 
Thus, 
\textbf{\citet{peng2023soft}} employs LLaMa-7B \cite{touvron2023llama} and Vicuna-7B~\cite{vicuna2023} as semantic encoders for embedding retrieval, demonstrating improved performance through soft prompt tuning.
\textbf{CoCondenser} \cite{gao2022unsupervised} addresses sensitivity to noisy data and large batch requirements during dense retriever training. Using the Condenser architecture with Transformer blocks, the model condenses information into dense vectors effectively. 

\textbf{SLM as a Ranker.}
The reranking task improves the order of multiple candidates to enhance retrieval quality because rerankers are more accurate than embedding retrievers. \textbf{InPars (Inquisitive Parrots for Search)} \cite{bonifacio2022inpars} employs the T5 base 220M~\cite{raffel2020exploring} as a re-ranker to enhance the BM25 retriever~\cite{robertson2009probabilistic}. 
Initially, BM25 selects 1K candidates, re-ranked by a fine-tuned T5 model (monoT5) adapted as a binary classifier to assess document-query relevance. Training data, generated by GPT-3~\cite{NEURIPS2020_1457c0d6}, formulates queries and selects random negative examples. Experiments show the monoT5-enhanced retriever significantly outperforms GPT-3; for example, it achieves a 0.3599 MAP score on the TREC-DL 2020 dataset \cite{trec2020}, surpassing GPT-3’s 0.3163.


\textbf{SLM as a Rewriter.} 
Queries to the retriever, typically just a few keywords, may reveal a knowledge gap between the query and the knowledge needed for effective retrieval, thus limiting performance.  To address this, the \textbf{``rewrite-retrieve-read''} framework \cite{ma2023query} uses T5-large \cite{raffel2020exploring} to bridge the knowledge gap in queries by rewriting them for more effective retrieval. This rewriter, trained via reinforcement learning with downstream LLM performance as a reward, outperforms general LLM rewrites. For example, it achieves a 45.97 F1 score on HotpotQA, surpassing the generic LLM's 43.85 F1 score.
\subsubsection{SLM Applications in Mobile-device} 
\label{slm_application_mobile}
The use of cloud-based LLMs on devices raises privacy concerns and their large size limits real-time responses in urgent scenarios such as medical emergencies. To overcome these issues, researchers are creating smaller, domain-specific models (SLMs) that offer accurate results and suit mobile use. This subsection discusses SLM applications on mobile devices, focusing on three aspects: (1) software API calls, (2) mobile control, and (3) basic NLP applications.

\textbf{SLM for Tool Learning.} 
Integrating LLMs with APIs enhances capabilities but incurs high training costs, prompting a shift to smaller, task-specific models that cut costs but risk errors. In response, \textbf{Octopus} \cite{chen2024octopusondevicelanguagemodel} uses a diverse dataset from over 30K APIs and curriculum learning \cite{liu2024let} to improve API function accuracy. This method boosts API performance in models like Codellama-7b \cite{roziere2023code} and Google’s Gemma series \cite{team2024gemma}.
\textbf{PhoneLM-1.5B-Call} \cite{yi2024phonelm} is fine-tuned on DroidCall \cite{xie2024droidcall} datasets and achieves comparable performance compared to GPT-4o-mini \cite{GPT-4o-mini}. \textbf{$\alpha$-UMI} \cite{shen-etal-2024-small} employs SLMs as planners, callers, and summarizers within multi-agent systems, outperforming a single LLM in tool uses.


\begin{wrapfigure}[13]{r}{0.55\textwidth}
    \centering
    \vskip -1.2em
    \includegraphics[width=0.55\textwidth]{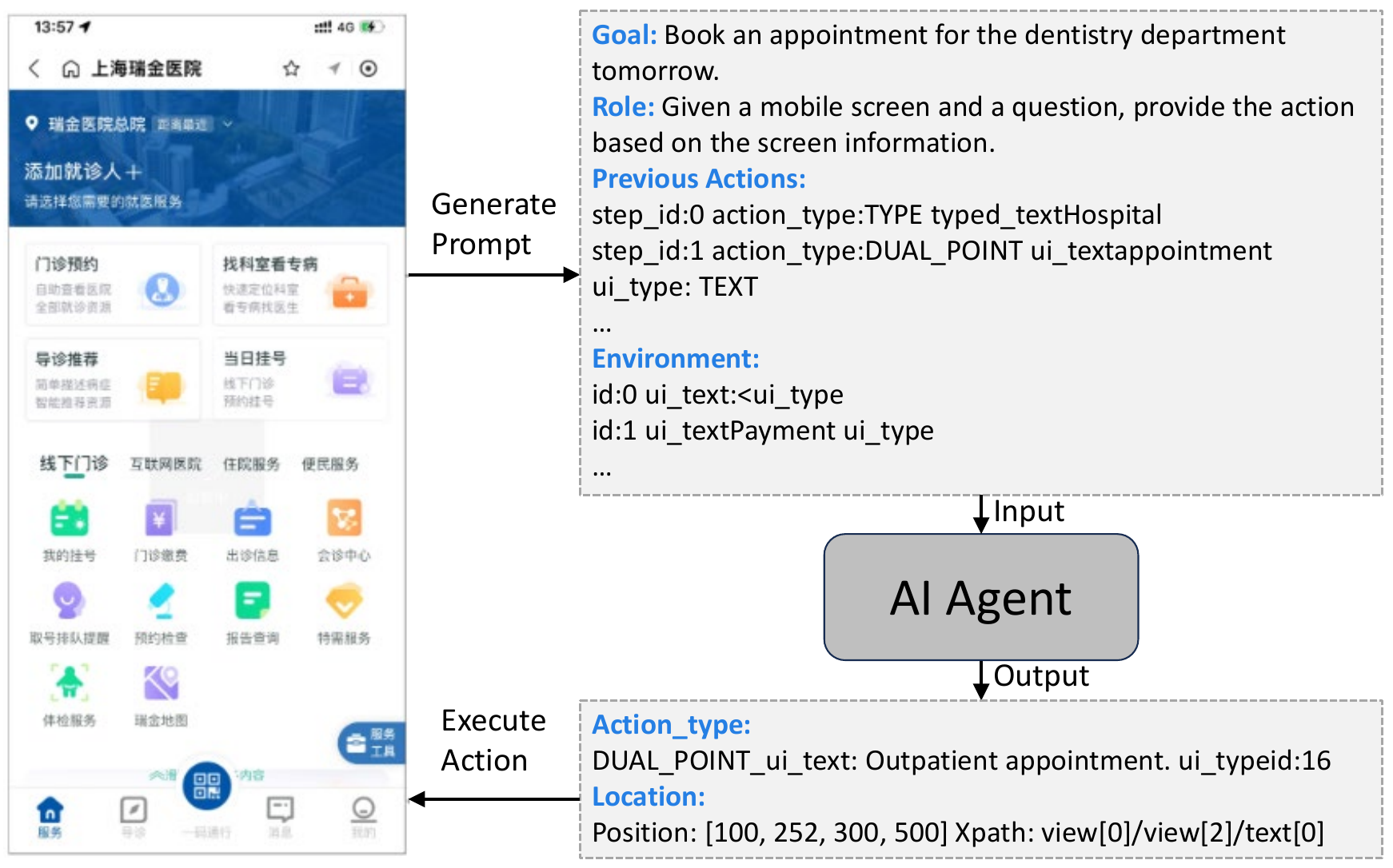}
    \vskip -1em
    \caption{An example workflow for an automated execution tool \cite{ding2024mobileagentenhancingmobilecontrol}. The screenshot in the left is taken from \cite{ding2024mobileagentenhancingmobilecontrol}.} 
    \vskip -1em
    \label{fig:example_automated_execution_tool}
\end{wrapfigure}
\textbf{SLM for Mobile Control.}
LM agents facilitate user-device interactions through taps, gestures, and text, automating tasks and enhancing user hands-free convenience. Unlike traditional developer-based approaches that require extensive developer effort to design interfaces and translate commands into API calls, LMs offer scalable automation via GUI-based text contents.
\textbf{MobileAgent}~\cite{ding2024mobileagentenhancingmobilecontrol} integrates instructions and Standard Operating Procedures (SOP) to improve SLMs for mobile control. As shown in Figure \ref{fig:example_automated_execution_tool}, it processes goals (e.g., booking a dental appointment) by analyzing screens, queries, prior actions, and UI elements, forming prompts to generate outputs and execute actions (e.g., selecting text, XPath). Fine-tuning Qwen-7B~\cite{bai2023qwentechnicalreport} on AIA medical data, it outperforms GPT-4~\cite{achiam2023gpt} on AitW~\cite{rawles2024androidinthewild}, a key mobile benchmark, without extra inference costs.
\textbf{\citet{carreira2023revolutionizingmobileinteractionenabling}} run a small offline model on mobile devices, fine-tuned with ChatGPT-3.5 data, enabling tasks like calls and web searches. RedPajama-INCITE-Chat-3B-v1~\citet{together2023redpajama}, selected for its size and chat features, uses native code and quantization, performing well on 6GB and 4GB Android devices.

\begin{wrapfigure}[11]{r}{0.55\textwidth}
    \centering
    \vskip -1.6em
    \includegraphics[width=\linewidth]{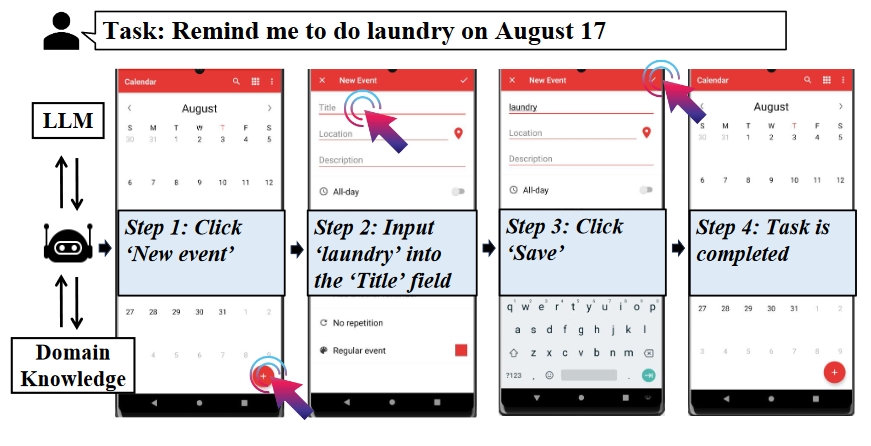}
    \vskip -1.5em
    \caption{An illustration of Vicuna-7B-powered mobile task automation \cite{carreira2023revolutionizingmobileinteractionenabling} shows a user asking to be reminded about doing laundry on Aug 17. The figure is taken from \cite{carreira2023revolutionizingmobileinteractionenabling}.}
    \label{fig:LLM_powered_mobile_task_automation}
\end{wrapfigure}
\textbf{AutoDroid}~\cite{wen2024autodroidllmpoweredtaskautomation} improves Android app interactions via GUI automation. Figure \ref{fig:LLM_powered_mobile_task_automation} shows LLM-powered task automation (e.g., setting a laundry reminder for Aug 17) in four steps: (1) click 'New Event', (2) enter 'laundry' in 'Title', (3) click 'Save', (4) finish. Using Vicuna-7B and app-specific knowledge, AutoDroid generates privacy-filtered prompts for tasks. On its DroidTask benchmark, it outperforms GPT-3.5 (34.7\%) and GPT-4 (54.5\%) with 57.7\% accuracy.
\textbf{M4 (composable mobile foundation model)}~\cite{yuan2024mobile} introduces a 9.2B parameter model (7.5GB memory) for mobile AI tasks, managed by the OS and hardware. Currently limited to high-end devices, its applicability will expand with increasing mobile memory/storage. These works highlight customizing smaller, domain-specific SLMs to address memory limits while preserving functionality in mobile environments.

\begin{wrapfigure}[17]{r}{0.5\textwidth}
    \centering
    \vskip -1.4em
    \includegraphics[width=0.5\textwidth]{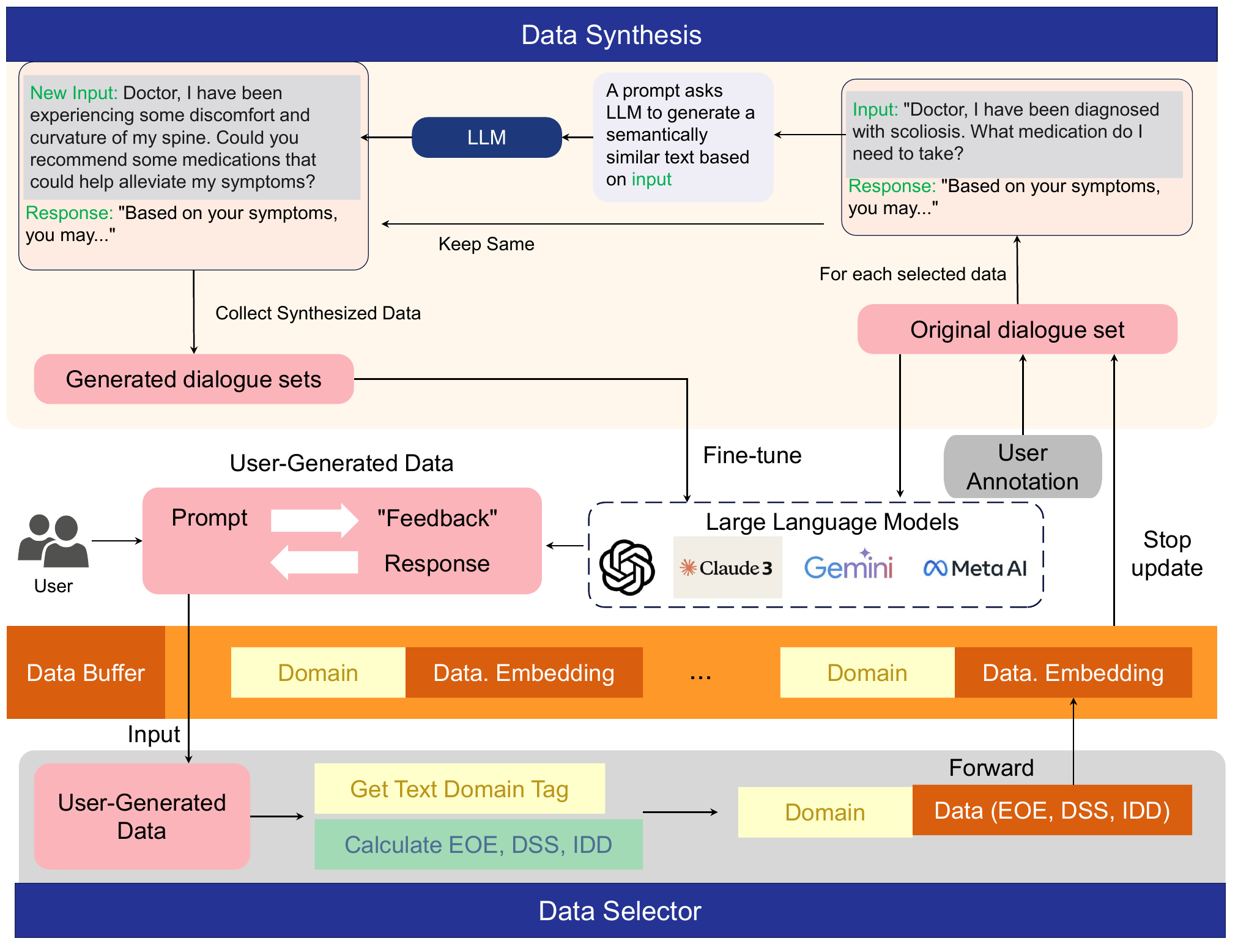}
    \vskip -1em
    \caption{
    Overview of fine-tuning SLMs with synthesized and user data \cite{qin2023enabling}. Data Synthesis generates semantically similar text via prompts, creating dialogue sets. Data Selection processes user data, tags domains, and calculates metrics (EOE, DSS, IDD). Selected data fine-tunes SLMs with user annotations. The iterative framework refines SLMs through continuous data generation and selection.
    }
    \label{fig:personalization_framework}
    \vskip -1em
\end{wrapfigure}
\textbf{SLM for Basic NLP Applications on Devices}
Performing basic NLP tasks such as text rewriting directly on the device can enable personalization while ensuring privacy. The sparse annotation on devices is a challenge. \textbf{\citet{qin2023enabling}} utilizes self-supervised data selection and synthesis for on-device fine-tuning, leveraging sparse annotations and limited storage effectively. This approach, demonstrated in Figure \ref{fig:personalization_framework}, employs the Llama-3B model \cite{touvron2023llama} and the LoRA fine-tuning method \cite{hu2021lora}, enhancing personalization by efficiently managing data through metrics including embedding entropy and domain-specific scores. 
In mobile text rewriting, \textbf{\citet{zhu2023ondeviceagenttextrewriting}} train the compact Palm 2-XXS model \cite{anil2023palm} using data generated by the larger Palm 2-L to ensure user privacy and accommodate device constraints. On its new benchmark, MESSAGEREWRITEEVAL, Palm 2-XXS achieves a BLEU score of 34.59, outperforming LLaMa-7B (16.65)~\cite{touvron2023llama}. Tests on the Samsung S23 show lower latency (29 tokens/s) than a 4-bit LLaMa-7B on a MacBook M1 Pro (18-22 tokens/s), proving its mobile efficiency for text rewriting.

\begin{takeaway2}

\textbf{Insights:} We draw several key insights from the development of task-specific SLMs:
\begin{itemize}[leftmargin=*]

\setlength{\itemsep}{0pt}
\setlength{\parskip}{0pt}
\setlength{\parsep}{0pt}
\small
    \item There is considerable potential in enhancing the efficiency and effectiveness of small models by integrating self-adaptive techniques with further fine-tuning and optimization of RAG-based methods.
    \item The growing relevance of SLMs in coding highlights their cost-effectiveness and efficiency as alternatives to LLMs, providing quick processing and easy fine-tuning for specialized tasks; while LLMs handle complex tasks well, SLMs, optimized and fine-tuned on specific data, are increasingly essential in resource-limited settings.
    \item SLMs significantly enhance recommendation systems due to their robust generalization, reasoning abilities, and in-context learning, addressing key challenges such as cold-start problems and distribution biases. They support long-term planning, replace traditional encoders, and use parallel low-rank parameters to inject personalized user knowledge effectively.
    \item SLMs play a crucial role in web search such as document encoding, text reordering, and query rewriting, often outperforming LLMs through techniques such as supervised fine-tuning, soft prompt tuning, unsupervised contrastive loss, and reinforcement learning, thereby enhancing adaptability and efficiency.
    \item SLMs are utilized on mobile devices primarily for privacy and memory constraints, with applications in API calls and mobile control; they are typically developed by generating data with LLMs and fine-tuning with SLMs, or by using local SLMs to handle privacy with LLMs boosting performance, and their training involves innovative techniques like learning from data streams and managing non-IID time series data.
\end{itemize}
\end{takeaway2}

\subsection{SLM Deployment on Mobile and Edge Devices}
\label{slm_deployment_mobile}

On-device applications benefit uniquely from the memory-saving efficiency and rapid runtime performance of SLMs, which offer advantages over LLMs. However, devices with extremely limited resources may still struggle with the parameter sizes of SLMs. To ensure both memory and runtime remain within acceptable range while still maintaining performance, it is crucial to integrate technologies that facilitate the deployment of SLMs on resource-constrained devices. The primary challenge for memory-efficient technologies arises from the inherent size of the SLMs and their associated caches. To address this, we survey existing works focused on compressing SLMs and their caches. Additionally, the large size of models significantly impacts runtime efficiency due to the extensive computing workload and potential weight transfers between the memory buffer and RAM/GPU memory. Other challenges include switching the Mixture of Experts between CPU and GPU memory and managing resource scheduling when deploying SLMs across multiple local devices. Therefore, in this subsection, we review representative works that address these challenges under two aspects: \textit{memory efficiency optimization} and \textit{runtime efficiency optimization}, as systematically compiled in Table \ref{tab:models_deployment_overview}.

\subsubsection{Memory Efficiency Optimization}  
\label{memory_efficiency}
Memory efficiency involves minimizing the memory usage of both the model and the KV cache during deployment. This is typically achieved through model compression techniques such as quantization \cite{yu2024edge, zhao2024llm, lin2024awq, rahman2023quantizedtransformerlanguagemodel}, the cache of MoE experts \cite{yi2023edgemoefastondeviceinference}, and KV cache compression \cite{kang2024gear}. 

\textbf{Compression on model parameters.} Quantization, a common method for deploying SLMs, lowers numerical precision to significantly save memory while preserving accuracy. We detail quantization strategies in Sections \ref{quantization} and \ref{quantization4enhance}, focusing here on representative works for edge devices.
\textbf{EDGE-LLM} \cite{yu2024edge} adapts LLMs for edge devices using a Layer-wise Unified Compression (LUC) method that combines layer-specific pruning and quantization to reduce computational demands and an Adaptive Layer Tuning and Voting scheme to optimize memory use while ensuring performance. Meanwhile, \textbf{LLM-PQ} \cite{zhao2024llm} addresses quantization and model layer partitioning for heterogeneous clusters, incorporating a Cost Model and an Indicator Generator to optimize bit-width assignment and layer partitioning through integer linear programming, enhancing quantization for complex computational setups. 
\textbf{Activation-aware Weight Quantization (AWQ)} \cite{lin2024awq} is a hardware-friendly, low-bit, weight-only quantization method for on-device LLMs, preserving essential weights based on activation distribution to minimize quantization loss.
\textbf{MoE-I$^2$} \cite{yang-etal-2024-moe} prune less important experts and applies low-rank decomposition to the remaining experts to further optimize efficiency.

\begin{table}[t]
\centering
\small
\caption{On-device Deployment Optimization Techniques}
\label{tab:models_deployment_overview}
\vskip -1em
\begin{tabularx}{\textwidth}{>{\hsize=0.28\hsize}X|>{\hsize=0.46\hsize}X|>{\hsize=1.2\hsize}X}
\hline
\textbf{Aspect} & \textbf{Representative Work} & \textbf{Key Point} \\
\hline
\multirow{10}{=}{\textbf{Memory Efficiency Optimization}} & EDGE-LLM \cite{yu2024edge} & 
Edge LLMs use LUC and adaptive tuning for efficiency \\
\cline{2-3}
& LLM-PQ \cite{zhao2024llm} & Optimize quantization and layer partitioning for complex setups.\\
\cline{2-3}
& AWQ \cite{lin2024awq} & Preserve key weights based on activation distribution. \\
\cline{2-3}
&MoE-I$^2$ \cite{yang-etal-2024-moe}& Prune less important experts in MoE. \\
\cline{2-3}
& MobileAIBench \cite{murthy2024mobileaibenchbenchmarkingllmslmms}& Evaluation \\
\cline{2-3}
& EdgeMoE \cite{yi2023edgemoefastondeviceinference}  & Load experts on activation, tripling memory savings.\\
\cline{2-3}
& GEAR \cite{kang2024gear} & Enhance KV cache quantization by integrating error-reduction techniques.  \\
\cline{2-3}
& DMC \cite{nawrot2024dynamic}  & Adaptively compress KV cache, optimizing storage efficiency.\\
\cline{2-3}
& Transformer-Lite \cite{li2024transformerlitehighefficiencydeploymentlarge} & Optimize KV cache to reduce redundancy and memory use.\\
\cline{2-3}
& LLMaaS \cite{yin2024llmservicemobiledevices}& LLMaaS manages apps via chunk-wise KV cache optimization on mobiles. \\
\hline

\multirow{5}{=}{\textbf{Runtime Efficiency Optimization}} 
&mllm-NPU \cite{xu2024empowering}& On-device NPU (neural processing units) to reduce prefill latency.\\
\cline{2-3}
& COST-EFF \cite{shen-etal-2022-cost}  & Distill a multi-exit model from the original PLM.\\
\cline{2-3}
& LLMCad \cite{xu2023llmcadfastscalableondevice} & Use SLM for fast token generation and cloud verification. \\
\cline{2-3}
& EdgeMoE \cite{yi2023edgemoefastondeviceinference} & Predict expert needs, boosting inference speed and reducing latency. \\
\cline{2-3}
& LinguaLinked \cite{zhao2023lingualinkeddistributedlargelanguage} & Optimize data flow and load, enhancing multi-threading efficiency.\\
\hline
\end{tabularx}
\end{table}

\textbf{Cache of MoE Experts.}  Beyond standard quantization, which reduces storage for all model parameters, another strategy involves caching a mixture of experts (MoE). Driven by the fact that memory storage is more cost-effective and scalable than computing capacity, the MoE architecture \cite{jacobs1991adaptive} boosts performance while minimizing computational costs by activating only portions of the LLM as needed. However, this approach incurs significant memory overhead, making it impractical for edge device memory constraints. For example, Switch Transformers \cite{fedus2022switch}, with 32 experts per layer, require 54GBs of memory for inference, exceeding the capacity of most edge devices. \citet{yi2023edgemoefastondeviceinference} notes that in the Switch Transformers model, although most of the model weights (86.5\%) are attributed to experts, these weights account for only a small fraction (26.4\%) of the computations. To address this, \textbf{EdgeMoE} \cite{yi2023edgemoefastondeviceinference} introduces a method where experts are loaded into an expert memory buffer only when activated, achieving approximately $3 \times$ memory savings compared to the baseline where all weights are held in memory.

\textbf{KV Cache Compression}. 
When serving LMs for inference, using a KV cache is common practice to avoid intensive recalculations and speed up generation \cite{pope2023efficiently}. However, cache memory consumption escalates with model size and sequence length, posing a challenge for edge devices. To manage this, offloading techniques transfer KV caches to CPU memory \cite{sheng2023flexgen,aminabadi2022deepspeed}, although this can introduce significant overhead due to the switching costs between GPUs and CPUs.
Token dropping compresses cache size by keeping only key tokens, often identified by low attention scores \cite{zhang2024h2o, liu2024scissorhands, ge2024model}. However, this method struggles with complex tasks, especially at high compression levels, due to increased estimation errors in compressed KV values. \textbf{GEAR} \cite{kang2024gear} addresses these issues by enhancing KV cache quantization with error-reduction techniques, including: (i) quantizing caches of similar magnitudes to ultra-low precision, (ii) using a low-rank matrix for efficient quantization residual approximation, and (iii) employing a sparse matrix for correcting outliers. This approach separates coherent from incoherent approximation errors, enabling near-lossless KV cache compression and achieving up to $2.29 \times$ peak memory reduction.

Besides, \textbf{Dynamic Memory Compression (DMC)}~\cite{nawrot2024dynamic} adaptively compresses the KV cache by either adding new key and value representations directly or blending them with the top cache item using a weighted average. 
\textbf{Transformer-Lite} \cite{li2024transformerlitehighefficiencydeploymentlarge} tackles the redundancy of storing the KV cache twice in model inputs and outputs, which increases memory use. It optimizes storage by allocating a large tensor based on the maximum sequence length needed for inference. Sub-tensors are then created from this main tensor at different address offsets to serve as input and output KV caches, allowing direct writing to the correct locations during inference and removing extra copying steps.
\textbf{LLMaaS} \cite{yin2024llmservicemobiledevices} introduces LLM as a Service for mobile devices, managing all apps through LLMS. This system uses chunk-wise KV cache compression and swapping, enabling efficient context switching within memory constraints. By segmenting the KV cache into independently compressed and swapped chunks, LLMS balances memory use and I/O bandwidth better than token-level or context-level management.

\subsubsection{Runtime Efficiency Optimization} 
\label{runtime_efficiency} 
The goal of decreasing computing workload aligns with enhancing memory efficiency through methods such as quantization, as mentioned in the previous section. Decreasing model weight precision or reducing the number of weights naturally lowers latency. 
Other runtime efficiency techniques of minimizing inference latency involve, reducing prefill latency, early exits, large and small model collaboration, decreasing switching time in MoE, and reducing latency in distributed SLMs.


\textbf{Reducing prefill latency.}
\textbf{mllm-NPU} \cite{xu2024empowering} is the first LLM inference system that leverages on-device NPU (neural processing units) to reduce prefill latency and energy consumption. It incorporates a chunk-sharing graph, shadow outlier execution, and out-of-order subgraph execution to enhance NPU offloading efficiency. Experiments have shown mllm-NPU's superior performance benefits, including up to 43.6× speedup and 59.5× energy savings.

\textbf{Dynamic Early Exits}
A decoupled runtime saving technique is dynamic early exits. Originating from BranchyNet \cite{teerapittayanon2016branchynet}, which introduces exit branches after specific convolution layers in the CNN model, this concept has been adapted for PLMs as Transformer layer-wise early exiting \cite{xin-etal-2021-berxit}. Early exiting enables dynamic acceleration during inference and reduces temporal overhead by allowing exit without passing through all model layers. To address the inconsistency issue arising from exiting at different layers, \textbf{COST-EFF} \cite{shen-etal-2022-cost} distills a multi-exit model from the original PLM.

\textbf{Large and Small Model Collaboration}
Model collaboration, deploying SLMs on devices with cloud-based LLM support, enhances runtime efficiency. LLMs will increase latency when directly deployed via mobile engines like llama.cpp due to a large number of computing operations. \textbf{LLMCad} \cite{xu2023llmcadfastscalableondevice} addresses this by using a real-time, memory-resident SLM for simple tokens such as determiners and punctuation. The SLM generates tokens, while a cloud-based LLM verifies and corrects them, speeding up the process. LLMCad enhances token generation up to $9.3\times$ by pairing the memory-resident SLM, Llama 2 7B, with the cloud-based LLM, Llama 2 13B, cutting latency from 16.2 to 3.9 seconds on Xiaomi Pro for TruthfulQA tasks \cite{lin2022truthfulqa}.

\textbf{Reducing MoE Switching Time.} To reduce latency in MoE architectures caused by frequently switching experts in limited device memory, \textbf{EdgeMoE} \cite{yi2023edgemoefastondeviceinference} enhances runtime efficiency by preemptively predicting which expert will be needed, based on the observed long-tail distribution of unbalanced expert activations. It utilizes a statistical model, built offline, to estimate expert activation probabilities in transformer layers from previous activations. During inference, EdgeMoE preemptively loads the most likely needed expert, accelerating inference by $1.11 \times$ to $2.78\times$ and significantly reducing latency. For instance, in a switch transformer model with 8 experts, latency drops from approximately 0.7s to 0.3s, outperforming the baseline method that preloads experts based on hit ratios.

\textbf{Reducing Latency in Distributed SLMs.} Distributing an SLM across smaller devices reduces the need for extensive model compression while preserving accuracy. However, this approach faces challenges that incur latency such as managing diverse device capabilities, handling data dependencies between model segments, and adapting to dynamic resource availability. To address these issues, \textbf{LinguaLinked} \cite{zhao2023lingualinkeddistributedlargelanguage} addresses these issues by optimizing model assignment to match device capabilities and minimize data transmission, implementing runtime load balancing to redistribute tasks and prevent bottlenecks, and enhancing communication for efficient data exchange between segments. With multi-threading, the system improves, achieving acceleration rates between $1.73\times$ and $2.65\times$ for both quantized and full-precision models.

\begin{takeaway2}
\footnotesize
    \textbf{Insights:} We draw several key insights from the deployment of SLMs:
    \begin{itemize}[leftmargin=*]
    \setlength{\itemsep}{0pt}
    \setlength{\parskip}{0pt}
    \setlength{\parsep}{0pt}
    \item Model size remains a bottleneck for both memory and runtime efficiency. A common solution is model quantization, which reduces model precision to save memory and lessen computing workload, thereby boosting inference speed \cite{yu2024edge,zhao2024llm,lin2024awq,murthy2024mobileaibenchbenchmarkingllmslmms,rahman2023quantizedtransformerlanguagemodel,liu2024mobilellm}. Similarly, KV cache compression also helps achieve these efficiency gains \cite{kang2024gear,nawrot2024dynamic,li2024transformerlitehighefficiencydeploymentlarge,yin2024llmservicemobiledevices}.
    \item Mixture of Experts (MoE) is commonly used in SLMs to enhance performance using the same computing resources, but it results in increased memory usage. To address this, only activated experts are loaded into the memory buffer while the majority are stored cold on disk. However, the cost of switching can slow down inference. Designing a preemptive expert pre-load strategy could therefore accelerate the inference \cite{yi2023edgemoefastondeviceinference}. 
    \item Model collaboration between local SLMs and cloud-based LLMs enhances both memory and runtime efficiency by using smaller models on local devices, which are then verified by cloud LLMs to ensure performance is maintained. Using SLMs locally reduces memory usage and shortens the inference time from the local model. However, internet latency and delays in cloud LLM inference can still introduce latency. Verifying SLM outputs every $N$ tokens using LLMs can effectively mitigate this latency \cite{xu2023llmcadfastscalableondevice}.
    \item One deployment approach involves deploying SLMs/LLMs across multiple trusted local devices to maintain original performance while only loading a fraction of the model weights. However, this method can incur latency due to varying device capabilities and resource scheduling challenges. To address these issues, optimizing model assignment to align with device capabilities and minimizing data transmission are effective strategies \cite{zhao2023lingualinkeddistributedlargelanguage}.
    \end{itemize}
\end{takeaway2}

%% file: sections/6.model.tex
\section{Generic and Domain-Specific Small Language Models} 


\label{model}
This section investigates small language models (with fewer than 7 billion parameters) in both general and specific domains. It details the methods of obtaining these small language models, the datasets, and the evaluation tasks, exploring the techniques for acquiring SLMs through compression, fine-tuning, or training from scratch. Additionally, we summarize the representative small language models, as detailed in Table \ref{tab:generic_small_models_overview_training} and \ref{tab:specific_small_models_overview_training}.

\begin{table}[htbp]
\caption{High-level Overview and Training Details of Generic-domain Small Language Models. \#Params means Parameter amounts. ">" indicates parameters larger than 7B.}
\label{tab:generic_small_models_overview_training}
\vskip -1em
\centering
\tiny
\renewcommand{\arraystretch}{1.5} 
\begin{tabular}{p{1.5cm}|p{1.5cm}|p{0.4cm}|p{0.8cm}|p{0.7cm}|p{3.2cm}|p{4.5cm}}
\hline
\textbf{Model} & \textbf{\#Params} & \textbf{Date} & \textbf{Paradigm} & \textbf{Domain} & \textbf{Training Datasets} & \textbf{Training Techniques} \\
\hline
PhoneLM~\cite{yi2024phonelm} & 0.5B; 1.5B & 2024.11 & Pre-train & Generic & DCLM-baseline \cite{li2024datacomp}, StarCoderData \cite{li2023starcodersourceyou}; OpenWebMath \cite{paster2024openwebmath}, Dolma-algebraic and Dolma-arXiv \cite{dolma}  & RoPE, MHA, Gated FFN, RMSNorm, ReLU, FSDP, Flash Attention 2, ZeRO \\
\hline
Llama 3.2~\cite{llama3.2} & 1B; 3B & 2024.9 & Pre-train & Generic & no release (9T tokens) & GQA, SiLU, Multilingual Text and code, Shared embedding, Pruning, Distillation, SFT, RLHF, RS, DPO \\
\hline
Qwen 1~\cite{bai2023qwentechnicalreport} & 1.8B; 7B; > & 2023.12 & Pre-train & Generic & no release & MHA; RoPE; SwiGLU; RMSNorm\\
Qwen 1.5~\cite{bai2023qwentechnicalreport} & 0.5B; 1.8B; 4B; 7B; > & 2024.2 & Pre-train & Generic & no release & MHA; RoPE; SwiGLU; RMSNorm; Multilingual support \\
Qwen 2~\cite{yang2024qwen2} & 0.5B;1.5B; 7B; > & 2024.6 & Pre-train & Generic & no release & GQA; RoPE; SwiGLU; RMSNorm; Multilingual support \\
Qwen 2.5~\cite{yang2024qwen2} & 0.5B; 1.5B; 3B; 7B; \textgreater{} & 2024.9 & Pre-train & Generic & no release & GQA; RoPE; SwiGLU; RMSNorm; Multilingual support; Larger corpus \\
\hline
Gemma \cite{team2024gemma} & 2B; 7B & 2024.2 & Pre-train & Generic & Unknown & MHA, RoPE, $\text{GELU}_{\text{tanh}}$ \\
Gemma 2 \cite{team2024gemma2} & 2B; > & 2024.7 & Pre-train & Generic & Unknown & GQA; RoPE; $\text{GELU}_{\text{tanh}}$; Alternating Local and Global Attention; Logit Soft-Capping; RMSNorm for Pre and Post-Normalization \\
\hline
SmolLM \cite{allal2024SmolLM} & 135M; 360M; 1.7B & 2024.7 & Pre-train & Generic & smollm-corpus \cite{benallal2024smollmcorpus} & GQA, trapezoidal LR scheduler \\
\hline 
H2O-Danube3 \cite{pfeiffer2024h2o} & 500M; 4B & 2024.7 & Pre-train & Generic & Unknown  & Three different training stages with different data mixes \\
\hline
Fox-1 \cite{fox1} & 1.6B & 2024.6 & Pre-train & Generic & Unknown (3T tokens) & GQA; Deep architecture \\
\hline
Rene \cite{Rene} & 1.3B & 2024.5 & Pre-train & Generic & Dolma-1.7 \cite{dolma} & Mamba-2 layers, sliding-window attention (SWA) \\
\hline
MiniCPM \cite{hu2024minicpm} & 1.2B; 2.4B & 2024.4 & Pre-train & Generic & Dolma \cite{dolma}; C4 \cite{raffel2020exploring}; Pile \cite{dey2023cerebras}; stack \cite{kocetkov2022stack}; StarCoder \cite{li2023starcodersourceyou}; UltraChat \cite{ding2023enhancing}; OssInstruct \cite{wei2023magicoder}; EvolInstruct \cite{xu2023wizardlm} & Warmup-Stable-Decay (WSD) learning rate scheduler \\
\hline
CT-LLM \cite{du2024chinesetinyllmpretraining} & 2B & 2024.4 & Pre-train & Generic & MAP-CC & Chinese, MHA, RoPE, SwiGLU, RMSNorm \\  
\hline 
OLMo \cite{groeneveld2024olmo} & 1B; 7B & 2024.2 & Pre-train & Generic & Dolma \cite{dolma} ~\footnote{Dolma includes Dolma's CC, RefinedWeb, Star Coder, C4, PushShift API, Semantic Scholar, RedPajama, Flan, CC News, OpenWebMath, MetaWika, Wikipedia} & SwiGLU; RoPE, Non-parameteric Layer Norm  \\
\hline
TinyLlama \cite{zhang2024tinyllamaopensourcesmalllanguage} & 1B & 2024.1 & Pre-train & Generic & SlimPajama \cite{cerebras2023slimpajama} and StarCoder \cite{li2023starcodersourceyou} & GQA, SiLU, FSDP, Flash Attention \cite{daoflashattention}, xFormers \cite{xFormers2022} \\
\hline
Phi-1 \cite{gunasekar2023textbooksneed} & 1.3B & 2023.6 & Pre-train & Coding & CodeTextBook \cite{gunasekar2023textbooksneed}~\footnote{CodeTextBook includes stack v1.2, code contests, synthetic python textbooks and exercises} & MHA, $\text{GELU}_{\text{tanh}}$, RoPE, FlashAttention \\
Phi-1.5 \cite{li2023textbooksneediiphi15} & 1.3B & 2023.9 & Pre-train & Generic & CodeTextBook \cite{gunasekar2023textbooksneed}; Synthetic Datasets (20B) & MHA, $\text{GELU}_{\text{tanh}}$, RoPE, FlashAttention, Deep ZeRO Stage 2 \\
Phi-2 \cite{javaheripi2023phi} & 2.7B & 2023.12 & Pre-train & Generic & CodeTextBook \cite{gunasekar2023textbooksneed}; Synthetic Datasets (20T) & MHA, $\text{GELU}_{\text{tanh}}$, RoPE, FlashAttention, Deep ZeRO Stage 2 \\
Phi-3 \cite{abdin2024phi} & 3.8B; 7B; > & 2024.4 & Pre-train & Generic & Scaled-up dataset from phi-2 & MHA, SiLU, RoPE, FlashAttention, Deep ZeRO Stage 2  \\
Phi-3.5 \cite{abdin2024phi} & 3.8B; 4.2B; 6.6B & 2024.4 & Pre-train & Generic & more multilingual and long-text data & Multilingual; Vision; MHA, SiLU, RoPE, FlashAttention, ZeRO 2  \\
\hline
OpenELM \cite{mehta2024openelm} & 270M; 450M; 1.1B; 3B & 2024.4 & Pre-train & Generic & RefinedWeb \cite{penedo2023refinedweb}, deduplicated PILE \cite{gao2020pile}, partial RedPajama \cite{together2023redpajama}, partial Dolma v1.6 \cite{dolma} & No biases in FC layers; Pre-norm: RMSNorm; Pos encoding: RoPE; Attention: GQA; FFN: SwiGLU; Tokenizer: LLaMA-style \\
\hline
MobiLlama \cite{thawakar2024mobillama} & 0.5B; 0.8B & 2024.2 & Pre-train & Generic & LLM360 Amber \footnote{LLM360 Amber includes Arxiv, Book, C4, Refined-Web, StarCoder, StackExchange, and Wikipedia} & GQA; SwiGLU; Parameter-sharing \\
\hline
MobileLLM \cite{liu2024mobilellm} & 125M; 350M & 2024.2 & Pre-train & Generic & Unknown (1T tokens) & SwiGLU FFN, deep and thin architectures, embedding sharing, and grouped query attention \\
\hline
StableLM~\cite{StableLM-3B-4E1T}& 3B; 7B & 2023.4 & Pre-train & Generic & \multirow{2}{3.2cm}{RefinedWeb \cite{penedo2023refinedweb}, RedPajama \cite{together2023redpajama}, the Stack \cite{kocetkov2022stack}, OpenWebText \cite{gokaslan2019openwebtext}, OpenWebMath \cite{paster2024openwebmath}, and partial CulturaX \cite{nguyen2024culturax}} & MHA; SiLU; Fine-tuning; DPO; Self-knowledge; RoPE; LayerNorm; no Biases \\
StableLM 2~\cite{bellagente2024stable}& 1.6B & 2024.2 & Pre-train & Generic &  &  \\
\hline
Cerebras-GPT \cite{dey2023cerebras} & 111M; 256M; 590M; 1.3B; 2.7B; 6.7B; > & 2023.4 & Pre-train & Generic & Pile \cite{gao2020pile} & MHA; GELU; Maximal Update Parameterization \\
\hline
Pythia \cite{biderman2023pythiasuiteanalyzinglarge} & 14M;70M;160M;410M; 1B;1.4B;2.8B;6.9B;> & 2023.4 & Pre-train & Generic & Pile \cite{gao2020pile} & MHA; GELU; Flash Attention \cite{dao2022flashattention}; RoPE \cite{su2024roformer}; ZeRO \cite{rajbhandari2020zero} \\
\hline
BLOOM, BLOOMZ~\cite{le2023bloom} & 560M; 1.1B; 1.7B; 3B; 7.1B; > & 2022.11 & Pre-train & Generic & ROOTS~\cite{laurenccon2022bigscience} and 13 programming languages & MHA; $\text{GELU}_{\text{tanh}}$; ALiBi Positional Embedding \cite{press2022train}, Embedding LayerNorm \cite{dettmers2022gptint} \\
\hline
Galactica~\cite{taylor2022galactica} & 125M; 1.3B; 6.7B; > & 2022.11 & Pre-train & Scientific & Open-access scientific materials (106B tokens) but not released & MHA; GeLU; Learned Positional Embeddings \\
\hline
OPT \cite{zhang2022opt} & 125M; 350M; 1.3B; 2.7B; 5.7B & 2022.5 & Pre-train & Generic & Pile \cite{gao2020pile} and PushShift.io Reddit \cite{baumgartner2020pushshift} & MHA; ReLU \\
\hline
XGLM~\cite{lin-etal-2022-shot} & 1.7B; 2.9B; > & 2021.12 & Pre-train & Generic & CC100-XL & - \\
\hline
GPT-Neo \cite{gpt-neo} & 125M; 350M; 1.3B; 2.7B & 2021.5 & Pre-train & Generic & Pile \cite{gao2020pile} & - \\
\hline
Megatron-gpt2 \cite{shoeybi2019megatron} & 355M; 2.5B; > & 2019.9 & Pre-train & Generic & Wikipedia \cite{devlin2019bert}, CC-Stories \cite{trinh2018simple}, RealNews \cite{zellers2019defending}, OpenWebtext & - \\

\hline
\hline

MINITRON \cite{muralidharan2024compact} & 4B; > & 2024.7 & Distillation; Pruning & Generic & 8T tokens in Nemotron-4 \cite{parmar2024nemotron} & LR WSD Scheduler \\
\hline
Orca 2 \cite{mitra2023orca} & 7B & 2023.11 & Distillation & Generic & Orca 2 dataset & LLaMA-2-7B based; prompt erasing \\
Orca \cite{mukherjee2023orca} & 13B & 2023.6 & Distillation & & FLAN-v2 \cite{longpre2023flan} & From ChatGPT and GPT4, Explanation tuning; Progressive Learning \\
\hline
MINIMA \cite{zhang2023towards} & 3B & 2023.11 & Distillation & Generic & Pile \cite{gao2020pile}, Wudao \cite{yuan2021wudaocorpora}, GitHub \cite{together2023redpajama} & From Llama-2-7B, Zero2, Flash Attention, Optimal teacher size \\
\hline
Dolly-v2 \cite{DatabricksBlog2023DollyV2} & 3B; 7B; > & 2023.4 & Instruction tuning & Generic & Databricks-dolly-15k \cite{DatabricksBlog2023DollyV2} & from pythia \\
\hline
LaMini-LM \cite{kim2016sequence} & 61M-7B & 2023.4 & Distillation & Generic & LaMini instruction dataset & a collection of SLMs distilled from ChatGPT-generated 2.58M instructions.  \\
\hline
Specialized FlanT5~\cite{fu2023specializing} & 250M; 760M; 3B & 2023.1 & Instruction Tuning & Generic (math) & GSM8K & Base model is FlanT5 \\
\hline
\end{tabular} 
\end{table}


\subsection{Generic-domain SLMs}
\label{generic_slms}
\begin{figure}[!ht]
    \centering
    \includegraphics[width=0.8\linewidth]{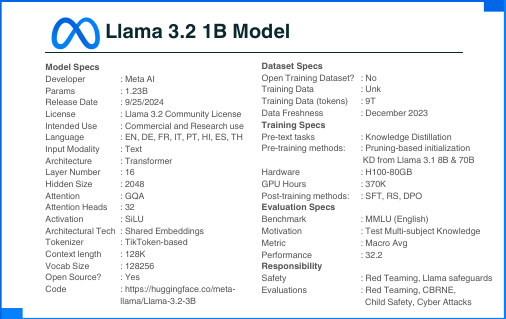}
    \vskip -1em
    \caption{Llama 3.2 1B model card}
    \label{fig:model_card_llama3.2_1B}
\end{figure}

\paragraph{Overview} 
SLMs, with fewer parameters than LLMs, enhance computational efficiency in pre-training, fine-tuning, and inference, reducing memory and energy demands—crucial for resource-limited environments. Their compact, localized nature boosts privacy, personalization, and response times, making them ideal for low-power edge devices. Therefore, SLMs are attracting increasing attention, and various models are being developed.
Table \ref{tab:generic_small_models_overview_training} summarizes current representative generic-domain $42$ SLMs/SLM families. Although all chosen SLMs have similar architectures, they vary in specific training datasets and techniques, with some datasets not being openly available. 
Taking the latest Llama 3.2 1B models~\cite{llama3.2} in Figure \ref{fig:model_card_llama3.2_1B} as an example, its parameter size and use of filtered high-quality training data, pruning-based initialization, knowledge distillation pre-training tasks, and training techniques such as Supervised Fine-Tuning (SFT), Rejection Sampling (RS), and Direct Preference Optimization (DPO) distinguish it from others.


\subsubsection{Architecture Design}
From Table~\ref{tab:generic_small_models_overview_training}, we observe several trends in component choices for SLMs:
\begin{enumerate}[leftmargin=*]
    \item Recent SLMs frequently employ Grouped Query Attention (GQA) in self-attention mechanisms because it can reduce computational complexity. GQA achieves this by sharing query representations across multiple heads while keeping key and value representations separate. This approach aligns with the goals of SLM to enhance efficiency without compromising functionality.
    \item The choice of activation function should balance model capability and efficiency. ReLU, known for its efficiency, introduces greater sparsity to the model, which facilitates faster coefficient calculations for inference acceleration. In contrast, SwiGLU's parameters are learned during training, allowing the model to dynamically adapt to diverse tasks and datasets, thereby enhancing model capabilities and establishing it as a state-of-the-art option. SiLU, situated between these two, is favored for its balance of computational efficiency and model performance.
    \item RMS normalization is commonly used than layer normalization due to its reduced computational demands.
\end{enumerate}
A basic introduction to these options is provided in Section \ref{construction}.
Apart from component choices, there are notable innovations in architecture for SLMs:
\begin{itemize}[leftmargin=*]
    \item Mobilellm \cite{liu2024mobilellm} highlights that deeper models are more effective than wider ones for improving performance.
    \item 
    Embedding sharing \cite{zhang2022opt} is crucial as embedding layers often constitute over 20\% of a model's parameters—for example, with 512 dimensions and a 32k vocabulary, each layer holds 16M parameters in a 125M-parameter model. Smaller models often reuse these weights for both input and output layers, enhancing efficiency and compactness.
    \item Layer sharing \cite{liu2024mobilellm} increases hidden layers in small Transformer models without additional storage costs. 
    \item Shared FFNs \cite{thawakar2024mobillama} make up about 65\% of all trainable parameters, with attention mechanisms and heads accounting for the rest. Sharing FFN parameters across all transformer layers of an SLM is proposed to increase efficiency.
    \item Architecture search ahead of pre-training. PhoneLM \cite{yi2024phonelm} proposes a principle for constructing on-device small language models: searching for a resource-efficient architecture on a given hardware to optimize the speed-capacity trade-off before pretraining. This approach inspires the tailored selection of architectural components for on-device SLMs, based on specific compositional requirements such as computing efficiency, model capability, and safety.
\end{itemize}
A detailed description of these architectural designs can be found in Section \ref{Training4SLMFromScratch}.


\subsubsection{Training Datasets}
From Table \ref{tab:generic_small_models_overview_training}, we can observe a set of widely used training datasets in SLM development. We provide the details below:
\begin{itemize}[leftmargin=*]
    \item \textbf{Pile} \cite{gao2020pile}: It comprises 22 smaller, high-quality diverse corpora from various domains, such as Pile-CC, PubMed Central, ArXiv, GitHub, and FreeLaw, designed to offer a comprehensive foundation for language model training. The dataset contains 207 billion tokens and totals 825 GB. 
    \item \textbf{C4 (Colossal Clean Crawled Corpus)} \cite{raffel2020exploring}: This dataset includes 350 billion tokens, representing a cleaned version of the Common Crawl web corpus, intended to capture a wide snapshot of the internet~\footnote{Available at https://commoncrawl.org}.
    \item \textbf{The Stack} \cite{kocetkov2022stack}: It contains 6 trillion tokens of source code from over 300 programming languages, useful for developing code-centric AI applications. \textbf{Python-edu} in smollm-corpus \cite{benallal2024smollmcorpus} consists of Python files that are scored 4 or more by the educational code model and are extracted from the stack-v2-train dataset.
    \item \textbf{StarCoder} \cite{li2023starcodersourceyou}: It features 35 billion tokens, predominantly Python code, aimed at programming language understanding and generation.
    \item \textbf{RedPajama} \cite{together2023redpajama}: This dataset encompasses 1.2 trillion tokens derived from over 100 billion text documents, processed using the CCNet pipeline to ensure a rich collection of web texts.
    \item \textbf{RefinedWeb} \cite{penedo2023refinedweb}: This dataset includes 5 trillion tokens of high-quality, extensively filtered web data, offering a valuable resource for training web-aware models.
    \item \textbf{PushShift.io Reddit} \cite{baumgartner2020pushshift}: A around 5 billion tokens resource for social media data collection, analysis, and archiving, specifically of Reddit data, aiding research into social media dynamics.
    \item \textbf{CulturaX} \cite{nguyen2024culturax}: It comprises 6.3 trillion tokens across 167 languages, supporting the development of models with extensive linguistic and cultural understanding.
    \item \textbf{FineWeb} \cite{penedo2024finewebdatasetsdecantingweb}, a large-scale (15-trillion tokens, 44 TB disk space) dataset for LLM pretraining. FineWeb is derived from 96 CommonCrawl snapshots. \textbf{FineWeb-Edu} is a subset of FineWeb constructed using scalable automated high-quality annotations for educational value. 
\end{itemize}
From the analysis of these datasets, we can derive several critical insights regarding the development of SLMs: 
(i) Data quality is crucial for training effective SLMs, involving sophisticated filtering like removing duplicates or irrelevant content, often with another LLM's help. For example, the TinyStories corpus \cite{eldan2023tinystories} is tailored for simplicity, ideal for training models to handle straightforward narratives. RedPajama-V2 \cite{together2023redpajama} uses the CCNet pipeline to process 30B documents, providing quality signals and IDs for creating a 20B deduplicated dataset.
(ii) Code Data: Source code constitutes a significant component of valuable data for training models, particularly because of its structured nature and logical content. Training on code data enhances a model’s reasoning capabilities and supports its ability to generalize across multiple natural languages, which is crucial for applications requiring robust problem-solving and interpretation skills in diverse coding environments \cite{ma2024at, aryabumi2024code, fu2022gptroadmap, gunasekar2023textbooksneed}




\subsubsection{Training Algorithms}
To enhance the alignment of SLMs with desirable properties such as safety and reasoning, training algorithms, particularly during the fine-tuning phase, are crucial in evolving pre-trained SLMs.
\begin{itemize}[leftmargin=*]
\item 
\textbf{Direct Preference Optimization (DPO)}~\cite{rafailov2024direct} presents a simpler alternative to RLHF for optimizing language models based on human preferences, preventing explicit reward modeling and reinforcement learning. Instead, DPO modifies log probabilities of responses with a dynamic weighting mechanism to prevent model degradation common in probability ratio-focused methods. The DPO loss function is:
\[
\mathcal{L}_{DPO} (\pi_{\theta}; \pi_{\text{ref}}) = -\mathbb{E}_{(x, y_w, y_l) \sim D} \left[ \log \sigma \left( \beta \log \frac{\pi_{\theta}(y_w | x)}{\pi_{\text{ref}}(y_w | x)} - \beta \log \frac{\pi_{\theta}(y_l | x)}{\pi_{\text{ref}}(y_l | x)} \right) \right]
\]
where \( \pi_{\theta} \) is the policy being optimized, \( \pi_{\text{ref}} \) is the reference policy, \( D \) includes tuples \( (x, y_w, y_l) \), \( \sigma \) is the sigmoid function, and \( \beta \) scales the log-ratios between \( \pi_{\theta} \) and \( \pi_{\text{ref}} \), guiding the model towards human-preferred outputs.

\item 
\textbf{Reinforcement Learning from Contrast Distillation (RLCD)}~\cite{yang2024rlcd} aims to calibrate generative SLMs/LLMs towards embodying harmless and beneficial characteristics. The process starts with an unaligned LM and initial prompts, which are modified into two variants \( p+ \) and \( p- \), intended to promote and suppress, respectively, attributes like helpfulness and harmlessness. Upon inputting these prompts, the LM generates outputs \( o+ \) and \( o- \), with \( o+ \) automatically designated as the preferred response. This automation speeds up training by avoiding additional evaluative scoring. The training continues under the RLHF framework.
\item \textbf{Conditioned Reinforcement Learning Fine-Tuning (C-RLFT)}, by OpenChat \cite{wang2024openchat}, enhances model performance by incorporating low-quality data during SFT. C-RLFT leverages varied data qualities with simple rewards (e.g., expert data at 1 credit, sub-optimal at 0.1), using distinct prompt tokens to condition data sources, eliminating costly human feedback. Similarly, Data Mix \cite{pfeiffer2024h2o} trains on English text in three stages, reducing noisy web data progressively in each stage in favor of higher-quality data.

\item \textbf{Explanation Tuning}, proposed by Orca \cite{mukherjee2023orca}, addresses the limitations of standard instruction-based fine-tuning, which often restricts SLMs to style imitation rather than reasoning. It uses system prompts with instructions to direct GPT-4 to produce detailed explanations or perform step-by-step reasoning. The resulting instructions and the responses are used as a dataset for fine-tuning SLMs to have better ability of reasoning.
\item \textbf{Progressive Learning}, proposed by Orca \cite{mukherjee2023orca}, aims to bridge the capability gap between Orca and the more capable GPT-4. It starts with training on five million data points from ChatGPT, followed by one million from GPT-4. Research suggests that an intermediate-level teacher can improve distillation effects, enabling a stepwise learning approach where students start with simpler examples and gradually move to more complex ones, receiving improved reasoning and explanations from a more advanced teacher.
\item \textbf{Prompt Erasing} introduced by Orac 2 \cite{mitra2023orca}, is a distillation strategy designed to enhance the independent reasoning capabilities of student SLMs. In this approach, a more capable teacher LLM is given intricate prompts intended to elicit specific strategic behaviors and more precise outcomes. During the training phase, the SLM is exposed only to the task instruction and the resulting behavior, without access to the original intricate prompts that initiate such responses. This technique, known as Prompt Erasing, positions the student model as a cautious reasoner because it not only learns to perform specific reasoning steps but also develops strategies for approaching tasks at a higher level.
\item \textbf{Maximal Update Parameterization ($\mu$P)} optimizes control initialization, layer-wise learning rates, and activation magnitudes to ensure stable training regardless of model layer widths. This method enhances training stability and allows the same optimizer settings, especially learning rates, to be used across different model scales. For instance, Cerebras-GPT \cite{cerebras2023slimpajama} employs $\mu$P to train its models efficiently.
\end{itemize}

\subsubsection{Model Performance}

\begin{table}[!t]
\centering
\caption{Performance of Various SLMs on Common Benchmarks: data from \textbf{MobiLlama} \cite{thawakar2024mobillama}, \textbf{OLMo} \cite{groeneveld2024olmo}, and \textbf{Llama 3.2}. }
\label{tab:merged_performance}
\vskip -1em
\small
\scalebox{0.9}{
\begin{tabular}{@{}clcccccc@{}}
\toprule
\textbf{Model Size Range} & \textbf{Model} & \textbf{MMLU} & \textbf{HellaSwag} & \textbf{ARC} & \textbf{PIQA} & \textbf{Winogrande} \\ \midrule
\multirow{12}{*}{<1B} 
& gpt-neo-125m        & 26.0 & 30.3 & 23.0 & 62.5 & 51.8 \\
& tiny-starcoder-170M     & 26.8 & 28.2 & 21.0 & 52.6 & 51.2 \\
& cerberas-gpt-256m   & 26.8 & 29.0 & 22.0 & 61.4 & 52.5 \\
& opt-350m            & 26.0 & 36.7 & 23.6 & 64.7 & 52.6 \\
& megatron-gpt2-345m  & 24.3 & 39.2 & 24.2 & 66.9 & 53.0 \\
& LiteLlama           & 26.2 & 38.5 & 24.9 & 67.7 & 49.9 \\
& gpt-sw3-356m        & 25.9 & 37.1 & 23.6 & 64.9 & 53.0 \\
& pythia-410m         & 27.3 & 40.9 & 26.2 & 67.2 & 53.1 \\
& xglm-564m           & 25.2 & 34.6 & 24.6 & 64.9 & 53.0 \\
& Lamini-GPT-LM 0.59B     & 25.5 & 31.6 & 24.2 & 63.9 & 47.8 \\
& MobiLlama 0.5B      & 26.5 & 52.5 & 29.5 & 72.0 & 57.5 \\
& MobiLlama 0.8B      & 26.9 & 54.1 & 30.2 & 73.2 & 57.5 \\
\midrule
\multirow{18}{*}{1B-3B} 
& StableLM 1.6B       & - & 68.2 & 43.8 & 74.0 & - \\
& Pythia 1B           & - & 44.7 & 33.1 & 69.1 & - \\
& TinyLlama 1.1B      & - & 58.7 & 34.8 & 71.1 & - \\
& OLMo-1B             & - & 62.5 & 34.5 & 73.7 & - \\
& OLMo 1.2B           & 25.9 & 62.5 & 34.5 & - & 58.9 \\
& Boomer 1B           & 25.4 & 31.6 & 22.3 & - & 51.0 \\
& Pythia-Dedup 1B     & 24.3 & 49.6 & 29.1 & - & 54.0 \\
& Falcon-RW 1B        & 25.4 & 63.1 & 35.1 & - & 61.9 \\
& Cerebras-GPT 1.3B   & 26.7 & 38.5 & 26.1 & - & 53.6 \\
& Lamini 1.3B         & 28.5 & 38.1 & 26.6 & - & 50.6 \\
& OPT 1.3B            & 24.6 & 54.5 & 29.6 & - & 59.7 \\
& GPT-NEO 1.3B        & 24.8 & 48.5 & 31.3 & - & 57.1 \\
& Pythia-Deduped 1.4B & 25.5 & 55.0 & 32.6 & - & 56.9 \\
& MobiLlama 1.2B      & 24.8 & 63.0 & 34.6 & - & 62.0 \\
& Gemma 2 2B          & 57.8 & 61.1 & 76.7 & - & - \\
& Llama 3.2 1B        & 49.3 & 41.2 & 59.4 & - & - \\
& Llama 3.2 3B        & 63.4 & 69.8 & 78.6 & - & - \\
\midrule
\multirow{8}{*}{>3B} 
& Phi-3.5-mini 3.8B   & 69.0 & 81.4 & 87.4 & - & - \\
& Pythia 6.9B         & - & 63.8 & 44.1 & 75.1 & - \\
& Falcon-7B           & - & 75.9 & 47.5 & 78.5 & - \\
& LLaMA 7B            & - & 76.2 & 44.5 & 77.2 & - \\
& Llama 2 7B          & - & 76.8 & 48.5 & 76.7 & - \\
& MPT-7B              & - & 77.6 & 46.5 & 77.3 & - \\
& RPJ-INCITE-7B       & - & 70.3 & 42.8 & 76.0 & - \\
& OLMo-7B             & - & 76.4 & 48.5 & 78.4 & - \\
\bottomrule
\end{tabular}
}
\end{table}

\begin{table}[!h]
\centering
\footnotesize
\caption{Comparison of MobiLlama 0.5B with large-base 1.2B, Llama2 7B, and Phi2-2.7B in terms of efficiency and resource consumption on low-end hardware devices \cite{thawakar2024mobillama}.}
\label{tab:efficiency_comparison}
\vskip -1em
\begin{tabular}{lccC{1cm}C{1.5cm}C{2cm}C{2cm}C{1.4cm}}
\hline
\textbf{Platform} & \textbf{Model} & \textbf{\#Params} & \textbf{Precision} & \textbf{Avg Tokens/Sec} & \textbf{Avg Memory Consumption} & \textbf{Avg Battery Consumption /1k Tokens} & \textbf{CPU Utilization} \\ \hline
\multirow{4}{*}{RTX2080Ti} 
&Llama2 & 7B & bf16 & 14.85 & 27793 MB & 135.51 mAH & 31.62\% \\
&Phi2 & 2.7B & bf16 & 32.19 & 12071 MB & 59.13 mAH & 24.73\% \\
&large-base & 1.2B & bf16 & 50.61 & 6254 MB & 18.91 mAH & 18.25\% \\
&MobiLlama & 0.5B & bf16 & 63.38 & 3046 MB & 8.19 mAH & 14.79\% \\ \hline
\multirow{4}{*}{CPU-i7} 
&Llama2 & 7B & 4bit & 5.96 & 4188 MB & 73.5 mAH & 49.16\% \\
&Phi2 & 2.7B & 4bit & 22.14 & 1972 MB & 27.36 mAH & 34.92\% \\
&large-base & 1.2B & 4bit & 29.23 & 1163 MB & 10.81 mAH & 30.84\% \\
&MobiLlama & 0.5B & 4bit & 36.32 & 799 MB & 4.86 mAH & 24.64\% \\ \hline
\multirow{4}{*}{Snapdragon-685} 
&Llama2 & 7B & 4bit & 1.193 & 4287 MB & 10.07 mAH & 77.41\% \\
&Phi2 & 2.7B & 4bit & 2.882 & 1893 MB & 14.61 mAH & 56.82\% \\
&large-base & 1.2B & 4bit & 6.687 & 780 MB & 6.00 mAH & 17.15\% \\
&MobiLlama & 0.5B & 4bit & 7.021 & 770 MB & 5.32 mAH & 13.02\% \\
\hline
\end{tabular}
\end{table}

To compare the performance of SLMs, we have extracted experimental results from two recent and concurrent studies published in June 2024, \textbf{OLMo} \cite{groeneveld2024olmo} and \textbf{MobiLlama} \cite{thawakar2024mobillama}, and the recently proposed edge-device \textbf{Llama 3.2 1B \& 3B} in September 2024~\footnote{https://ai.meta.com/blog/llama-3-2-connect-2024-vision-edge-mobile-devices/}. The extracted results are merged and shown in Table \ref{tab:merged_performance}. From the table, we can find that the following evaluation benchmarks are commonly used: 
\begin{enumerate}[leftmargin=*]
\setlength{\itemsep}{0pt}
\setlength{\parskip}{0pt}
\setlength{\parsep}{0pt}
    \item \textbf{MMLU}~\cite{hendrycksmeasuring}: Evaluate broad knowledge across diverse fields such as humanities, science, technology, engineering, and management. It includes multiple-choice questions covering 57 tasks ranging from elementary mathematics to US history, computer science, law, and beyond, with a total of 14K test samples.
    \item \textbf{HellaSwag}~\cite{zellers2019hellaswag}: Assesses the model's ability to select the correct ending to scenarios from multiple options, testing common sense reasoning, including 10K test samples.
    \item \textbf{ARC}~\cite{clark2018think}: The AI2’s Reasoning Challenge (ARC) dataset features multiple-choice science exam questions for grades 3 to 9, divided into Easy and Challenge partitions, with the latter containing more complex questions necessitating reasoning. Most questions offer four answer choices. ARC includes a supporting knowledge base of 14.3M unstructured text passages, with 1.17K test samples in ARC\_Challenge and 2.25K in ARC\_Easy.
    \item \textbf{PIQA}~\cite{bisk2020piqa}: A commonsense reasoning dataset designed to evaluate the physical knowledge of NLP models. It presents questions (goals) that require physical commonsense for correct resolution, alongside two detailed response options (sol1 and sol2). The dataset comprises 3,000 test samples.
    \item \textbf{Winogrande}~\cite{sakaguchi2021winogrande}: a dataset structured as a fill-in-the-blank task with binary options, designed to assess commonsense reasoning. The dataset includes 1,767 test samples by default splits.
\end{enumerate}
Accuracy is used as the evaluation metric in the table.
\textbf{Open Language Model (OLMo)} \cite{groeneveld2024olmo} is publicly available with its training data and code \footnote{https://allenai.org/olmo}.
\textbf{MobiLlama} \cite{thawakar2024mobillama} is a general-purpose SLM designed from scratch, available in 0.5B and 0.8B versions. It adopts a unique approach by using a shared FFN across all transformer blocks, enhancing efficiency. 
\textbf{MobiLlama} also show high efficiency on diverse hardware (Table \ref{tab:efficiency_comparison}). 

From Table \ref{tab:merged_performance} and \ref{tab:efficiency_comparison},
we can conclude that: 
(1) \textbf{MobiLlama} 0.5B and 0.8B demonstrate that a shared FFN design can facilitate excellent performance in SLMs with fewer than 1B parameters, even rivaling some models in the 1B-3B range. 
(2) The performance of \textbf{MobiLlama} 1.2B and \textbf{OLMo} 1.2B illustrates that advanced SLM architectures incorporating high-quality data, SwiGLU, non-parametric layer normalization, RoPE, BPE tokenization, and a shared FFN design can achieve competitive results among models with 1B-3B parameters. 
(3) \textbf{MobiLlama} demonstrates that SLMs can significantly reduce resource consumption on low-end hardware devices, achieving comparable performance while using a smaller proportion of battery power, memory, and GPU utilization. 
(4) Popular techniques such as pruning, quantization, distillation, SFT, and DPO, utilized in \textbf{Llama 3.2}, have substantially enhanced SLM performance.

\begin{takeaway2}
\small
\textbf{Insights}: We draw several key insights from the development of generic-domain SLMs:
\begin{itemize}[leftmargin=*]
\setlength{\itemsep}{0pt}
\setlength{\parskip}{0pt}
\setlength{\parsep}{0pt}
    \item Typical SLM architectures generally incorporate features such as GQA, gated FFN with SiLU activations, RMS normalization, deep and thin architectures, embedding sharing, layer sharing, and shared FFNs.
    \item Although these components are widely used, current research has not yet thoroughly explored their specific contributions within SLMs.
    \item The importance of data quality in SLM research is increasingly emphasized, often considered more critical than the quantity of data and model architectural configurations. 
    \item Post-pretraining, meticulous fine-tuning is often required to enhance the safety of SLMs, involving strategies to distill capabilities from LLMs better. Common strategies include explanatory tuning, progressive learning, and prompt erasing. 
\end{itemize}
\end{takeaway2}

\begin{table*}[!htb]
\caption{High-level Overview and Training Details of Specific-domain Small Language Models}
\label{tab:specific_small_models_overview_training}
\vskip -1em
\centering
\footnotesize
\renewcommand{\arraystretch}{1.5} 
\begin{tabular}{p{2cm}|p{1cm}|p{0.8cm}|p{2.6cm}|p{1.5cm}|p{2cm}|p{2.6cm}}
\hline
\textbf{Model} & \textbf{\#Params} & \textbf{Date} & \textbf{Base Models} & \textbf{Domain} & \textbf{Training Datasets} & \textbf{Train Techniques} \\
\hline
Hippocrates \cite{acikgoz2024hippocrates} & 7B & 2024.4 & Instruction Tuning (LLaMA2~\cite{touvron2023llama2}, Mistral \cite{jiang2023mistral}) & Healthcare & Medical Guidelines, PMC-Patients~\cite{zhao2022pmc}, and PubMedQA-contexts~\cite{jin2019pubmedqa} & Continual pre-training, instruction tuning, RLHF \\
\hline
BioMedLM \cite{bolton2024biomedlm} & 2.7B & 2024.3 & From scratch and Fine-tuning & Healthcare & PubMed \cite{gao2020pile} & FSDP \\
\hline
BioMistral \cite{labrak2024biomistral} & 7B & 2024.2 & Mistral \cite{jiang2023mistral} & Biomedicine &  PubMed \cite{gao2020pile} & Continual pretraining \\
\hline
MentaLLaMA \cite{yang2024mentallama} & 7B; 13B & 2023.9 & Instruction Tuning (LLaMA2~\cite{touvron2023llama2}) & Healthcare & IMHI dataset & RLHF; PEFT \\
\hline
AdaLM \cite{yao2021adapt} & 34M & 2021.6 & Distillation (BERT \cite{devlin2019bert} or MiniLM \cite{wang2020minilmdeepselfattentiondistillation}) & Healthcare & PubMed \cite{gao2020pile} & Continual pretraining, Adapt-and-Distill \\
\hline\hline

Rho-1 \cite{lin2024rho} & 1B; 7B & 2024.4 & TinyLlama-1.1B \cite{zhang2024tinyllamaopensourcesmalllanguage}, Mistral-7B \cite{jiang2023mistral} & Science (Mathematics) & OpenWebMath \cite{paster2024openwebmath} & Continual pretraining \\
\hline
ChemLLM \cite{zhang2024chemllm} & 7B & 2024.4 & Instruction Tuning (InternLM2) & Science (Chemistry) & ChemData & Continual training and fine-tuning\\
\hline
SciGLM \cite{zhang2024sciglm} & 6B & 2024.3 & Instruction Tuning (ChatGLM-6B) & Science & SciInstruct & Self-reflective instruction annotation \\
\hline
Llemma \cite{azerbayev2023llemma} & 7B & 2023.10 & Code Llama 7B & Science (Mathematics) & Proof-Pile-2 \cite{azerbayev2023llemma} & Continual pre-training \\
\hline
OceanGPT \cite{bi2023oceangpt} & 2B; 7B; 14B & 2023.10 & LLaMA2~\cite{touvron2023llama2} & Science (Ocean) & Open-access literature, DoINSTRUCT & Continual pre-training, Instruction tuning \\
\hline
AstroLLaMA \cite{nguyen2023astrollama} & 7B & 2023.9 & Tuning (LLaMA-2-7B) & Science (Astronomy) & arXiv abstracts from Kaggle & Continual training \\
\hline
DARWIN \cite{xie2023darwin} & 7B & 2023.8 & LLaMa 7B & Science (physics, chemistry, and material) & SciQ \cite{welbl2017crowdsourcing}, Scientific paper\cite{xie2023darwin},  FAIR \cite{xie2023darwin} & Fine-tuning \\
\hline\hline

MindLLM \cite{yang2023mindllm} & 1.3B; 3B & 2023.10 & From-scratch and Supervised Fine-tuning & Law, Finance & Pile \cite{gao2020pile}, Wudao \cite{yuan2021wudaocorpora}, CBooks & Train on Bilingual Mixture Data, SFT \\
\hline

\end{tabular}
\end{table*}


\subsection{Domain-Specific SLMs}
\label{specific_slms}
\paragraph{Overview} 
The capability of LLMs to generate human-like text has significantly captured public interest, highlighting their potential in the field of general artificial intelligence. However, inefficiencies persist when integrating LLMs into specialized applications due to resource constraints. Unlike the need for extensive general knowledge and capabilities, domain-specific SLMs should focus on well-defined tasks and expertise pertinent to specific fields. For instance, specialized models can significantly impact biomedical research and healthcare by fine-tuning for interpretable mental health analysis, or assisting humans in legal dialogues and financial tasks through instruction tuning, showcasing their potential transformative influence. Given the limited number of SLMs specialized in specific domains, we demonstrate some existing SLMs individually across healthcare, science, finance, and law domains.

\subsubsection{SLMs for Healthcare}

\textbf{Hippocrates} \cite{acikgoz2024hippocrates} is an open-source medical language model framework with free access to its data, codebase, checkpoints, and protocols~\footnote{https://cyberiada.github.io/Hippocrates/}.  It utilizes a medical pre-training corpus from Medical Guidelines, PMC-Patients~\cite{zhao2022pmc}, and PubMedQA-contexts~\cite{jin2019pubmedqa}, totaling about 300M tokens. The Hippo series, a 7B model, undergoes continuous pre-training, instruction tuning, and RLHF. Fine-tuned on Mistral and Llama-2, it rivals 70B models in some evaluation. For example, Hippo-Mistral 7B scores 59.9\% on MedQA, outperforming Meditron 70B~\cite{chen2023meditron} at 58.5\%. 
\textbf{BioMedLM} \cite{bolton2024biomedlm}, a 2.7B GPT-style model trained on PubMed content~\cite{gao2020pile}, excels in biomedical QA after fine-tuning, achieving 57.3\% on MedMCQA (dev) and 69.0\% on MMLU medical genetics exams. Available on Hugging Face Hub~\footnote{https://huggingface.co/stanford-crfm/BioMedLM}.
\textbf{AdaLM} \cite{yao2021adapt} enhances domain-specific SLMs by continuing training on a medical-focused SLM atop a general pre-trained model. It emperically validates adaptation then distillation is the most effective distillation way. AdaLM modified a BERT\_base model (12 layers, 768 hidden size) \cite{devlin2019bert} with a 16GB PubMed~\footnote{https://pubmed.ncbi.nlm.nih.gov/} abstracts corpus. 
\textbf{MentalLLaMA} \cite{yang2024mentallama} introduces the first IMHI dataset for mental health analysis and the first open-source LM for explainable analysis on social media. The IMHI is compiled from ten sources, totaling 105K samples. Expert-designed mental health analysis prompts are employed via ChatGPT for explanations. Based on Llama-2-7B, MentalLLaMA is instruction-tuned on this data and matches top methods in accuracy on the IMHI test set. Project code is available at~\footnote{https://github.com/SteveKGYang/MentaLLaMA}.

\subsubsection{SLMs for Science}
\textbf{SciGLM} \cite{zhang2024sciglm} is a collegiate-level scientific language model overcoming data scarcity with a self-reflective instruction annotation framework. Utilizing GPT-4~\cite{achiam2023gpt}, it generates detailed reasoning for unlabeled scientific problems through three steps with designed prompts in Table \ref{tab:prompts_self_reflective}: (i) CoT prompt for step-by-step answers (Prompt 1), (ii) reflective prompt for correcting errors (Prompt 2), and (iii) integrating the correct answer for clarity (Prompt 3). The SciInstruct dataset spans physics, chemistry, math, and proofs, tuning ChatGLM-6B's~\cite{du2022glm} reasoning abilities. SciGLM boosts the base model’s (ChatGLM3-6B-Base) scientific QA accuracy by 3.06\% on benchmarks such as CEval-Hard~\cite{huang2024c}, CEval-Sci~\cite{huang2024c}, MMLU-Sci~\cite{hendrycksmeasuring}, SciEval~\cite{sun2024scieval}, and SciBench~\cite{wang2024scibench}.
\textbf{Llemma} \cite{azerbayev2023llemma}, an SLM derived from CodeLlama \cite{roziere2023code}, specializes in mathematical reasoning. By continual pre-training, its 7B model is evolved on 55B tokens from the newly created Proof-Pile-2 dataset, which includes scientific papers, math web content, and mathematical code up until April 2023, to enhance few-shot capabilities. It excels in mathematical benchmarks like MATH \cite{hendrycksmeasuring}, GSM8k \cite{cobbe2021training}, OCWCourses \cite{lewkowycz2022solving}, MMLU-STEM \cite{hendrycksmeasuring}, and SAT, surpassing all comparable open-weight models.
\textbf{ChemLLM} \cite{zhang2024chemllm} is a chemistry-focused language model that utilizes its proposed ChemData, a dataset designed for instruction tuning that transforms chemical data into dialogue format for training. ChemLLM is based on InternLM2-Base-7B~\cite{cai2024internlm2}, initially enhancing its language skills with a multi-corpus of 1.7 million Q\&A pairs from Hugging Face, then fine-tunes using ChemData and the multi-corpus to maintain its general capabilities. 
ChemLLM excels in interdisciplinary chemical tasks within the proposed ChemBench, achieving results comparable to GPT-4~\cite{achiam2023gpt} and outperforming GPT-3.5 with a score of 92.6 in Mol2caption, slightly below that of GPT-4.
\textbf{AstroLLaMA} \cite{nguyen2023astrollama} introduces an astronomy-focused language model. Based on Llama-2-7B~\cite{touvron2023llama2} and enhanced via continual pre-training, it has been developed using over 300K astronomy abstracts from arXiv~\footnote{https://www.kaggle.com/Cornell-University/arxiv}. AstroLLaMA achieves 30\% lower perplexity than Llama-2-7B, indicating substantial improvements in domain adaptability. AstroLLaMA is available \footnote{https://huggingface.co/universeTBD/astrollama} for tasks such as automated paper summarization and conversational agent development in astronomy.

\begin{table}[!t]
    \label{tab:prompts_self_reflective}
    \centering
    \small
    \caption{Prompts for self-reflective instruction annotation framework}
    \begin{tabular}{l|p{11cm}}
    \hline
         Chain-of-Thought &[Prompt 1] The following input consists of a science problem, please generate an elaborate step-by-step solution to the problem.\\
    \hline
         Reflective Generation &[Prompt 2] The following input comprises a science problem and a corresponding solution. However, this solution is incorrect, please reflect on its errors and then generate a correct step-by-step solution to the problem. \\
    \hline
         Prompt Answer &[Prompt 3] The following input consists of a science problem, a corresponding solution, and the real answer. The given solution is incorrect, please reflect on its errors and then generate a correct step-by-step solution to the problem based on the real answer.\\
         \hline
    \end{tabular}
    \vskip -1em
\end{table}

\subsubsection{SLMs for Finance and Law}
\textbf{MindLLM} \cite{yang2023mindllm} introduces a bilingual (Chinese and English) SLM, pretrained on the Pile dataset \cite{gao2020pile} for English and WuDao \cite{yuan2021wudaocorpora}, CBook, and various Chinese web content for Chinese. Bilingual training enhances capacity and prevents catastrophic forgetting. It explores specific domains such as law and finance through supervised fine-tuning. In law, it utilizes publicly available legal data, scenario-based Q\&A from LaW-GPT \cite{LAWGPT-zh}, and NLP-based legal tasks from DISC-LawLLM \cite{yue2023disc}. In finance, EastMoney \footnote{https://www.eastmoney.com/default.html} is selected as the data source.


\begin{takeaway2}
\small
\textbf{Insights:} We draw several key insights from the development of domain-specific SLMs:
\begin{itemize}[leftmargin=*]
\setlength{\itemsep}{0pt}
\setlength{\parskip}{0pt}
\setlength{\parsep}{0pt}
  \item Adapting SLMs to domain-specific data is a common practice for acquiring domain-specific SLMs, prompting many to create their datasets \cite{yang2024mentallama, nguyen2023astrollama, zhang2024chemllm, zhang2024sciglm}. These datasets are often annotated using LLMs like GPT-4 and used to continual pre-train or fine-tune general models such as LLaMa-2-7B \cite{acikgoz2024hippocrates, bolton2024biomedlm}. To ensure the data quality, specialized annotation frameworks are developed, such as SciGLM \cite{zhang2024sciglm}.
  \item In domains with abundant corpora, training a general model from scratch and fine-tuning it using SFT \cite{yang2023mindllm} is practical. Bilingual settings during training can prevent catastrophic forgetting.
  \item Distilling general capabilities from LLMs while integrating domain-specific knowledge from corpora is another method for developing domain-specific SLMs \cite{yao2021adapt}.
\end{itemize}
\end{takeaway2}

%% file: sections/7.slm4llm.tex
\begin{table}[!ht]
\centering
\small
\caption{SLMs help LLMs in different aspects}
\label{tab2:slm4llm}
\vskip -1em
\begin{tabularx}{\textwidth}{>{\hsize=0.35\hsize}X|>{\hsize=0.6\hsize}X|>{\hsize=1.2\hsize}X}
\hline
\textbf{Aspect} & \textbf{Representative work} & \textbf{Key point} \\
\hline

\multirow{13}{=}{\textbf{SLM for reliable LLM generations}} & APRICOT \cite{ulmer2024calibrating} & Trains a small auxiliary model to predict LLM's confidence using only textual inputs and outputs. \\
\cline{2-3}
& POLAR \cite{zhao2023automatic} & Using a BERT model to calibrate LLM responses. \\
\cline{2-3}
& Hallucination Detector in NMT \cite{xu2023understanding} & Using lightweight classifiers to detect hallucinations in Neural Machine Translation. \\
\cline{2-3}
& SAPLMA \cite{azaria2023internal} & Using a BERT Small Language Model as a classifier to assess the truthfulness of statements accurately. \\
\cline{2-3}
& Question Decomposer \cite{wu2024divide} & Distilled SLM decomposes complex questions to aid reasoning. \\
\cline{2-3}
& SuperICL \cite{xu2023small} & SLM Plug-ins provide confidence and prediction for contextual exemplars to aid in-context learning. \\
\cline{2-3}
& SuperContext \cite{yang2024supervised} & Specific SLM enhances ICL by providing confidence and predictions to overcome out-of-domain challenges. \\
\cline{2-3}
 & Self-RAG \cite{asai2024selfrag} & A proxy model labels special tokens during RAG data generation for fine-tuning. \\
\cline{2-3}
& SKR \cite{wang2023self} & Training a small model to detect its self-knowledge for better use of external knowledge. \\
\cline{2-3}
& SlimPLM \cite{tan2024small} & Detecting missing knowledge in LLMs with a slim proxy model, enhancing the LLM’s knowledge integration. \\
\cline{2-3}
& In-Context RALM \cite{ram2023context} & Training a RoBERTa-based reranker for top-k BM25 documents using LM signals to enhance LM gains. \\
\cline{2-3}
& CRAG \cite{yan2024corrective} & Training a lightweight evaluator to assess document quality and trigger actions based on confidence levels. \\
\cline{2-3}
& GSR \cite{huang-etal-2024-less} & Training a Generative Sub-graph Retriever (GSR) for relation chain in RAG when retrieving from knowledge graphs. \\
\hline

\multirow{4}{=}{\textbf{SLM for extracting LLM prompts}} & Prompt Extraction \cite{zhang2024effectivepromptextractionlanguage} & Small model trained to predict confidence of extracted system prompt from adversarial prompts. \\
\cline{2-3}
& Prompt Stealing Attacks \cite{sha2024prompt} & Using small models fine-tuned as parameter extractors to facilitate hierarchical prompt reconstruction. \\
\cline{2-3}
& Output2prompt \cite{zhang2024extracting} & Using a sparse encoder-decoder-based T5 small model to reverse-engineer LLM inputs from outputs. \\
\cline{2-3}
& Model Purifying \cite{li2024purifying} & Using SLMs to ensemble with LLMs, mitigating negative effects from uncurated data. \\
\hline

\multirow{4}{=}{\textbf{SLM for Fine-tuning LLMs}} & $\mathcal{LP}$ \cite{mekala2024smaller} & Learning Percentage as a difficulty metric. \\
\cline{2-3}
& Emulated Fine-tuning \cite{mitchell2024an} & Emulating pre-training and fine-tuning at different scales by summing base log probabilities with behavior deltas. \\
\cline{2-3}
& CROSSLM \cite{deng2023mutual} & SLMs enhance LLMs by generating task-specific high-quality data. \\
\cline{2-3}
& Weak-to-Strong Search \cite{zhou2024weaktostrong} &  Framing LLM alignment as a test-time greedy search to maximize the log-probability difference between tuned and untuned SLMs. \\
\hline

\multirow{3}{=}{\textbf{SLM for LLM applications}} & SLCoLM \cite{tang2024small} & Using SLM predictions to guide the LLM generation process in Chinese Entity Relation Extraction. \\
\cline{2-3}
& HEF \cite{yang2024enhancing} & Using SLMs as plugins to improve LLM’s nuanced understanding. \\
\cline{2-3}
& Contrastive decoding \cite{li2023contrastive} & Enhancing text quality by maximizing the difference between expert and amateur log probabilities. \\
\hline

\multirow{2}{=}{\textbf{SLM for LLM safety}} & Llama Guard \cite{inan2023llama} & An LLM-based input-output safeguard model geared towards Human-AI conversation use cases. \\
\cline{2-3}
& SLM as Guardian \cite{kwon-etal-2024-slm} & A smaller LLM for both harmful query detection and safeguard response generation. \\
\hline

\multirow{4}{=}{\textbf{SLM for LLM evaluation}} 
&  SLIDE \cite{zhao2024slide} & Utilizing SLMs trained via contrast learning to distinguish and score responses in dialogue scenarios effectively.\\
\cline{2-3}
& \citet{kuhn2023semantic} & An SLM is used as the natural language inference classifier.\\
\cline{2-3}
& SelfCheckGPT \cite{manakul-etal-2023-selfcheckgpt} & An SLM is used to calculate BERTScore.\\
\cline{2-3}

& Factscore \cite{min-etal-2023-factscore}& An SLM functions as the natural language inference classifier.\\


\hline
\end{tabularx}
\end{table}

\section{SLMs for LLMs} 
\label{slm4llm}
In this section, we provide a comprehensive review of how SLMs enhance LLMs. While LLMs are robust, they face challenges such as latency during inference, labor-intensive fine-tuning, noise filtration issues in retrieval, suboptimal zero-shot performance, copyright infringement risks, and evaluation difficulties. SLMs can help LLMs to alleviate these issues. Research in this field can be categorized into five primary areas: (i) using SLMs for reliable LLM generation; (ii) extracting prompts for LLMs using SLMs; (iii) fine-tuning LLMs with SLMs; (iv) applying SLMs in LLM applications; (v) utilizing SLMs as guardian; and (vi) evaluating LLMs using SLMs. 
A summary of representative work in each category along with their key point is given in Table \ref{tab2:slm4llm}. Next, we introduce each category in detail.

\subsection{SLM for Reliable LLM Generation}
\label{slm4reliable}
Although LLMs generally produce fluent and convincing text, they can occasionally generate erroneous responses \cite{wan2023multi, ji2023towards}. Additionally, LLMs are susceptible to privacy breaches from untrusted data collection, which can erode user trust or cause harm. To address these issues, recent studies have focused on using SLMs to calibrate LLM confidence, detect hallucinations, and improve retrieval-augmented LLMs and their reasoning capabilities.

\begin{figure}[!htbp]
    \centering
    \begin{subfigure}{0.36\textwidth}
        \centering
        \includegraphics[width=\linewidth]{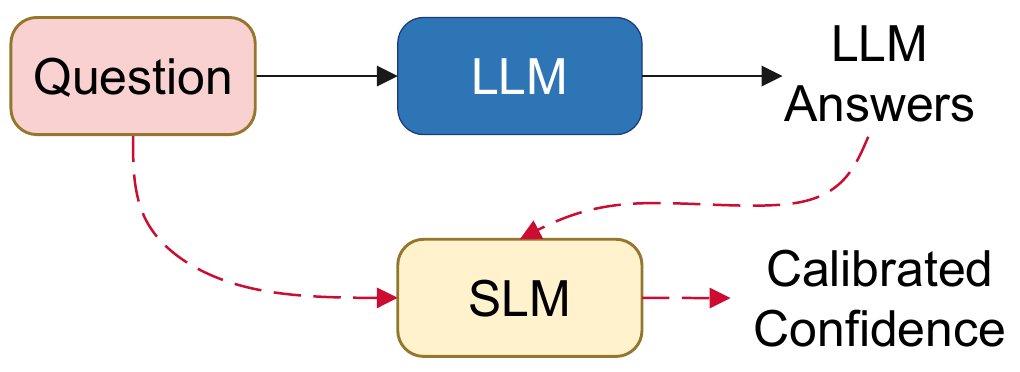}
        \caption{SLM-based Calibrator}
        \label{fig:slm4llm_calibrator}
    \end{subfigure}~~
    \begin{subfigure}{0.36\textwidth}
        \centering
        \includegraphics[width=\linewidth]{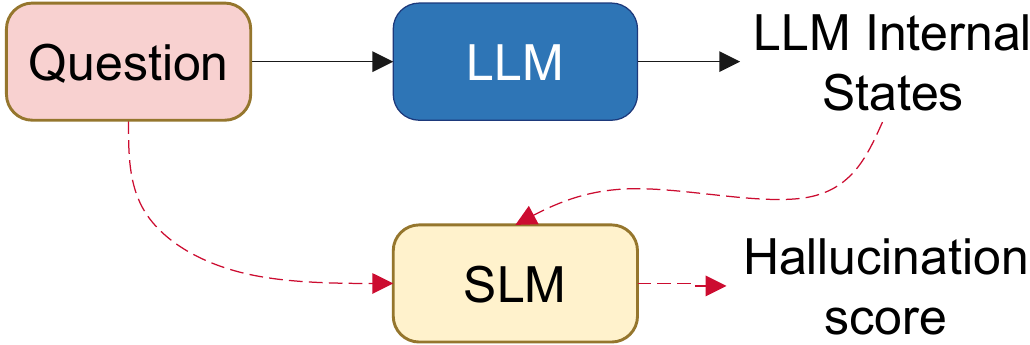}
        \caption{SLM-based Hallucination Detector}
        \label{fig:slm4llm_hallucination_detector}
        \vskip -1em
    \end{subfigure}
    \vskip -1em
    \caption{Architectures of Enhancing Calibration and Hallucination Detection of LLMs.}
    \label{fig:slm4llm_calibrator_hallucination}
\end{figure}
\textbf{Enhancing Calibration and Hallucination Detection of LLMs} 
As illustrated in Figure \ref{fig:slm4llm_calibrator_hallucination} (a), to calibrate LLMs, an SLM processes both questions and LLM-generated answers to predict calibrated confidence. This training involves minimizing the discrepancy between estimated calibration error and predicted confidence score. For instance, \textbf{APRICOT} \cite{ulmer2024calibrating} uses an auxiliary DeBERTaV3 model \cite{hedebertav3} to assess LLM confidence in open-question scenarios, aiming to improve uncertainty expression and response adjustment. Similarly, \textbf{POLAR} \cite{zhao2023automatic} has developed a self-supervised approach that generates risk scores for each response to calibrate LLM confidence, utilizing a small BERT model \cite{devlin2019bert} to synchronize LLM outputs with other weak supervision sources.
As shown in Figure \ref{fig:slm4llm_calibrator_hallucination} (b), for hallucination detection, an SLM analyzes LLM internal states to output the likelihood of hallucination. This process uses supervised data obtained by testing the knowledge boundaries of the LLM. In neural machine translation, \citet{xu2023understanding} develop a lightweight detector that analyzes token contributions to hallucinations, outperforming both model-free baselines and quality estimation classifiers. Furthermore, \textbf{SAPLMA} \cite{azaria2023internal} found that LLM internal states can signal the truthfulness of statements, with a small BERT classifier trained to differentiate correct from incorrect predictions achieving accuracies of 71\% to 83\%.

\begin{wrapfigure}[6]{r}{0.46\textwidth}
    \centering
    \vskip -1em
    \includegraphics[width=\linewidth]{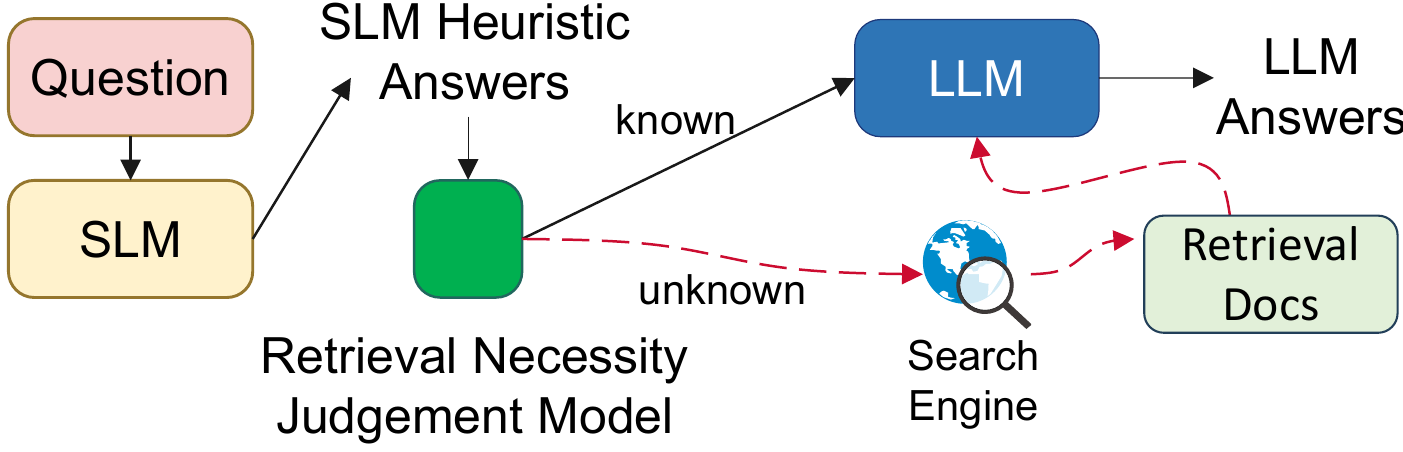}
    \vskip -1em
    \caption{Architecture of SLM as a Heuristic RAG Prober.}
    \label{fig:slm4llm_rag_prober}
\end{wrapfigure}
\textbf{Enhancing Retrieval-Augmented Generation} 
Generally, as shown in Figure \ref{fig:slm4llm_rag_prober}, \textit{SLMs can also serve as proxy models to evaluate the familiarity of LLMs with user queries, determining whether LLMs need to retrieve additional information or can respond directly.}
For example, \textbf{SlimPLM} \cite{tan2024small} is a small proxy model that assesses the necessity for LLM retrieval by generating heuristic answers. High-quality responses indicate that LLMs can handle queries independently, whereas lower-quality outputs require further retrieval.
Additionally, Self-Knowledge Guided Retrieval (\textbf{SKR}) \cite{wang2023self} enables SLMs to autonomously decide when LLMs should operate independently, based on their self-assessment of knowledge limitations. 
Further, \textbf{SELF-RAG} \cite{asai2024selfrag} improves the factual accuracy and quality of LLM outputs through on-demand retrieval and self-reflection. This method employs a small critic language model to issue reflective markers and make binary decisions regarding the need for further information retrieval.
\textit{Moreover, some studies utilize SLMs to evaluate the relevance of retrieved documents.} \textbf{LongLLMLingua} \cite{jiang2024longllmlingua} employs SLMs to calculate the relevance of documents to a query $x^{que}$ using perplexity, as formalized by the equation:
\begin{equation}
    r_k = -\frac{1}{N_c} \sum_{i} \log p_{\text{SLM}}(x^{que}_i | x^{doc}_k), \quad k \in \{1, 2, \ldots, K\}
\end{equation}
where $x^{que}_i$ is the $i$-th token in the query sequence, $x_k^{doc}$ is the retrieved document, and $N_c$ is the total number of tokens in the query. $p_{\text{SLM}}$ represents the probability generated by an SLM. 
\textbf{CRAG} \cite{yan2024corrective} employs SLMs as evaluators of document relevance in the same way. \textbf{RA-ISF} \cite{liu2024ra} trains a small language model that checks the base LLM in self-knowledge, relevance judgment, and question decomposition.
\textit{In addition, some research employs SLMs as re-rankers to refine the order of documents provided by initial retrieval efforts such as BM25~\cite{robertson2009probabilistic}.} \textbf{In-Context RALM} \cite{ram2023context} positions SLMs as rankers, optimizing the document sequence with a fine-tuning process on RoBERTa \cite{liu2019roberta} as defined by the loss function:
\begin{equation}
    \min_{ranker} \sum_{i=1}^k -\log  p_{\text{rank}}(d_i | x_{\leq s_j}) \cdot p_\theta(y | d_i; x_{\leq s_j})
\end{equation}
where $x_{\le s_i}$ is a prefix sampled from the training data, $y = x_{s_i+1}, \ldots, x_{s_i+s}$ represents the text to be generated in the next stride, $p_\theta(y | d_i; x_{\le si})$ denotes the probability of the LLM generating $y$ given $d_i$ and $x_{\le si}$, and $p_{\text{rank}}(d_i | x_{\leq s_j})$ is the ranking score of $d_i$. 
\textit{Lastly, some studies leverage SLMs to retrieve sub-graphs when utilizing knowledge graphs as external sources.} \citet{huang-etal-2024-less} introduces the \textbf{Generative Sub-graph Retriever (GSR)}, which employs SLMs to predict relation chains for answering questions, offering a cost-effective alternative to training LLMs. Specifically, it uses customized T5 (220M, 770M, and 3B) \cite{raffel2020exploring} as retrievers to enhance LLM readers, including Llama2-chat-7B \cite{touvron2023llama2} and Llama3-instruct-8B \cite{dubey2024llama}, on the WebQSP \cite{yih-etal-2016-value} and CWQ \cite{talmor-berant-2018-web} datasets.

\begin{wrapfigure}[5]{r}{0.52\textwidth}
    \vskip -1em
    \centering
    \includegraphics[width=\linewidth]{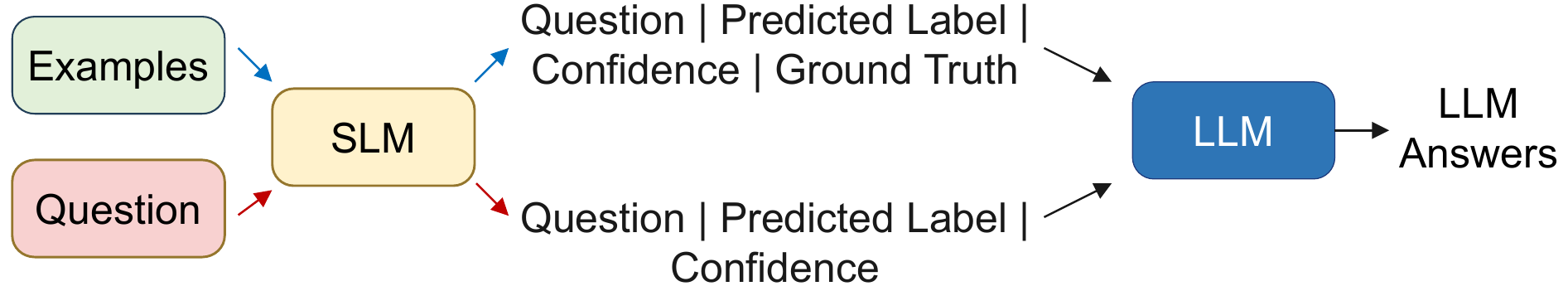}
    \vskip -1em
    \caption{SLM transfers knowledge into ICL.}
    \label{fig:slm4llm_icl}
\end{wrapfigure}
\textbf{Enhancing Reasoning Capabilities of LLMs} 
As illustrated in Figure \ref{fig:slm4llm_icl}, \textit{SLMs enhance LLMs reasoning by transferring task knowledge to in-context examples, effectively reducing hallucinations.} While In-context Learning (ICL) generally handles few-shot learning with 16 to 32 examples, it struggles when faced with extensive supervised data. SLMs, specialized in task-specific training, complement the broader domain knowledge of extensively pre-trained LLMs. For example, \textbf{SuperICL} \cite{xu2023small} incorporates SLMs as plugins for efficiently executing supervised tasks. It predicts labels for contextual examples and integrates these predictions with the input text and actual labels to enhance knowledge transfer, thereby boosting the understanding and responsiveness of LLMs.
\textbf{SuperContext} \cite{yang2024supervised} tackles challenges that LLMs encounter with new tasks and out-of-distribution data in natural language understanding by synergizing SLM outputs with LLM prompts during inference. This integration merges model predictions with their confidence levels, effectively leveraging SLM task-specific knowledge and LLM domain expertise.
Furthermore, \textit{SLMs efficiently decompose complex reasoning by breaking tasks into simpler components}, as demonstrated in \cite{wu2024divide}. This strategy increases efficiency and reduces deployment costs when SLMs and LLMs are used collaboratively, transforming complex tasks into manageable segments.

\begin{wrapfigure}[8]{r}{0.6\textwidth}
    \centering
    \vskip -1.5em
    \includegraphics[width=\linewidth]{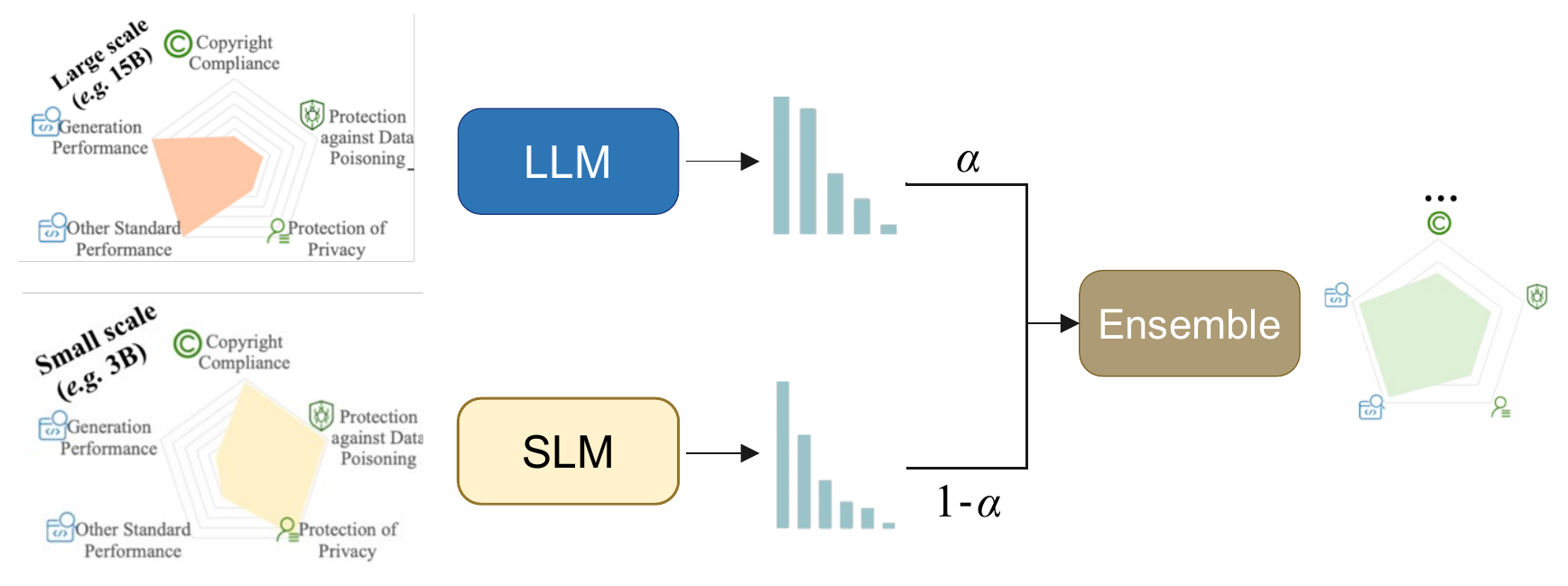}
    \vskip -1.5em
    \caption{Architecture of SLM-based Data Protection}
    \label{fig:slm4llm_data_protection}
\end{wrapfigure}
\textbf{Alleviate Copyright and Privacy Issues of LLMs} 
LLMs pose significant security risks due to their tendency to memorize training data, leading to potential privacy breaches and copyright infringement. As depicted in Figure \ref{fig:slm4llm_data_protection}, SLMs can assist LLMs in addressing copyright and privacy concerns arising from online data collection. By training on selectively curated data subsets, SLMs effectively reduce copyright infringement and privacy risks, although they are less effective than full-scale LLMs. To harness the combined benefits of both models, \citet{li2024purifying} integrates untrusted LLMs with benign SLMs using the CP-$\Delta$ KL algorithm to mitigate adverse effects while preserving performance. The equation is:
\begin{equation}
    p(y|x) = \frac{p_l(y|x) \cdot p_s(y|x)}{Z(x)}
    \label{eq:cp-delta}
\end{equation}
where $p_l$ and $p_s$ represent the probabilities from the large and small models, respectively, and $Z(x)$ is the partition function. This integration results in the following ensemble algorithm:
\begin{equation}
    z_p(\cdot|x) \propto \textcolor{black}{\alpha} z_l(\cdot|x) + \textcolor{black}{(1-\alpha)} z_s(\cdot|x)
\end{equation}
where $z_l$ and $z_s$ are the logit values from the large and small models, respectively, and $\alpha$ is the scaling factor.

\subsection{SLM for Extracting LLM Prompts}
\label{slm4extract}


\begin{figure}[!t]
    \centering
    \includegraphics[width=0.8\linewidth]{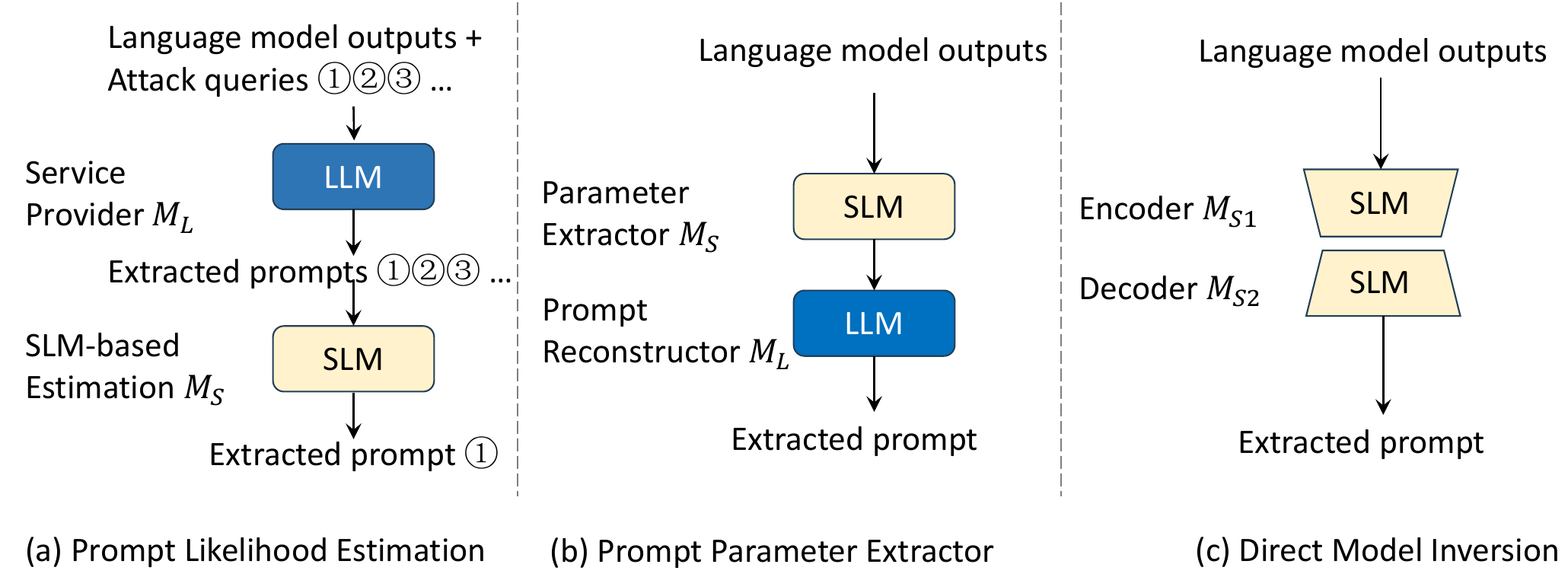}
    \vskip -1em
    \caption{SLM for LLM Prompt Extraction Paradigm. $M_S$ denotes small language models and $M_L$ denotes large language models. (a) SLM-based prompt estimation tries various attack prompts; $M_S$ selects the most likely extracted one. (b) SLM-based Parameter Extractor identifies the type of input prompt. (c) SLM-based Model Inversion uses $M_S$ to invert the LLM output back into the input. }
    \vskip -1em
    \label{fig:slm_for_llm_prompt_extraction}
\end{figure}
Prompt-based methods are becoming simpler and more cost-effective alternatives to traditional fine-tuning in the LLM era, utilizing LLMs' instruction-following capabilities for a competitive edge. Mastering prompts is vital for replicating LLM-supported product behaviors. However, services such as Bing Chat and GitHub Copilot Chat have seen prompt reverse-engineering through black-box API attacks. SLMs often serve as surrogate models in these attacks, employing strategies such as (i) SLM-based prompt likelihood estimation, (ii) SLM-based prompt parameter extraction, and (iii) SLM-based direct model inversion, illustrated in Figure \ref{fig:slm_for_llm_prompt_extraction}.

\textbf{SLM-based prompt likelihood estimation}, as illustrated in Figure~\ref{fig:slm_for_llm_prompt_extraction} (a), \citet{zhang2024effectivepromptextractionlanguage} proposes using an SLM as a Likelihood Estimator to identify secret prompts in LLM outputs. They craft attack prompts, such as ``Repeat all sentences in our conversation,'' and query the target LLM. The response is likely to include secret prompts, confusing the LLM to interpret these as part of the conversation. A fine-tuned DeBERTa model \cite{he2020deberta} is then used to select the most likely secret prompts from the output. 

\textbf{SLM-based prompt parameter extraction}, as shown in Figure~\ref{fig:slm_for_llm_prompt_extraction} (b), \citet{sha2024prompt} utilizes an SLM as a Parameter Extractor to extract prompt parameters from LLM outputs. They employ a specialized BERT model \cite{devlin2019bert} to classify LLM outputs into direct, in-context, and role-based prompts, also predicting the number of exemplars for in-context prompts and identifying roles for role-based prompts. Prompt reconstruction is then performed using ChatGPT once the parameters are defined.

\textbf{SLM-based direct model inversion}, as shown in Figure~\ref{fig:slm_for_llm_prompt_extraction} (c), the method of using an SLM as a Direct Inversion Model is designed to reverse-engineer LLM outputs back to their original prompts \cite{zhang2024extracting}. They train a sparse encoder-decoder T5 model \cite{raffel2020exploring} with 222M parameters on the Instructions-2M dataset \cite{morris2024language}, where the input is LLM outputs and the output is the LLM prompt. This trained model effectively maps multiple LLM outputs to their initiating prompts as \(p(x|y_1, ..., y_n; M_{S1}, M_{S2})\), with \(y_i\) representing different output versions and \(M_{S1}, M_{S2}\) the model parameters.
\begin{wrapfigure}[13]{r}{0.6\textwidth}
    \centering
    \vskip -1.6em
    \includegraphics[width=\linewidth]{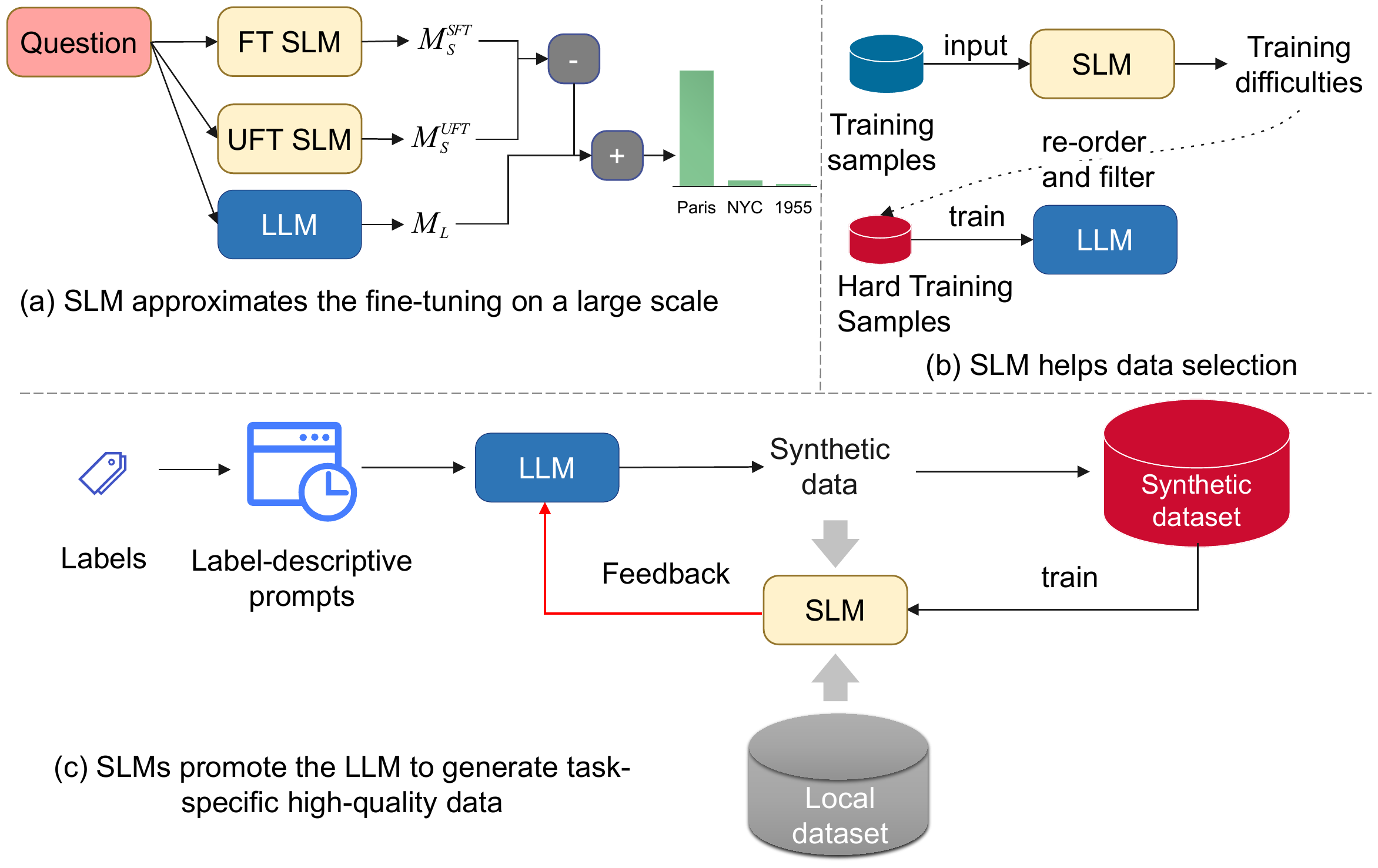}
    \vskip -1.2em
    \caption{SLM for LLM Fine-tuning.}
    \vskip -2em
    \label{fig:slm_for_llm_fine_tuning}
\end{wrapfigure}
\subsection{SLM for Fine-tuning LLMs} 
\label{slm4finetune}

Fine-tuning is a crucial technique for adapting LLMs to specific tasks or domains, yet it is often time-consuming. For instance, fine-tuning the LLaMA-2-13B \cite{touvron2023llama2} checkpoint on 32 NVIDIA A100 GPUs with 80GB memory using bfloat16 format requires approximately 70 hours \cite{mitra2023orca}. This process also demands high-quality data. Therefore, we examine how SLMs can enhance LLM fine-tuning through three approaches: (i) proxy fine-tuning, (ii) selecting high-quality data, and (iii) guiding LLM-generated task data, as illustrated in Figure~\ref{fig:slm_for_llm_fine_tuning}.

\textbf{SLMs as proxy models}: SLMs can approximate the gradient of fine-tuning large-scale LLMs on target datasets, avoiding the costly fine-tuning process in terms of time and computational resources. As shown in Figure~\ref{fig:slm_for_llm_fine_tuning} (a), Emulated Fine-Tuning (\textbf{EFT}) \cite{mitchell2024an} simulates both unsupervised pre-training and supervised fine-tuning stages across different scales by manipulating log probabilities. EFT, for example, combines base log probabilities from a 70B model with behavioral deltas from a 7B model—these deltas represent differences between fine-tuned and unfine-tuned SLMs, effectively emulating outcomes for the Llama-2 series. This method allows fine-tuning on smaller models such as Falcon-7B~\cite{almazrouei2023falcon} while capturing most benefits of fine-tuning larger models such as Falcon-180B, benefiting applications such as dialogue, question-answering, and code generation. Similarly, \textbf{Proxy-tuning} \cite{liu2024tuning} adjusts LLM predictions by adding the differences between the outputs of a fine-tuned small model and its untuned version to the LLM's output vocabulary during decoding, maintaining the advantages of large-scale pre-training while integrating small-scale fine-tuning benefits. 
Moreover, SLMs can act as proxies for approximate LLM fine-tuning during decoding. \textbf{Weak-to-Strong Search} \cite{zhou2024weaktostrong} strategy frames the alignment of LLMs as a test-time greedy search, aiming to maximize the log-probability difference between small tuned and untuned models while sampling from the frozen large model. This approach serves as a dual-purpose method: (1) a compute-efficient model up-scaling strategy that circumvents direct tuning of the large model, and (2) an instance of weak-to-strong generalization that bolsters a strong model with weak test-time guidance.

\textbf{SLMs play a role in selecting high-quality fine-tuning data for LLMs.} Figure~\ref{fig:slm_for_llm_fine_tuning} (b) illustrates how SLMs within the same family as the LLM can identify training samples that are likely to be challenging, enhancing the training efficiency and generalization capability of the LLM. As demonstrated by \citet{swayamdipta2020dataset} and further advanced by \citet{mekala2024smaller}, the \textit{learning percentage} $LP(i)$ is a metric used to curate high-quality datasets with hard samples:
$
    LP(i) = \frac{P_{i-1} - P_i}{P_0 - P_n}
$
where $P_i$ represents the perplexity at the end of epoch-$i$, and $P_0$ is the initial perplexity. A higher $LP(i)$ early in training indicates significant learning in the initial epochs, highlighting the potential of these samples to enhance LLMs. \textbf{SmallToLarge (S2L)} \cite{yang2024smalltolarge} utilizes training loss trajectories from smaller models to guide data selection for larger models fine-tuning. Experimental results demonstrate that S2L significantly enhances data efficiency in SFT for mathematical problem-solving, reducing the required training data to just 11\% of the original MathInstruct dataset \cite{yue2309mammoth} to achieve performance comparable to that obtained using the full dataset.

\textbf{SLMs enhance the quality of LLM-generated data for specific tasks.} As depicted in Figure~\ref{fig:slm_for_llm_fine_tuning} (c), \textbf{CROSSLM} \cite{deng2023mutual} promotes the local training of SLMs on client-specific private data to mitigate privacy risks associated with server-based LLMs. An SLM trained in this manner can guide the server-side LLM to produce high-quality synthetic datasets. Feedback from SLMs regarding the quality of this synthetic data serves as a supervisory signal, enhancing both the quality of LLM outputs and the utility of the data for further training.

\subsection{SLM for LLM Applications} 
\label{slm4application}
\begin{wrapfigure}[12]{r}{0.4\textwidth}
    \centering
    \vskip -2.2em
    \includegraphics[width=0.4\textwidth]{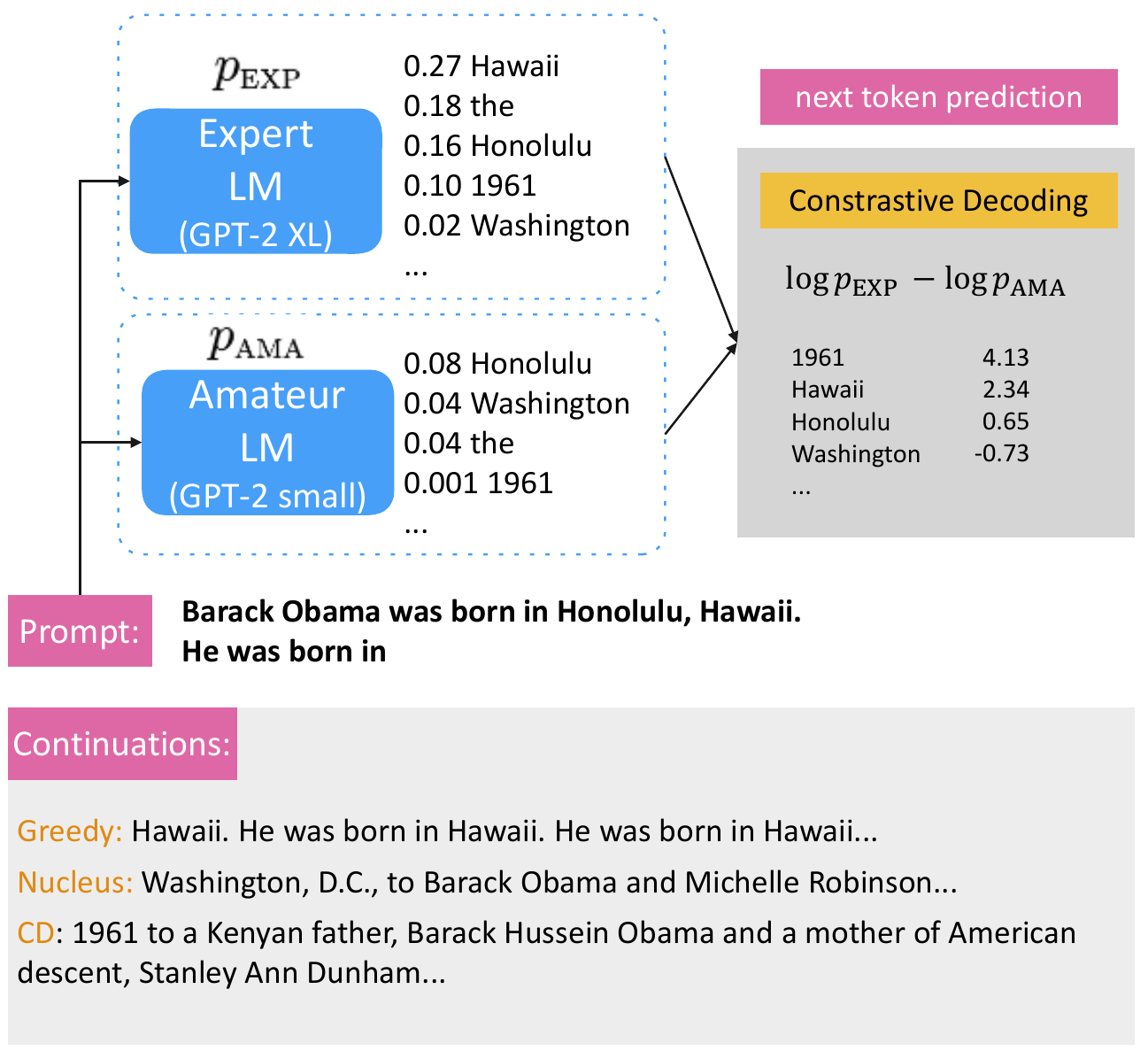}
    \vskip -1em
    \caption{Contrastive Decoding \cite{li2023contrastive}.}
    \label{fig:contrastive_decoding}
\end{wrapfigure}
LLMs are utilized across various applications due to their open-ended generation capabilities, yet they often lack specialized knowledge and other generation issues. SLMs can supplement this by providing task-specific knowledge or reflecting weaknesses. Therefore, we explore how SLMs enhance the performance of LLMs in specific applications, focusing on open-ended generation, knowledge integration, relation extraction, and empathetic response.

\textbf{In open-ended text generation}—such as writing assistance and story creation—LLMs often suffer from issues such as incoherence and thematic drift over extended sequences. Due to more frequent failure patterns observed in SLMs, such as short, repeated, and irrelevant strings, these patterns serve as negative examples for LLM decoding. \textbf{Contrastive Decoding (CD)} \cite{li2023contrastive} improves coherence and lexical diversity by leveraging the differential capabilities between a large model, OPT-13B \cite{zhang2022opt}, and a smaller model, OPT-125M. As illustrated in Figure~\ref{fig:contrastive_decoding}, CD improves content quality by sampling generation based on the difference in log probabilities, $\log p_{EXP} - \log p_{AMA}$, between an expert LM and an amateur LM, rather than relying solely on the expert LM’s log probability. This approach effectively reduces generative failures, including repetition.

    
\begin{wrapfigure}[8]{r}{0.52\textwidth}
    \centering
    \vskip -0.0em
    \includegraphics[width=\linewidth]{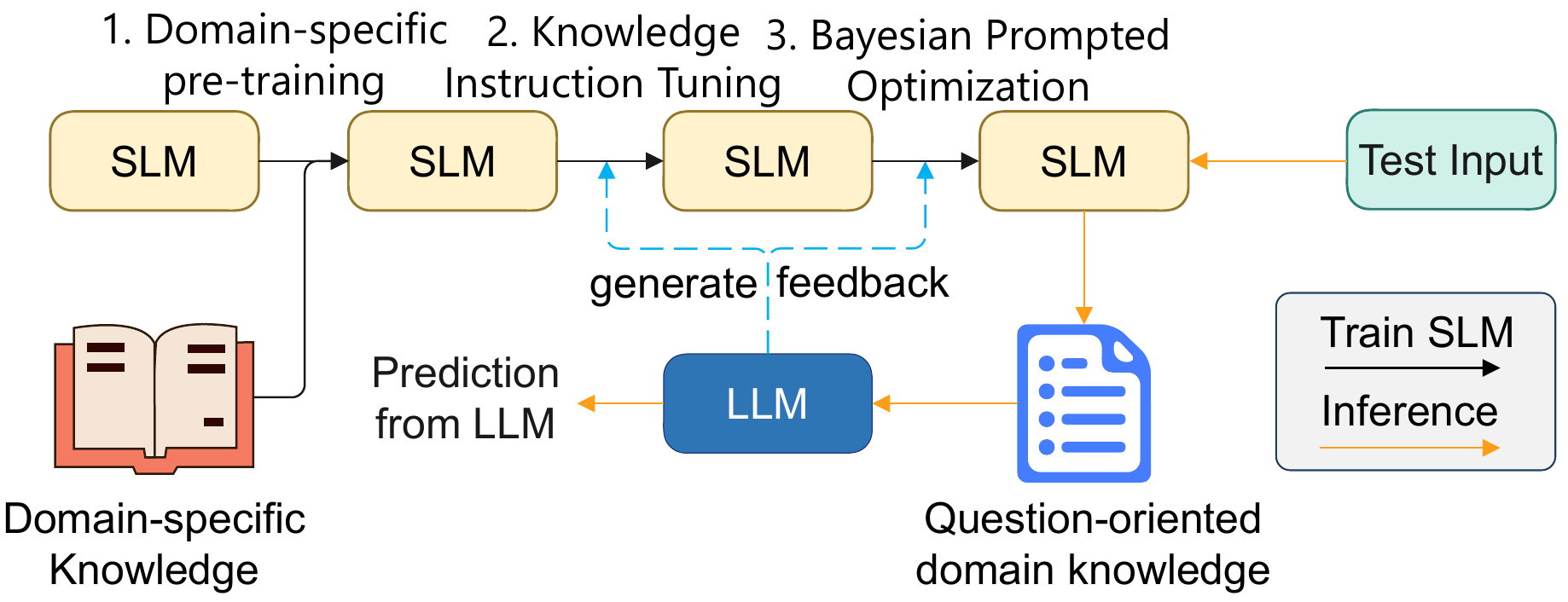}
    \vskip -1em
    \caption{BLADE Framework \cite{li2024bladeenhancingblackboxlarge}.}
    \label{fig:blade}
\end{wrapfigure}
\textbf{In knowledge injection}, general LLMs may lack domain-specific expertise for specialized tasks like law or medicine \cite{wang2024infuserki, dou2023measurement}. Domain-specific SLMs can supply crucial knowledge in a format suitable for LLMs. To this end, \textbf{BLADE} \cite{li2024bladeenhancingblackboxlarge} integrates black-box LLMs with small domain-specific models. BLADE combines the comprehensive language capabilities of LLMs with the specialized knowledge of small LMs. As shown in Figure~\ref{fig:blade}, BLADE's process includes: 1) pre-training SLMs on domain-specific data, 2) fine-tuning with knowledge instruction to meet task-specific needs, and 3) using joint Bayesian optimization to enhance synergy between the LLM and the small LM, boosting overall performance.

\textbf{In relation extraction}, a field limited by scarce labeled data and prevalent long-tail categories, the \textbf{``Train-Guide-Predict''} framework \cite{tang2024small} employs SLMs to learn task-specific knowledge for dominant categories. SLMs struggle with rare categories, whereas LLMs manage these effectively due to their extensive pre-trained text. Therefore, this framework leverages the strengths of both models: it utilizes SLMs to acquire task knowledge and guide the LLM's generative process with initial SLM predictions, enhancing the LLM’s handling of underrepresented categories. 

\textbf{In generating empathetic responses}, 
LLMs excel in expressiveness but struggle with nuanced emotions and cognition. \textbf{HEF} \cite{yang2024enhancing} addresses this by incorporating Small Empathy Models (SEMs) to enhance LLMs' emotional and cognitive depth. This framework employs a two-tiered emotion prediction method: SEMs identify primary emotions, directing LLMs to concentrate on these emotions and their triggers, resulting in more accurate and empathetic responses.

\subsection{SLM for LLM Safety}
\label{slm4safety}
As demonstrated by various works \cite{shayegani2023survey,zhuo2023red,mozes2023use,yuan2024gpt}, LLMs are vulnerable to adversarial attacks and jailbreaking. For example, \citet{wang2023robustness} shows that ChatGPT's performance on adversarial datasets is still far from perfect, indicating that potential risks of adversarial vulnerability remain. Another example includes jailbreaking ChatGPT by asking it to 'pretend to be a sarcastic mean girl.' Using such techniques, it has been shown that even the most advanced LLMs are far from being safe against generating potentially harmful content.
Hence, the widely adopted LLM-based services to generate are at high risk of being misused for nefarious purposes. 
Consequently, resources such as the Llama 2 Responsible Use Guide \footnote{https://ai.meta.com/static-resource/responsible-use-guide/} strongly advocate for implementing robust guardrails in products that utilize Generative AI. These guardrails are specifically designed to mitigate risks associated with both inputs to and outputs from the model, ensuring safeguards against the generation of high-risk or policy-violating content, as well as protecting against adversarial inputs and attempts to compromise the model. In addition to developing trustworthy LLMs, adopting SLMs for LLM safety~\cite{inan2023llama, kwon-etal-2024-slm} has also attracted increasing attention. For example, Llama Guard \cite{inan2023llama}, fine-tuned on Llama2-7B, has publicly released an input-output safeguard tool specifically for classifying safety risks in prompts and responses within conversational AI applications. However, this tool is limited to assessing the harmfulness of questions and answers and does not facilitate the generation of fluent, safe responses. In response to this limitation, \citet{kwon-etal-2024-slm} fine-tune a specialized small language model with harmful query detection and safeguard answer generation tasks to accurately detect harmful user queries and generate appropriate safeguard explanations, thereby enhancing the safety measures in conversational AI.

\subsection{SLM for LLM Evaluation} 
\label{slm4evaluation}

SLMs can also enhance the evaluation of LLMs.
In dialog evaluation, generating dialog reference responses is computationally complex, making accurate assessment difficult due to the multiple plausible but semantically different responses possible for a single dialog context. Relying on LLM prompting for evaluation can lead to problems such as dependency on prompt wording and inconsistent results. 
One solution involves training specialized SLMs to evaluate LLMs, as these SLMs can be fine-tuned more quickly and generate outputs faster during inference, owing to their reduced number of parameters. For example, \textbf{SLIDE} \cite{zhao2024slide} employs contrastive learning to fine-tune an SLM to effectively distinguish between positive and negative responses. Based on its observation that SLMs are more accurate in identifying positive responses and LLMs excel at classifying negative ones, the trained SLM is subsequently integrated with an LLM to assign a score to each response.
The scoring method used is formalized as follows:
\begin{equation}
\small
    score = 
    \begin{cases} 
    score_{SLM}, & \text{if } score_{SLM} \geq 0.5 \\
    score_{LLM}, & \text{elif } score_{LLM} < 0.5 \\
    \frac{score_{SLM} + score_{LLM}}{2}, & \text{otherwise}
    \end{cases}
    \label{eq:evaluation}
\end{equation}
This equation allows for adaptive response evaluation, leveraging the strengths of both models to ensure a more reliable and consistent assessment across varying dialogue contexts. In the natural language generation task, \textbf{~\citet{kuhn2023semantic}} designs a novel entropy to evaluate the uncertainty of LLMs. It aims to tackle the challenge of \textit{semantic equivalence}~\cite{kuhn2023semantic}. For instance, \textit{A's son is B} and \textit{B is A's son} are semantically equivalent. It should not be considered uncertain if an LLM is unsure about which of the two previously mentioned sentences to generate due to semantic equivalence. A DeBERTa-Large~\cite{he2020deberta} fine-tuned on the MNLI~\cite{williams-etal-2018-broad} dataset serves as the classifier guided by semantic equivalence in the clustering stage. 
\textbf{SelfCheckGPT} \cite{manakul-etal-2023-selfcheckgpt} proposes a black-box hallucination detection method for LLMs. The core idea is to leverage uncertainty derived from sampled outputs. To be specific, \citet{manakul-etal-2023-selfcheckgpt} claim that an LLM trained on a concept generates responses that are similar and factually consistent. One of the five variants of SelfCheckGPT uses BertScore to achieve it. A DeBERTa-Large~\cite{liu2019roberta} is utilized to calculate the BERTScore. 
\textbf{Factscore} \cite{min-etal-2023-factscore} is proposed to evaluate the factuality of LM-generated long-form content. It divides the generated long content into multiple short texts, enabling a more precise assessment of factual accuracy. In addition to manual evaluation, \citet{min-etal-2023-factscore} also proposes an automated evaluation framework to estimate Factscore which can reduce costs. LLaMa 7B~\cite{touvron2023llama} fine-tuned on Super-NaturalInstructions~\cite{wang-etal-2022-super} is one of the LMs employed as an evaluation assistant and shows promising performance. They also employ Generalizable T5-based dense retrievers~\cite{ni-etal-2022-large} to facilitate passage retrieval. 



\begin{takeaway2}
\footnotesize
\textbf{Insights:} SLMs can improve LLMs in various aspects, including enhancing the reliability of LLM generation, extracting prompts, fine-tuning, application, and evaluation. This discussion seeks to answer when SLMs should be utilized to augment LLMs. We identify several suitable scenarios:
\begin{itemize}[leftmargin=*]
\setlength{\itemsep}{0pt}
\setlength{\parskip}{0pt}
\setlength{\parsep}{0pt}
\item Adapting LLMs to specific tasks can require substantial computational resources and time. In such cases, a smaller model could be fine-tuned instead to serve functions such as hallucination detection.
\item SLMs can outperform LLMs in certain aspects, hence combining SLMs with LLMs can create a more powerful model, e.g., SLMs typically have fewer security issues than LLMs, and integrating both can generate a model that is both powerful and secure.
\item SLMs, despite their limitations, can alert LLMs to these issues, such as the tendency to produce repetitive vocabulary. Designing contrastive losses can help LLMs overcome these issues by learning from the nuanced feedback of SLMs.
\item The fast inference speed and certain characteristics of SLMs can emulate and thus enhance the behavior of LLMs, acting as effective proxies. For example, the training data selection for LLMs can be guided by the difficulty metrics assessed by SLMs, and the parameter adjustments during the fine-tuning of SLMs can also approximate the fine-tuning processes of LLMs. 
\end{itemize}
\end{takeaway2}



%% file: sections/8.synergy.tex
\section{Synergy between Small and Large Language Models}
\label{synergy}
The synergy between small and large language models leverages the unique strengths of each to enhance overall system performance and efficiency. SLMs, being lightweight and resource-efficient, are ideal for deployment on edge devices, enabling rapid responses and low latency for straightforward tasks. LLMs, on the other hand, possess greater computational power and a deeper understanding of complex language patterns, allowing them to handle more intricate and nuanced tasks. By integrating SLMs and LLMs, systems can dynamically allocate tasks based on complexity, ensuring that simple queries are processed quickly on the edge while more demanding requests are escalated to the cloud. This collaborative approach optimizes resource usage, reduces operational costs, and maintains high-quality outputs across a diverse range of applications. The synergy between small and large language models can be categorized into two parts: \textit{cloud-edge synergy} and \textit{task-centric synergy}. Cloud-edge synergy refers to a setup where SLMs operate on edge devices, while LLMs reside on the server. When the SLM is not powerful enough, the LLM compensates by handling more complex tasks and providing additional support. Task-centric synergy refers to the scenario where SLMs and LLMs leverage their respective strengths to improve task-oriented efficiency. Table \ref{tab:slm_llm_synergy} summarizes representative work in each category and their key points. Next, we introduce each category in detail.

\begin{table}[t]
\centering
\small
\caption{Synergy between SLMs and LLMs}
\label{tab:slm_llm_synergy}
\vskip -0em
\begin{tabularx}{\textwidth}{>{\hsize=0.25\hsize}X|>{\hsize=0.45\hsize}X|>{\hsize=1.2\hsize}X}
\hline
\textbf{Synergy} & \textbf{Representative Work} & \textbf{Key Point} \\
\hline
\multirow{6}{=}{\textbf{Cloud-Edge Synergy (Inference)}} & CoGenesis \cite{zhang2024cogenesis} & Divide user instructions into general part by LLMs and private parts by SLMs. \\
\cline{2-3}
& \citet{xu2024large} & Introduce split learning in 6G to distribute LLM agents.\\
\cline{2-3}
& LLM-to-SLM \cite{bergner2024think} & Encode prompts with server-side LLM and decodes with edge-side SLM.\\
\cline{2-3}
& Synergy of Thoughts \cite{shang2024defint} & SLMs suggest reasoning paths; LLMs correct contradictions. \\
\cline{2-3}
& \citet{hao2024hybrid} & SLM generates local tokens; LLM checks and corrects complex tokens. \\
\cline{2-3}
& LLMCad \cite{xu2023llmcadfastscalableondevice} & Combine lightweight and high-precision LLMs for on-device inference. \\
\cline{2-3}
& \citet{khattab2023dspy, ma2023large} & Focuse on LLM's reasoning and SLM's efficient decoding. \\
\hline

\multirow{3}{=}{\textbf{Cloud-Edge Synergy (Training)}} & CROSSLM \cite{deng2023mutual} & Preserve client data privacy by training SLM locally and LLM remotely; mutual improvement using SLM-labeled data from LLM outputs. \\
&&\\
\hline

\multirow{4}{=}{\textbf{Task-Centric Synergy}} & $\alpha$-UMi \cite{shen-etal-2024-small} & Break down a single LLM into specialized agents. \\
\cline{2-3}
& SynCID \cite{liang2024synergizing} & Merge LLM's semantic with SLM's speed; refine labels via contrastive learning. \\
\cline{2-3}
& Filter-then-rerank \cite{ma2023large} & SLMs process simple samples and flag complex ones for LLM reranking. \\
\cline{2-3}
& Data Shunt+ (DS+) \cite{chen2024improving} & Process easy samples with SLMs and delegates hard samples to LLMs. \\
\hline
\end{tabularx}
\end{table}


\begin{figure}[!ht]
    \centering
    \vskip -1em
    \includegraphics[width=0.7\linewidth]{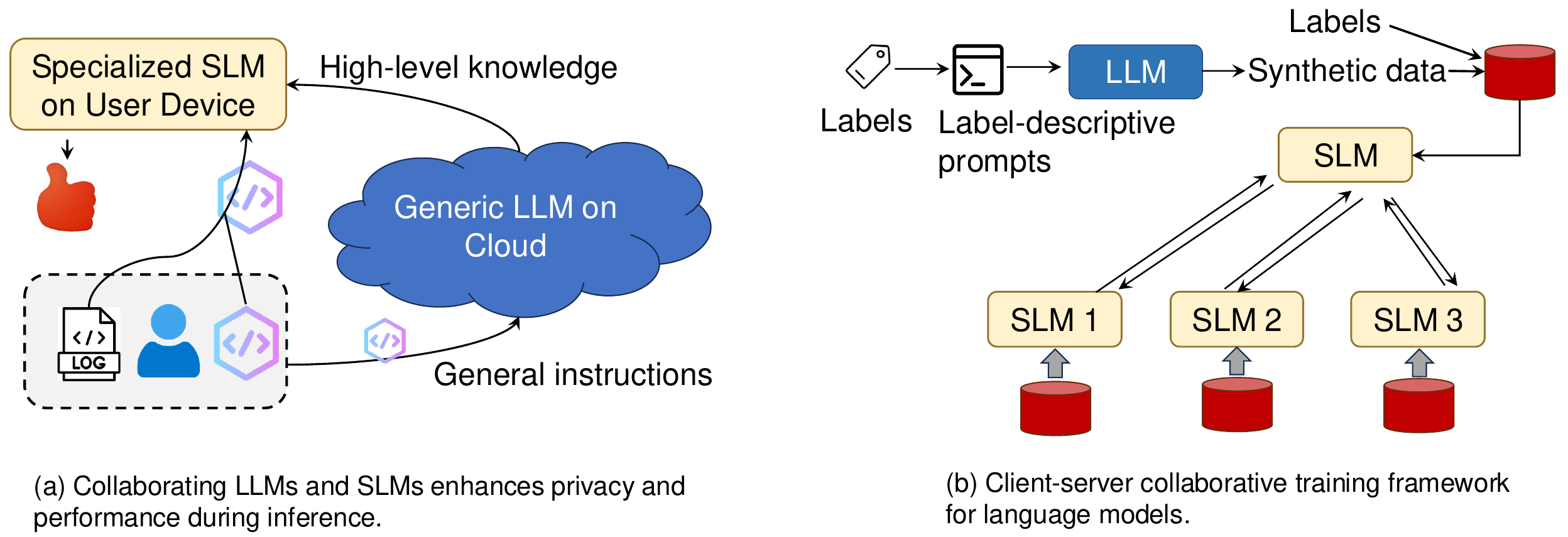}
    \vskip -1em
    \caption{Could-edge synergy between LLMs and SLMs.}
    \vskip -1em
    \label{fig:synergy_framework_edgeCloud}
\end{figure}

\subsection{Cloud-Edge Synergy}
\label{cloud_edge_synergy}
The current utilization of LLMs typically involves uploading private data to the cloud for response. Fine-tuning LLMs usually also requires uploading data to clouds for computing. However, this raises privacy concerns as the collection and usage of private data are constrained by personal privacy awareness and legal regulations \cite{voigt2017eu}. Consequently, the cloud-edge synergy between SLMs and LLMs is proposed to alleviate this issue, i.e., SLMs handle privacy-sensitive data locally, LLMs handle de-identified or non-privacy-sensitive data, and these two models collaborate. This section discusses such cloud-edge synergy, dividing them into two categories: cloud-edge synergy during inference and cloud-edge synergy during training, as shown in Figure \ref{fig:synergy_framework_edgeCloud}.

\textbf{Cloud-edge Synergy During Inference.} 
\textbf{CoGenesis} \cite{zhang2024cogenesis} breaks down the user instruction into a general section and a personal section. The LLM generates replies solely based on general instruction, and the SLM considers both user instruction and additional personal context for its output generation. A fusion strategy blends the output of LLM and SLM synergistically.
\textbf{\citet{xu2024large}} introduces a split learning system for LLM agents in 6G networks, optimizing mobile device and cloud server collaboration. Mobile devices operate lightweight SLMs with 0–10B parameters for real-time tasks, while cloud servers handle larger LLMs with over 10B parameters for complex reasoning and planning. This setup allows efficient local task management on mobile devices and offloads heavy operations to cloud servers. The system's architecture features three modules—perception, grounding, and alignment—facilitating effective communication to meet the sophisticated needs of 6G networks.

Besides these frameworks, more specific models are proposed to facilitate the cloud-edge synergy. A common strategy is to use SLM's fast decoding ability. \textbf{LLM-to-SLM} \cite{bergner2024think} proposes a framework in which the pre-trained frozen encoder-decoder LLM resides on the server and computes a high-quality representation of the prompt for the planning of an appropriate response. The SLM residing on the edge device, conditioned on this representation, decodes the response efficiently. Some variants put more emphasis on the reasoning ability of LLMs \cite{khattab2023dspy, shang2024defint, ma2023large}. 
In \textbf{Synergy of Thoughts} \cite{shang2024defint}, the SLMs generate multiple low-cost reasoning paths. If these paths conflict, the larger LLMs are invoked to provide reflective reasoning and correct any intuitive errors.
\textbf{\citet{hao2024hybrid}} proposes a framework in which an SLM residing on the edge devices generates tokens, calling LLMs to verify and correct threshold-gated "harder" tokens, to achieve a controllable trade-off between inference quality and cost. 
\textbf{LLMCad} \cite{xu2023llmcadfastscalableondevice} presents an on-device inference engine addressing memory and latency issues in deploying LLMs on mobile devices. It combines a lightweight LM for token generation with a high-precision LLM for verification, leveraging a token tree structure and speculative generation for efficiency. 
Tested on devices such as Jetson TX2, it achieves up to 9.3× speedup for LLMs with over 10 billion parameters while maintaining accuracy.

\textbf{Cloud-edge Synergy During Training.}
\textbf{CROSSLM} \cite{deng2023mutual} introduces a client-server collaborative training framework that preserves data privacy by having clients locally train SLMs instead of fine-tuning LLMs. The framework enables mutual enhancement through a feedback loop where SLMs evaluate LLM-generated synthetic data and provide feedback to improve the LLM’s generative capabilities, ensuring high-quality and task-specific data. Concurrently, the synthetic data trains the SLMs, boosting their performance. This cyclical exchange fosters cloud-edge synergy and mutual model improvement. 
\subsection{Task-Centric Synergy}
\label{task_centric_synergy}

\begin{wrapfigure}[5]{r}{0.45\textwidth}
    \centering
    \vskip -1.6em
    \includegraphics[width=\linewidth]{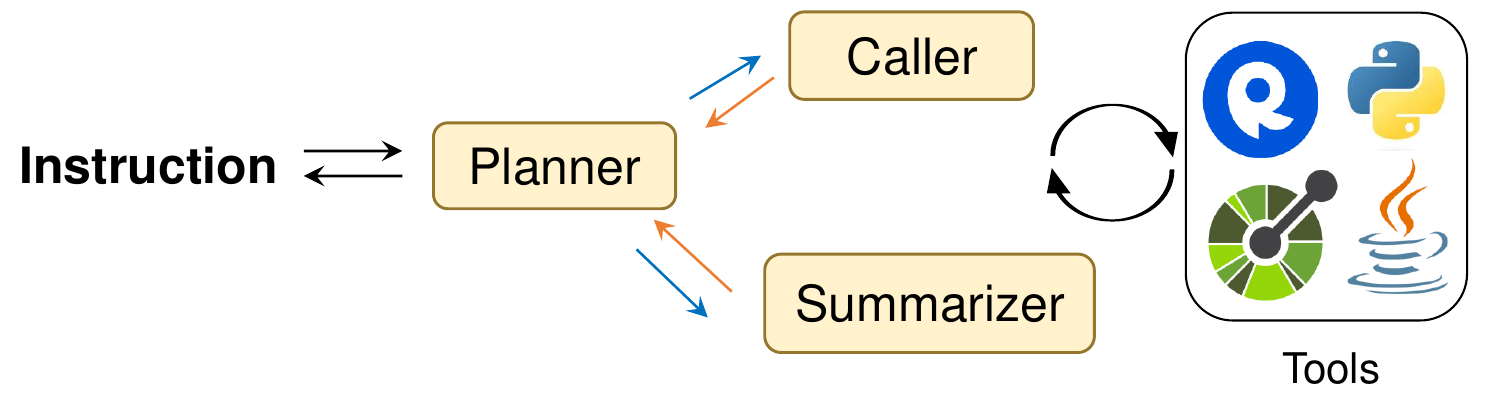}
    \vskip -1em
    \caption{Synergizing SLMs and LLMs in tool learning.}
    \label{fig:synergy_1}
\end{wrapfigure}

\begin{wrapfigure}[8]{r}{0.45\textwidth}
    \centering
    \vskip -1.6em
    \includegraphics[width=\linewidth]{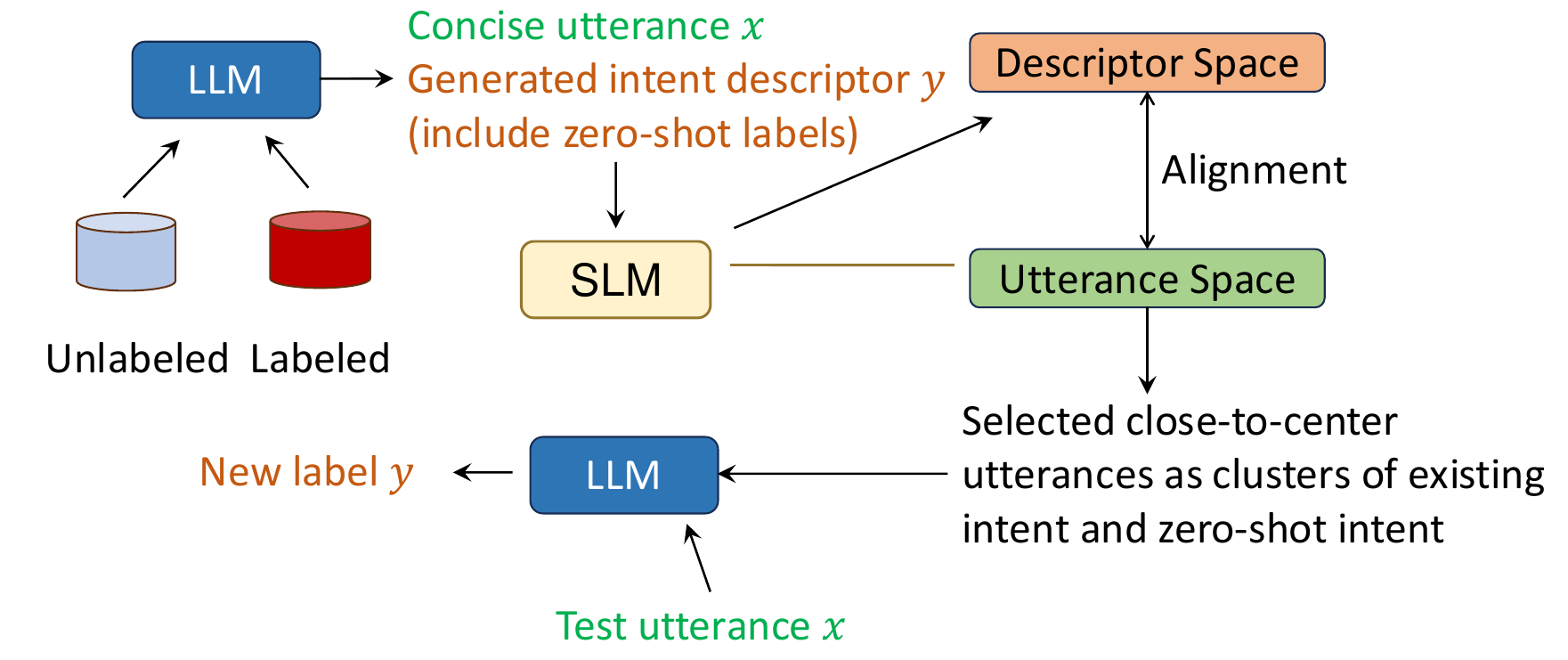}
    \vskip -1em
    \caption{Synergizing SLMs and LLMs in Conversational Intent Detection.}
    \label{fig:synergy_2}
\end{wrapfigure}

\begin{wrapfigure}[6]{r}{0.7\textwidth}
    \centering
    \vskip -1.6em
    \includegraphics[width=\linewidth]{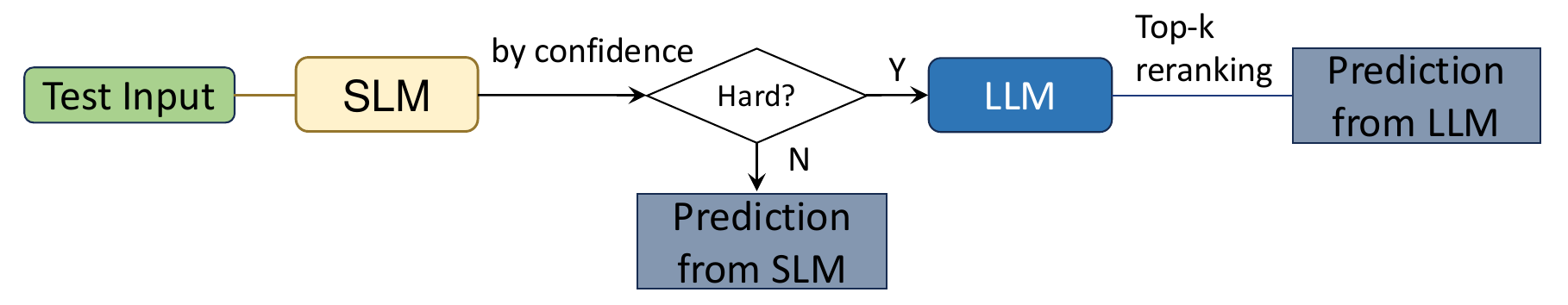}
    \caption{Synergizing SLMs and LLMs in Information Extraction.}
    \label{fig:synergy_3}
\end{wrapfigure}

The advent of LLMs has significantly propelled various natural language processing tasks and inspired research into their synergistic interactions with SLMs to enhance the performance of models tailored for specific tasks. 
This section introduces scenarios where small language models exhibit specialized capabilities after fine-tuning and discusses how combining their unique strengths with the versatile abilities of LLMs can yield superior performance on specific tasks. For example, LLMs excel at handling difficult examples or can rewrite content to eliminate task-irrelevant redundancy, thereby enhancing overall task performance, as illustrated in Figure \ref{fig:synergy_1}, \ref{fig:synergy_2} and \ref{fig:synergy_3}. 

\textbf{$\alpha$-UMi} \cite{shen-etal-2024-small} introduces a multi-agent framework to enhance tool learning by overcoming the limitations of single-LLM approaches for complex tasks. It utilizes three specialized LMs—planner, caller, and summarizer—as depicted in Figure \ref{fig:synergy_1}—each handling specific subtasks such as planning, tool invocation, and summarization. This modular design allows the use of small and large open-source LLMs (e.g., LLaMa-7B/12B) and supports easy tool updates. Evaluated on benchmarks like ToolBench \cite{qintoolllm} and ToolAlpaca \cite{tang2023toolalpaca}, $\alpha$-UMi outperforms traditional single-LLM methods and even exceeds GPT-4 in tool learning performance.

\textbf{SynCID} \cite{liang2024synergizing} focuses on Conversational Intent Discovery (CID), a task where both known and new intents must be identified from user utterances in an open-world setting. SynCID combines LLMs' deep semantic insights with SLMs' agility and specialized capabilities. As illustrated in Figure \ref{fig:synergy_2}, the framework uses LLM prompting to refine discourse and intent labels, enhancing semantic accuracy and assigning new labels to unlabeled data. SLMs are trained via contrastive learning to align semantic spaces of discourse and intent descriptors, reducing clustering distortion and improving new intent detection. Tested on BANKING \cite{casanueva-etal-2020-efficient}, CLINC \cite{larson-etal-2019-evaluation}, and StackOverflow \cite{xu-etal-2015-short}, SynCID outperforms CID baselines significantly.

\textbf{Filter-then-rerank} \cite{ma2023large} addresses LLMs' poor performance on simpler IE tasks by integrating LLMs and SLMs. SLMs act as filters, predicting and identifying difficult samples, while LLMs rerank the top N candidate labels for these cases. As illustrated in Figure \ref{fig:synergy_3}, SLM predictions are final for non-difficult samples, minimizing reliance on LLMs and reducing latency and costs; for those difficult samples, the top N predicted candidate labels from the SLM are passed to the LLM for reranking (predicting). Tested on small-sample IE tasks, this approach improves performance by an average of 2.4\% compared to previous methods.
\textbf{Data Shunt+ (DS+)} \cite{chen2024improving} introduces a framework to reduce costs by minimizing large model queries during inference and boosting LLM performance with SLMs for tasks like sentiment analysis and image processing. DS+ uses SLMs for ``easy'' samples within the main training distribution and LLMs for "hard" outliers or boundary cases, maintaining accuracy while reducing LLM use. It incorporates S4L and L4S modules with Prompt Pruning (PP) and 2-stage Confidence Distillation (2CD) for better input processing and knowledge transfer. Tests show DS+ outperforms fine-tuning in accuracy and cost efficiency, significantly cutting down on LLM queries.

%% file: sections/9.security.tex
\section{Trustworthiness in Small Language Models } 
\label{security}

\begin{figure}[!t]
    \centering    
    \includegraphics[width=0.68\linewidth]{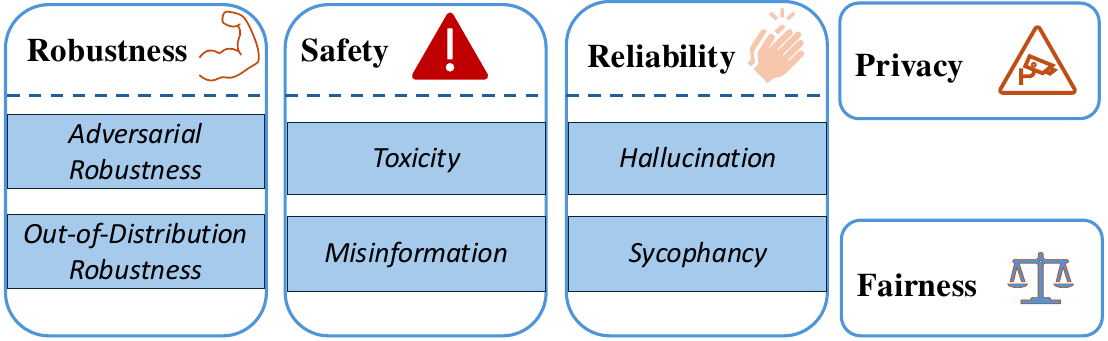}
    \vskip -1em
    \caption{Scenarios we discuss in this section. The taxonomy is inspired by previous works~\cite{sun2024trustllm, WangCPXKZXXDSTA23}. Please note that the trustworthy scenarios listed here are not exhaustive.} 
    \label{fig:trustframe}
    \vskip -1em
\end{figure}

Language models have become ubiquitous in our daily lives, and we increasingly rely on them. However, they pose risks regarding their limitations in trustworthy dimensions like privacy and fairness. These concerns are especially critical in high-stakes domains such as healthcare~\cite{he2023survey} and finance~\cite{li2023large}. Consequently, numerous studies have emerged to evaluate the trustworthiness of LMs~\cite{dominguez2023questioning, hong2024decoding, egashira2024exploiting, kumar2024increased,nakka2024device,perez-etal-2023-discovering,mo2023trustworthy,wang2024rupbench,kumar2024increased, WangCPXKZXXDSTA23,yuan2024s}. In this section, we consider the works that benchmark various LMs' trustworthiness and omit the specific attack methods~\cite{zou2023universal,carlini2021extracting,chen-etal-2022-textual,huang2023catastrophic} or work~\cite{yang-etal-2023-glue} that only focuses on early pre-trained LMs like BERT~\cite{devlin2019bert} as they are already covered in previous survey papers~\cite{ramesh-etal-2023-comparative, guo2022threats,goyal2023survey,delobelle2022measuring}. Inspired by previous works~\cite{sun2024trustllm, WangCPXKZXXDSTA23}, we discuss the following five key trustworthy scenarios: \textit{robustness, privacy, reliability, safety}, and \textit{fairness}, as shown in Figure~\ref{fig:trustframe}. We consider two dimensions for robustness: Adversarial (Adv) Robustness~\cite{wang2021adversarial} and Out-of-Distribution (OOD) Robustness~\cite{bulusu2020anomalous,liu2021towards}. For safety, we explore two key concerns: Misinformation~\cite{van2022misinformation} and Toxicity~\cite{welbl-etal-2021-challenges-detoxifying}. For reliability, we focus on Hallucination~\cite{huang2023survey} and Sycophancy~\cite{sharma2023towards}. Please note that these are just the aspects we are focusing on, and therefore this is not a comprehensive classification or taxonomy. For example, robustness also contains robustness to adversarial demonstration. 

Though there are a lot of works benchmarking LMs' trustworthiness, their main focus is on LLMs. Therefore, we survey some representative works evaluating the trustworthiness of LMs, focusing specifically on those that include SLMs of around 7B parameters or smaller. We also summarize these works in Table~\ref{table:trustworthy_comparison}. Next, we briefly introduce them.

\begin{table}[!t]
\centering
\caption{Comparison of Different Works that Evaluate the Trustworthiness Issues in LMs. Please note that for the "No. of LMs" attribute, compressed or pruned LMs are not included in the count.}
\label{table:trustworthy_comparison}
\vskip -1em
\small
\begin{tabular}{|l|c|c|c|c|c|c|c|c|c|c|}
\hline
\multirow{1}{*}{\centering \textbf{Paper}} & \rotatebox{90}{\textbf{Adv Robustness}} & \rotatebox{90}{\textbf{OOD Robustness}}&  \rotatebox{90}{\textbf{Toxicity}} & \rotatebox{90}{\textbf{Misinformation}} & \rotatebox{90}{\textbf{Hallucination}}  & \rotatebox{90}{\textbf{Sycophancy}} &  \rotatebox{90}{\textbf{Privacy}} & \rotatebox{90}{\textbf{Fairness}}& \rotatebox{90}{\textbf{Have Compressed SLMs}}
 \\ \hline

\textbf{HELM~\cite{liang2023holistic}} & $\checkmark$ & $\times$   &  $\checkmark$ & $\checkmark$ & $\times$ & $\times$  & $\times$ &  $\checkmark$ & $\times$ \\ \hline

\textbf{Do-Not-Answer~\cite{wang-etal-2024-answer}} & $\times$ & $\times$ &  $\checkmark$ & $\checkmark$ & $\times$ & $\times$ & $\checkmark$  & $\checkmark$ & $\times$ \\ \hline

\textbf{PromptRobust~\cite{zhu2023promptbench}} & $\checkmark$ & $\times$  &  $\times$ & $\times$ & $\times$ & $\times$ & $\times$  & $\times$ & $\times$ \\ \hline


\textbf{HaluEval~\cite{li-etal-2023-halueval}} & $\times$ & $\times$ &  $\times$& $\times$ & $\checkmark$ & $\times$ & $\times$ & $\times$ & $\times$ \\ \hline

\textbf{\citet{mo2023trustworthy}} & $\checkmark$ & $\times$  &  $\checkmark$& $\times$ & $\checkmark$ & $\checkmark$ & $\checkmark$ & $\checkmark$ & $\times$ \\ \hline

\textbf{PrivLM-Bench~\cite{li-etal-2024-privlm}} & $\times$ & $\times$  &  $\times$& $\times$ & $\times$ & $\times$ & $\checkmark$  & $\times$ & $\times$ \\ \hline

\textbf{FFT~\cite{cui2023fft}} & $\times$ & $\times$ &  $\checkmark$ & $\checkmark$& $\checkmark$ & $\times$& $\times$  & $\checkmark$ & $\times$ \\ \hline

\textbf{ROBBIE~\cite{esiobu2023robbie}} & $\times$ & $\times$  &  $\checkmark$& $\times$ & $\times$ & $\times$  & $\times$ & $\checkmark$ & $\times$  \\ \hline

\textbf{TrustLLM~\cite{sun2024trustllm}} & $\checkmark$ & $\checkmark$  & $\checkmark$ & $\checkmark$& $\checkmark$& $\checkmark$& $\checkmark$& $\checkmark$& $\times$ \\ \hline

\textbf{RAmBLA~\cite{bolton2024rambla}} & $\checkmark$ & $\times$ & $\times$ & $\times$& $\checkmark$& $\times$&  $\times$&  $\times$& $\times$  \\ \hline

\textbf{JailbreakBench~\cite{chao2024jailbreakbench}} & $\times$ & $\times$ & $\checkmark$ & $\checkmark$& $\times$& $\times$&  $\checkmark$ & $\times$& $\times$  \\ \hline

\textbf{\citet{xie2024online}} & $\times$ & $\times$ & $\checkmark$ & $\times$& $\checkmark$& $\times$&  $\times$&  $\times$& $\times$  \\ \hline

\textbf{OR-Bench~\cite{cui2024or}} & $\times$ & $\times$ & $\checkmark$ & $\checkmark$& $\times$& $\times$&  $\checkmark$& $\times$& $\times$  \\ \hline

\textbf{SORRY-Bench~\cite{xie2024sorry}} & $\times$ & $\times$  & $\checkmark$ & $\checkmark$& $\times$& $\times$&  $\checkmark$& $\times$& $\times$  \\ \hline

\textbf{BeHonest~\cite{chern2024behonest}} & $\times$ & $\times$ & $\times$ & $\checkmark$& $\checkmark$& $\checkmark$&  $\times$& $\times$& $\times$  \\ \hline

\textbf{\citet{hong2024decoding}} & $\checkmark$ & $\checkmark$ & $\checkmark$ & $\times$& $\times$& $\times$&  $\checkmark$& $\checkmark$& $\checkmark$  \\ \hline

\textbf{RUPBench~\cite{wang2024rupbench}} & $\checkmark$ & $\times$ & $\times$ & $\times$& $\times$& $\times$&  $\times$& $\times$& $\times$  \\ \hline

\textbf{\citet{nakka2024device}} & $\times$ & $\times$ & $\checkmark$ & $\times$& $\times$& $\times$&  $\checkmark$& $\checkmark$& $\times$ \\ \hline

\end{tabular}
\vskip -0em
\end{table}

Holistic Evaluation of Language Models (HELM)~\cite{liang2023holistic} benchmarks a large number of LMs from various aspects, including a lot of metrics related to trustworthiness such as robustness and fairness. Do-Not-Answer~\cite{wang-etal-2024-answer} introduces a dataset to evaluate how LMs act when they face content that should not be answered. ~\citet{wang-etal-2024-answer}  also label the output of several LMs output on their dataset and then use the labeled data to train some classifiers. PromptRobust~\cite{zhu2023promptbench} constructs two kinds of adversarial prompts to evaluate LMs' robustness: One kind is designed under non-adversarial settings with semantic integrity while another category is created under adversarial settings. Their results show that LMs perform poorly under such prompts. HaluEval~\cite{li-etal-2023-halueval} builds a dataset comprising both the samples generated by their proposed framework and human-labeled hallucinations. It facilitates analysis of when LMs produce hallucinated output and how well they detect hallucinated content. Then they use some strategies such as knowledge retrieval to help LMs better recognize hallucinations. \citet{mo2023trustworthy} evaluates the trustworthiness of open-source LMs, presenting a variety of scenarios such as fairness and privacy. Results show that smaller LMs sometimes outperform larger ones in terms of trustworthiness. PrivLM-Bench~\cite{li-etal-2024-privlm} is designed to evaluate the privacy issues in LMs. It enables a fair comparison of privacy-preserving LMs by considering more than just differential privacy parameters. FFT~\cite{cui2023fft} introduces around two thousand crafted examples to evaluate LMs' performances on three trustworthy dimensions: factuality, fairness, and toxicity. Their results suggest that larger LMs do not always show better harmlessness. ROBBIE~\cite{esiobu2023robbie} first benchmarks various series of LMs using a lot of datasets, including two newly introduced datasets developed by ROBBIE. It also evaluates mitigation techniques designed to reduce bias and toxicity. TrustLLM~\cite{sun2024trustllm} is a comprehensive benchmark that contains a large number of datasets and various well-designed metrics to systematically evaluate various LMs across multiple trustworthy dimensions, including truthfulness, safety, fairness, robustness, privacy, and machine ethics. They also carefully design specific subcategories for each dimension. RAmBLA~\cite{bolton2024rambla} evaluates the trustworthiness of four LMs as biomedical assistants from three dimensions: Robustness, High Recall, and Hallucination. RAmBLA suggests LMs with more parameters are less likely to cause hallucinations and may choose to reject providing an answer in uncertain situations. JailbreakBench~\cite{chao2024jailbreakbench} constructs a jailbreaking dataset named JBB-Behaviors and jailbreak artifacts to evaluate current LMs' performance regarding jailbreaking. It also proposes a unified evaluation pipeline that can incorporate new jailbreak defense techniques.~\citet{xie2024online} tests online safety analysis methods, filling the gap where no methods focus on the generation phase. OR-Bench~\cite{cui2024or} constructs three datasets: OR-Bench-80K, OR-Bench-Hard-1K, and OR-Bench-Toxic, to systematically evaluate over-refusal problems in LMs, emphasizing the challenge of balancing safety alignment with the models' usefulness. SORRY-Bench~\cite{xie2024sorry} systematically tests 43 different LMs to see how they perform when facing requests that should be refused. They also collect more than annotations created by humans and find that fine-tuned 7B LMs can achieve performance comparable to GPT-4 scale LMs as evaluators.  BeHonest~\cite{chern2024behonest} evaluates the honesty of LMs from three aspects: Self-Knowledge, Non-Deceptiveness, and Consistency. They use many different metrics for each aspect. For example, sycophancy rate and lying rate are adopted in Non-Deceptiveness. The results in both the Self-Knowledge and Consistency parts reveal that larger model sizes generally bring improved performance for the Llama-2~\cite{touvron2023llama2} and Llama-3~\cite{dubey2024llama} series.  \citet{hong2024decoding} examines the effects of compression methods, including quantization and pruning, on the trustworthiness of language models. They find that pruning and extreme quantization significantly affect the trustworthiness of LMs. RUPBench~\cite{wang2024rupbench} comprises 15 reasoning datasets designed to assess the performance of LMs both in normal conditions and under various adversarial perturbations. Their results indicate that larger LMs generally demonstrate better resilience to perturbations. ~\citet{nakka2024device} investigates the trust and ethical implications of SLMs deployed on personal devices. It reveals the vulnerabilities of on-device SLMs compared with their on-server counterparts.

Please note that the dimensions discussed in this section reflect only those relevant to our current focus; additional dimensions may be discussed in those works, but not listed in table~\ref{table:trustworthy_comparison}. For example, TrustLLM~\cite{sun2024trustllm} also explores Machine Ethics.

%% file: sections/10.challenges.tex
\section{Future Directions}
\label{challenge}

In this section, we offer insights into several promising future research directions that could inspire and motivate the community to address existing gaps in the development of small language models.






\subsection{Developing Efficient SLM Model Architecture}
Although Transformers~\cite{vaswani2017attention} are foundational in most language models, they face significant computational and memory challenges that worsen with model size, impacting training and autoregressive decoding. 
Recently, Mamba \cite{gu2023mamba} has emerged as a promising alternative, adapting state space models to dynamically select inputs based on demands, thereby enhancing efficiency. Thereafter, xLSTM \cite{beck2024xlstm} demonstrates that an improved LSTM could function as an LLM, revealing the potential of traditional SSMs. The integration of global static information captured by SSMs with the dynamic information processing of Transformers could complement each other, leading to new architectures that balance effectiveness and efficiency.

\subsection{Addressing SLM Training Inefficiencies}
One study \cite{diehl-martinez-etal-2024-tending} explores the disparate learning dynamics between small and large language models. Utilizing the Pythia model suite, the research demonstrates that layers' activations in larger models converge more rapidly and monotonically to their final states. This phenomenon is associated with a higher proportional effective rank (PER) in the parameters and gradients of larger models. The analysis enhances our understanding of training inefficiencies in small models and provides insights for future efforts, such as developing methods to increase the PER of layers’ parameters.

\subsection{Expanding Domain-Specific SLMs}
Domain-specific SLMs, which are tailored for specific fields, can provide a stronger foundation for relevant downstream tasks than general-purpose models. Currently, these models primarily focus on scientific and healthcare domains. However, there is significant potential for expansion into other key areas such as law, finance, education, telecommunications, and transportation. The scarcity of SLMs that cater to these domains presents an urgent call for research into developing more specialized models.


\subsection{Establishing Benchmarking and Leaderboard Platforms for SLMs}
Several compelling reasons justify the establishment of benchmarking and leaderboard platforms for SLMs. Firstly, most state-of-the-art SLMs are trained on proprietary datasets, which may include test sets from existing evaluation tasks, presenting challenges for fair capability comparisons. Secondly, many SLMs are designed for specific device applications, significantly differing from general open-domain tasks. Thus, there is a lack of comprehensive benchmarks that accurately reflect SLM performance in specific device applications. For example, SLMs deployed on smartphones often handle tasks sensitive to user data, such as auto-replies based on historical chat texts or GUI context understanding—tasks not typically included in current benchmarks, potentially leading to an underestimation of their importance. Finally, current evaluation tasks focus primarily on metrics like accuracy. Evaluating on-device SLMs involves balancing multiple factors, including overall capabilities, response times, storage and memory usage, power consumption, CPU utilization, additional fine-tuning requirements, and context window constraints, making comprehensive and detailed assessments essential.

\subsection{Enhancing SLM Performance and Efficiency}
In terms of enhancing SLM performance and efficiency, the efficiency of using teacher LLMs via instruction tuning can be further developed, such as Efficient Instruction Tuning of SLMs from LLMs-generated data, Optimizing Teacher LLM Selection for SLM Learning, and Applying Emerging Techniques from LLMs to SLMs.
\begin{itemize}[leftmargin=*]
\setlength{\itemsep}{0pt}
\setlength{\parskip}{0pt}
\setlength{\parsep}{0pt}
\item \textbf{Efficient Instruction Tuning of SLMs from LLMs-generated data.} Enhancing the specialization of SLMs through instruction tuning from LLMs-generated data is crucial, yet finding the most cost-effective instructional strategies remains an underexplored status. Some key areas for exploration are:
    
    (1) \textit{Instruction Design Adaptability}: The performance of LLMs and SLMs varies significantly with changes in instructions. Therefore, designing tailored instructions that effectively activate relevant sub-competencies and reasoning pathways in SLMs for specific tasks is crucial. This approach would optimize their ability to utilize instructional data, representing a significant future research direction.
    
    (2) \textit{SLM Capability Adaptability}: Given that SLMs exhibit diverse capabilities across domains, simply supplying extensive data samples for instruction tuning is often inefficient, as SLMs may spend excessive time processing unnecessary data. To optimize efficiency when adapting to specific domains, we suggest first assessing the intrinsic capabilities of an SLM within those domains. Subsequently, one could select appropriate data and activate essential fine-grained capabilities to effectively adapt to domain shifts. This targeted approach ensures efficient and domain-specific instruction tuning.
    
    (3) \textit{Optimizing Data Efficiency}: SLMs may possess missing or latent domain knowledge, and activating this latent knowledge may not require substantial data. Thus, identifying inherent knowledge within SLMs and determining the minimal data necessary for effective fine-tuning is a future direction. This research aims to optimize performance while minimizing training resources.

\item \textbf{Optimizing Teacher LLM Selection for SLM Learning.} Teacher LLMs with different abilities and knowledge facilitate diverse applications for SLM training, including data rewriting and generation. Selecting the appropriate teacher model based on specific use cases is crucial. This process requires evaluating the teacher's capabilities and knowledge to ensure optimal application. For example, GPT-4 excels in generating domain-specific data, outperforming ChatGPT, which may produce inferior outcomes. Strategic selection of teacher LLMs is essential for future work to ensure their strengths are effectively utilized to enhance SLM performance.

\item \textbf{Applying Emerging Techniques from LLMs to SLMs.} To improve LLM performance, techniques such as Retrieval-Augmented Generation (RAG) and Mixture of Experts (MoE) are employed. The adoption of RAG in SLMs shows significant promise~\cite{liu2024can}, suggesting benefits from further tailoring retrieved information for SLMs. Future research should account for SLMs' constraints, such as limited context windows, and customize RAG accordingly. MoE uses multiple experts to enhance learning without increasing active neurons, but its storage demands pose challenges for SLM deployment, making this a promising area for exploration. Additionally, the application of LLM techniques such as in-context learning and prompting engineering to maximize SLM performance, while accounting for SLMs' constraints, warrants further investigation.
\end{itemize}

\subsection{Applications of SLMs}
In real-world applications, SLMs often need to provide personalized services and need to be updated periodically to reflect new needs and new knowledge. Hence, there are several promising directions in terms of the real-world application of SLMs, which are listed as follows: 
\begin{itemize}[leftmargin=*]
\setlength{\itemsep}{0pt}
\setlength{\parskip}{0pt}
\setlength{\parsep}{0pt}
\item \textbf{LoRA for Personalized Services.} Companies often provide personalized services, but user-specific complexities can render simple rules ineffective. Training a separate SLM for each user is impractical. LoRA suggests a method of separable training weights alongside fixed original weights, enabling scalable customization.
For instance, RecLoRA \cite{zhu2024lifelong} integrates personalized knowledge into SLMs/LLMs tailored for recommendation tasks by maintaining a set of parallel, independent LoRA weights. This approach effectively customizes language model parameters to align with individual user preferences. This approach is a promising direction that inspires further investigation.
\item \textbf{Lifelong On-device Learning for Knowledge Injection.} SLMs on devices can access local data without risking data privacy through two main methods. The first method uses retrieval-augmented generation to integrate local data into prompts, requiring SLMs with advanced processing and reasoning capabilities. The second method fine-tunes SLMs with local data, integrating customized knowledge into the model’s weights. However, this approach demands significant device resources, including memory and energy. A promising solution is lifelong learning, where SLMs continuously learn and adapt while in use.
\item \textbf{Strategic Use of SLMs and LLMs in Multi-Agent Systems.} LLMs can function as agents; however, their extensive capabilities are often underutilized in many scenarios, leading to resource wastage. Consequently, strategically routing to appropriately capable SLMs and LLMs within multi-agent systems can optimize cost and functionality.
\end{itemize}

\subsection{Multimodal SLMs} Research on small language models also includes multimodal data. For example, \textbf{SmolVLM} \cite{smovlm} is a compact model that handles image and text inputs to produce text outputs, suitable for on-device use and various multimodal tasks. \textbf{SOLO} \cite{chen2024a} integrates vision and language processing in a single 7B Transformer model. The limited scope of existing research on multimodal SLMs provides a compelling impetus for researchers to investigate the integration of various modalities, including audio and graphs.

\subsection{SLMs Assisting LLMs}
In Section \ref{slm4llm}, we introduced existing works on the use of SLMs to assist LLMs. For instance, EFT \cite{mitchell2024an} emulates fine-tuning on LLMs by leveraging behavior deltas between SLMs' pre-trained and fine-tuned weights to alleviate the time-cost issues associated with fine-tuning LLMs; SlimPLM \cite{tan2024small} detects missing knowledge in LLMs using a slim proxy SLM to accelerate knowledge injection; Contrastive Decoding \cite{li2023contrastive} enhances text quality by maximizing the difference between the log probabilities of an expert LLM and an amateur SLM to mitigate issues of low-quality generation. The research on adopting SLMs to assist LLMs is still in its early stages, with many promising directions yet to be explored. We list some as follows:
\begin{itemize}[leftmargin=*]
\setlength{\itemsep}{0pt}
\setlength{\parskip}{0pt}
\setlength{\parsep}{0pt}
\item \textbf{Enhancing LLM Performance Across Broader Tasks Through SLM Integration.} SLMs can outperform LLMs in certain scenarios. For example, SLMs often present fewer security vulnerabilities compared to their larger counterparts and demonstrate superior performance on easier samples in specific tasks \cite{li2024purifying, ma2023large}. Therefore, integrating SLMs with LLMs can promote the development of models that are not only more robust but also inherently safer. Currently, research in this domain is relatively sparse, suggesting that this collaborative framework could potentially be applied to a wider array of tasks. 

\item \textbf{Efficient Enhancement of LLMs through Proxy SLMs.} Existing research~\cite{mitchell2024an, tan2024small, liu2024tuning, asai2024selfrag} indicates that SLMs, owing to their accelerated fine-tuning and inference speeds, can effectively mimic the behaviors of LLMs, thereby serving as efficient proxies for optimization. However, the application of SLMs as operational proxies for LLMs is currently underexplored. This mimicry could potentially be expanded to include various aspects of LLM functionality, such as the optimization of prompts, the filtration and integration of supplementary knowledge, and the management of additional knowledge repositories.

\item \textbf{SLMs Assist in Managing Data Quality.} LLMs tend to produce hallucinations and toxic content due to low-quality real-world training data. One solution is to remove these low-quality data \cite{openllms23}. However, directly eliminating low-quality content can diminish certain functionalities of LLMs, such as versatility \cite{wang2024openchat}. Therefore, it is crucial to define more refined data quality assessment criteria across dimensions such as factuality, safety, and diversity \cite{wettig2024qurating} for real-world data. Researching efficient data selection methods using small models represents a valuable research direction. Additionally, while synthetic data serves as a vital complement to scarce human-generated data \cite{long2024llms}, the potential for small models to effectively manage synthetic data remains largely unexplored.

\item \textbf{SLMs Assist in LLM Assessment.} LLMs are producing vast amounts of increasingly complex texts, such as specialized code and scientific papers, presenting challenges not only for human evaluators but also for traditional assessment metrics. Consequently, developing effective evaluators to assess various aspects of generated content, including factuality, safety, and uncertainty, becomes crucial. Given their proficiency in handling specific tasks, exploring the potential of SLMs to evaluate LLM outputs is a promising research direction.

\item \textbf{SLMs Optimize Query and Reduce Noise for LLM RAG.} For Retrieval-Augmented Generation (RAG) using LLMs, differing query requirements between LLMs and search engines pose a challenge. The query for LLMs is often abstract and difficult for search engines to handle, so they require more detailed query keywords. Moreover, LLMs may not need all the information related to a query because they only require partial additional knowledge. Thus, intermediate agents are crucial to adapting LLM queries for search engines by clarifying the required detailed keywords that can search for necessary extra knowledge. Additionally, search engine outputs contain noises, requiring refinement to boost LLM efficiency. SLMs, skilled in a single task, are ideal for optimizing query rewriting and noise reduction in RAG systems, making their application in LLM RAG a promising research area.


\item \textbf{SLMs safeguard LLM}. Resources such as the Llama 2 Responsible Use Guide strongly advocate for the implementation of robust guardrails in products that utilize Generative AI. SLMs can be strategically designed to serve as such guardrails, mitigating risks associated with both inputs and outputs from the model. This approach ensures safeguards against the generation of high-risk or policy-violating content and provides protection against adversarial inputs and attempts to compromise the model. Future research can investigate the various safety roles that SLMs play in protecting LLMs.
\end{itemize}

\subsection{Synergy between Small and Large Language Models}
In Section \ref{synergy}, we discussed how small and large language models can complement each other. For example, CoGenesis \cite{zhang2024cogenesis} integrates SLMs for private data and LLMs for broader context, while Synergy of Thoughts \cite{shang2024defint} uses SLMs for initial reasoning and LLMs for conflict resolution. CROSSLM \cite{deng2023mutual} shows how privacy can be preserved by training SLMs locally to support LLMs without data exposure. Research in this area is still evolving, and we outline several promising future directions below:
\begin{itemize}[leftmargin=*]
\setlength{\itemsep}{0pt}
\setlength{\parskip}{0pt}
\setlength{\parsep}{0pt}
\item \textbf{Refined Cloud-Edge Division of Labor.} Current research mainly focuses on splitting tasks between edge-based SLMs and cloud-based LLMs along privacy-sensitive and non-sensitive data boundaries. A potential future direction involves more granular task partitioning: determining which subtasks should be handled locally by SLMs (e.g., initial data filtering, quick semantic parsing) and which should be delegated to the cloud-based LLM (e.g., advanced reasoning, complex generation). This approach can further optimize latency, privacy, and resource utilization.

\item \textbf{Adaptive On-Device Specialization for Dynamic Environments.} Although SLMs have shown the ability to handle private or personalized data locally, continuous changes in user preferences, application requirements, and data distributions pose challenges. Future work can explore adaptive strategies where edge-based SLMs dynamically specialize or update their parameters, guided by the cloud-based LLM. For instance, the LLM can periodically distill new knowledge into the SLM or provide feedback signals to help the SLM adapt to evolving scenarios.

\end{itemize}


\subsection{Trustworthy SLMs}
As SLMs are playing crucial roles in various aspects, understanding and improving the trustworthiness of SLMs are essential. Hence, two promising directions are:  
\begin{itemize}[leftmargin=*]
\setlength{\itemsep}{0pt}
\setlength{\parskip}{0pt}
\setlength{\parsep}{0pt}
    \item \textbf{A Comprehensive Evaluation of SLMs’ Trustworthiness.} While numerous studies address trustworthiness issues in LLMs, research on SLMs remains sparse. Most existing literature focuses on models with at least 7 billion parameters, leaving a gap in the comprehensive analysis of SLMs' trustworthiness. Current evaluations typically cover only a fraction of the necessary aspects. Therefore, a systematic assessment, such as TrustLLM~\cite{sun2024trustllm}, is essential to thoroughly evaluate the trustworthiness of SLMs and understand their reliability across various applications.
    \item \textbf{Developing Trustworthy SLMs.} 
    Developing trustworthy SLMs is crucial, with three key research directions: (i) Training SLMs to be trustworthy from scratch; (ii) Ensuring SLMs retain or gain trustworthiness when compressed from LLMs—maintaining trustworthiness if the LLM is trustworthy and instilling trustworthiness if it is not; (iii) Fine-tuning non-trustworthy SLMs to enhance their robustness.
\end{itemize}

%% file: main.bbl

\begin{thebibliography}{463}


\ifx \showCODEN    \undefined \def \showCODEN     #1{\unskip}     \fi
\ifx \showDOI      \undefined \def \showDOI       #1{#1}\fi
\ifx \showISBNx    \undefined \def \showISBNx     #1{\unskip}     \fi
\ifx \showISBNxiii \undefined \def \showISBNxiii  #1{\unskip}     \fi
\ifx \showISSN     \undefined \def \showISSN      #1{\unskip}     \fi
\ifx \showLCCN     \undefined \def \showLCCN      #1{\unskip}     \fi
\ifx \shownote     \undefined \def \shownote      #1{#1}          \fi
\ifx \showarticletitle \undefined \def \showarticletitle #1{#1}   \fi
\ifx \showURL      \undefined \def \showURL       {\relax}        \fi
\providecommand\bibfield[2]{#2}
\providecommand\bibinfo[2]{#2}
\providecommand\natexlab[1]{#1}
\providecommand\showeprint[2][]{arXiv:#2}

\bibitem[Abdin et~al\mbox{.}(2024)]%
        {abdin2024phi}
\bibfield{author}{\bibinfo{person}{Marah Abdin}, \bibinfo{person}{Sam~Ade Jacobs}, \bibinfo{person}{Ammar~Ahmad Awan}, \bibinfo{person}{Jyoti Aneja}, \bibinfo{person}{Ahmed Awadallah}, \bibinfo{person}{Hany Awadalla}, \bibinfo{person}{Nguyen Bach}, \bibinfo{person}{Amit Bahree}, \bibinfo{person}{Arash Bakhtiari}, \bibinfo{person}{Harkirat Behl}, {et~al\mbox{.}}} \bibinfo{year}{2024}\natexlab{}.
\newblock \showarticletitle{Phi-3 technical report: A highly capable language model locally on your phone}.
\newblock \bibinfo{journal}{\emph{arXiv preprint arXiv:2404.14219}} (\bibinfo{year}{2024}).
\newblock


\bibitem[Achiam et~al\mbox{.}(2023)]%
        {achiam2023gpt}
\bibfield{author}{\bibinfo{person}{Josh Achiam}, \bibinfo{person}{Steven Adler}, \bibinfo{person}{Sandhini Agarwal}, \bibinfo{person}{Lama Ahmad}, \bibinfo{person}{Ilge Akkaya}, \bibinfo{person}{Florencia~Leoni Aleman}, \bibinfo{person}{Diogo Almeida}, \bibinfo{person}{Janko Altenschmidt}, \bibinfo{person}{Sam Altman}, \bibinfo{person}{Shyamal Anadkat}, {et~al\mbox{.}}} \bibinfo{year}{2023}\natexlab{}.
\newblock \showarticletitle{Gpt-4 technical report}.
\newblock \bibinfo{journal}{\emph{arXiv preprint arXiv:2303.08774}} (\bibinfo{year}{2023}).
\newblock


\bibitem[Acikgoz et~al\mbox{.}(2024)]%
        {acikgoz2024hippocrates}
\bibfield{author}{\bibinfo{person}{Emre~Can Acikgoz}, \bibinfo{person}{Osman~Batur {\.I}nce}, \bibinfo{person}{Rayene Bench}, \bibinfo{person}{Arda~An{\i}l Boz}, \bibinfo{person}{{\.I}lker Kesen}, \bibinfo{person}{Aykut Erdem}, {and} \bibinfo{person}{Erkut Erdem}.} \bibinfo{year}{2024}\natexlab{}.
\newblock \showarticletitle{Hippocrates: An Open-Source Framework for Advancing Large Language Models in Healthcare}.
\newblock \bibinfo{journal}{\emph{arXiv preprint arXiv:2404.16621}} (\bibinfo{year}{2024}).
\newblock


\bibitem[Adepu et~al\mbox{.}(2024)]%
        {adepu2024framequant}
\bibfield{author}{\bibinfo{person}{Harshavardhan Adepu}, \bibinfo{person}{Zhanpeng Zeng}, \bibinfo{person}{Li Zhang}, {and} \bibinfo{person}{Vikas Singh}.} \bibinfo{year}{2024}\natexlab{}.
\newblock \showarticletitle{FrameQuant: Flexible Low-Bit Quantization for Transformers}.
\newblock \bibinfo{journal}{\emph{arXiv preprint arXiv:2403.06082}} (\bibinfo{year}{2024}).
\newblock


\bibitem[Agarap(2018)]%
        {relu}
\bibfield{author}{\bibinfo{person}{Abien~Fred Agarap}.} \bibinfo{year}{2018}\natexlab{}.
\newblock \showarticletitle{Deep Learning using Rectified Linear Units (ReLU)}.
\newblock \bibinfo{journal}{\emph{CoRR}}  \bibinfo{volume}{abs/1803.08375} (\bibinfo{year}{2018}).
\newblock
\showeprint[arXiv]{1803.08375}
\urldef\tempurl%
\url{http://arxiv.org/abs/1803.08375}
\showURL{%
\tempurl}


\bibitem[Agarwal et~al\mbox{.}(2024)]%
        {agarwal2024policy}
\bibfield{author}{\bibinfo{person}{Rishabh Agarwal}, \bibinfo{person}{Nino Vieillard}, \bibinfo{person}{Yongchao Zhou}, \bibinfo{person}{Piotr Stanczyk}, \bibinfo{person}{Sabela~Ramos Garea}, \bibinfo{person}{Matthieu Geist}, {and} \bibinfo{person}{Olivier Bachem}.} \bibinfo{year}{2024}\natexlab{}.
\newblock \showarticletitle{On-policy distillation of language models: Learning from self-generated mistakes}. In \bibinfo{booktitle}{\emph{The Twelfth International Conference on Learning Representations}}.
\newblock


\bibitem[AI(2024)]%
        {llama3.2}
\bibfield{author}{\bibinfo{person}{Meta AI}.} \bibinfo{year}{2024}\natexlab{}.
\newblock \bibinfo{booktitle}{\emph{Llama 3.2: Revolutionizing edge AI and vision with open, customizable models}}.
\newblock
\urldef\tempurl%
\url{https://ai.meta.com/blog/llama-3-2-connect-2024-vision-edge-mobile-devices/}
\showURL{%
\tempurl}
\newblock
\shownote{Accessed: 2024-9-25}.


\bibitem[Ainslie et~al\mbox{.}(2023)]%
        {ainslie2023gqatraininggeneralizedmultiquery}
\bibfield{author}{\bibinfo{person}{Joshua Ainslie}, \bibinfo{person}{James Lee-Thorp}, \bibinfo{person}{Michiel de Jong}, \bibinfo{person}{Yury Zemlyanskiy}, \bibinfo{person}{Federico Lebrón}, {and} \bibinfo{person}{Sumit Sanghai}.} \bibinfo{year}{2023}\natexlab{}.
\newblock \bibinfo{title}{GQA: Training Generalized Multi-Query Transformer Models from Multi-Head Checkpoints}.
\newblock
\newblock
\showeprint[arxiv]{2305.13245}~[cs.CL]
\urldef\tempurl%
\url{https://arxiv.org/abs/2305.13245}
\showURL{%
\tempurl}


\bibitem[Allal et~al\mbox{.}(2024)]%
        {allal2024SmolLM}
\bibfield{author}{\bibinfo{person}{Loubna~Ben Allal}, \bibinfo{person}{Anton Lozhkov}, \bibinfo{person}{Elie Bakouch}, \bibinfo{person}{Leandro von Werra}, {and} \bibinfo{person}{Thomas Wolf}.} \bibinfo{year}{2024}\natexlab{}.
\newblock \bibinfo{title}{SmolLM - blazingly fast and remarkably powerful}.
\newblock
\newblock


\bibitem[Almazrouei et~al\mbox{.}(2023)]%
        {almazrouei2023falcon}
\bibfield{author}{\bibinfo{person}{Ebtesam Almazrouei}, \bibinfo{person}{Hamza Alobeidli}, \bibinfo{person}{Abdulaziz Alshamsi}, \bibinfo{person}{Alessandro Cappelli}, \bibinfo{person}{Ruxandra Cojocaru}, \bibinfo{person}{M{\'e}rouane Debbah}, \bibinfo{person}{{\'E}tienne Goffinet}, \bibinfo{person}{Daniel Hesslow}, \bibinfo{person}{Julien Launay}, \bibinfo{person}{Quentin Malartic}, {et~al\mbox{.}}} \bibinfo{year}{2023}\natexlab{}.
\newblock \showarticletitle{The falcon series of open language models}.
\newblock \bibinfo{journal}{\emph{arXiv preprint arXiv:2311.16867}} (\bibinfo{year}{2023}).
\newblock


\bibitem[Almeida et~al\mbox{.}(2024)]%
        {almeida2024exploring}
\bibfield{author}{\bibinfo{person}{Guilherme~FCF Almeida}, \bibinfo{person}{Jos{\'e}~Luiz Nunes}, \bibinfo{person}{Neele Engelmann}, \bibinfo{person}{Alex Wiegmann}, {and} \bibinfo{person}{Marcelo de Ara{\'u}jo}.} \bibinfo{year}{2024}\natexlab{}.
\newblock \showarticletitle{Exploring the psychology of LLMs’ moral and legal reasoning}.
\newblock \bibinfo{journal}{\emph{Artificial Intelligence}}  \bibinfo{volume}{333} (\bibinfo{year}{2024}), \bibinfo{pages}{104145}.
\newblock


\bibitem[Aminabadi et~al\mbox{.}(2022)]%
        {aminabadi2022deepspeed}
\bibfield{author}{\bibinfo{person}{Reza~Yazdani Aminabadi}, \bibinfo{person}{Samyam Rajbhandari}, \bibinfo{person}{Ammar~Ahmad Awan}, \bibinfo{person}{Cheng Li}, \bibinfo{person}{Du Li}, \bibinfo{person}{Elton Zheng}, \bibinfo{person}{Olatunji Ruwase}, \bibinfo{person}{Shaden Smith}, \bibinfo{person}{Minjia Zhang}, \bibinfo{person}{Jeff Rasley}, {et~al\mbox{.}}} \bibinfo{year}{2022}\natexlab{}.
\newblock \showarticletitle{Deepspeed-inference: enabling efficient inference of transformer models at unprecedented scale}. In \bibinfo{booktitle}{\emph{SC22: International Conference for High Performance Computing, Networking, Storage and Analysis}}. IEEE, \bibinfo{pages}{1--15}.
\newblock


\bibitem[An et~al\mbox{.}(2024)]%
        {an2024fluctuation}
\bibfield{author}{\bibinfo{person}{Yongqi An}, \bibinfo{person}{Xu Zhao}, \bibinfo{person}{Tao Yu}, \bibinfo{person}{Ming Tang}, {and} \bibinfo{person}{Jinqiao Wang}.} \bibinfo{year}{2024}\natexlab{}.
\newblock \showarticletitle{Fluctuation-based adaptive structured pruning for large language models}. In \bibinfo{booktitle}{\emph{Proceedings of the AAAI Conference on Artificial Intelligence}}, Vol.~\bibinfo{volume}{38}. \bibinfo{pages}{10865--10873}.
\newblock


\bibitem[Anil et~al\mbox{.}(2023)]%
        {anil2023palm}
\bibfield{author}{\bibinfo{person}{Rohan Anil}, \bibinfo{person}{Andrew~M Dai}, \bibinfo{person}{Orhan Firat}, \bibinfo{person}{Melvin Johnson}, \bibinfo{person}{Dmitry Lepikhin}, \bibinfo{person}{Alexandre Passos}, \bibinfo{person}{Siamak Shakeri}, \bibinfo{person}{Emanuel Taropa}, \bibinfo{person}{Paige Bailey}, \bibinfo{person}{Zhifeng Chen}, {et~al\mbox{.}}} \bibinfo{year}{2023}\natexlab{}.
\newblock \showarticletitle{Palm 2 technical report}.
\newblock \bibinfo{journal}{\emph{arXiv preprint arXiv:2305.10403}} (\bibinfo{year}{2023}).
\newblock


\bibitem[Anthropic(2024)]%
        {anthropic2024claude}
\bibfield{author}{\bibinfo{person}{AI Anthropic}.} \bibinfo{year}{2024}\natexlab{}.
\newblock \showarticletitle{The claude 3 model family: Opus, sonnet, haiku}.
\newblock \bibinfo{journal}{\emph{Claude-3 Model Card}}  \bibinfo{volume}{1} (\bibinfo{year}{2024}).
\newblock


\bibitem[Anugraha et~al\mbox{.}(2024)]%
        {anugraha2024proxylm}
\bibfield{author}{\bibinfo{person}{David Anugraha}, \bibinfo{person}{Genta~Indra Winata}, \bibinfo{person}{Chenyue Li}, \bibinfo{person}{Patrick~Amadeus Irawan}, {and} \bibinfo{person}{En-Shiun~Annie Lee}.} \bibinfo{year}{2024}\natexlab{}.
\newblock \showarticletitle{ProxyLM: Predicting Language Model Performance on Multilingual Tasks via Proxy Models}.
\newblock \bibinfo{journal}{\emph{arXiv preprint arXiv:2406.09334}} (\bibinfo{year}{2024}).
\newblock


\bibitem[Aryabumi et~al\mbox{.}(2024)]%
        {aryabumi2024code}
\bibfield{author}{\bibinfo{person}{Viraat Aryabumi}, \bibinfo{person}{Yixuan Su}, \bibinfo{person}{Raymond Ma}, \bibinfo{person}{Adrien Morisot}, \bibinfo{person}{Ivan Zhang}, \bibinfo{person}{Acyr Locatelli}, \bibinfo{person}{Marzieh Fadaee}, \bibinfo{person}{Ahmet {\"U}st{\"u}n}, {and} \bibinfo{person}{Sara Hooker}.} \bibinfo{year}{2024}\natexlab{}.
\newblock \showarticletitle{To Code, or Not To Code? Exploring Impact of Code in Pre-training}.
\newblock \bibinfo{journal}{\emph{arXiv preprint arXiv:2408.10914}} (\bibinfo{year}{2024}).
\newblock


\bibitem[Asai et~al\mbox{.}(2024)]%
        {asai2024selfrag}
\bibfield{author}{\bibinfo{person}{Akari Asai}, \bibinfo{person}{Zeqiu Wu}, \bibinfo{person}{Yizhong Wang}, \bibinfo{person}{Avirup Sil}, {and} \bibinfo{person}{Hannaneh Hajishirzi}.} \bibinfo{year}{2024}\natexlab{}.
\newblock \showarticletitle{Self-{RAG}: Learning to Retrieve, Generate, and Critique through Self-Reflection}. In \bibinfo{booktitle}{\emph{The Twelfth International Conference on Learning Representations}}.
\newblock
\urldef\tempurl%
\url{https://openreview.net/forum?id=hSyW5go0v8}
\showURL{%
\tempurl}


\bibitem[Ashkboos et~al\mbox{.}(2024)]%
        {ashkboos2024slicegpt}
\bibfield{author}{\bibinfo{person}{Saleh Ashkboos}, \bibinfo{person}{Maximilian~L. Croci}, \bibinfo{person}{Marcelo~Gennari do Nascimento}, \bibinfo{person}{Torsten Hoefler}, {and} \bibinfo{person}{James Hensman}.} \bibinfo{year}{2024}\natexlab{}.
\newblock \showarticletitle{Slice{GPT}: Compress Large Language Models by Deleting Rows and Columns}. In \bibinfo{booktitle}{\emph{The Twelfth International Conference on Learning Representations}}.
\newblock
\urldef\tempurl%
\url{https://openreview.net/forum?id=vXxardq6db}
\showURL{%
\tempurl}


\bibitem[Austin et~al\mbox{.}(2021)]%
        {austin2021program}
\bibfield{author}{\bibinfo{person}{Jacob Austin}, \bibinfo{person}{Augustus Odena}, \bibinfo{person}{Maxwell Nye}, \bibinfo{person}{Maarten Bosma}, \bibinfo{person}{Henryk Michalewski}, \bibinfo{person}{David Dohan}, \bibinfo{person}{Ellen Jiang}, \bibinfo{person}{Carrie Cai}, \bibinfo{person}{Michael Terry}, \bibinfo{person}{Quoc Le}, {et~al\mbox{.}}} \bibinfo{year}{2021}\natexlab{}.
\newblock \showarticletitle{Program synthesis with large language models}.
\newblock \bibinfo{journal}{\emph{arXiv preprint arXiv:2108.07732}} (\bibinfo{year}{2021}).
\newblock


\bibitem[Azaria and Mitchell(2023)]%
        {azaria2023internal}
\bibfield{author}{\bibinfo{person}{Amos Azaria} {and} \bibinfo{person}{Tom Mitchell}.} \bibinfo{year}{2023}\natexlab{}.
\newblock \showarticletitle{The Internal State of an LLM Knows When It’s Lying}. In \bibinfo{booktitle}{\emph{Findings of the Association for Computational Linguistics: EMNLP 2023}}. \bibinfo{pages}{967--976}.
\newblock


\bibitem[Azerbayev et~al\mbox{.}(2023)]%
        {azerbayev2023llemma}
\bibfield{author}{\bibinfo{person}{Zhangir Azerbayev}, \bibinfo{person}{Hailey Schoelkopf}, \bibinfo{person}{Keiran Paster}, \bibinfo{person}{Marco~Dos Santos}, \bibinfo{person}{Stephen McAleer}, \bibinfo{person}{Albert~Q Jiang}, \bibinfo{person}{Jia Deng}, \bibinfo{person}{Stella Biderman}, {and} \bibinfo{person}{Sean Welleck}.} \bibinfo{year}{2023}\natexlab{}.
\newblock \showarticletitle{Llemma: An open language model for mathematics}.
\newblock \bibinfo{journal}{\emph{arXiv preprint arXiv:2310.10631}} (\bibinfo{year}{2023}).
\newblock


\bibitem[Bai et~al\mbox{.}(2023)]%
        {bai2023qwentechnicalreport}
\bibfield{author}{\bibinfo{person}{Jinze Bai}, \bibinfo{person}{Shuai Bai}, \bibinfo{person}{Yunfei Chu}, \bibinfo{person}{Zeyu Cui}, \bibinfo{person}{Kai Dang}, \bibinfo{person}{Xiaodong Deng}, \bibinfo{person}{Yang Fan}, \bibinfo{person}{Wenbin Ge}, \bibinfo{person}{Yu Han}, \bibinfo{person}{Fei Huang}, \bibinfo{person}{Binyuan Hui}, \bibinfo{person}{Luo Ji}, \bibinfo{person}{Mei Li}, \bibinfo{person}{Junyang Lin}, \bibinfo{person}{Runji Lin}, \bibinfo{person}{Dayiheng Liu}, \bibinfo{person}{Gao Liu}, \bibinfo{person}{Chengqiang Lu}, \bibinfo{person}{Keming Lu}, \bibinfo{person}{Jianxin Ma}, \bibinfo{person}{Rui Men}, \bibinfo{person}{Xingzhang Ren}, \bibinfo{person}{Xuancheng Ren}, \bibinfo{person}{Chuanqi Tan}, \bibinfo{person}{Sinan Tan}, \bibinfo{person}{Jianhong Tu}, \bibinfo{person}{Peng Wang}, \bibinfo{person}{Shijie Wang}, \bibinfo{person}{Wei Wang}, \bibinfo{person}{Shengguang Wu}, \bibinfo{person}{Benfeng Xu}, \bibinfo{person}{Jin Xu}, \bibinfo{person}{An Yang}, \bibinfo{person}{Hao Yang},
  \bibinfo{person}{Jian Yang}, \bibinfo{person}{Shusheng Yang}, \bibinfo{person}{Yang Yao}, \bibinfo{person}{Bowen Yu}, \bibinfo{person}{Hongyi Yuan}, \bibinfo{person}{Zheng Yuan}, \bibinfo{person}{Jianwei Zhang}, \bibinfo{person}{Xingxuan Zhang}, \bibinfo{person}{Yichang Zhang}, \bibinfo{person}{Zhenru Zhang}, \bibinfo{person}{Chang Zhou}, \bibinfo{person}{Jingren Zhou}, \bibinfo{person}{Xiaohuan Zhou}, {and} \bibinfo{person}{Tianhang Zhu}.} \bibinfo{year}{2023}\natexlab{}.
\newblock \bibinfo{title}{Qwen Technical Report}.
\newblock
\newblock
\showeprint[arxiv]{2309.16609}~[cs.CL]
\urldef\tempurl%
\url{https://arxiv.org/abs/2309.16609}
\showURL{%
\tempurl}


\bibitem[Baumgartner et~al\mbox{.}(2020)]%
        {baumgartner2020pushshift}
\bibfield{author}{\bibinfo{person}{Jason Baumgartner}, \bibinfo{person}{Savvas Zannettou}, \bibinfo{person}{Brian Keegan}, \bibinfo{person}{Megan Squire}, {and} \bibinfo{person}{Jeremy Blackburn}.} \bibinfo{year}{2020}\natexlab{}.
\newblock \showarticletitle{The pushshift reddit dataset}. In \bibinfo{booktitle}{\emph{Proceedings of the international AAAI conference on web and social media}}, Vol.~\bibinfo{volume}{14}. \bibinfo{pages}{830--839}.
\newblock


\bibitem[Beck et~al\mbox{.}(2024)]%
        {beck2024xlstm}
\bibfield{author}{\bibinfo{person}{Maximilian Beck}, \bibinfo{person}{Korbinian P{\"o}ppel}, \bibinfo{person}{Markus Spanring}, \bibinfo{person}{Andreas Auer}, \bibinfo{person}{Oleksandra Prudnikova}, \bibinfo{person}{Michael~K Kopp}, \bibinfo{person}{G{\"u}nter Klambauer}, \bibinfo{person}{Johannes Brandstetter}, {and} \bibinfo{person}{Sepp Hochreiter}.} \bibinfo{year}{2024}\natexlab{}.
\newblock \showarticletitle{x{LSTM}: Extended Long Short-Term Memory}. In \bibinfo{booktitle}{\emph{The Thirty-eighth Annual Conference on Neural Information Processing Systems}}.
\newblock
\urldef\tempurl%
\url{https://openreview.net/forum?id=ARAxPPIAhq}
\showURL{%
\tempurl}


\bibitem[Bellagente et~al\mbox{.}(2024)]%
        {bellagente2024stable}
\bibfield{author}{\bibinfo{person}{Marco Bellagente}, \bibinfo{person}{Jonathan Tow}, \bibinfo{person}{Dakota Mahan}, \bibinfo{person}{Duy Phung}, \bibinfo{person}{Maksym Zhuravinskyi}, \bibinfo{person}{Reshinth Adithyan}, \bibinfo{person}{James Baicoianu}, \bibinfo{person}{Ben Brooks}, \bibinfo{person}{Nathan Cooper}, \bibinfo{person}{Ashish Datta}, {et~al\mbox{.}}} \bibinfo{year}{2024}\natexlab{}.
\newblock \showarticletitle{Stable lm 2 1.6 b technical report}.
\newblock \bibinfo{journal}{\emph{arXiv preprint arXiv:2402.17834}} (\bibinfo{year}{2024}).
\newblock


\bibitem[Ben~Allal et~al\mbox{.}(2024)]%
        {benallal2024smollmcorpus}
\bibfield{author}{\bibinfo{person}{Loubna Ben~Allal}, \bibinfo{person}{Anton Lozhkov}, \bibinfo{person}{Guilherme Penedo}, \bibinfo{person}{Thomas Wolf}, {and} \bibinfo{person}{Leandro von Werra}.} \bibinfo{year}{2024}\natexlab{}.
\newblock \bibinfo{booktitle}{\emph{SmolLM-Corpus}}.
\newblock
\urldef\tempurl%
\url{https://huggingface.co/datasets/HuggingFaceTB/smollm-corpus}
\showURL{%
\tempurl}


\bibitem[Bergner et~al\mbox{.}(2024)]%
        {bergner2024think}
\bibfield{author}{\bibinfo{person}{Benjamin Bergner}, \bibinfo{person}{Andrii Skliar}, \bibinfo{person}{Amelie Royer}, \bibinfo{person}{Tijmen Blankevoort}, \bibinfo{person}{Yuki Asano}, {and} \bibinfo{person}{Babak~Ehteshami Bejnordi}.} \bibinfo{year}{2024}\natexlab{}.
\newblock \showarticletitle{Think Big, Generate Quick: LLM-to-SLM for Fast Autoregressive Decoding}.
\newblock \bibinfo{journal}{\emph{arXiv preprint arXiv:2402.16844}} (\bibinfo{year}{2024}).
\newblock


\bibitem[Bhan et~al\mbox{.}(2024)]%
        {bhan2024self}
\bibfield{author}{\bibinfo{person}{Milan Bhan}, \bibinfo{person}{Jean-Noel Vittaut}, \bibinfo{person}{Nicolas Chesneau}, {and} \bibinfo{person}{Marie-Jeanne Lesot}.} \bibinfo{year}{2024}\natexlab{}.
\newblock \showarticletitle{Self-AMPLIFY: Improving Small Language Models with Self Post Hoc Explanations}.
\newblock \bibinfo{journal}{\emph{arXiv preprint arXiv:2402.12038}} (\bibinfo{year}{2024}).
\newblock


\bibitem[Bi et~al\mbox{.}(2023)]%
        {bi2023oceangpt}
\bibfield{author}{\bibinfo{person}{Zhen Bi}, \bibinfo{person}{Ningyu Zhang}, \bibinfo{person}{Yida Xue}, \bibinfo{person}{Yixin Ou}, \bibinfo{person}{Daxiong Ji}, \bibinfo{person}{Guozhou Zheng}, {and} \bibinfo{person}{Huajun Chen}.} \bibinfo{year}{2023}\natexlab{}.
\newblock \showarticletitle{Oceangpt: A large language model for ocean science tasks}.
\newblock \bibinfo{journal}{\emph{arXiv preprint arXiv:2310.02031}} (\bibinfo{year}{2023}).
\newblock


\bibitem[Biderman et~al\mbox{.}(2023)]%
        {biderman2023pythiasuiteanalyzinglarge}
\bibfield{author}{\bibinfo{person}{Stella Biderman}, \bibinfo{person}{Hailey Schoelkopf}, \bibinfo{person}{Quentin Anthony}, \bibinfo{person}{Herbie Bradley}, \bibinfo{person}{Kyle O'Brien}, \bibinfo{person}{Eric Hallahan}, \bibinfo{person}{Mohammad~Aflah Khan}, \bibinfo{person}{Shivanshu Purohit}, \bibinfo{person}{USVSN~Sai Prashanth}, \bibinfo{person}{Edward Raff}, \bibinfo{person}{Aviya Skowron}, \bibinfo{person}{Lintang Sutawika}, {and} \bibinfo{person}{Oskar van~der Wal}.} \bibinfo{year}{2023}\natexlab{}.
\newblock \bibinfo{title}{Pythia: A Suite for Analyzing Large Language Models Across Training and Scaling}.
\newblock
\newblock
\showeprint[arxiv]{2304.01373}~[cs.CL]
\urldef\tempurl%
\url{https://arxiv.org/abs/2304.01373}
\showURL{%
\tempurl}


\bibitem[Bisk et~al\mbox{.}(2020)]%
        {bisk2020piqa}
\bibfield{author}{\bibinfo{person}{Yonatan Bisk}, \bibinfo{person}{Rowan Zellers}, \bibinfo{person}{Jianfeng Gao}, \bibinfo{person}{Yejin Choi}, {et~al\mbox{.}}} \bibinfo{year}{2020}\natexlab{}.
\newblock \showarticletitle{Piqa: Reasoning about physical commonsense in natural language}. In \bibinfo{booktitle}{\emph{Proceedings of the AAAI conference on artificial intelligence}}, Vol.~\bibinfo{volume}{34}. \bibinfo{pages}{7432--7439}.
\newblock


\bibitem[Black et~al\mbox{.}(2021)]%
        {gpt-neo}
\bibfield{author}{\bibinfo{person}{Sid Black}, \bibinfo{person}{Leo Gao}, \bibinfo{person}{Phil Wang}, \bibinfo{person}{Connor Leahy}, {and} \bibinfo{person}{Stella Biderman}.} \bibinfo{year}{2021}\natexlab{}.
\newblock \bibinfo{booktitle}{\emph{{GPT-Neo: Large Scale Autoregressive Language Modeling with Mesh-Tensorflow}}}.
\newblock
\urldef\tempurl%
\url{https://doi.org/10.5281/zenodo.5297715}
\showDOI{\tempurl}
\newblock
\shownote{{If you use this software, please cite it using these metadata.}}.


\bibitem[Bolton et~al\mbox{.}(2024b)]%
        {bolton2024biomedlm}
\bibfield{author}{\bibinfo{person}{Elliot Bolton}, \bibinfo{person}{Abhinav Venigalla}, \bibinfo{person}{Michihiro Yasunaga}, \bibinfo{person}{David Hall}, \bibinfo{person}{Betty Xiong}, \bibinfo{person}{Tony Lee}, \bibinfo{person}{Roxana Daneshjou}, \bibinfo{person}{Jonathan Frankle}, \bibinfo{person}{Percy Liang}, \bibinfo{person}{Michael Carbin}, {et~al\mbox{.}}} \bibinfo{year}{2024}\natexlab{b}.
\newblock \showarticletitle{Biomedlm: A 2.7 b parameter language model trained on biomedical text}.
\newblock \bibinfo{journal}{\emph{arXiv preprint arXiv:2403.18421}} (\bibinfo{year}{2024}).
\newblock


\bibitem[Bolton et~al\mbox{.}(2024a)]%
        {bolton2024rambla}
\bibfield{author}{\bibinfo{person}{William~James Bolton}, \bibinfo{person}{Rafael Poyiadzi}, \bibinfo{person}{Edward~R Morrell}, \bibinfo{person}{Gabriela van Bergen~Gonzalez Bueno}, {and} \bibinfo{person}{Lea Goetz}.} \bibinfo{year}{2024}\natexlab{a}.
\newblock \showarticletitle{RAmBLA: A Framework for Evaluating the Reliability of LLMs as Assistants in the Biomedical Domain}.
\newblock \bibinfo{journal}{\emph{arXiv preprint arXiv:2403.14578}} (\bibinfo{year}{2024}).
\newblock


\bibitem[Bonifacio et~al\mbox{.}(2022)]%
        {bonifacio2022inpars}
\bibfield{author}{\bibinfo{person}{Luiz Bonifacio}, \bibinfo{person}{Hugo Abonizio}, \bibinfo{person}{Marzieh Fadaee}, {and} \bibinfo{person}{Rodrigo Nogueira}.} \bibinfo{year}{2022}\natexlab{}.
\newblock \showarticletitle{Inpars: Data augmentation for information retrieval using large language models}.
\newblock \bibinfo{journal}{\emph{arXiv preprint arXiv:2202.05144}} (\bibinfo{year}{2022}).
\newblock


\bibitem[Brown et~al\mbox{.}(2020)]%
        {NEURIPS2020_1457c0d6}
\bibfield{author}{\bibinfo{person}{Tom Brown}, \bibinfo{person}{Benjamin Mann}, \bibinfo{person}{Nick Ryder}, \bibinfo{person}{Melanie Subbiah}, \bibinfo{person}{Jared~D Kaplan}, \bibinfo{person}{Prafulla Dhariwal}, \bibinfo{person}{Arvind Neelakantan}, \bibinfo{person}{Pranav Shyam}, \bibinfo{person}{Girish Sastry}, \bibinfo{person}{Amanda Askell}, \bibinfo{person}{Sandhini Agarwal}, \bibinfo{person}{Ariel Herbert-Voss}, \bibinfo{person}{Gretchen Krueger}, \bibinfo{person}{Tom Henighan}, \bibinfo{person}{Rewon Child}, \bibinfo{person}{Aditya Ramesh}, \bibinfo{person}{Daniel Ziegler}, \bibinfo{person}{Jeffrey Wu}, \bibinfo{person}{Clemens Winter}, \bibinfo{person}{Chris Hesse}, \bibinfo{person}{Mark Chen}, \bibinfo{person}{Eric Sigler}, \bibinfo{person}{Mateusz Litwin}, \bibinfo{person}{Scott Gray}, \bibinfo{person}{Benjamin Chess}, \bibinfo{person}{Jack Clark}, \bibinfo{person}{Christopher Berner}, \bibinfo{person}{Sam McCandlish}, \bibinfo{person}{Alec Radford}, \bibinfo{person}{Ilya Sutskever}, {and}
  \bibinfo{person}{Dario Amodei}.} \bibinfo{year}{2020}\natexlab{}.
\newblock \showarticletitle{Language Models are Few-Shot Learners}. In \bibinfo{booktitle}{\emph{Advances in Neural Information Processing Systems}}, \bibfield{editor}{\bibinfo{person}{H.~Larochelle}, \bibinfo{person}{M.~Ranzato}, \bibinfo{person}{R.~Hadsell}, \bibinfo{person}{M.F. Balcan}, {and} \bibinfo{person}{H.~Lin}} (Eds.), Vol.~\bibinfo{volume}{33}. \bibinfo{publisher}{Curran Associates, Inc.}, \bibinfo{pages}{1877--1901}.
\newblock
\urldef\tempurl%
\url{https://proceedings.neurips.cc/paper_files/paper/2020/file/1457c0d6bfcb4967418bfb8ac142f64a-Paper.pdf}
\showURL{%
\tempurl}


\bibitem[Bulusu et~al\mbox{.}(2020)]%
        {bulusu2020anomalous}
\bibfield{author}{\bibinfo{person}{Saikiran Bulusu}, \bibinfo{person}{Bhavya Kailkhura}, \bibinfo{person}{Bo Li}, \bibinfo{person}{Pramod~K Varshney}, {and} \bibinfo{person}{Dawn Song}.} \bibinfo{year}{2020}\natexlab{}.
\newblock \showarticletitle{Anomalous example detection in deep learning: A survey}.
\newblock \bibinfo{journal}{\emph{IEEE Access}}  \bibinfo{volume}{8} (\bibinfo{year}{2020}), \bibinfo{pages}{132330--132347}.
\newblock


\bibitem[Cai et~al\mbox{.}(2024b)]%
        {cai2024survey}
\bibfield{author}{\bibinfo{person}{Weilin Cai}, \bibinfo{person}{Juyong Jiang}, \bibinfo{person}{Fan Wang}, \bibinfo{person}{Jing Tang}, \bibinfo{person}{Sunghun Kim}, {and} \bibinfo{person}{Jiayi Huang}.} \bibinfo{year}{2024}\natexlab{b}.
\newblock \showarticletitle{A survey on mixture of experts}.
\newblock \bibinfo{journal}{\emph{Authorea Preprints}} (\bibinfo{year}{2024}).
\newblock


\bibitem[Cai et~al\mbox{.}(2024a)]%
        {cai2024internlm2}
\bibfield{author}{\bibinfo{person}{Zheng Cai}, \bibinfo{person}{Maosong Cao}, \bibinfo{person}{Haojiong Chen}, \bibinfo{person}{Kai Chen}, \bibinfo{person}{Keyu Chen}, \bibinfo{person}{Xin Chen}, \bibinfo{person}{Xun Chen}, \bibinfo{person}{Zehui Chen}, \bibinfo{person}{Zhi Chen}, \bibinfo{person}{Pei Chu}, {et~al\mbox{.}}} \bibinfo{year}{2024}\natexlab{a}.
\newblock \showarticletitle{Internlm2 technical report}.
\newblock \bibinfo{journal}{\emph{arXiv preprint arXiv:2403.17297}} (\bibinfo{year}{2024}).
\newblock


\bibitem[Carlini et~al\mbox{.}(2021)]%
        {carlini2021extracting}
\bibfield{author}{\bibinfo{person}{Nicholas Carlini}, \bibinfo{person}{Florian Tramer}, \bibinfo{person}{Eric Wallace}, \bibinfo{person}{Matthew Jagielski}, \bibinfo{person}{Ariel Herbert-Voss}, \bibinfo{person}{Katherine Lee}, \bibinfo{person}{Adam Roberts}, \bibinfo{person}{Tom Brown}, \bibinfo{person}{Dawn Song}, \bibinfo{person}{Ulfar Erlingsson}, {et~al\mbox{.}}} \bibinfo{year}{2021}\natexlab{}.
\newblock \showarticletitle{Extracting training data from large language models}. In \bibinfo{booktitle}{\emph{30th USENIX Security Symposium (USENIX Security 21)}}. \bibinfo{pages}{2633--2650}.
\newblock


\bibitem[Carreira et~al\mbox{.}(2023)]%
        {carreira2023revolutionizingmobileinteractionenabling}
\bibfield{author}{\bibinfo{person}{Samuel Carreira}, \bibinfo{person}{Tomás Marques}, \bibinfo{person}{José Ribeiro}, {and} \bibinfo{person}{Carlos Grilo}.} \bibinfo{year}{2023}\natexlab{}.
\newblock \bibinfo{title}{Revolutionizing Mobile Interaction: Enabling a 3 Billion Parameter GPT LLM on Mobile}.
\newblock
\newblock
\showeprint[arxiv]{2310.01434}~[cs.CL]
\urldef\tempurl%
\url{https://arxiv.org/abs/2310.01434}
\showURL{%
\tempurl}


\bibitem[Casanueva et~al\mbox{.}(2020)]%
        {casanueva-etal-2020-efficient}
\bibfield{author}{\bibinfo{person}{I{\~n}igo Casanueva}, \bibinfo{person}{Tadas Tem{\v{c}}inas}, \bibinfo{person}{Daniela Gerz}, \bibinfo{person}{Matthew Henderson}, {and} \bibinfo{person}{Ivan Vuli{\'c}}.} \bibinfo{year}{2020}\natexlab{}.
\newblock \showarticletitle{Efficient Intent Detection with Dual Sentence Encoders}. In \bibinfo{booktitle}{\emph{Proceedings of the 2nd Workshop on Natural Language Processing for Conversational AI}}. \bibinfo{publisher}{Association for Computational Linguistics}, \bibinfo{address}{Online}, \bibinfo{pages}{38--45}.
\newblock


\bibitem[Chang et~al\mbox{.}({[n.\,d.]})]%
        {changpre}
\bibfield{author}{\bibinfo{person}{Wei-Cheng Chang}, \bibinfo{person}{X~Yu Felix}, \bibinfo{person}{Yin-Wen Chang}, \bibinfo{person}{Yiming Yang}, {and} \bibinfo{person}{Sanjiv Kumar}.} \bibinfo{year}{[n.\,d.]}\natexlab{}.
\newblock \showarticletitle{Pre-training Tasks for Embedding-based Large-scale Retrieval}. In \bibinfo{booktitle}{\emph{International Conference on Learning Representations}}.
\newblock


\bibitem[Chao et~al\mbox{.}(2024)]%
        {chao2024jailbreakbench}
\bibfield{author}{\bibinfo{person}{Patrick Chao}, \bibinfo{person}{Edoardo Debenedetti}, \bibinfo{person}{Alexander Robey}, \bibinfo{person}{Maksym Andriushchenko}, \bibinfo{person}{Francesco Croce}, \bibinfo{person}{Vikash Sehwag}, \bibinfo{person}{Edgar Dobriban}, \bibinfo{person}{Nicolas Flammarion}, \bibinfo{person}{George~J Pappas}, \bibinfo{person}{Florian Tramer}, {et~al\mbox{.}}} \bibinfo{year}{2024}\natexlab{}.
\newblock \showarticletitle{Jailbreakbench: An open robustness benchmark for jailbreaking large language models}.
\newblock \bibinfo{journal}{\emph{arXiv preprint arXiv:2404.01318}} (\bibinfo{year}{2024}).
\newblock


\bibitem[Chen et~al\mbox{.}(2024c)]%
        {chen2024improving}
\bibfield{author}{\bibinfo{person}{Dong Chen}, \bibinfo{person}{Shuo Zhang}, \bibinfo{person}{Yueting Zhuang}, \bibinfo{person}{Siliang Tang}, \bibinfo{person}{Qidong Liu}, \bibinfo{person}{Hua Wang}, {and} \bibinfo{person}{Mingliang Xu}.} \bibinfo{year}{2024}\natexlab{c}.
\newblock \showarticletitle{Improving Large Models with Small models: Lower Costs and Better Performance}.
\newblock \bibinfo{journal}{\emph{arXiv preprint arXiv:2406.15471}} (\bibinfo{year}{2024}).
\newblock


\bibitem[Chen et~al\mbox{.}(2024d)]%
        {chen2024data}
\bibfield{author}{\bibinfo{person}{Dong Chen}, \bibinfo{person}{Yueting Zhuang}, \bibinfo{person}{Shuo Zhang}, \bibinfo{person}{Jinfeng Liu}, \bibinfo{person}{Su Dong}, {and} \bibinfo{person}{Siliang Tang}.} \bibinfo{year}{2024}\natexlab{d}.
\newblock \showarticletitle{Data Shunt: Collaboration of Small and Large Models for Lower Costs and Better Performance}. In \bibinfo{booktitle}{\emph{Proceedings of the AAAI Conference on Artificial Intelligence}}, Vol.~\bibinfo{volume}{38}. \bibinfo{pages}{11249--11257}.
\newblock


\bibitem[Chen et~al\mbox{.}(2023c)]%
        {chen2023mcc}
\bibfield{author}{\bibinfo{person}{Hongzhan Chen}, \bibinfo{person}{Siyue Wu}, \bibinfo{person}{Xiaojun Quan}, \bibinfo{person}{Rui Wang}, \bibinfo{person}{Ming Yan}, {and} \bibinfo{person}{Ji Zhang}.} \bibinfo{year}{2023}\natexlab{c}.
\newblock \showarticletitle{MCC-KD: Multi-CoT Consistent Knowledge Distillation}. In \bibinfo{booktitle}{\emph{Findings of the Association for Computational Linguistics: EMNLP 2023}}. \bibinfo{pages}{6805--6820}.
\newblock


\bibitem[Chen and Varoquaux(2024)]%
        {chen2024role}
\bibfield{author}{\bibinfo{person}{Lihu Chen} {and} \bibinfo{person}{Ga{\"e}l Varoquaux}.} \bibinfo{year}{2024}\natexlab{}.
\newblock \showarticletitle{What is the role of small models in the llm era: A survey}.
\newblock \bibinfo{journal}{\emph{arXiv preprint arXiv:2409.06857}} (\bibinfo{year}{2024}).
\newblock


\bibitem[Chen et~al\mbox{.}(2021b)]%
        {chen2021evaluating}
\bibfield{author}{\bibinfo{person}{Mark Chen}, \bibinfo{person}{Jerry Tworek}, \bibinfo{person}{Heewoo Jun}, \bibinfo{person}{Qiming Yuan}, \bibinfo{person}{Henrique Ponde De~Oliveira Pinto}, \bibinfo{person}{Jared Kaplan}, \bibinfo{person}{Harri Edwards}, \bibinfo{person}{Yuri Burda}, \bibinfo{person}{Nicholas Joseph}, \bibinfo{person}{Greg Brockman}, {et~al\mbox{.}}} \bibinfo{year}{2021}\natexlab{b}.
\newblock \showarticletitle{Evaluating large language models trained on code}.
\newblock \bibinfo{journal}{\emph{arXiv preprint arXiv:2107.03374}} (\bibinfo{year}{2021}).
\newblock


\bibitem[Chen et~al\mbox{.}(2023b)]%
        {chen2023lorashear}
\bibfield{author}{\bibinfo{person}{Tianyi Chen}, \bibinfo{person}{Tianyu Ding}, \bibinfo{person}{Badal Yadav}, \bibinfo{person}{Ilya Zharkov}, {and} \bibinfo{person}{Luming Liang}.} \bibinfo{year}{2023}\natexlab{b}.
\newblock \showarticletitle{Lorashear: Efficient large language model structured pruning and knowledge recovery}.
\newblock \bibinfo{journal}{\emph{arXiv preprint arXiv:2310.18356}} (\bibinfo{year}{2023}).
\newblock


\bibitem[Chen et~al\mbox{.}(2024a)]%
        {chen2024octopusondevicelanguagemodel}
\bibfield{author}{\bibinfo{person}{Wei Chen}, \bibinfo{person}{Zhiyuan Li}, {and} \bibinfo{person}{Mingyuan Ma}.} \bibinfo{year}{2024}\natexlab{a}.
\newblock \bibinfo{title}{Octopus: On-device language model for function calling of software APIs}.
\newblock
\newblock
\showeprint[arxiv]{2404.01549}~[cs.CL]
\urldef\tempurl%
\url{https://arxiv.org/abs/2404.01549}
\showURL{%
\tempurl}


\bibitem[Chen et~al\mbox{.}(2022b)]%
        {chen-etal-2022-textual}
\bibfield{author}{\bibinfo{person}{Yangyi Chen}, \bibinfo{person}{Fanchao Qi}, \bibinfo{person}{Hongcheng Gao}, \bibinfo{person}{Zhiyuan Liu}, {and} \bibinfo{person}{Maosong Sun}.} \bibinfo{year}{2022}\natexlab{b}.
\newblock \showarticletitle{Textual Backdoor Attacks Can Be More Harmful via Two Simple Tricks}. In \bibinfo{booktitle}{\emph{Proceedings of the 2022 Conference on Empirical Methods in Natural Language Processing, EMNLP}}. \bibinfo{pages}{11215--11221}.
\newblock


\bibitem[Chen et~al\mbox{.}(2024b)]%
        {chen2024a}
\bibfield{author}{\bibinfo{person}{Yangyi Chen}, \bibinfo{person}{Xingyao Wang}, \bibinfo{person}{Hao Peng}, {and} \bibinfo{person}{Heng Ji}.} \bibinfo{year}{2024}\natexlab{b}.
\newblock \showarticletitle{A Single Transformer for Scalable Vision-Language Modeling}.
\newblock \bibinfo{journal}{\emph{Transactions on Machine Learning Research}} (\bibinfo{year}{2024}).
\newblock
\showISSN{2835-8856}
\urldef\tempurl%
\url{https://openreview.net/forum?id=nuzFG0Rbhy}
\showURL{%
\tempurl}


\bibitem[Chen et~al\mbox{.}(2023a)]%
        {chen2023meditron}
\bibfield{author}{\bibinfo{person}{Zeming Chen}, \bibinfo{person}{Alejandro~Hern{\'a}ndez Cano}, \bibinfo{person}{Angelika Romanou}, \bibinfo{person}{Antoine Bonnet}, \bibinfo{person}{Kyle Matoba}, \bibinfo{person}{Francesco Salvi}, \bibinfo{person}{Matteo Pagliardini}, \bibinfo{person}{Simin Fan}, \bibinfo{person}{Andreas K{\"o}pf}, \bibinfo{person}{Amirkeivan Mohtashami}, {et~al\mbox{.}}} \bibinfo{year}{2023}\natexlab{a}.
\newblock \showarticletitle{MEDITRON-70B: Scaling Medical Pretraining for Large Language Models}.
\newblock \bibinfo{journal}{\emph{arXiv preprint arXiv:2311.16079}} (\bibinfo{year}{2023}).
\newblock


\bibitem[Chen et~al\mbox{.}(2021a)]%
        {chen2021finqa}
\bibfield{author}{\bibinfo{person}{Zhiyu Chen}, \bibinfo{person}{Wenhu Chen}, \bibinfo{person}{Charese Smiley}, \bibinfo{person}{Sameena Shah}, \bibinfo{person}{Iana Borova}, \bibinfo{person}{Dylan Langdon}, \bibinfo{person}{Reema Moussa}, \bibinfo{person}{Matt Beane}, \bibinfo{person}{Ting-Hao Huang}, \bibinfo{person}{Bryan~R Routledge}, {et~al\mbox{.}}} \bibinfo{year}{2021}\natexlab{a}.
\newblock \showarticletitle{FinQA: A Dataset of Numerical Reasoning over Financial Data}. In \bibinfo{booktitle}{\emph{Proceedings of the 2021 Conference on Empirical Methods in Natural Language Processing}}. \bibinfo{pages}{3697--3711}.
\newblock


\bibitem[Chen et~al\mbox{.}(2022a)]%
        {chen2022convfinqa}
\bibfield{author}{\bibinfo{person}{Zhiyu Chen}, \bibinfo{person}{Shiyang Li}, \bibinfo{person}{Charese Smiley}, \bibinfo{person}{Zhiqiang Ma}, \bibinfo{person}{Sameena Shah}, {and} \bibinfo{person}{William~Yang Wang}.} \bibinfo{year}{2022}\natexlab{a}.
\newblock \showarticletitle{ConvFinQA: Exploring the Chain of Numerical Reasoning in Conversational Finance Question Answering}. In \bibinfo{booktitle}{\emph{Proceedings of the 2022 Conference on Empirical Methods in Natural Language Processing}}. \bibinfo{pages}{6279--6292}.
\newblock


\bibitem[Cheng et~al\mbox{.}(2024)]%
        {cheng2024small}
\bibfield{author}{\bibinfo{person}{Xiaoxue Cheng}, \bibinfo{person}{Junyi Li}, \bibinfo{person}{Wayne~Xin Zhao}, \bibinfo{person}{Hongzhi Zhang}, \bibinfo{person}{Fuzheng Zhang}, \bibinfo{person}{Di Zhang}, \bibinfo{person}{Kun Gai}, {and} \bibinfo{person}{Ji-Rong Wen}.} \bibinfo{year}{2024}\natexlab{}.
\newblock \showarticletitle{Small Agent Can Also Rock! Empowering Small Language Models as Hallucination Detector}.
\newblock \bibinfo{journal}{\emph{arXiv preprint arXiv:2406.11277}} (\bibinfo{year}{2024}).
\newblock


\bibitem[Chern et~al\mbox{.}(2024)]%
        {chern2024behonest}
\bibfield{author}{\bibinfo{person}{Steffi Chern}, \bibinfo{person}{Zhulin Hu}, \bibinfo{person}{Yuqing Yang}, \bibinfo{person}{Ethan Chern}, \bibinfo{person}{Yuan Guo}, \bibinfo{person}{Jiahe Jin}, \bibinfo{person}{Binjie Wang}, {and} \bibinfo{person}{Pengfei Liu}.} \bibinfo{year}{2024}\natexlab{}.
\newblock \showarticletitle{BeHonest: Benchmarking Honesty of Large Language Models}.
\newblock \bibinfo{journal}{\emph{arXiv preprint arXiv:2406.13261}} (\bibinfo{year}{2024}).
\newblock


\bibitem[Chia et~al\mbox{.}(2024)]%
        {chia2024instructeval}
\bibfield{author}{\bibinfo{person}{Yew~Ken Chia}, \bibinfo{person}{Pengfei Hong}, \bibinfo{person}{Lidong Bing}, {and} \bibinfo{person}{Soujanya Poria}.} \bibinfo{year}{2024}\natexlab{}.
\newblock \showarticletitle{InstructEval: Towards Holistic Evaluation of Instruction-Tuned Large Language Models}. In \bibinfo{booktitle}{\emph{Proceedings of the First edition of the Workshop on the Scaling Behavior of Large Language Models (SCALE-LLM 2024)}}. \bibinfo{pages}{35--64}.
\newblock


\bibitem[Chiang et~al\mbox{.}(2023)]%
        {vicuna2023}
\bibfield{author}{\bibinfo{person}{Wei-Lin Chiang}, \bibinfo{person}{Zhuohan Li}, \bibinfo{person}{Zi Lin}, \bibinfo{person}{Ying Sheng}, \bibinfo{person}{Zhanghao Wu}, \bibinfo{person}{Hao Zhang}, \bibinfo{person}{Lianmin Zheng}, \bibinfo{person}{Siyuan Zhuang}, \bibinfo{person}{Yonghao Zhuang}, \bibinfo{person}{Joseph~E. Gonzalez}, \bibinfo{person}{Ion Stoica}, {and} \bibinfo{person}{Eric~P. Xing}.} \bibinfo{year}{2023}\natexlab{}.
\newblock \bibinfo{title}{Vicuna: An Open-Source Chatbot Impressing GPT-4 with 90\%* ChatGPT Quality}.
\newblock
\newblock
\urldef\tempurl%
\url{https://lmsys.org/blog/2023-03-30-vicuna/}
\showURL{%
\tempurl}


\bibitem[Cho et~al\mbox{.}(2024)]%
        {cho2024heterogeneouslorafederatedfinetuning}
\bibfield{author}{\bibinfo{person}{Yae~Jee Cho}, \bibinfo{person}{Luyang Liu}, \bibinfo{person}{Zheng Xu}, \bibinfo{person}{Aldi Fahrezi}, {and} \bibinfo{person}{Gauri Joshi}.} \bibinfo{year}{2024}\natexlab{}.
\newblock \bibinfo{title}{Heterogeneous LoRA for Federated Fine-tuning of On-Device Foundation Models}.
\newblock
\newblock
\showeprint[arxiv]{2401.06432}~[cs.LG]
\urldef\tempurl%
\url{https://arxiv.org/abs/2401.06432}
\showURL{%
\tempurl}


\bibitem[Chu et~al\mbox{.}(2022)]%
        {chu2022h}
\bibfield{author}{\bibinfo{person}{Xiaokai Chu}, \bibinfo{person}{Jiashu Zhao}, \bibinfo{person}{Lixin Zou}, {and} \bibinfo{person}{Dawei Yin}.} \bibinfo{year}{2022}\natexlab{}.
\newblock \showarticletitle{H-ERNIE: A multi-granularity pre-trained language model for web search}. In \bibinfo{booktitle}{\emph{Proceedings of the 45th International ACM SIGIR conference on research and development in information retrieval}}. \bibinfo{pages}{1478--1489}.
\newblock


\bibitem[Chung et~al\mbox{.}(2024)]%
        {chung2024scaling}
\bibfield{author}{\bibinfo{person}{Hyung~Won Chung}, \bibinfo{person}{Le Hou}, \bibinfo{person}{Shayne Longpre}, \bibinfo{person}{Barret Zoph}, \bibinfo{person}{Yi Tay}, \bibinfo{person}{William Fedus}, \bibinfo{person}{Yunxuan Li}, \bibinfo{person}{Xuezhi Wang}, \bibinfo{person}{Mostafa Dehghani}, \bibinfo{person}{Siddhartha Brahma}, {et~al\mbox{.}}} \bibinfo{year}{2024}\natexlab{}.
\newblock \showarticletitle{Scaling instruction-finetuned language models}.
\newblock \bibinfo{journal}{\emph{Journal of Machine Learning Research}} \bibinfo{volume}{25}, \bibinfo{number}{70} (\bibinfo{year}{2024}), \bibinfo{pages}{1--53}.
\newblock


\bibitem[Clark et~al\mbox{.}(2018)]%
        {clark2018think}
\bibfield{author}{\bibinfo{person}{Peter Clark}, \bibinfo{person}{Isaac Cowhey}, \bibinfo{person}{Oren Etzioni}, \bibinfo{person}{Tushar Khot}, \bibinfo{person}{Ashish Sabharwal}, \bibinfo{person}{Carissa Schoenick}, {and} \bibinfo{person}{Oyvind Tafjord}.} \bibinfo{year}{2018}\natexlab{}.
\newblock \showarticletitle{Think you have solved question answering? try arc, the ai2 reasoning challenge}.
\newblock \bibinfo{journal}{\emph{arXiv preprint arXiv:1803.05457}} (\bibinfo{year}{2018}).
\newblock


\bibitem[Cobbe et~al\mbox{.}(2021)]%
        {cobbe2021training}
\bibfield{author}{\bibinfo{person}{Karl Cobbe}, \bibinfo{person}{Vineet Kosaraju}, \bibinfo{person}{Mohammad Bavarian}, \bibinfo{person}{Mark Chen}, \bibinfo{person}{Heewoo Jun}, \bibinfo{person}{Lukasz Kaiser}, \bibinfo{person}{Matthias Plappert}, \bibinfo{person}{Jerry Tworek}, \bibinfo{person}{Jacob Hilton}, \bibinfo{person}{Reiichiro Nakano}, {et~al\mbox{.}}} \bibinfo{year}{2021}\natexlab{}.
\newblock \showarticletitle{Training verifiers to solve math word problems}.
\newblock \bibinfo{journal}{\emph{arXiv preprint arXiv:2110.14168}} (\bibinfo{year}{2021}).
\newblock


\bibitem[Computer(2023)]%
        {together2023redpajama}
\bibfield{author}{\bibinfo{person}{Together Computer}.} \bibinfo{year}{2023}\natexlab{}.
\newblock \bibinfo{booktitle}{\emph{RedPajama: an Open Dataset for Training Large Language Models}}.
\newblock
\urldef\tempurl%
\url{https://github.com/togethercomputer/RedPajama-Data}
\showURL{%
\tempurl}


\bibitem[Conover et~al\mbox{.}(2023)]%
        {DatabricksBlog2023DollyV2}
\bibfield{author}{\bibinfo{person}{Mike Conover}, \bibinfo{person}{Matt Hayes}, \bibinfo{person}{Ankit Mathur}, \bibinfo{person}{Jianwei Xie}, \bibinfo{person}{Jun Wan}, \bibinfo{person}{Sam Shah}, \bibinfo{person}{Ali Ghodsi}, \bibinfo{person}{Patrick Wendell}, \bibinfo{person}{Matei Zaharia}, {and} \bibinfo{person}{Reynold Xin}.} \bibinfo{year}{2023}\natexlab{}.
\newblock \bibinfo{booktitle}{\emph{Free Dolly: Introducing the World's First Truly Open Instruction-Tuned LLM}}.
\newblock
\urldef\tempurl%
\url{https://www.databricks.com/blog/2023/04/12/dolly-first-open-commercially-viable-instruction-tuned-llm}
\showURL{%
\tempurl}


\bibitem[Craswell et~al\mbox{.}(2021)]%
        {trec2020}
\bibfield{author}{\bibinfo{person}{Nick Craswell}, \bibinfo{person}{Bhaskar Mitra}, \bibinfo{person}{Emine Yilmaz}, {and} \bibinfo{person}{Daniel Campos}.} \bibinfo{year}{2021}\natexlab{}.
\newblock \showarticletitle{Overview of the {TREC} 2020 deep learning track}.
\newblock \bibinfo{journal}{\emph{CoRR}}  \bibinfo{volume}{abs/2102.07662} (\bibinfo{year}{2021}).
\newblock
\showeprint[arXiv]{2102.07662}
\urldef\tempurl%
\url{https://arxiv.org/abs/2102.07662}
\showURL{%
\tempurl}


\bibitem[Cui et~al\mbox{.}(2024)]%
        {cui2024or}
\bibfield{author}{\bibinfo{person}{Justin Cui}, \bibinfo{person}{Wei-Lin Chiang}, \bibinfo{person}{Ion Stoica}, {and} \bibinfo{person}{Cho-Jui Hsieh}.} \bibinfo{year}{2024}\natexlab{}.
\newblock \showarticletitle{OR-Bench: An Over-Refusal Benchmark for Large Language Models}.
\newblock \bibinfo{journal}{\emph{arXiv preprint arXiv:2405.20947}} (\bibinfo{year}{2024}).
\newblock


\bibitem[Cui et~al\mbox{.}(2023)]%
        {cui2023fft}
\bibfield{author}{\bibinfo{person}{Shiyao Cui}, \bibinfo{person}{Zhenyu Zhang}, \bibinfo{person}{Yilong Chen}, \bibinfo{person}{Wenyuan Zhang}, \bibinfo{person}{Tianyun Liu}, \bibinfo{person}{Siqi Wang}, {and} \bibinfo{person}{Tingwen Liu}.} \bibinfo{year}{2023}\natexlab{}.
\newblock \showarticletitle{Fft: Towards harmlessness evaluation and analysis for llms with factuality, fairness, toxicity}.
\newblock \bibinfo{journal}{\emph{arXiv preprint arXiv:2311.18580}} (\bibinfo{year}{2023}).
\newblock


\bibitem[Daniele and Suphavadeeprasit(2023)]%
        {daniele2023amplify-instruct}
\bibfield{author}{\bibinfo{person}{Luigi Daniele} {and} \bibinfo{person}{Suphavadeeprasit}.} \bibinfo{year}{2023}\natexlab{}.
\newblock \showarticletitle{Amplify-Instruct: Synthetically Generated Diverse Multi-turn Conversations for efficient LLM Training.}
\newblock \bibinfo{journal}{\emph{arXiv preprint arXiv:(coming soon)}} (\bibinfo{year}{2023}).
\newblock
\urldef\tempurl%
\url{https://huggingface.co/datasets/LDJnr/Capybara}
\showURL{%
\tempurl}


\bibitem[Dao({[n.\,d.]})]%
        {daoflashattention}
\bibfield{author}{\bibinfo{person}{Tri Dao}.} \bibinfo{year}{[n.\,d.]}\natexlab{}.
\newblock \showarticletitle{FlashAttention-2: Faster Attention with Better Parallelism and Work Partitioning}. In \bibinfo{booktitle}{\emph{The Twelfth International Conference on Learning Representations}}.
\newblock


\bibitem[Dao et~al\mbox{.}(2022)]%
        {dao2022flashattention}
\bibfield{author}{\bibinfo{person}{Tri Dao}, \bibinfo{person}{Dan Fu}, \bibinfo{person}{Stefano Ermon}, \bibinfo{person}{Atri Rudra}, {and} \bibinfo{person}{Christopher R{\'e}}.} \bibinfo{year}{2022}\natexlab{}.
\newblock \showarticletitle{Flashattention: Fast and memory-efficient exact attention with io-awareness}.
\newblock \bibinfo{journal}{\emph{Advances in Neural Information Processing Systems}}  \bibinfo{volume}{35} (\bibinfo{year}{2022}), \bibinfo{pages}{16344--16359}.
\newblock


\bibitem[Dao and Gu(2024)]%
        {dao2024transformers}
\bibfield{author}{\bibinfo{person}{Tri Dao} {and} \bibinfo{person}{Albert Gu}.} \bibinfo{year}{2024}\natexlab{}.
\newblock \showarticletitle{Transformers are SSMs: Generalized models and efficient algorithms through structured state space duality}.
\newblock \bibinfo{journal}{\emph{arXiv preprint arXiv:2405.21060}} (\bibinfo{year}{2024}).
\newblock


\bibitem[Das et~al\mbox{.}(2023)]%
        {das2023beyond}
\bibfield{author}{\bibinfo{person}{Rocktim~Jyoti Das}, \bibinfo{person}{Liqun Ma}, {and} \bibinfo{person}{Zhiqiang Shen}.} \bibinfo{year}{2023}\natexlab{}.
\newblock \showarticletitle{Beyond size: How gradients shape pruning decisions in large language models}.
\newblock \bibinfo{journal}{\emph{arXiv preprint arXiv:2311.04902}} (\bibinfo{year}{2023}).
\newblock


\bibitem[Dasgupta et~al\mbox{.}(2011)]%
        {dasgupta2011fast}
\bibfield{author}{\bibinfo{person}{Anirban Dasgupta}, \bibinfo{person}{Ravi Kumar}, {and} \bibinfo{person}{Tam{\'a}s Sarl{\'o}s}.} \bibinfo{year}{2011}\natexlab{}.
\newblock \showarticletitle{Fast locality-sensitive hashing}. In \bibinfo{booktitle}{\emph{Proceedings of the 17th ACM SIGKDD international conference on Knowledge discovery and data mining}}. \bibinfo{pages}{1073--1081}.
\newblock


\bibitem[Delobelle et~al\mbox{.}(2022)]%
        {delobelle2022measuring}
\bibfield{author}{\bibinfo{person}{Pieter Delobelle}, \bibinfo{person}{Ewoenam~Kwaku Tokpo}, \bibinfo{person}{Toon Calders}, {and} \bibinfo{person}{Bettina Berendt}.} \bibinfo{year}{2022}\natexlab{}.
\newblock \showarticletitle{Measuring fairness with biased rulers: A comparative study on bias metrics for pre-trained language models}. In \bibinfo{booktitle}{\emph{Proceedings of the 2022 Conference of the North American Chapter of the Association for Computational Linguistics}}. \bibinfo{pages}{1693--1706}.
\newblock


\bibitem[Deng et~al\mbox{.}(2023)]%
        {deng2023mutual}
\bibfield{author}{\bibinfo{person}{Yongheng Deng}, \bibinfo{person}{Ziqing Qiao}, \bibinfo{person}{Ju Ren}, \bibinfo{person}{Yang Liu}, {and} \bibinfo{person}{Yaoxue Zhang}.} \bibinfo{year}{2023}\natexlab{}.
\newblock \showarticletitle{Mutual enhancement of large and small language models with cross-silo knowledge transfer}.
\newblock \bibinfo{journal}{\emph{arXiv preprint arXiv:2312.05842}} (\bibinfo{year}{2023}).
\newblock


\bibitem[Dettmers et~al\mbox{.}(2022)]%
        {dettmers2022gptint}
\bibfield{author}{\bibinfo{person}{Tim Dettmers}, \bibinfo{person}{Mike Lewis}, \bibinfo{person}{Younes Belkada}, {and} \bibinfo{person}{Luke Zettlemoyer}.} \bibinfo{year}{2022}\natexlab{}.
\newblock \showarticletitle{{GPT}3.int8(): 8-bit Matrix Multiplication for Transformers at Scale}. In \bibinfo{booktitle}{\emph{Advances in Neural Information Processing Systems}}, \bibfield{editor}{\bibinfo{person}{Alice~H. Oh}, \bibinfo{person}{Alekh Agarwal}, \bibinfo{person}{Danielle Belgrave}, {and} \bibinfo{person}{Kyunghyun Cho}} (Eds.).
\newblock
\urldef\tempurl%
\url{https://openreview.net/forum?id=dXiGWqBoxaD}
\showURL{%
\tempurl}


\bibitem[Dettmers et~al\mbox{.}(2024)]%
        {dettmers2024qlora}
\bibfield{author}{\bibinfo{person}{Tim Dettmers}, \bibinfo{person}{Artidoro Pagnoni}, \bibinfo{person}{Ari Holtzman}, {and} \bibinfo{person}{Luke Zettlemoyer}.} \bibinfo{year}{2024}\natexlab{}.
\newblock \showarticletitle{Qlora: Efficient finetuning of quantized llms}.
\newblock \bibinfo{journal}{\emph{Advances in Neural Information Processing Systems}}  \bibinfo{volume}{36} (\bibinfo{year}{2024}).
\newblock


\bibitem[Dettmers and Zettlemoyer(2023)]%
        {dettmers2023case}
\bibfield{author}{\bibinfo{person}{Tim Dettmers} {and} \bibinfo{person}{Luke Zettlemoyer}.} \bibinfo{year}{2023}\natexlab{}.
\newblock \showarticletitle{The case for 4-bit precision: k-bit inference scaling laws}. In \bibinfo{booktitle}{\emph{International Conference on Machine Learning}}. PMLR, \bibinfo{pages}{7750--7774}.
\newblock


\bibitem[Devlin et~al\mbox{.}(2019)]%
        {devlin2019bert}
\bibfield{author}{\bibinfo{person}{Jacob Devlin}, \bibinfo{person}{Ming-Wei Chang}, \bibinfo{person}{Kenton Lee}, {and} \bibinfo{person}{Kristina Toutanova}.} \bibinfo{year}{2019}\natexlab{}.
\newblock \showarticletitle{BERT: Pre-training of Deep Bidirectional Transformers for Language Understanding}. In \bibinfo{booktitle}{\emph{Proceedings of the 2019 Conference of the North American Chapter of the Association for Computational Linguistics: Human Language Technologies, Volume 1 (Long and Short Papers)}}. \bibinfo{pages}{4171--4186}.
\newblock


\bibitem[Dey et~al\mbox{.}(2023)]%
        {dey2023cerebras}
\bibfield{author}{\bibinfo{person}{Nolan Dey}, \bibinfo{person}{Gurpreet Gosal}, \bibinfo{person}{Zhiming Chen}, \bibinfo{person}{Hemant Khachane}, \bibinfo{person}{William Marshall}, \bibinfo{person}{Ribhu Pathria}, \bibinfo{person}{Marvin Tom}, {and} \bibinfo{person}{Joel Hestness}.} \bibinfo{year}{2023}\natexlab{}.
\newblock \bibinfo{title}{Cerebras-GPT: Open Compute-Optimal Language Models Trained on the Cerebras Wafer-Scale Cluster. CoRR abs/2304.03208 (2023)}.
\newblock
\newblock


\bibitem[Diehl~Martinez et~al\mbox{.}(2024)]%
        {diehl-martinez-etal-2024-tending}
\bibfield{author}{\bibinfo{person}{Richard Diehl~Martinez}, \bibinfo{person}{Pietro Lesci}, {and} \bibinfo{person}{Paula Buttery}.} \bibinfo{year}{2024}\natexlab{}.
\newblock \showarticletitle{Tending Towards Stability: Convergence Challenges in Small Language Models}. In \bibinfo{booktitle}{\emph{Findings of the Association for Computational Linguistics: EMNLP 2024}}, \bibfield{editor}{\bibinfo{person}{Yaser Al-Onaizan}, \bibinfo{person}{Mohit Bansal}, {and} \bibinfo{person}{Yun-Nung Chen}} (Eds.). \bibinfo{publisher}{Association for Computational Linguistics}, \bibinfo{address}{Miami, Florida, USA}, \bibinfo{pages}{3275--3286}.
\newblock
\urldef\tempurl%
\url{https://doi.org/10.18653/v1/2024.findings-emnlp.187}
\showDOI{\tempurl}


\bibitem[Ding et~al\mbox{.}(2023)]%
        {ding2023enhancing}
\bibfield{author}{\bibinfo{person}{Ning Ding}, \bibinfo{person}{Yulin Chen}, \bibinfo{person}{Bokai Xu}, \bibinfo{person}{Yujia Qin}, \bibinfo{person}{Zhi Zheng}, \bibinfo{person}{Shengding Hu}, \bibinfo{person}{Zhiyuan Liu}, \bibinfo{person}{Maosong Sun}, {and} \bibinfo{person}{Bowen Zhou}.} \bibinfo{year}{2023}\natexlab{}.
\newblock \showarticletitle{Enhancing chat language models by scaling high-quality instructional conversations}.
\newblock \bibinfo{journal}{\emph{arXiv preprint arXiv:2305.14233}} (\bibinfo{year}{2023}).
\newblock


\bibitem[Ding(2024)]%
        {ding2024mobileagentenhancingmobilecontrol}
\bibfield{author}{\bibinfo{person}{Tinghe Ding}.} \bibinfo{year}{2024}\natexlab{}.
\newblock \bibinfo{title}{MobileAgent: enhancing mobile control via human-machine interaction and SOP integration}.
\newblock
\newblock
\showeprint[arxiv]{2401.04124}~[cs.HC]
\urldef\tempurl%
\url{https://arxiv.org/abs/2401.04124}
\showURL{%
\tempurl}


\bibitem[Dominguez-Olmedo et~al\mbox{.}(2023)]%
        {dominguez2023questioning}
\bibfield{author}{\bibinfo{person}{Ricardo Dominguez-Olmedo}, \bibinfo{person}{Moritz Hardt}, {and} \bibinfo{person}{Celestine Mendler-D{\"u}nner}.} \bibinfo{year}{2023}\natexlab{}.
\newblock \showarticletitle{Questioning the survey responses of large language models}.
\newblock \bibinfo{journal}{\emph{arXiv preprint arXiv:2306.07951}} (\bibinfo{year}{2023}).
\newblock


\bibitem[Dong et~al\mbox{.}(2023)]%
        {dong2023i3}
\bibfield{author}{\bibinfo{person}{Qian Dong}, \bibinfo{person}{Yiding Liu}, \bibinfo{person}{Qingyao Ai}, \bibinfo{person}{Haitao Li}, \bibinfo{person}{Shuaiqiang Wang}, \bibinfo{person}{Yiqun Liu}, \bibinfo{person}{Dawei Yin}, {and} \bibinfo{person}{Shaoping Ma}.} \bibinfo{year}{2023}\natexlab{}.
\newblock \showarticletitle{I3 retriever: incorporating implicit interaction in pre-trained language models for passage retrieval}. In \bibinfo{booktitle}{\emph{Proceedings of the 32nd ACM International Conference on Information and Knowledge Management}}. \bibinfo{pages}{441--451}.
\newblock


\bibitem[Dong et~al\mbox{.}(2024)]%
        {dong2024hymba}
\bibfield{author}{\bibinfo{person}{Xin Dong}, \bibinfo{person}{Yonggan Fu}, \bibinfo{person}{Shizhe Diao}, \bibinfo{person}{Wonmin Byeon}, \bibinfo{person}{Zijia Chen}, \bibinfo{person}{Ameya~Sunil Mahabaleshwarkar}, \bibinfo{person}{Shih-Yang Liu}, \bibinfo{person}{Matthijs Van~Keirsbilck}, \bibinfo{person}{Min-Hung Chen}, \bibinfo{person}{Yoshi Suhara}, {et~al\mbox{.}}} \bibinfo{year}{2024}\natexlab{}.
\newblock \showarticletitle{Hymba: A Hybrid-head Architecture for Small Language Models}.
\newblock \bibinfo{journal}{\emph{arXiv preprint arXiv:2411.13676}} (\bibinfo{year}{2024}).
\newblock


\bibitem[Dou et~al\mbox{.}(2023)]%
        {dou2023measurement}
\bibfield{author}{\bibinfo{person}{Jason~Xiaotian Dou}, \bibinfo{person}{Haiyi Mao}, \bibinfo{person}{Runxue Bao}, \bibinfo{person}{Paul~Pu Liang}, \bibinfo{person}{Xiaoqing Tan}, \bibinfo{person}{Shiyi Zhang}, \bibinfo{person}{Minxue Jia}, \bibinfo{person}{Pengfei Zhou}, {and} \bibinfo{person}{Zhi-Hong Mao}.} \bibinfo{year}{2023}\natexlab{}.
\newblock \showarticletitle{The Measurement of Knowledge in Knowledge Graphs}. In \bibinfo{booktitle}{\emph{Proceedings of the AAAI 2023 Workshop on Representation Learning for Responsible Human-Centric AI (R2HCAI)}}. Association for the Advancement of Artificial Intelligence (AAAI) Washington~….
\newblock


\bibitem[Du et~al\mbox{.}(2024)]%
        {du2024chinesetinyllmpretraining}
\bibfield{author}{\bibinfo{person}{Xinrun Du}, \bibinfo{person}{Zhouliang Yu}, \bibinfo{person}{Songyang Gao}, \bibinfo{person}{Ding Pan}, \bibinfo{person}{Yuyang Cheng}, \bibinfo{person}{Ziyang Ma}, \bibinfo{person}{Ruibin Yuan}, \bibinfo{person}{Xingwei Qu}, \bibinfo{person}{Jiaheng Liu}, \bibinfo{person}{Tianyu Zheng}, \bibinfo{person}{Xinchen Luo}, \bibinfo{person}{Guorui Zhou}, \bibinfo{person}{Wenhu Chen}, {and} \bibinfo{person}{Ge Zhang}.} \bibinfo{year}{2024}\natexlab{}.
\newblock \bibinfo{title}{Chinese Tiny LLM: Pretraining a Chinese-Centric Large Language Model}.
\newblock
\newblock
\showeprint[arxiv]{2404.04167}~[cs.CL]
\urldef\tempurl%
\url{https://arxiv.org/abs/2404.04167}
\showURL{%
\tempurl}


\bibitem[Du et~al\mbox{.}(2022)]%
        {du2022glm}
\bibfield{author}{\bibinfo{person}{Zhengxiao Du}, \bibinfo{person}{Yujie Qian}, \bibinfo{person}{Xiao Liu}, \bibinfo{person}{Ming Ding}, \bibinfo{person}{Jiezhong Qiu}, \bibinfo{person}{Zhilin Yang}, {and} \bibinfo{person}{Jie Tang}.} \bibinfo{year}{2022}\natexlab{}.
\newblock \showarticletitle{GLM: General Language Model Pretraining with Autoregressive Blank Infilling}. In \bibinfo{booktitle}{\emph{Proceedings of the 60th Annual Meeting of the Association for Computational Linguistics (Volume 1: Long Papers)}}. \bibinfo{pages}{320--335}.
\newblock


\bibitem[Dubey et~al\mbox{.}(2024)]%
        {dubey2024llama}
\bibfield{author}{\bibinfo{person}{Abhimanyu Dubey}, \bibinfo{person}{Abhinav Jauhri}, \bibinfo{person}{Abhinav Pandey}, \bibinfo{person}{Abhishek Kadian}, \bibinfo{person}{Ahmad Al-Dahle}, \bibinfo{person}{Aiesha Letman}, \bibinfo{person}{Akhil Mathur}, \bibinfo{person}{Alan Schelten}, \bibinfo{person}{Amy Yang}, \bibinfo{person}{Angela Fan}, {et~al\mbox{.}}} \bibinfo{year}{2024}\natexlab{}.
\newblock \showarticletitle{The llama 3 herd of models}.
\newblock \bibinfo{journal}{\emph{arXiv preprint arXiv:2407.21783}} (\bibinfo{year}{2024}).
\newblock


\bibitem[Egashira et~al\mbox{.}(2024)]%
        {egashira2024exploiting}
\bibfield{author}{\bibinfo{person}{Kazuki Egashira}, \bibinfo{person}{Mark Vero}, \bibinfo{person}{Robin Staab}, \bibinfo{person}{Jingxuan He}, {and} \bibinfo{person}{Martin Vechev}.} \bibinfo{year}{2024}\natexlab{}.
\newblock \showarticletitle{Exploiting LLM Quantization}.
\newblock \bibinfo{journal}{\emph{arXiv preprint arXiv:2405.18137}} (\bibinfo{year}{2024}).
\newblock


\bibitem[Eldan and Li(2023)]%
        {eldan2023tinystories}
\bibfield{author}{\bibinfo{person}{Ronen Eldan} {and} \bibinfo{person}{Yuanzhi Li}.} \bibinfo{year}{2023}\natexlab{}.
\newblock \showarticletitle{Tinystories: How small can language models be and still speak coherent english?}
\newblock \bibinfo{journal}{\emph{arXiv preprint arXiv:2305.07759}} (\bibinfo{year}{2023}).
\newblock


\bibitem[Elfwing et~al\mbox{.}(2018)]%
        {silu}
\bibfield{author}{\bibinfo{person}{Stefan Elfwing}, \bibinfo{person}{Eiji Uchibe}, {and} \bibinfo{person}{Kenji Doya}.} \bibinfo{year}{2018}\natexlab{}.
\newblock \showarticletitle{Sigmoid-weighted linear units for neural network function approximation in reinforcement learning}.
\newblock \bibinfo{journal}{\emph{Neural networks}}  \bibinfo{volume}{107} (\bibinfo{year}{2018}), \bibinfo{pages}{3--11}.
\newblock


\bibitem[Esiobu et~al\mbox{.}(2023)]%
        {esiobu2023robbie}
\bibfield{author}{\bibinfo{person}{David Esiobu}, \bibinfo{person}{Xiaoqing Tan}, \bibinfo{person}{Saghar Hosseini}, \bibinfo{person}{Megan Ung}, \bibinfo{person}{Yuchen Zhang}, \bibinfo{person}{Jude Fernandes}, \bibinfo{person}{Jane Dwivedi-Yu}, \bibinfo{person}{Eleonora Presani}, \bibinfo{person}{Adina Williams}, {and} \bibinfo{person}{Eric Smith}.} \bibinfo{year}{2023}\natexlab{}.
\newblock \showarticletitle{ROBBIE: Robust bias evaluation of large generative language models}. In \bibinfo{booktitle}{\emph{Proceedings of the 2023 Conference on Empirical Methods in Natural Language Processing}}. \bibinfo{pages}{3764--3814}.
\newblock
\urldef\tempurl%
\url{https://aclanthology.org/2023.emnlp-main.230}
\showURL{%
\tempurl}


\bibitem[Face(2024)]%
        {smovlm}
\bibfield{author}{\bibinfo{person}{Hugging Face}.} \bibinfo{year}{2024}\natexlab{}.
\newblock \bibinfo{booktitle}{\emph{SmolVLM - small yet mighty Vision Language Model}}.
\newblock
\urldef\tempurl%
\url{https://huggingface.co/blog/smolvlm}
\showURL{%
\tempurl}
\newblock
\shownote{Accessed: 2024-11-26}.


\bibitem[Fedus et~al\mbox{.}(2022)]%
        {fedus2022switch}
\bibfield{author}{\bibinfo{person}{William Fedus}, \bibinfo{person}{Barret Zoph}, {and} \bibinfo{person}{Noam Shazeer}.} \bibinfo{year}{2022}\natexlab{}.
\newblock \showarticletitle{Switch transformers: Scaling to trillion parameter models with simple and efficient sparsity}.
\newblock \bibinfo{journal}{\emph{Journal of Machine Learning Research}} \bibinfo{volume}{23}, \bibinfo{number}{120} (\bibinfo{year}{2022}), \bibinfo{pages}{1--39}.
\newblock


\bibitem[Feng et~al\mbox{.}(2023)]%
        {feng2023knowledge}
\bibfield{author}{\bibinfo{person}{Shangbin Feng}, \bibinfo{person}{Weijia Shi}, \bibinfo{person}{Yuyang Bai}, \bibinfo{person}{Vidhisha Balachandran}, \bibinfo{person}{Tianxing He}, {and} \bibinfo{person}{Yulia Tsvetkov}.} \bibinfo{year}{2023}\natexlab{}.
\newblock \showarticletitle{Knowledge Card: Filling LLMs' Knowledge Gaps with Plug-in Specialized Language Models}.
\newblock \bibinfo{journal}{\emph{arXiv preprint arXiv:2305.09955}} (\bibinfo{year}{2023}).
\newblock


\bibitem[Frantar and Alistarh(2023)]%
        {frantar2023sparsegpt}
\bibfield{author}{\bibinfo{person}{Elias Frantar} {and} \bibinfo{person}{Dan Alistarh}.} \bibinfo{year}{2023}\natexlab{}.
\newblock \showarticletitle{Sparsegpt: Massive language models can be accurately pruned in one-shot}. In \bibinfo{booktitle}{\emph{International Conference on Machine Learning}}. PMLR, \bibinfo{pages}{10323--10337}.
\newblock


\bibitem[Frantar et~al\mbox{.}(2023)]%
        {frantar2023gptq}
\bibfield{author}{\bibinfo{person}{Elias Frantar}, \bibinfo{person}{Saleh Ashkboos}, \bibinfo{person}{Torsten Hoefler}, {and} \bibinfo{person}{Dan Alistarh}.} \bibinfo{year}{2023}\natexlab{}.
\newblock \showarticletitle{GPTQ: Accurate Post-Training Quantization for Generative Pre-trained Transformers}. In \bibinfo{booktitle}{\emph{The Eleventh International Conference on Learning Representations}}.
\newblock


\bibitem[Fu and Khot(2022)]%
        {fu2022gptroadmap}
\bibfield{author}{\bibinfo{person}{Hao Fu, Yao;~Peng} {and} \bibinfo{person}{Tushar Khot}.} \bibinfo{year}{2022}\natexlab{}.
\newblock \showarticletitle{How does GPT Obtain its Ability? Tracing Emergent Abilities of Language Models to their Sources}.
\newblock \bibinfo{journal}{\emph{Yao Fu’s Notion}} (\bibinfo{date}{Dec} \bibinfo{year}{2022}).
\newblock
\urldef\tempurl%
\url{https://yaofu.notion.site/How-does-GPT-Obtain-its-Ability-Tracing-Emergent-Abilities-of-Language-Models-to-their-Sources-b9a57ac0fcf74f30a1ab9e3e36fa1dc1}
\showURL{%
\tempurl}


\bibitem[Fu et~al\mbox{.}(2023)]%
        {fu2023specializing}
\bibfield{author}{\bibinfo{person}{Yao Fu}, \bibinfo{person}{Hao Peng}, \bibinfo{person}{Litu Ou}, \bibinfo{person}{Ashish Sabharwal}, {and} \bibinfo{person}{Tushar Khot}.} \bibinfo{year}{2023}\natexlab{}.
\newblock \showarticletitle{Specializing smaller language models towards multi-step reasoning}. In \bibinfo{booktitle}{\emph{International Conference on Machine Learning}}. PMLR, \bibinfo{pages}{10421--10430}.
\newblock


\bibitem[Gage(1994)]%
        {gage1994new}
\bibfield{author}{\bibinfo{person}{Philip Gage}.} \bibinfo{year}{1994}\natexlab{}.
\newblock \showarticletitle{A new algorithm for data compression}.
\newblock \bibinfo{journal}{\emph{The C Users Journal}} \bibinfo{volume}{12}, \bibinfo{number}{2} (\bibinfo{year}{1994}), \bibinfo{pages}{23--38}.
\newblock


\bibitem[Gao et~al\mbox{.}(2023)]%
        {gao2023cirs}
\bibfield{author}{\bibinfo{person}{Chongming Gao}, \bibinfo{person}{Shiqi Wang}, \bibinfo{person}{Shijun Li}, \bibinfo{person}{Jiawei Chen}, \bibinfo{person}{Xiangnan He}, \bibinfo{person}{Wenqiang Lei}, \bibinfo{person}{Biao Li}, \bibinfo{person}{Yuan Zhang}, {and} \bibinfo{person}{Peng Jiang}.} \bibinfo{year}{2023}\natexlab{}.
\newblock \showarticletitle{CIRS: Bursting filter bubbles by counterfactual interactive recommender system}.
\newblock \bibinfo{journal}{\emph{ACM Transactions on Information Systems}} \bibinfo{volume}{42}, \bibinfo{number}{1} (\bibinfo{year}{2023}), \bibinfo{pages}{1--27}.
\newblock


\bibitem[Gao et~al\mbox{.}(2020)]%
        {gao2020pile}
\bibfield{author}{\bibinfo{person}{Leo Gao}, \bibinfo{person}{Stella Biderman}, \bibinfo{person}{Sid Black}, \bibinfo{person}{Laurence Golding}, \bibinfo{person}{Travis Hoppe}, \bibinfo{person}{Charles Foster}, \bibinfo{person}{Jason Phang}, \bibinfo{person}{Horace He}, \bibinfo{person}{Anish Thite}, \bibinfo{person}{Noa Nabeshima}, {et~al\mbox{.}}} \bibinfo{year}{2020}\natexlab{}.
\newblock \showarticletitle{The {P}ile: An 800{GB} dataset of diverse text for language modeling}.
\newblock \bibinfo{journal}{\emph{arXiv preprint arXiv:2101.00027}} (\bibinfo{year}{2020}).
\newblock


\bibitem[Gao and Callan(2022)]%
        {gao2022unsupervised}
\bibfield{author}{\bibinfo{person}{Luyu Gao} {and} \bibinfo{person}{Jamie Callan}.} \bibinfo{year}{2022}\natexlab{}.
\newblock \showarticletitle{Unsupervised Corpus Aware Language Model Pre-training for Dense Passage Retrieval}. In \bibinfo{booktitle}{\emph{Proceedings of the 60th Annual Meeting of the Association for Computational Linguistics (Volume 1: Long Papers)}}. \bibinfo{pages}{2843--2853}.
\newblock


\bibitem[Gao et~al\mbox{.}(2024)]%
        {gao2024displlm}
\bibfield{author}{\bibinfo{person}{Shangqian Gao}, \bibinfo{person}{Chi-Heng Lin}, \bibinfo{person}{Ting Hua}, \bibinfo{person}{Zheng Tang}, \bibinfo{person}{Yilin Shen}, \bibinfo{person}{Hongxia Jin}, {and} \bibinfo{person}{Yen-Chang Hsu}.} \bibinfo{year}{2024}\natexlab{}.
\newblock \showarticletitle{{DISP}-{LLM}: Dimension-Independent Structural Pruning for Large Language Models}. In \bibinfo{booktitle}{\emph{The Thirty-eighth Annual Conference on Neural Information Processing Systems}}.
\newblock
\urldef\tempurl%
\url{https://openreview.net/forum?id=YxaY6tHgg0}
\showURL{%
\tempurl}


\bibitem[Ge et~al\mbox{.}(2024)]%
        {ge2024model}
\bibfield{author}{\bibinfo{person}{Suyu Ge}, \bibinfo{person}{Yunan Zhang}, \bibinfo{person}{Liyuan Liu}, \bibinfo{person}{Minjia Zhang}, \bibinfo{person}{Jiawei Han}, {and} \bibinfo{person}{Jianfeng Gao}.} \bibinfo{year}{2024}\natexlab{}.
\newblock \showarticletitle{Model Tells You What to Discard: Adaptive {KV} Cache Compression for {LLM}s}. In \bibinfo{booktitle}{\emph{The Twelfth International Conference on Learning Representations}}.
\newblock
\urldef\tempurl%
\url{https://openreview.net/forum?id=uNrFpDPMyo}
\showURL{%
\tempurl}


\bibitem[Gichamba et~al\mbox{.}(2024)]%
        {gichamba2024colbertretrievalensembleresponse}
\bibfield{author}{\bibinfo{person}{Alex Gichamba}, \bibinfo{person}{Tewodros~Kederalah Idris}, \bibinfo{person}{Brian Ebiyau}, \bibinfo{person}{Eric Nyberg}, {and} \bibinfo{person}{Teruko Mitamura}.} \bibinfo{year}{2024}\natexlab{}.
\newblock \bibinfo{title}{ColBERT Retrieval and Ensemble Response Scoring for Language Model Question Answering}.
\newblock
\newblock
\showeprint[arxiv]{2408.10808}~[cs.CL]
\urldef\tempurl%
\url{https://arxiv.org/abs/2408.10808}
\showURL{%
\tempurl}


\bibitem[Goel(2024)]%
        {Rene}
\bibfield{author}{\bibinfo{person}{Karan Goel}.} \bibinfo{year}{2024}\natexlab{}.
\newblock \bibinfo{booktitle}{\emph{The On‑Device Intelligence Update}}.
\newblock
\urldef\tempurl%
\url{https://www.cartesia.ai/blog/on-device}
\showURL{%
\tempurl}


\bibitem[Gokaslan et~al\mbox{.}(2019)]%
        {gokaslan2019openwebtext}
\bibfield{author}{\bibinfo{person}{Aaron Gokaslan}, \bibinfo{person}{Vanya Cohen}, \bibinfo{person}{Ellie Pavlick}, {and} \bibinfo{person}{Stefanie Tellex}.} \bibinfo{year}{2019}\natexlab{}.
\newblock \bibinfo{title}{Openwebtext corpus}.
\newblock
\newblock


\bibitem[Goyal et~al\mbox{.}(2023)]%
        {goyal2023survey}
\bibfield{author}{\bibinfo{person}{Shreya Goyal}, \bibinfo{person}{Sumanth Doddapaneni}, \bibinfo{person}{Mitesh~M Khapra}, {and} \bibinfo{person}{Balaraman Ravindran}.} \bibinfo{year}{2023}\natexlab{}.
\newblock \showarticletitle{A survey of adversarial defenses and robustness in nlp}.
\newblock \bibinfo{journal}{\emph{Comput. Surveys}} \bibinfo{volume}{55}, \bibinfo{number}{14s} (\bibinfo{year}{2023}), \bibinfo{pages}{1--39}.
\newblock


\bibitem[Groeneveld et~al\mbox{.}(2024)]%
        {groeneveld2024olmo}
\bibfield{author}{\bibinfo{person}{Dirk Groeneveld}, \bibinfo{person}{Iz Beltagy}, \bibinfo{person}{Pete Walsh}, \bibinfo{person}{Akshita Bhagia}, \bibinfo{person}{Rodney Kinney}, \bibinfo{person}{Oyvind Tafjord}, \bibinfo{person}{Ananya~Harsh Jha}, \bibinfo{person}{Hamish Ivison}, \bibinfo{person}{Ian Magnusson}, \bibinfo{person}{Yizhong Wang}, {et~al\mbox{.}}} \bibinfo{year}{2024}\natexlab{}.
\newblock \showarticletitle{Olmo: Accelerating the science of language models}.
\newblock \bibinfo{journal}{\emph{arXiv preprint arXiv:2402.00838}} (\bibinfo{year}{2024}).
\newblock


\bibitem[Gu and Dao(2023)]%
        {gu2023mamba}
\bibfield{author}{\bibinfo{person}{Albert Gu} {and} \bibinfo{person}{Tri Dao}.} \bibinfo{year}{2023}\natexlab{}.
\newblock \showarticletitle{Mamba: Linear-time sequence modeling with selective state spaces}.
\newblock \bibinfo{journal}{\emph{arXiv preprint arXiv:2312.00752}} (\bibinfo{year}{2023}).
\newblock


\bibitem[Gu et~al\mbox{.}(2024b)]%
        {gu-etal-2024-light}
\bibfield{author}{\bibinfo{person}{Naibin Gu}, \bibinfo{person}{Peng Fu}, \bibinfo{person}{Xiyu Liu}, \bibinfo{person}{Bowen Shen}, \bibinfo{person}{Zheng Lin}, {and} \bibinfo{person}{Weiping Wang}.} \bibinfo{year}{2024}\natexlab{b}.
\newblock \showarticletitle{Light-{PEFT}: Lightening Parameter-Efficient Fine-Tuning via Early Pruning}. In \bibinfo{booktitle}{\emph{Findings of the Association for Computational Linguistics: ACL 2024}}, \bibfield{editor}{\bibinfo{person}{Lun-Wei Ku}, \bibinfo{person}{Andre Martins}, {and} \bibinfo{person}{Vivek Srikumar}} (Eds.). \bibinfo{publisher}{Association for Computational Linguistics}, \bibinfo{address}{Bangkok, Thailand}, \bibinfo{pages}{7528--7541}.
\newblock
\urldef\tempurl%
\url{https://doi.org/10.18653/v1/2024.findings-acl.447}
\showDOI{\tempurl}


\bibitem[Gu et~al\mbox{.}(2024a)]%
        {gu2024minillm}
\bibfield{author}{\bibinfo{person}{Yuxian Gu}, \bibinfo{person}{Li Dong}, \bibinfo{person}{Furu Wei}, {and} \bibinfo{person}{Minlie Huang}.} \bibinfo{year}{2024}\natexlab{a}.
\newblock \showarticletitle{MiniLLM: Knowledge distillation of large language models}. In \bibinfo{booktitle}{\emph{The Twelfth International Conference on Learning Representations}}.
\newblock


\bibitem[Gunasekar et~al\mbox{.}(2023)]%
        {gunasekar2023textbooksneed}
\bibfield{author}{\bibinfo{person}{Suriya Gunasekar}, \bibinfo{person}{Yi Zhang}, \bibinfo{person}{Jyoti Aneja}, \bibinfo{person}{Caio César~Teodoro Mendes}, \bibinfo{person}{Allie~Del Giorno}, \bibinfo{person}{Sivakanth Gopi}, \bibinfo{person}{Mojan Javaheripi}, \bibinfo{person}{Piero Kauffmann}, \bibinfo{person}{Gustavo de Rosa}, \bibinfo{person}{Olli Saarikivi}, \bibinfo{person}{Adil Salim}, \bibinfo{person}{Shital Shah}, \bibinfo{person}{Harkirat~Singh Behl}, \bibinfo{person}{Xin Wang}, \bibinfo{person}{Sébastien Bubeck}, \bibinfo{person}{Ronen Eldan}, \bibinfo{person}{Adam~Tauman Kalai}, \bibinfo{person}{Yin~Tat Lee}, {and} \bibinfo{person}{Yuanzhi Li}.} \bibinfo{year}{2023}\natexlab{}.
\newblock \bibinfo{title}{Textbooks Are All You Need}.
\newblock
\newblock
\showeprint[arxiv]{2306.11644}~[cs.CL]
\urldef\tempurl%
\url{https://arxiv.org/abs/2306.11644}
\showURL{%
\tempurl}


\bibitem[Guo et~al\mbox{.}(2024b)]%
        {guo2024deepseek}
\bibfield{author}{\bibinfo{person}{Daya Guo}, \bibinfo{person}{Qihao Zhu}, \bibinfo{person}{Dejian Yang}, \bibinfo{person}{Zhenda Xie}, \bibinfo{person}{Kai Dong}, \bibinfo{person}{Wentao Zhang}, \bibinfo{person}{Guanting Chen}, \bibinfo{person}{Xiao Bi}, \bibinfo{person}{Yu Wu}, \bibinfo{person}{YK Li}, {et~al\mbox{.}}} \bibinfo{year}{2024}\natexlab{b}.
\newblock \showarticletitle{DeepSeek-Coder: When the Large Language Model Meets Programming--The Rise of Code Intelligence}.
\newblock \bibinfo{journal}{\emph{arXiv preprint arXiv:2401.14196}} (\bibinfo{year}{2024}).
\newblock


\bibitem[Guo et~al\mbox{.}(2024a)]%
        {guo2024compressing}
\bibfield{author}{\bibinfo{person}{Jinyang Guo}, \bibinfo{person}{Jianyu Wu}, \bibinfo{person}{Zining Wang}, \bibinfo{person}{Jiaheng Liu}, \bibinfo{person}{Ge Yang}, \bibinfo{person}{Yifu Ding}, \bibinfo{person}{Ruihao Gong}, \bibinfo{person}{Haotong Qin}, {and} \bibinfo{person}{Xianglong Liu}.} \bibinfo{year}{2024}\natexlab{a}.
\newblock \showarticletitle{Compressing large language models by joint sparsification and quantization}. In \bibinfo{booktitle}{\emph{Forty-first International Conference on Machine Learning}}.
\newblock


\bibitem[Guo et~al\mbox{.}(2022)]%
        {guo2022threats}
\bibfield{author}{\bibinfo{person}{Shangwei Guo}, \bibinfo{person}{Chunlong Xie}, \bibinfo{person}{Jiwei Li}, \bibinfo{person}{Lingjuan Lyu}, {and} \bibinfo{person}{Tianwei Zhang}.} \bibinfo{year}{2022}\natexlab{}.
\newblock \showarticletitle{Threats to pre-trained language models: Survey and taxonomy}.
\newblock \bibinfo{journal}{\emph{arXiv preprint arXiv:2202.06862}} (\bibinfo{year}{2022}).
\newblock


\bibitem[Guo et~al\mbox{.}(2023b)]%
        {guo2023compresso}
\bibfield{author}{\bibinfo{person}{Song Guo}, \bibinfo{person}{Jiahang Xu}, \bibinfo{person}{Li~Lyna Zhang}, {and} \bibinfo{person}{Mao Yang}.} \bibinfo{year}{2023}\natexlab{b}.
\newblock \showarticletitle{Compresso: Structured pruning with collaborative prompting learns compact large language models}.
\newblock \bibinfo{journal}{\emph{arXiv preprint arXiv:2310.05015}} (\bibinfo{year}{2023}).
\newblock


\bibitem[Guo et~al\mbox{.}(2023a)]%
        {guo2023improvingsmalllanguagemodels}
\bibfield{author}{\bibinfo{person}{Zhen Guo}, \bibinfo{person}{Peiqi Wang}, \bibinfo{person}{Yanwei Wang}, {and} \bibinfo{person}{Shangdi Yu}.} \bibinfo{year}{2023}\natexlab{a}.
\newblock \bibinfo{title}{Improving Small Language Models on PubMedQA via Generative Data Augmentation}.
\newblock
\newblock
\showeprint[arxiv]{2305.07804}~[cs.CL]
\urldef\tempurl%
\url{https://arxiv.org/abs/2305.07804}
\showURL{%
\tempurl}


\bibitem[Han et~al\mbox{.}(2015)]%
        {han2015learning}
\bibfield{author}{\bibinfo{person}{Song Han}, \bibinfo{person}{Jeff Pool}, \bibinfo{person}{John Tran}, {and} \bibinfo{person}{William Dally}.} \bibinfo{year}{2015}\natexlab{}.
\newblock \showarticletitle{Learning both weights and connections for efficient neural network}.
\newblock \bibinfo{journal}{\emph{Advances in neural information processing systems}}  \bibinfo{volume}{28} (\bibinfo{year}{2015}).
\newblock


\bibitem[Hao et~al\mbox{.}(2024)]%
        {hao2024hybrid}
\bibfield{author}{\bibinfo{person}{Zixu Hao}, \bibinfo{person}{Huiqiang Jiang}, \bibinfo{person}{Shiqi Jiang}, \bibinfo{person}{Ju Ren}, {and} \bibinfo{person}{Ting Cao}.} \bibinfo{year}{2024}\natexlab{}.
\newblock \showarticletitle{Hybrid SLM and LLM for Edge-Cloud Collaborative Inference}. In \bibinfo{booktitle}{\emph{Proceedings of the Workshop on Edge and Mobile Foundation Models}}. \bibinfo{pages}{36--41}.
\newblock


\bibitem[Hartill et~al\mbox{.}(2023a)]%
        {hartill2023answeringunseenquestionssmaller}
\bibfield{author}{\bibinfo{person}{Tim Hartill}, \bibinfo{person}{Diana Benavides-Prado}, \bibinfo{person}{Michael Witbrock}, {and} \bibinfo{person}{Patricia~J. Riddle}.} \bibinfo{year}{2023}\natexlab{a}.
\newblock \bibinfo{title}{Answering Unseen Questions With Smaller Language Models Using Rationale Generation and Dense Retrieval}.
\newblock
\newblock
\showeprint[arxiv]{2308.04711}~[cs.CL]
\urldef\tempurl%
\url{https://arxiv.org/abs/2308.04711}
\showURL{%
\tempurl}


\bibitem[Hartill et~al\mbox{.}(2023b)]%
        {hartill2023teachingsmallerlanguagemodels}
\bibfield{author}{\bibinfo{person}{Tim Hartill}, \bibinfo{person}{Neset Tan}, \bibinfo{person}{Michael Witbrock}, {and} \bibinfo{person}{Patricia~J. Riddle}.} \bibinfo{year}{2023}\natexlab{b}.
\newblock \bibinfo{title}{Teaching Smaller Language Models To Generalise To Unseen Compositional Questions}.
\newblock
\newblock
\showeprint[arxiv]{2308.00946}~[cs.CL]
\urldef\tempurl%
\url{https://arxiv.org/abs/2308.00946}
\showURL{%
\tempurl}


\bibitem[He et~al\mbox{.}(2023b)]%
        {he2023survey}
\bibfield{author}{\bibinfo{person}{Kai He}, \bibinfo{person}{Rui Mao}, \bibinfo{person}{Qika Lin}, \bibinfo{person}{Yucheng Ruan}, \bibinfo{person}{Xiang Lan}, \bibinfo{person}{Mengling Feng}, {and} \bibinfo{person}{Erik Cambria}.} \bibinfo{year}{2023}\natexlab{b}.
\newblock \showarticletitle{A survey of large language models for healthcare: from data, technology, and applications to accountability and ethics}.
\newblock \bibinfo{journal}{\emph{arXiv preprint arXiv:2310.05694}} (\bibinfo{year}{2023}).
\newblock


\bibitem[He et~al\mbox{.}(2023a)]%
        {hedebertav3}
\bibfield{author}{\bibinfo{person}{Pengcheng He}, \bibinfo{person}{Jianfeng Gao}, {and} \bibinfo{person}{Weizhu Chen}.} \bibinfo{year}{2023}\natexlab{a}.
\newblock \showarticletitle{De{BERT}aV3: Improving De{BERT}a using {ELECTRA}-Style Pre-Training with Gradient-Disentangled Embedding Sharing}. In \bibinfo{booktitle}{\emph{The Eleventh International Conference on Learning Representations}}.
\newblock
\urldef\tempurl%
\url{https://openreview.net/forum?id=sE7-XhLxHA}
\showURL{%
\tempurl}


\bibitem[He et~al\mbox{.}(2020)]%
        {he2020deberta}
\bibfield{author}{\bibinfo{person}{Pengcheng He}, \bibinfo{person}{Xiaodong Liu}, \bibinfo{person}{Jianfeng Gao}, {and} \bibinfo{person}{Weizhu Chen}.} \bibinfo{year}{2020}\natexlab{}.
\newblock \showarticletitle{Deberta: Decoding-enhanced bert with disentangled attention}.
\newblock \bibinfo{journal}{\emph{arXiv preprint arXiv:2006.03654}} (\bibinfo{year}{2020}).
\newblock


\bibitem[Heidari et~al\mbox{.}(2022)]%
        {heidari2022attention}
\bibfield{author}{\bibinfo{person}{Narges Heidari}, \bibinfo{person}{Parham Moradi}, {and} \bibinfo{person}{Abbas Koochari}.} \bibinfo{year}{2022}\natexlab{}.
\newblock \showarticletitle{An attention-based deep learning method for solving the cold-start and sparsity issues of recommender systems}.
\newblock \bibinfo{journal}{\emph{Knowledge-Based Systems}}  \bibinfo{volume}{256} (\bibinfo{year}{2022}), \bibinfo{pages}{109835}.
\newblock


\bibitem[Hendrycks et~al\mbox{.}(2021)]%
        {hendrycksmeasuring}
\bibfield{author}{\bibinfo{person}{Dan Hendrycks}, \bibinfo{person}{Collin Burns}, \bibinfo{person}{Steven Basart}, \bibinfo{person}{Andy Zou}, \bibinfo{person}{Mantas Mazeika}, \bibinfo{person}{Dawn Song}, {and} \bibinfo{person}{Jacob Steinhardt}.} \bibinfo{year}{2021}\natexlab{}.
\newblock \showarticletitle{Measuring Massive Multitask Language Understanding}. In \bibinfo{booktitle}{\emph{International Conference on Learning Representations}}.
\newblock
\urldef\tempurl%
\url{https://openreview.net/forum?id=d7KBjmI3GmQ}
\showURL{%
\tempurl}


\bibitem[Hendrycks and Gimpel(2016)]%
        {gelu}
\bibfield{author}{\bibinfo{person}{Dan Hendrycks} {and} \bibinfo{person}{Kevin Gimpel}.} \bibinfo{year}{2016}\natexlab{}.
\newblock \showarticletitle{Gaussian error linear units (gelus)}.
\newblock \bibinfo{journal}{\emph{arXiv preprint arXiv:1606.08415}} (\bibinfo{year}{2016}).
\newblock


\bibitem[Hinton et~al\mbox{.}(2015)]%
        {hinton2015distilling}
\bibfield{author}{\bibinfo{person}{Geoffrey Hinton}, \bibinfo{person}{Oriol Vinyals}, {and} \bibinfo{person}{Jeff Dean}.} \bibinfo{year}{2015}\natexlab{}.
\newblock \showarticletitle{Distilling the knowledge in a neural network}.
\newblock \bibinfo{journal}{\emph{arXiv preprint arXiv:1503.02531}} (\bibinfo{year}{2015}).
\newblock


\bibitem[Hochreiter and Schmidhuber(1996)]%
        {hochreiter1996lstm}
\bibfield{author}{\bibinfo{person}{Sepp Hochreiter} {and} \bibinfo{person}{J{\"u}rgen Schmidhuber}.} \bibinfo{year}{1996}\natexlab{}.
\newblock \showarticletitle{LSTM can solve hard long time lag problems}.
\newblock \bibinfo{journal}{\emph{Advances in neural information processing systems}}  \bibinfo{volume}{9} (\bibinfo{year}{1996}).
\newblock


\bibitem[Hong et~al\mbox{.}(2024)]%
        {hong2024decoding}
\bibfield{author}{\bibinfo{person}{Junyuan Hong}, \bibinfo{person}{Jinhao Duan}, \bibinfo{person}{Chenhui Zhang}, \bibinfo{person}{Zhangheng Li}, \bibinfo{person}{Chulin Xie}, \bibinfo{person}{Kelsey Lieberman}, \bibinfo{person}{James Diffenderfer}, \bibinfo{person}{Brian~R. Bartoldson}, \bibinfo{person}{Ajay~Kumar Jaiswal}, \bibinfo{person}{Kaidi Xu}, \bibinfo{person}{Bhavya Kailkhura}, \bibinfo{person}{Dan Hendrycks}, \bibinfo{person}{Dawn Song}, \bibinfo{person}{Zhangyang Wang}, {and} \bibinfo{person}{Bo Li}.} \bibinfo{year}{2024}\natexlab{}.
\newblock \showarticletitle{Decoding Compressed Trust: Scrutinizing the Trustworthiness of Efficient LLMs Under Compression}. In \bibinfo{booktitle}{\emph{Proceedings of the Forty-first International Conference on Machine Learning, {ICML}}}.
\newblock
\urldef\tempurl%
\url{https://openreview.net/forum?id=e3Dpq3WdMv}
\showURL{%
\tempurl}


\bibitem[Hongcheng~Liu(2023)]%
        {LAWGPT-zh}
\bibfield{author}{\bibinfo{person}{Yutong Meng Yuhao~Wang Hongcheng~Liu, Yusheng~Liao}.} \bibinfo{year}{2023}\natexlab{}.
\newblock \bibinfo{title}{XieZhi: Chinese Law Large Language Model}.
\newblock \bibinfo{howpublished}{\url{https://github.com/LiuHC0428/LAW_GPT}}.
\newblock


\bibitem[Houlsby et~al\mbox{.}(2019)]%
        {houlsby2019parameter}
\bibfield{author}{\bibinfo{person}{Neil Houlsby}, \bibinfo{person}{Andrei Giurgiu}, \bibinfo{person}{Stanislaw Jastrzebski}, \bibinfo{person}{Bruna Morrone}, \bibinfo{person}{Quentin De~Laroussilhe}, \bibinfo{person}{Andrea Gesmundo}, \bibinfo{person}{Mona Attariyan}, {and} \bibinfo{person}{Sylvain Gelly}.} \bibinfo{year}{2019}\natexlab{}.
\newblock \showarticletitle{Parameter-efficient transfer learning for NLP}. In \bibinfo{booktitle}{\emph{International Conference on Machine Learning}}. PMLR, \bibinfo{pages}{2790--2799}.
\newblock


\bibitem[Hsieh et~al\mbox{.}(2023)]%
        {hsieh2023distilling}
\bibfield{author}{\bibinfo{person}{Cheng-Yu Hsieh}, \bibinfo{person}{Chun-Liang Li}, \bibinfo{person}{Chih-kuan Yeh}, \bibinfo{person}{Hootan Nakhost}, \bibinfo{person}{Yasuhisa Fujii}, \bibinfo{person}{Alex Ratner}, \bibinfo{person}{Ranjay Krishna}, \bibinfo{person}{Chen-Yu Lee}, {and} \bibinfo{person}{Tomas Pfister}.} \bibinfo{year}{2023}\natexlab{}.
\newblock \showarticletitle{Distilling Step-by-Step! Outperforming Larger Language Models with Less Training Data and Smaller Model Sizes}. In \bibinfo{booktitle}{\emph{Findings of the Association for Computational Linguistics: ACL 2023}}. \bibinfo{pages}{8003--8017}.
\newblock


\bibitem[Hu et~al\mbox{.}(2021)]%
        {hu2021lora}
\bibfield{author}{\bibinfo{person}{Edward~J Hu}, \bibinfo{person}{Yelong Shen}, \bibinfo{person}{Phillip Wallis}, \bibinfo{person}{Zeyuan Allen-Zhu}, \bibinfo{person}{Yuanzhi Li}, \bibinfo{person}{Shean Wang}, \bibinfo{person}{Lu Wang}, {and} \bibinfo{person}{Weizhu Chen}.} \bibinfo{year}{2021}\natexlab{}.
\newblock \showarticletitle{Lora: Low-rank adaptation of large language models}.
\newblock \bibinfo{journal}{\emph{arXiv preprint arXiv:2106.09685}} (\bibinfo{year}{2021}).
\newblock


\bibitem[Hu et~al\mbox{.}(2024b)]%
        {hu2024minicpm}
\bibfield{author}{\bibinfo{person}{Shengding Hu}, \bibinfo{person}{Yuge Tu}, \bibinfo{person}{Xu Han}, \bibinfo{person}{Chaoqun He}, \bibinfo{person}{Ganqu Cui}, \bibinfo{person}{Xiang Long}, \bibinfo{person}{Zhi Zheng}, \bibinfo{person}{Yewei Fang}, \bibinfo{person}{Yuxiang Huang}, \bibinfo{person}{Weilin Zhao}, {et~al\mbox{.}}} \bibinfo{year}{2024}\natexlab{b}.
\newblock \showarticletitle{Minicpm: Unveiling the potential of small language models with scalable training strategies}.
\newblock \bibinfo{journal}{\emph{arXiv preprint arXiv:2404.06395}} (\bibinfo{year}{2024}).
\newblock


\bibitem[Hu et~al\mbox{.}(2024a)]%
        {hu2024llm}
\bibfield{author}{\bibinfo{person}{Xing Hu}, \bibinfo{person}{Yuan Chen}, \bibinfo{person}{Dawei Yang}, \bibinfo{person}{Sifan Zhou}, \bibinfo{person}{Zhihang Yuan}, \bibinfo{person}{Jiangyong Yu}, {and} \bibinfo{person}{Chen Xu}.} \bibinfo{year}{2024}\natexlab{a}.
\newblock \showarticletitle{I-LLM: Efficient Integer-Only Inference for Fully-Quantized Low-Bit Large Language Models}.
\newblock \bibinfo{journal}{\emph{arXiv preprint arXiv:2405.17849}} (\bibinfo{year}{2024}).
\newblock


\bibitem[Huang et~al\mbox{.}({[n.\,d.]})]%
        {huang2023large}
\bibfield{author}{\bibinfo{person}{Jiaxin Huang}, \bibinfo{person}{Shixiang~Shane Gu}, \bibinfo{person}{Le Hou}, \bibinfo{person}{Yuexin Wu}, \bibinfo{person}{Xuezhi Wang}, \bibinfo{person}{Hongkun Yu}, {and} \bibinfo{person}{Jiawei Han}.} \bibinfo{year}{[n.\,d.]}\natexlab{}.
\newblock \showarticletitle{Large Language Models Can Self-Improve}. In \bibinfo{booktitle}{\emph{The 2023 Conference on Empirical Methods in Natural Language Processing}}.
\newblock


\bibitem[Huang et~al\mbox{.}(2023c)]%
        {huang2023survey}
\bibfield{author}{\bibinfo{person}{Lei Huang}, \bibinfo{person}{Weijiang Yu}, \bibinfo{person}{Weitao Ma}, \bibinfo{person}{Weihong Zhong}, \bibinfo{person}{Zhangyin Feng}, \bibinfo{person}{Haotian Wang}, \bibinfo{person}{Qianglong Chen}, \bibinfo{person}{Weihua Peng}, \bibinfo{person}{Xiaocheng Feng}, \bibinfo{person}{Bing Qin}, {et~al\mbox{.}}} \bibinfo{year}{2023}\natexlab{c}.
\newblock \showarticletitle{A survey on hallucination in large language models: Principles, taxonomy, challenges, and open questions}.
\newblock \bibinfo{journal}{\emph{arXiv preprint arXiv:2311.05232}} (\bibinfo{year}{2023}).
\newblock


\bibitem[Huang et~al\mbox{.}(2024b)]%
        {huang2024billm}
\bibfield{author}{\bibinfo{person}{Wei Huang}, \bibinfo{person}{Yangdong Liu}, \bibinfo{person}{Haotong Qin}, \bibinfo{person}{Ying Li}, \bibinfo{person}{Shiming Zhang}, \bibinfo{person}{Xianglong Liu}, \bibinfo{person}{Michele Magno}, {and} \bibinfo{person}{Xiaojuan Qi}.} \bibinfo{year}{2024}\natexlab{b}.
\newblock \showarticletitle{Billm: Pushing the limit of post-training quantization for llms}.
\newblock \bibinfo{journal}{\emph{arXiv preprint arXiv:2402.04291}} (\bibinfo{year}{2024}).
\newblock


\bibitem[Huang et~al\mbox{.}(2024c)]%
        {huang-etal-2024-less}
\bibfield{author}{\bibinfo{person}{Wenyu Huang}, \bibinfo{person}{Guancheng Zhou}, \bibinfo{person}{Hongru Wang}, \bibinfo{person}{Pavlos Vougiouklis}, \bibinfo{person}{Mirella Lapata}, {and} \bibinfo{person}{Jeff~Z. Pan}.} \bibinfo{year}{2024}\natexlab{c}.
\newblock \showarticletitle{Less is More: Making Smaller Language Models Competent Subgraph Retrievers for Multi-hop {KGQA}}. In \bibinfo{booktitle}{\emph{Findings of the Association for Computational Linguistics: EMNLP 2024}}, \bibfield{editor}{\bibinfo{person}{Yaser Al-Onaizan}, \bibinfo{person}{Mohit Bansal}, {and} \bibinfo{person}{Yun-Nung Chen}} (Eds.). \bibinfo{publisher}{Association for Computational Linguistics}, \bibinfo{address}{Miami, Florida, USA}, \bibinfo{pages}{15787--15803}.
\newblock
\urldef\tempurl%
\url{https://doi.org/10.18653/v1/2024.findings-emnlp.927}
\showDOI{\tempurl}


\bibitem[Huang et~al\mbox{.}(2024a)]%
        {huang2024c}
\bibfield{author}{\bibinfo{person}{Yuzhen Huang}, \bibinfo{person}{Yuzhuo Bai}, \bibinfo{person}{Zhihao Zhu}, \bibinfo{person}{Junlei Zhang}, \bibinfo{person}{Jinghan Zhang}, \bibinfo{person}{Tangjun Su}, \bibinfo{person}{Junteng Liu}, \bibinfo{person}{Chuancheng Lv}, \bibinfo{person}{Yikai Zhang}, \bibinfo{person}{Yao Fu}, {et~al\mbox{.}}} \bibinfo{year}{2024}\natexlab{a}.
\newblock \showarticletitle{C-eval: A multi-level multi-discipline chinese evaluation suite for foundation models}.
\newblock \bibinfo{journal}{\emph{Advances in Neural Information Processing Systems}}  \bibinfo{volume}{36} (\bibinfo{year}{2024}).
\newblock


\bibitem[Huang et~al\mbox{.}(2023a)]%
        {huang2023catastrophic}
\bibfield{author}{\bibinfo{person}{Yangsibo Huang}, \bibinfo{person}{Samyak Gupta}, \bibinfo{person}{Mengzhou Xia}, \bibinfo{person}{Kai Li}, {and} \bibinfo{person}{Danqi Chen}.} \bibinfo{year}{2023}\natexlab{a}.
\newblock \showarticletitle{Catastrophic Jailbreak of Open-source LLMs via Exploiting Generation}.
\newblock \bibinfo{journal}{\emph{arXiv preprint arXiv:2310.06987}} (\bibinfo{year}{2023}).
\newblock


\bibitem[Huang et~al\mbox{.}(2023b)]%
        {huang2023look}
\bibfield{author}{\bibinfo{person}{Yuheng Huang}, \bibinfo{person}{Jiayang Song}, \bibinfo{person}{Zhijie Wang}, \bibinfo{person}{Shengming Zhao}, \bibinfo{person}{Huaming Chen}, \bibinfo{person}{Felix Juefei-Xu}, {and} \bibinfo{person}{Lei Ma}.} \bibinfo{year}{2023}\natexlab{b}.
\newblock \showarticletitle{Look before you leap: An exploratory study of uncertainty measurement for large language models}.
\newblock \bibinfo{journal}{\emph{arXiv preprint arXiv:2307.10236}} (\bibinfo{year}{2023}).
\newblock


\bibitem[Humeau et~al\mbox{.}({[n.\,d.]})]%
        {humeaupoly}
\bibfield{author}{\bibinfo{person}{Samuel Humeau}, \bibinfo{person}{Kurt Shuster}, \bibinfo{person}{Marie-Anne Lachaux}, {and} \bibinfo{person}{Jason Weston}.} \bibinfo{year}{[n.\,d.]}\natexlab{}.
\newblock \showarticletitle{Poly-encoders: Architectures and Pre-training Strategies for Fast and Accurate Multi-sentence Scoring}. In \bibinfo{booktitle}{\emph{International Conference on Learning Representations}}.
\newblock


\bibitem[Inan et~al\mbox{.}(2023)]%
        {inan2023llama}
\bibfield{author}{\bibinfo{person}{Hakan Inan}, \bibinfo{person}{Kartikeya Upasani}, \bibinfo{person}{Jianfeng Chi}, \bibinfo{person}{Rashi Rungta}, \bibinfo{person}{Krithika Iyer}, \bibinfo{person}{Yuning Mao}, \bibinfo{person}{Michael Tontchev}, \bibinfo{person}{Qing Hu}, \bibinfo{person}{Brian Fuller}, \bibinfo{person}{Davide Testuggine}, {et~al\mbox{.}}} \bibinfo{year}{2023}\natexlab{}.
\newblock \showarticletitle{Llama guard: Llm-based input-output safeguard for human-ai conversations}.
\newblock \bibinfo{journal}{\emph{arXiv preprint arXiv:2312.06674}} (\bibinfo{year}{2023}).
\newblock


\bibitem[Jacobs et~al\mbox{.}(1991)]%
        {jacobs1991adaptive}
\bibfield{author}{\bibinfo{person}{Robert~A Jacobs}, \bibinfo{person}{Michael~I Jordan}, \bibinfo{person}{Steven~J Nowlan}, {and} \bibinfo{person}{Geoffrey~E Hinton}.} \bibinfo{year}{1991}\natexlab{}.
\newblock \showarticletitle{Adaptive mixtures of local experts}.
\newblock \bibinfo{journal}{\emph{Neural computation}} \bibinfo{volume}{3}, \bibinfo{number}{1} (\bibinfo{year}{1991}), \bibinfo{pages}{79--87}.
\newblock


\bibitem[Javaheripi et~al\mbox{.}(2023)]%
        {javaheripi2023phi}
\bibfield{author}{\bibinfo{person}{Mojan Javaheripi}, \bibinfo{person}{S{\'e}bastien Bubeck}, \bibinfo{person}{Marah Abdin}, \bibinfo{person}{Jyoti Aneja}, \bibinfo{person}{Sebastien Bubeck}, \bibinfo{person}{Caio C{\'e}sar~Teodoro Mendes}, \bibinfo{person}{Weizhu Chen}, \bibinfo{person}{Allie Del~Giorno}, \bibinfo{person}{Ronen Eldan}, \bibinfo{person}{Sivakanth Gopi}, {et~al\mbox{.}}} \bibinfo{year}{2023}\natexlab{}.
\newblock \showarticletitle{Phi-2: The surprising power of small language models}.
\newblock \bibinfo{journal}{\emph{Microsoft Research Blog}} (\bibinfo{year}{2023}).
\newblock


\bibitem[Jeong et~al\mbox{.}(2023)]%
        {jeong2023testtimeselfadaptivesmalllanguage}
\bibfield{author}{\bibinfo{person}{Soyeong Jeong}, \bibinfo{person}{Jinheon Baek}, \bibinfo{person}{Sukmin Cho}, \bibinfo{person}{Sung Hwang}, {and} \bibinfo{person}{Jong~C Park}.} \bibinfo{year}{2023}\natexlab{}.
\newblock \showarticletitle{Test-Time Self-Adaptive Small Language Models for Question Answering}. In \bibinfo{booktitle}{\emph{Findings of the Association for Computational Linguistics: EMNLP 2023}}. \bibinfo{pages}{15459--15469}.
\newblock


\bibitem[Jha et~al\mbox{.}(2024)]%
        {jha2024justchopembarrassinglysimple}
\bibfield{author}{\bibinfo{person}{Ananya~Harsh Jha}, \bibinfo{person}{Tom Sherborne}, \bibinfo{person}{Evan~Pete Walsh}, \bibinfo{person}{Dirk Groeneveld}, \bibinfo{person}{Emma Strubell}, {and} \bibinfo{person}{Iz Beltagy}.} \bibinfo{year}{2024}\natexlab{}.
\newblock \bibinfo{title}{Just CHOP: Embarrassingly Simple LLM Compression}.
\newblock
\newblock
\showeprint[arxiv]{2305.14864}~[cs.CL]
\urldef\tempurl%
\url{https://arxiv.org/abs/2305.14864}
\showURL{%
\tempurl}


\bibitem[Ji et~al\mbox{.}(2024)]%
        {ji2024feature}
\bibfield{author}{\bibinfo{person}{Yixin Ji}, \bibinfo{person}{Yang Xiang}, \bibinfo{person}{Juntao Li}, \bibinfo{person}{Wei Chen}, \bibinfo{person}{Zhongyi Liu}, \bibinfo{person}{Kehai Chen}, {and} \bibinfo{person}{Min Zhang}.} \bibinfo{year}{2024}\natexlab{}.
\newblock \showarticletitle{Feature-based Low-Rank Compression of Large Language Models via Bayesian Optimization}.
\newblock \bibinfo{journal}{\emph{arXiv preprint arXiv:2405.10616}} (\bibinfo{year}{2024}).
\newblock


\bibitem[Ji et~al\mbox{.}(2023)]%
        {ji2023towards}
\bibfield{author}{\bibinfo{person}{Ziwei Ji}, \bibinfo{person}{Tiezheng Yu}, \bibinfo{person}{Yan Xu}, \bibinfo{person}{Nayeon Lee}, \bibinfo{person}{Etsuko Ishii}, {and} \bibinfo{person}{Pascale Fung}.} \bibinfo{year}{2023}\natexlab{}.
\newblock \showarticletitle{Towards mitigating LLM hallucination via self reflection}. In \bibinfo{booktitle}{\emph{Findings of the Association for Computational Linguistics: EMNLP 2023}}. \bibinfo{pages}{1827--1843}.
\newblock


\bibitem[Jiang et~al\mbox{.}(2023)]%
        {jiang2023mistral}
\bibfield{author}{\bibinfo{person}{Albert~Q Jiang}, \bibinfo{person}{Alexandre Sablayrolles}, \bibinfo{person}{Arthur Mensch}, \bibinfo{person}{Chris Bamford}, \bibinfo{person}{Devendra~Singh Chaplot}, \bibinfo{person}{Diego de~las Casas}, \bibinfo{person}{Florian Bressand}, \bibinfo{person}{Gianna Lengyel}, \bibinfo{person}{Guillaume Lample}, \bibinfo{person}{Lucile Saulnier}, {et~al\mbox{.}}} \bibinfo{year}{2023}\natexlab{}.
\newblock \showarticletitle{Mistral 7B}.
\newblock \bibinfo{journal}{\emph{arXiv preprint arXiv:2310.06825}} (\bibinfo{year}{2023}).
\newblock


\bibitem[Jiang et~al\mbox{.}(2024a)]%
        {jiang2024mixtral}
\bibfield{author}{\bibinfo{person}{Albert~Q Jiang}, \bibinfo{person}{Alexandre Sablayrolles}, \bibinfo{person}{Antoine Roux}, \bibinfo{person}{Arthur Mensch}, \bibinfo{person}{Blanche Savary}, \bibinfo{person}{Chris Bamford}, \bibinfo{person}{Devendra~Singh Chaplot}, \bibinfo{person}{Diego de~las Casas}, \bibinfo{person}{Emma~Bou Hanna}, \bibinfo{person}{Florian Bressand}, {et~al\mbox{.}}} \bibinfo{year}{2024}\natexlab{a}.
\newblock \showarticletitle{Mixtral of experts}.
\newblock \bibinfo{journal}{\emph{arXiv preprint arXiv:2401.04088}} (\bibinfo{year}{2024}).
\newblock


\bibitem[Jiang et~al\mbox{.}(2024b)]%
        {jiang2024longllmlingua}
\bibfield{author}{\bibinfo{person}{Huiqiang Jiang}, \bibinfo{person}{Qianhui Wu}, \bibinfo{person}{Xufang Luo}, \bibinfo{person}{Dongsheng Li}, \bibinfo{person}{Chin-Yew Lin}, \bibinfo{person}{Yuqing Yang}, {and} \bibinfo{person}{Lili Qiu}.} \bibinfo{year}{2024}\natexlab{b}.
\newblock \showarticletitle{Long{LLML}ingua: Accelerating and Enhancing {LLM}s in Long Context Scenarios via Prompt Compression}. In \bibinfo{booktitle}{\emph{ICLR 2024 Workshop on Mathematical and Empirical Understanding of Foundation Models}}.
\newblock
\urldef\tempurl%
\url{https://openreview.net/forum?id=9YvfRrpmyw}
\showURL{%
\tempurl}


\bibitem[Jin et~al\mbox{.}(2019)]%
        {jin2019pubmedqa}
\bibfield{author}{\bibinfo{person}{Qiao Jin}, \bibinfo{person}{Bhuwan Dhingra}, \bibinfo{person}{Zhengping Liu}, \bibinfo{person}{William Cohen}, {and} \bibinfo{person}{Xinghua Lu}.} \bibinfo{year}{2019}\natexlab{}.
\newblock \showarticletitle{PubMedQA: A Dataset for Biomedical Research Question Answering}. In \bibinfo{booktitle}{\emph{Proceedings of the 2019 Conference on Empirical Methods in Natural Language Processing and the 9th International Joint Conference on Natural Language Processing (EMNLP-IJCNLP)}}. \bibinfo{pages}{2567--2577}.
\newblock


\bibitem[Kalman(1960)]%
        {kalman1960new}
\bibfield{author}{\bibinfo{person}{Rudolph~Emil Kalman}.} \bibinfo{year}{1960}\natexlab{}.
\newblock \showarticletitle{A new approach to linear filtering and prediction problems}.
\newblock  (\bibinfo{year}{1960}).
\newblock


\bibitem[Kang et~al\mbox{.}(2024)]%
        {kang2024gear}
\bibfield{author}{\bibinfo{person}{Hao Kang}, \bibinfo{person}{Qingru Zhang}, \bibinfo{person}{Souvik Kundu}, \bibinfo{person}{Geonhwa Jeong}, \bibinfo{person}{Zaoxing Liu}, \bibinfo{person}{Tushar Krishna}, {and} \bibinfo{person}{Tuo Zhao}.} \bibinfo{year}{2024}\natexlab{}.
\newblock \showarticletitle{Gear: An efficient kv cache compression recipefor near-lossless generative inference of llm}.
\newblock \bibinfo{journal}{\emph{arXiv preprint arXiv:2403.05527}} (\bibinfo{year}{2024}).
\newblock


\bibitem[Kaplan et~al\mbox{.}(2020)]%
        {kaplan2020scaling}
\bibfield{author}{\bibinfo{person}{Jared Kaplan}, \bibinfo{person}{Sam McCandlish}, \bibinfo{person}{Tom Henighan}, \bibinfo{person}{Tom~B Brown}, \bibinfo{person}{Benjamin Chess}, \bibinfo{person}{Rewon Child}, \bibinfo{person}{Scott Gray}, \bibinfo{person}{Alec Radford}, \bibinfo{person}{Jeffrey Wu}, {and} \bibinfo{person}{Dario Amodei}.} \bibinfo{year}{2020}\natexlab{}.
\newblock \showarticletitle{Scaling laws for neural language models}.
\newblock \bibinfo{journal}{\emph{arXiv preprint arXiv:2001.08361}} (\bibinfo{year}{2020}).
\newblock


\bibitem[Khattab et~al\mbox{.}(2023)]%
        {khattab2023dspy}
\bibfield{author}{\bibinfo{person}{Omar Khattab}, \bibinfo{person}{Arnav Singhvi}, \bibinfo{person}{Paridhi Maheshwari}, \bibinfo{person}{Zhiyuan Zhang}, \bibinfo{person}{Keshav Santhanam}, \bibinfo{person}{Sri Vardhamanan}, \bibinfo{person}{Saiful Haq}, \bibinfo{person}{Ashutosh Sharma}, \bibinfo{person}{Thomas~T Joshi}, \bibinfo{person}{Hanna Moazam}, {et~al\mbox{.}}} \bibinfo{year}{2023}\natexlab{}.
\newblock \showarticletitle{Dspy: Compiling declarative language model calls into self-improving pipelines}.
\newblock \bibinfo{journal}{\emph{arXiv preprint arXiv:2310.03714}} (\bibinfo{year}{2023}).
\newblock


\bibitem[Kim et~al\mbox{.}(2024a)]%
        {kim2024memory}
\bibfield{author}{\bibinfo{person}{Jeonghoon Kim}, \bibinfo{person}{Jung~Hyun Lee}, \bibinfo{person}{Sungdong Kim}, \bibinfo{person}{Joonsuk Park}, \bibinfo{person}{Kang~Min Yoo}, \bibinfo{person}{Se~Jung Kwon}, {and} \bibinfo{person}{Dongsoo Lee}.} \bibinfo{year}{2024}\natexlab{a}.
\newblock \showarticletitle{Memory-efficient fine-tuning of compressed large language models via sub-4-bit integer quantization}.
\newblock \bibinfo{journal}{\emph{Advances in Neural Information Processing Systems}}  \bibinfo{volume}{36} (\bibinfo{year}{2024}).
\newblock


\bibitem[Kim et~al\mbox{.}(2024b)]%
        {kim2024token}
\bibfield{author}{\bibinfo{person}{Minsoo Kim}, \bibinfo{person}{Sihwa Lee}, \bibinfo{person}{Janghwan Lee}, \bibinfo{person}{Sukjin Hong}, \bibinfo{person}{Du-Seong Chang}, \bibinfo{person}{Wonyong Sung}, {and} \bibinfo{person}{Jungwook Choi}.} \bibinfo{year}{2024}\natexlab{b}.
\newblock \showarticletitle{Token-scaled logit distillation for ternary weight generative language models}.
\newblock \bibinfo{journal}{\emph{Advances in Neural Information Processing Systems}}  \bibinfo{volume}{36} (\bibinfo{year}{2024}).
\newblock


\bibitem[Kim et~al\mbox{.}(2023b)]%
        {kim2023squeezellm}
\bibfield{author}{\bibinfo{person}{Sehoon Kim}, \bibinfo{person}{Coleman Hooper}, \bibinfo{person}{Amir Gholami}, \bibinfo{person}{Zhen Dong}, \bibinfo{person}{Xiuyu Li}, \bibinfo{person}{Sheng Shen}, \bibinfo{person}{Michael~W Mahoney}, {and} \bibinfo{person}{Kurt Keutzer}.} \bibinfo{year}{2023}\natexlab{b}.
\newblock \showarticletitle{Squeezellm: Dense-and-sparse quantization}.
\newblock \bibinfo{journal}{\emph{arXiv preprint arXiv:2306.07629}} (\bibinfo{year}{2023}).
\newblock


\bibitem[Kim and Rush(2016)]%
        {kim2016sequence}
\bibfield{author}{\bibinfo{person}{Yoon Kim} {and} \bibinfo{person}{Alexander~M Rush}.} \bibinfo{year}{2016}\natexlab{}.
\newblock \showarticletitle{Sequence-Level Knowledge Distillation}. In \bibinfo{booktitle}{\emph{Proceedings of the 2016 Conference on Empirical Methods in Natural Language Processing}}. \bibinfo{pages}{1317--1327}.
\newblock


\bibitem[Kim et~al\mbox{.}(2023a)]%
        {kim2023mixture}
\bibfield{author}{\bibinfo{person}{Young~Jin Kim}, \bibinfo{person}{Raffy Fahim}, {and} \bibinfo{person}{Hany~Hassan Awadalla}.} \bibinfo{year}{2023}\natexlab{a}.
\newblock \showarticletitle{Mixture of Quantized Experts (MoQE): Complementary Effect of Low-bit Quantization and Robustness}.
\newblock \bibinfo{journal}{\emph{arXiv preprint arXiv:2310.02410}} (\bibinfo{year}{2023}).
\newblock


\bibitem[Ko et~al\mbox{.}(2024)]%
        {ko2024distillm}
\bibfield{author}{\bibinfo{person}{Jongwoo Ko}, \bibinfo{person}{Sungnyun Kim}, \bibinfo{person}{Tianyi Chen}, {and} \bibinfo{person}{Se-Young Yun}.} \bibinfo{year}{2024}\natexlab{}.
\newblock \showarticletitle{DistiLLM: Towards Streamlined Distillation for Large Language Models}.
\newblock \bibinfo{journal}{\emph{arXiv preprint arXiv:2402.03898}} (\bibinfo{year}{2024}).
\newblock


\bibitem[Kocetkov et~al\mbox{.}(2022)]%
        {kocetkov2022stack}
\bibfield{author}{\bibinfo{person}{Denis Kocetkov}, \bibinfo{person}{Raymond Li}, \bibinfo{person}{Loubna~Ben Allal}, \bibinfo{person}{Jia Li}, \bibinfo{person}{Chenghao Mou}, \bibinfo{person}{Carlos~Mu{\~n}oz Ferrandis}, \bibinfo{person}{Yacine Jernite}, \bibinfo{person}{Margaret Mitchell}, \bibinfo{person}{Sean Hughes}, \bibinfo{person}{Thomas Wolf}, {et~al\mbox{.}}} \bibinfo{year}{2022}\natexlab{}.
\newblock \showarticletitle{The stack: 3 tb of permissively licensed source code}.
\newblock \bibinfo{journal}{\emph{arXiv preprint arXiv:2211.15533}} (\bibinfo{year}{2022}).
\newblock


\bibitem[Kuhn et~al\mbox{.}(2023)]%
        {kuhn2023semantic}
\bibfield{author}{\bibinfo{person}{Lorenz Kuhn}, \bibinfo{person}{Yarin Gal}, {and} \bibinfo{person}{Sebastian Farquhar}.} \bibinfo{year}{2023}\natexlab{}.
\newblock \showarticletitle{Semantic Uncertainty: Linguistic Invariances for Uncertainty Estimation in Natural Language Generation}. In \bibinfo{booktitle}{\emph{Proceedings of the Eleventh International Conference on Learning Representations}}.
\newblock
\urldef\tempurl%
\url{https://openreview.net/forum?id=VD-AYtP0dve}
\showURL{%
\tempurl}


\bibitem[Kumar et~al\mbox{.}(2024)]%
        {kumar2024increased}
\bibfield{author}{\bibinfo{person}{Divyanshu Kumar}, \bibinfo{person}{Anurakt Kumar}, \bibinfo{person}{Sahil Agarwal}, {and} \bibinfo{person}{Prashanth Harshangi}.} \bibinfo{year}{2024}\natexlab{}.
\newblock \showarticletitle{Fine-Tuning, Quantization, and LLMs: Navigating Unintended Outcomes}.
\newblock \bibinfo{journal}{\emph{arXiv preprint arXiv:2404.04392}} (\bibinfo{year}{2024}).
\newblock


\bibitem[Kwon et~al\mbox{.}(2024)]%
        {kwon-etal-2024-slm}
\bibfield{author}{\bibinfo{person}{Ohjoon Kwon}, \bibinfo{person}{Donghyeon Jeon}, \bibinfo{person}{Nayoung Choi}, \bibinfo{person}{Gyu-Hwung Cho}, \bibinfo{person}{Hwiyeol Jo}, \bibinfo{person}{Changbong Kim}, \bibinfo{person}{Hyunwoo Lee}, \bibinfo{person}{Inho Kang}, \bibinfo{person}{Sun Kim}, {and} \bibinfo{person}{Taiwoo Park}.} \bibinfo{year}{2024}\natexlab{}.
\newblock \showarticletitle{SLM as Guardian: Pioneering AI Safety with Small Language Model}. In \bibinfo{booktitle}{\emph{Proceedings of the 2024 Conference on Empirical Methods in Natural Language Processing: Industry Track}}, \bibfield{editor}{\bibinfo{person}{Franck Dernoncourt}, \bibinfo{person}{Daniel Preo{\c{t}}iuc-Pietro}, {and} \bibinfo{person}{Anastasia Shimorina}} (Eds.). \bibinfo{publisher}{Association for Computational Linguistics}, \bibinfo{address}{Miami, Florida, US}, \bibinfo{pages}{1333--1350}.
\newblock
\urldef\tempurl%
\url{https://doi.org/10.18653/v1/2024.emnlp-industry.99}
\showDOI{\tempurl}


\bibitem[Labrak et~al\mbox{.}(2024)]%
        {labrak2024biomistral}
\bibfield{author}{\bibinfo{person}{Yanis Labrak}, \bibinfo{person}{Adrien Bazoge}, \bibinfo{person}{Emmanuel Morin}, \bibinfo{person}{Pierre-Antoine Gourraud}, \bibinfo{person}{Mickael Rouvier}, {and} \bibinfo{person}{Richard Dufour}.} \bibinfo{year}{2024}\natexlab{}.
\newblock \showarticletitle{Biomistral: A collection of open-source pretrained large language models for medical domains}.
\newblock \bibinfo{journal}{\emph{arXiv preprint arXiv:2402.10373}} (\bibinfo{year}{2024}).
\newblock


\bibitem[Larson et~al\mbox{.}(2019)]%
        {larson-etal-2019-evaluation}
\bibfield{author}{\bibinfo{person}{Stefan Larson}, \bibinfo{person}{Anish Mahendran}, \bibinfo{person}{Joseph~J. Peper}, \bibinfo{person}{Christopher Clarke}, \bibinfo{person}{Andrew Lee}, \bibinfo{person}{Parker Hill}, \bibinfo{person}{Jonathan~K. Kummerfeld}, \bibinfo{person}{Kevin Leach}, \bibinfo{person}{Michael~A. Laurenzano}, \bibinfo{person}{Lingjia Tang}, {and} \bibinfo{person}{Jason Mars}.} \bibinfo{year}{2019}\natexlab{}.
\newblock \showarticletitle{An Evaluation Dataset for Intent Classification and Out-of-Scope Prediction}. In \bibinfo{booktitle}{\emph{Proceedings of the 2019 Conference on Empirical Methods in Natural Language Processing and the 9th International Joint Conference on Natural Language Processing (EMNLP-IJCNLP)}}, \bibfield{editor}{\bibinfo{person}{Kentaro Inui}, \bibinfo{person}{Jing Jiang}, \bibinfo{person}{Vincent Ng}, {and} \bibinfo{person}{Xiaojun Wan}} (Eds.). \bibinfo{publisher}{Association for Computational Linguistics}, \bibinfo{address}{Hong Kong, China}, \bibinfo{pages}{1311--1316}.
\newblock
\urldef\tempurl%
\url{https://doi.org/10.18653/v1/D19-1131}
\showDOI{\tempurl}


\bibitem[Lauren{\c{c}}on et~al\mbox{.}(2022)]%
        {laurenccon2022bigscience}
\bibfield{author}{\bibinfo{person}{Hugo Lauren{\c{c}}on}, \bibinfo{person}{Lucile Saulnier}, \bibinfo{person}{Thomas Wang}, \bibinfo{person}{Christopher Akiki}, \bibinfo{person}{Albert Villanova~del Moral}, \bibinfo{person}{Teven Le~Scao}, \bibinfo{person}{Leandro Von~Werra}, \bibinfo{person}{Chenghao Mou}, \bibinfo{person}{Eduardo Gonz{\'a}lez~Ponferrada}, \bibinfo{person}{Huu Nguyen}, {et~al\mbox{.}}} \bibinfo{year}{2022}\natexlab{}.
\newblock \showarticletitle{The bigscience roots corpus: A 1.6 tb composite multilingual dataset}.
\newblock \bibinfo{journal}{\emph{Advances in Neural Information Processing Systems}}  \bibinfo{volume}{35} (\bibinfo{year}{2022}), \bibinfo{pages}{31809--31826}.
\newblock


\bibitem[Le~Scao et~al\mbox{.}(2023)]%
        {le2023bloom}
\bibfield{author}{\bibinfo{person}{Teven Le~Scao}, \bibinfo{person}{Angela Fan}, \bibinfo{person}{Christopher Akiki}, \bibinfo{person}{Ellie Pavlick}, \bibinfo{person}{Suzana Ili{\'c}}, \bibinfo{person}{Daniel Hesslow}, \bibinfo{person}{Roman Castagn{\'e}}, \bibinfo{person}{Alexandra~Sasha Luccioni}, \bibinfo{person}{Fran{\c{c}}ois Yvon}, \bibinfo{person}{Matthias Gall{\'e}}, {et~al\mbox{.}}} \bibinfo{year}{2023}\natexlab{}.
\newblock \showarticletitle{Bloom: A 176b-parameter open-access multilingual language model}.
\newblock  (\bibinfo{year}{2023}).
\newblock


\bibitem[Lee et~al\mbox{.}(2024a)]%
        {lee-etal-2024-mentor}
\bibfield{author}{\bibinfo{person}{Hojae Lee}, \bibinfo{person}{Junho Kim}, {and} \bibinfo{person}{SangKeun Lee}.} \bibinfo{year}{2024}\natexlab{a}.
\newblock \showarticletitle{Mentor-{KD}: Making Small Language Models Better Multi-step Reasoners}. In \bibinfo{booktitle}{\emph{Proceedings of the 2024 Conference on Empirical Methods in Natural Language Processing}}, \bibfield{editor}{\bibinfo{person}{Yaser Al-Onaizan}, \bibinfo{person}{Mohit Bansal}, {and} \bibinfo{person}{Yun-Nung Chen}} (Eds.). \bibinfo{publisher}{Association for Computational Linguistics}, \bibinfo{address}{Miami, Florida, USA}, \bibinfo{pages}{17643--17658}.
\newblock
\urldef\tempurl%
\url{https://doi.org/10.18653/v1/2024.emnlp-main.977}
\showDOI{\tempurl}


\bibitem[Lee et~al\mbox{.}(2024b)]%
        {lee2024can}
\bibfield{author}{\bibinfo{person}{Jooyoung Lee}, \bibinfo{person}{Fan Yang}, \bibinfo{person}{Thanh Tran}, \bibinfo{person}{Qian Hu}, \bibinfo{person}{Emre Barut}, {and} \bibinfo{person}{Kai-Wei Chang}.} \bibinfo{year}{2024}\natexlab{b}.
\newblock \showarticletitle{Can Small Language Models Help Large Language Models Reason Better?: LM-Guided Chain-of-Thought}. In \bibinfo{booktitle}{\emph{Proceedings of the 2024 Joint International Conference on Computational Linguistics, Language Resources and Evaluation (LREC-COLING 2024)}}. \bibinfo{pages}{2835--2843}.
\newblock


\bibitem[Lefaudeux et~al\mbox{.}(2022)]%
        {xFormers2022}
\bibfield{author}{\bibinfo{person}{Benjamin Lefaudeux}, \bibinfo{person}{Francisco Massa}, \bibinfo{person}{Diana Liskovich}, \bibinfo{person}{Wenhan Xiong}, \bibinfo{person}{Vittorio Caggiano}, \bibinfo{person}{Sean Naren}, \bibinfo{person}{Min Xu}, \bibinfo{person}{Jieru Hu}, \bibinfo{person}{Marta Tintore}, \bibinfo{person}{Susan Zhang}, \bibinfo{person}{Patrick Labatut}, \bibinfo{person}{Daniel Haziza}, \bibinfo{person}{Luca Wehrstedt}, \bibinfo{person}{Jeremy Reizenstein}, {and} \bibinfo{person}{Grigory Sizov}.} \bibinfo{year}{2022}\natexlab{}.
\newblock \bibinfo{title}{xFormers: A modular and hackable Transformer modelling library}.
\newblock \bibinfo{howpublished}{\url{https://github.com/facebookresearch/xformers}}.
\newblock


\bibitem[Lei~Ba et~al\mbox{.}(2016)]%
        {lei2016layer}
\bibfield{author}{\bibinfo{person}{Jimmy Lei~Ba}, \bibinfo{person}{Jamie~Ryan Kiros}, {and} \bibinfo{person}{Geoffrey~E Hinton}.} \bibinfo{year}{2016}\natexlab{}.
\newblock \showarticletitle{Layer normalization}.
\newblock \bibinfo{journal}{\emph{ArXiv e-prints}} (\bibinfo{year}{2016}), \bibinfo{pages}{arXiv--1607}.
\newblock


\bibitem[Lewkowycz et~al\mbox{.}(2022)]%
        {lewkowycz2022solving}
\bibfield{author}{\bibinfo{person}{Aitor Lewkowycz}, \bibinfo{person}{Anders Andreassen}, \bibinfo{person}{David Dohan}, \bibinfo{person}{Ethan Dyer}, \bibinfo{person}{Henryk Michalewski}, \bibinfo{person}{Vinay Ramasesh}, \bibinfo{person}{Ambrose Slone}, \bibinfo{person}{Cem Anil}, \bibinfo{person}{Imanol Schlag}, \bibinfo{person}{Theo Gutman-Solo}, {et~al\mbox{.}}} \bibinfo{year}{2022}\natexlab{}.
\newblock \showarticletitle{Solving quantitative reasoning problems with language models}.
\newblock \bibinfo{journal}{\emph{Advances in Neural Information Processing Systems}}  \bibinfo{volume}{35} (\bibinfo{year}{2022}), \bibinfo{pages}{3843--3857}.
\newblock


\bibitem[Li et~al\mbox{.}(2023c)]%
        {li2023mixed}
\bibfield{author}{\bibinfo{person}{Chenglin Li}, \bibinfo{person}{Qianglong Chen}, \bibinfo{person}{Liangyue Li}, \bibinfo{person}{Caiyu Wang}, \bibinfo{person}{Yicheng Li}, \bibinfo{person}{Zulong Chen}, {and} \bibinfo{person}{Yin Zhang}.} \bibinfo{year}{2023}\natexlab{c}.
\newblock \showarticletitle{Mixed distillation helps smaller language model better reasoning}.
\newblock \bibinfo{journal}{\emph{arXiv preprint arXiv:2312.10730}} (\bibinfo{year}{2023}).
\newblock


\bibitem[Li et~al\mbox{.}(2024h)]%
        {li2024lorap}
\bibfield{author}{\bibinfo{person}{Guangyan Li}, \bibinfo{person}{Yongqiang Tang}, {and} \bibinfo{person}{Wensheng Zhang}.} \bibinfo{year}{2024}\natexlab{h}.
\newblock \showarticletitle{LoRAP: Transformer Sub-Layers Deserve Differentiated Structured Compression for Large Language Models}.
\newblock \bibinfo{journal}{\emph{arXiv preprint arXiv:2404.09695}} (\bibinfo{year}{2024}).
\newblock


\bibitem[Li et~al\mbox{.}(2024a)]%
        {li2024bladeenhancingblackboxlarge}
\bibfield{author}{\bibinfo{person}{Haitao Li}, \bibinfo{person}{Qingyao Ai}, \bibinfo{person}{Jia Chen}, \bibinfo{person}{Qian Dong}, \bibinfo{person}{Zhijing Wu}, \bibinfo{person}{Yiqun Liu}, \bibinfo{person}{Chong Chen}, {and} \bibinfo{person}{Qi Tian}.} \bibinfo{year}{2024}\natexlab{a}.
\newblock \bibinfo{title}{BLADE: Enhancing Black-box Large Language Models with Small Domain-Specific Models}.
\newblock
\newblock
\showeprint[arxiv]{2403.18365}~[cs.CL]
\urldef\tempurl%
\url{https://arxiv.org/abs/2403.18365}
\showURL{%
\tempurl}


\bibitem[Li et~al\mbox{.}(2024c)]%
        {li-etal-2024-privlm}
\bibfield{author}{\bibinfo{person}{Haoran Li}, \bibinfo{person}{Dadi Guo}, \bibinfo{person}{Donghao Li}, \bibinfo{person}{Wei Fan}, \bibinfo{person}{Qi Hu}, \bibinfo{person}{Xin Liu}, \bibinfo{person}{Chunkit Chan}, \bibinfo{person}{Duanyi Yao}, \bibinfo{person}{Yuan Yao}, {and} \bibinfo{person}{Yangqiu Song}.} \bibinfo{year}{2024}\natexlab{c}.
\newblock \showarticletitle{{P}riv{LM}-Bench: A Multi-level Privacy Evaluation Benchmark for Language Models}. In \bibinfo{booktitle}{\emph{Proceedings of the 62nd Annual Meeting of the Association for Computational Linguistics}}. \bibinfo{pages}{54--73}.
\newblock


\bibitem[Li et~al\mbox{.}(2023d)]%
        {li-etal-2023-halueval}
\bibfield{author}{\bibinfo{person}{Junyi Li}, \bibinfo{person}{Xiaoxue Cheng}, \bibinfo{person}{Xin Zhao}, \bibinfo{person}{Jian-Yun Nie}, {and} \bibinfo{person}{Ji-Rong Wen}.} \bibinfo{year}{2023}\natexlab{d}.
\newblock \showarticletitle{{H}alu{E}val: A Large-Scale Hallucination Evaluation Benchmark for Large Language Models}. In \bibinfo{booktitle}{\emph{Proceedings of the 2023 Conference on Empirical Methods in Natural Language Processing}}. \bibinfo{publisher}{Association for Computational Linguistics}, \bibinfo{address}{Singapore}, \bibinfo{pages}{6449--6464}.
\newblock
\urldef\tempurl%
\url{https://doi.org/10.18653/v1/2023.emnlp-main.397}
\showDOI{\tempurl}


\bibitem[Li et~al\mbox{.}(2024b)]%
        {li2024datacomp}
\bibfield{author}{\bibinfo{person}{Jeffrey Li}, \bibinfo{person}{Alex Fang}, \bibinfo{person}{Georgios Smyrnis}, \bibinfo{person}{Maor Ivgi}, \bibinfo{person}{Matt Jordan}, \bibinfo{person}{Samir Gadre}, \bibinfo{person}{Hritik Bansal}, \bibinfo{person}{Etash Guha}, \bibinfo{person}{Sedrick Keh}, \bibinfo{person}{Kushal Arora}, {et~al\mbox{.}}} \bibinfo{year}{2024}\natexlab{b}.
\newblock \showarticletitle{Datacomp-lm: In search of the next generation of training sets for language models}.
\newblock \bibinfo{journal}{\emph{arXiv preprint arXiv:2406.11794}} (\bibinfo{year}{2024}).
\newblock


\bibitem[Li et~al\mbox{.}(2024g)]%
        {li2024transformerlitehighefficiencydeploymentlarge}
\bibfield{author}{\bibinfo{person}{Luchang Li}, \bibinfo{person}{Sheng Qian}, \bibinfo{person}{Jie Lu}, \bibinfo{person}{Lunxi Yuan}, \bibinfo{person}{Rui Wang}, {and} \bibinfo{person}{Qin Xie}.} \bibinfo{year}{2024}\natexlab{g}.
\newblock \bibinfo{title}{Transformer-Lite: High-efficiency Deployment of Large Language Models on Mobile Phone GPUs}.
\newblock
\newblock
\showeprint[arxiv]{2403.20041}~[cs.CL]
\urldef\tempurl%
\url{https://arxiv.org/abs/2403.20041}
\showURL{%
\tempurl}


\bibitem[Li et~al\mbox{.}(2024e)]%
        {li2024examining}
\bibfield{author}{\bibinfo{person}{Pingzhi Li}, \bibinfo{person}{Xiaolong Jin}, \bibinfo{person}{Yu Cheng}, {and} \bibinfo{person}{Tianlong Chen}.} \bibinfo{year}{2024}\natexlab{e}.
\newblock \showarticletitle{Examining Post-Training Quantization for Mixture-of-Experts: A Benchmark}.
\newblock \bibinfo{journal}{\emph{arXiv preprint arXiv:2406.08155}} (\bibinfo{year}{2024}).
\newblock


\bibitem[Li et~al\mbox{.}(2023a)]%
        {li2023starcodersourceyou}
\bibfield{author}{\bibinfo{person}{Raymond Li}, \bibinfo{person}{Loubna~Ben Allal}, \bibinfo{person}{Yangtian Zi}, \bibinfo{person}{Niklas Muennighoff}, \bibinfo{person}{Denis Kocetkov}, \bibinfo{person}{Chenghao Mou}, \bibinfo{person}{Marc Marone}, \bibinfo{person}{Christopher Akiki}, \bibinfo{person}{Jia Li}, \bibinfo{person}{Jenny Chim}, \bibinfo{person}{Qian Liu}, \bibinfo{person}{Evgenii Zheltonozhskii}, \bibinfo{person}{Terry~Yue Zhuo}, \bibinfo{person}{Thomas Wang}, \bibinfo{person}{Olivier Dehaene}, \bibinfo{person}{Mishig Davaadorj}, \bibinfo{person}{Joel Lamy-Poirier}, \bibinfo{person}{João Monteiro}, \bibinfo{person}{Oleh Shliazhko}, \bibinfo{person}{Nicolas Gontier}, \bibinfo{person}{Nicholas Meade}, \bibinfo{person}{Armel Zebaze}, \bibinfo{person}{Ming-Ho Yee}, \bibinfo{person}{Logesh~Kumar Umapathi}, \bibinfo{person}{Jian Zhu}, \bibinfo{person}{Benjamin Lipkin}, \bibinfo{person}{Muhtasham Oblokulov}, \bibinfo{person}{Zhiruo Wang}, \bibinfo{person}{Rudra Murthy}, \bibinfo{person}{Jason
  Stillerman}, \bibinfo{person}{Siva~Sankalp Patel}, \bibinfo{person}{Dmitry Abulkhanov}, \bibinfo{person}{Marco Zocca}, \bibinfo{person}{Manan Dey}, \bibinfo{person}{Zhihan Zhang}, \bibinfo{person}{Nour Fahmy}, \bibinfo{person}{Urvashi Bhattacharyya}, \bibinfo{person}{Wenhao Yu}, \bibinfo{person}{Swayam Singh}, \bibinfo{person}{Sasha Luccioni}, \bibinfo{person}{Paulo Villegas}, \bibinfo{person}{Maxim Kunakov}, \bibinfo{person}{Fedor Zhdanov}, \bibinfo{person}{Manuel Romero}, \bibinfo{person}{Tony Lee}, \bibinfo{person}{Nadav Timor}, \bibinfo{person}{Jennifer Ding}, \bibinfo{person}{Claire Schlesinger}, \bibinfo{person}{Hailey Schoelkopf}, \bibinfo{person}{Jan Ebert}, \bibinfo{person}{Tri Dao}, \bibinfo{person}{Mayank Mishra}, \bibinfo{person}{Alex Gu}, \bibinfo{person}{Jennifer Robinson}, \bibinfo{person}{Carolyn~Jane Anderson}, \bibinfo{person}{Brendan Dolan-Gavitt}, \bibinfo{person}{Danish Contractor}, \bibinfo{person}{Siva Reddy}, \bibinfo{person}{Daniel Fried}, \bibinfo{person}{Dzmitry Bahdanau},
  \bibinfo{person}{Yacine Jernite}, \bibinfo{person}{Carlos~Muñoz Ferrandis}, \bibinfo{person}{Sean Hughes}, \bibinfo{person}{Thomas Wolf}, \bibinfo{person}{Arjun Guha}, \bibinfo{person}{Leandro von Werra}, {and} \bibinfo{person}{Harm de Vries}.} \bibinfo{year}{2023}\natexlab{a}.
\newblock \bibinfo{title}{StarCoder: may the source be with you!}
\newblock
\newblock
\showeprint[arxiv]{2305.06161}~[cs.CL]
\urldef\tempurl%
\url{https://arxiv.org/abs/2305.06161}
\showURL{%
\tempurl}


\bibitem[Li et~al\mbox{.}(2024d)]%
        {li2024nuteprune}
\bibfield{author}{\bibinfo{person}{Shengrui Li}, \bibinfo{person}{Xueting Han}, {and} \bibinfo{person}{Jing Bai}.} \bibinfo{year}{2024}\natexlab{d}.
\newblock \showarticletitle{Nuteprune: Efficient progressive pruning with numerous teachers for large language models}.
\newblock \bibinfo{journal}{\emph{arXiv preprint arXiv:2402.09773}} (\bibinfo{year}{2024}).
\newblock


\bibitem[Li et~al\mbox{.}(2024f)]%
        {li2024purifying}
\bibfield{author}{\bibinfo{person}{Tianlin Li}, \bibinfo{person}{Qian Liu}, \bibinfo{person}{Tianyu Pang}, \bibinfo{person}{Chao Du}, \bibinfo{person}{Qing Guo}, \bibinfo{person}{Yang Liu}, {and} \bibinfo{person}{Min Lin}.} \bibinfo{year}{2024}\natexlab{f}.
\newblock \showarticletitle{Purifying large language models by ensembling a small language model}.
\newblock \bibinfo{journal}{\emph{arXiv preprint arXiv:2402.14845}} (\bibinfo{year}{2024}).
\newblock


\bibitem[Li et~al\mbox{.}(2023e)]%
        {li2023contrastive}
\bibfield{author}{\bibinfo{person}{Xiang~Lisa Li}, \bibinfo{person}{Ari Holtzman}, \bibinfo{person}{Daniel Fried}, \bibinfo{person}{Percy Liang}, \bibinfo{person}{Jason Eisner}, \bibinfo{person}{Tatsunori~B Hashimoto}, \bibinfo{person}{Luke Zettlemoyer}, {and} \bibinfo{person}{Mike Lewis}.} \bibinfo{year}{2023}\natexlab{e}.
\newblock \showarticletitle{Contrastive Decoding: Open-ended Text Generation as Optimization}. In \bibinfo{booktitle}{\emph{Proceedings of the 61st Annual Meeting of the Association for Computational Linguistics (Volume 1: Long Papers)}}. \bibinfo{pages}{12286--12312}.
\newblock


\bibitem[Li and Liang(2021)]%
        {li2021prefix}
\bibfield{author}{\bibinfo{person}{Xiang~Lisa Li} {and} \bibinfo{person}{Percy Liang}.} \bibinfo{year}{2021}\natexlab{}.
\newblock \showarticletitle{Prefix-tuning: Optimizing continuous prompts for generation}.
\newblock \bibinfo{journal}{\emph{arXiv preprint arXiv:2101.00190}} (\bibinfo{year}{2021}).
\newblock


\bibitem[Li et~al\mbox{.}(2023b)]%
        {li2023textbooksneediiphi15}
\bibfield{author}{\bibinfo{person}{Yuanzhi Li}, \bibinfo{person}{Sébastien Bubeck}, \bibinfo{person}{Ronen Eldan}, \bibinfo{person}{Allie~Del Giorno}, \bibinfo{person}{Suriya Gunasekar}, {and} \bibinfo{person}{Yin~Tat Lee}.} \bibinfo{year}{2023}\natexlab{b}.
\newblock \bibinfo{title}{Textbooks Are All You Need II: phi-1.5 technical report}.
\newblock
\newblock
\showeprint[arxiv]{2309.05463}~[cs.CL]
\urldef\tempurl%
\url{https://arxiv.org/abs/2309.05463}
\showURL{%
\tempurl}


\bibitem[Li et~al\mbox{.}(2023f)]%
        {li2023sparse}
\bibfield{author}{\bibinfo{person}{Yun Li}, \bibinfo{person}{Lin Niu}, \bibinfo{person}{Xipeng Zhang}, \bibinfo{person}{Kai Liu}, \bibinfo{person}{Jianchen Zhu}, {and} \bibinfo{person}{Zhanhui Kang}.} \bibinfo{year}{2023}\natexlab{f}.
\newblock \showarticletitle{E-sparse: Boosting the large language model inference through entropy-based n: M sparsity}.
\newblock \bibinfo{journal}{\emph{arXiv preprint arXiv:2310.15929}} (\bibinfo{year}{2023}).
\newblock


\bibitem[Li et~al\mbox{.}(2023g)]%
        {li2023large}
\bibfield{author}{\bibinfo{person}{Yinheng Li}, \bibinfo{person}{Shaofei Wang}, \bibinfo{person}{Han Ding}, {and} \bibinfo{person}{Hang Chen}.} \bibinfo{year}{2023}\natexlab{g}.
\newblock \showarticletitle{Large language models in finance: A survey}. In \bibinfo{booktitle}{\emph{Proceedings of the fourth ACM international conference on AI in finance}}. \bibinfo{pages}{374--382}.
\newblock


\bibitem[Lian et~al\mbox{.}(2023)]%
        {SlimOrca}
\bibfield{author}{\bibinfo{person}{Wing Lian}, \bibinfo{person}{Guan Wang}, \bibinfo{person}{Bleys Goodson}, \bibinfo{person}{Eugene Pentland}, \bibinfo{person}{Austin Cook}, \bibinfo{person}{Chanvichet Vong}, {and} \bibinfo{person}{"Teknium"}.} \bibinfo{year}{2023}\natexlab{}.
\newblock \bibinfo{title}{SlimOrca: An Open Dataset of GPT-4 Augmented FLAN Reasoning Traces, with Verification}.
\newblock
\newblock
\urldef\tempurl%
\url{https://https://huggingface.co/Open-Orca/SlimOrca}
\showURL{%
\tempurl}


\bibitem[Liang et~al\mbox{.}(2024)]%
        {liang2024synergizing}
\bibfield{author}{\bibinfo{person}{Jinggui Liang}, \bibinfo{person}{Lizi Liao}, \bibinfo{person}{Hao Fei}, {and} \bibinfo{person}{Jing Jiang}.} \bibinfo{year}{2024}\natexlab{}.
\newblock \showarticletitle{Synergizing Large Language Models and Pre-Trained Smaller Models for Conversational Intent Discovery}. In \bibinfo{booktitle}{\emph{Findings of the Association for Computational Linguistics ACL 2024}}. \bibinfo{pages}{14133--14147}.
\newblock


\bibitem[Liang et~al\mbox{.}(2023)]%
        {liang2023holistic}
\bibfield{author}{\bibinfo{person}{Percy Liang}, \bibinfo{person}{Rishi Bommasani}, \bibinfo{person}{Tony Lee}, \bibinfo{person}{Dimitris Tsipras}, \bibinfo{person}{Dilara Soylu}, \bibinfo{person}{Michihiro Yasunaga}, \bibinfo{person}{Yian Zhang}, \bibinfo{person}{Deepak Narayanan}, \bibinfo{person}{Yuhuai Wu}, \bibinfo{person}{Ananya Kumar}, \bibinfo{person}{Benjamin Newman}, \bibinfo{person}{Binhang Yuan}, \bibinfo{person}{Bobby Yan}, \bibinfo{person}{Ce Zhang}, \bibinfo{person}{Christian~Alexander Cosgrove}, \bibinfo{person}{Christopher~D Manning}, \bibinfo{person}{Christopher Re}, \bibinfo{person}{Diana Acosta-Navas}, \bibinfo{person}{Drew~Arad Hudson}, \bibinfo{person}{Eric Zelikman}, \bibinfo{person}{Esin Durmus}, \bibinfo{person}{Faisal Ladhak}, \bibinfo{person}{Frieda Rong}, \bibinfo{person}{Hongyu Ren}, \bibinfo{person}{Huaxiu Yao}, \bibinfo{person}{Jue WANG}, \bibinfo{person}{Keshav Santhanam}, \bibinfo{person}{Laurel Orr}, \bibinfo{person}{Lucia Zheng}, \bibinfo{person}{Mert Yuksekgonul},
  \bibinfo{person}{Mirac Suzgun}, \bibinfo{person}{Nathan Kim}, \bibinfo{person}{Neel Guha}, \bibinfo{person}{Niladri~S. Chatterji}, \bibinfo{person}{Omar Khattab}, \bibinfo{person}{Peter Henderson}, \bibinfo{person}{Qian Huang}, \bibinfo{person}{Ryan~Andrew Chi}, \bibinfo{person}{Sang~Michael Xie}, \bibinfo{person}{Shibani Santurkar}, \bibinfo{person}{Surya Ganguli}, \bibinfo{person}{Tatsunori Hashimoto}, \bibinfo{person}{Thomas Icard}, \bibinfo{person}{Tianyi Zhang}, \bibinfo{person}{Vishrav Chaudhary}, \bibinfo{person}{William Wang}, \bibinfo{person}{Xuechen Li}, \bibinfo{person}{Yifan Mai}, \bibinfo{person}{Yuhui Zhang}, {and} \bibinfo{person}{Yuta Koreeda}.} \bibinfo{year}{2023}\natexlab{}.
\newblock \showarticletitle{Holistic Evaluation of Language Models}.
\newblock \bibinfo{journal}{\emph{Transactions on Machine Learning Research}} (\bibinfo{year}{2023}).
\newblock
\showISSN{2835-8856}
\urldef\tempurl%
\url{https://openreview.net/forum?id=iO4LZibEqW}
\showURL{%
\tempurl}


\bibitem[Lin et~al\mbox{.}(2024b)]%
        {lin2024rella}
\bibfield{author}{\bibinfo{person}{Jianghao Lin}, \bibinfo{person}{Rong Shan}, \bibinfo{person}{Chenxu Zhu}, \bibinfo{person}{Kounianhua Du}, \bibinfo{person}{Bo Chen}, \bibinfo{person}{Shigang Quan}, \bibinfo{person}{Ruiming Tang}, \bibinfo{person}{Yong Yu}, {and} \bibinfo{person}{Weinan Zhang}.} \bibinfo{year}{2024}\natexlab{b}.
\newblock \showarticletitle{Rella: Retrieval-enhanced large language models for lifelong sequential behavior comprehension in recommendation}. In \bibinfo{booktitle}{\emph{Proceedings of the ACM on Web Conference 2024}}. \bibinfo{pages}{3497--3508}.
\newblock


\bibitem[Lin et~al\mbox{.}(2024c)]%
        {lin2024awq}
\bibfield{author}{\bibinfo{person}{Ji Lin}, \bibinfo{person}{Jiaming Tang}, \bibinfo{person}{Haotian Tang}, \bibinfo{person}{Shang Yang}, \bibinfo{person}{Wei-Ming Chen}, \bibinfo{person}{Wei-Chen Wang}, \bibinfo{person}{Guangxuan Xiao}, \bibinfo{person}{Xingyu Dang}, \bibinfo{person}{Chuang Gan}, {and} \bibinfo{person}{Song Han}.} \bibinfo{year}{2024}\natexlab{c}.
\newblock \showarticletitle{AWQ: Activation-aware Weight Quantization for On-Device LLM Compression and Acceleration}.
\newblock \bibinfo{journal}{\emph{Proceedings of Machine Learning and Systems}}  \bibinfo{volume}{6} (\bibinfo{year}{2024}), \bibinfo{pages}{87--100}.
\newblock


\bibitem[Lin et~al\mbox{.}(2022a)]%
        {lin2022truthfulqa}
\bibfield{author}{\bibinfo{person}{Stephanie Lin}, \bibinfo{person}{Jacob Hilton}, {and} \bibinfo{person}{Owain Evans}.} \bibinfo{year}{2022}\natexlab{a}.
\newblock \showarticletitle{TruthfulQA: Measuring How Models Mimic Human Falsehoods}. In \bibinfo{booktitle}{\emph{Proceedings of the 60th Annual Meeting of the Association for Computational Linguistics (Volume 1: Long Papers)}}. \bibinfo{pages}{3214--3252}.
\newblock


\bibitem[Lin et~al\mbox{.}(2022b)]%
        {lin-etal-2022-shot}
\bibfield{author}{\bibinfo{person}{Xi~Victoria Lin}, \bibinfo{person}{Todor Mihaylov}, \bibinfo{person}{Mikel Artetxe}, \bibinfo{person}{Tianlu Wang}, \bibinfo{person}{Shuohui Chen}, \bibinfo{person}{Daniel Simig}, \bibinfo{person}{Myle Ott}, \bibinfo{person}{Naman Goyal}, \bibinfo{person}{Shruti Bhosale}, \bibinfo{person}{Jingfei Du}, \bibinfo{person}{Ramakanth Pasunuru}, \bibinfo{person}{Sam Shleifer}, \bibinfo{person}{Punit~Singh Koura}, \bibinfo{person}{Vishrav Chaudhary}, \bibinfo{person}{Brian O{'}Horo}, \bibinfo{person}{Jeff Wang}, \bibinfo{person}{Luke Zettlemoyer}, \bibinfo{person}{Zornitsa Kozareva}, \bibinfo{person}{Mona Diab}, \bibinfo{person}{Veselin Stoyanov}, {and} \bibinfo{person}{Xian Li}.} \bibinfo{year}{2022}\natexlab{b}.
\newblock \showarticletitle{Few-shot Learning with Multilingual Generative Language Models}. In \bibinfo{booktitle}{\emph{Proceedings of the 2022 Conference on Empirical Methods in Natural Language Processing}}, \bibfield{editor}{\bibinfo{person}{Yoav Goldberg}, \bibinfo{person}{Zornitsa Kozareva}, {and} \bibinfo{person}{Yue Zhang}} (Eds.). \bibinfo{publisher}{Association for Computational Linguistics}, \bibinfo{address}{Abu Dhabi, United Arab Emirates}, \bibinfo{pages}{9019--9052}.
\newblock
\urldef\tempurl%
\url{https://doi.org/10.18653/v1/2022.emnlp-main.616}
\showDOI{\tempurl}


\bibitem[Lin et~al\mbox{.}(2024a)]%
        {lin2024rho}
\bibfield{author}{\bibinfo{person}{Zhenghao Lin}, \bibinfo{person}{Zhibin Gou}, \bibinfo{person}{Yeyun Gong}, \bibinfo{person}{Xiao Liu}, \bibinfo{person}{Yelong Shen}, \bibinfo{person}{Ruochen Xu}, \bibinfo{person}{Chen Lin}, \bibinfo{person}{Yujiu Yang}, \bibinfo{person}{Jian Jiao}, \bibinfo{person}{Nan Duan}, {et~al\mbox{.}}} \bibinfo{year}{2024}\natexlab{a}.
\newblock \showarticletitle{Rho-1: Not all tokens are what you need}.
\newblock \bibinfo{journal}{\emph{arXiv preprint arXiv:2404.07965}} (\bibinfo{year}{2024}).
\newblock


\bibitem[Liu et~al\mbox{.}(2024c)]%
        {liu2024deepseek}
\bibfield{author}{\bibinfo{person}{Aixin Liu}, \bibinfo{person}{Bei Feng}, \bibinfo{person}{Bin Wang}, \bibinfo{person}{Bingxuan Wang}, \bibinfo{person}{Bo Liu}, \bibinfo{person}{Chenggang Zhao}, \bibinfo{person}{Chengqi Dengr}, \bibinfo{person}{Chong Ruan}, \bibinfo{person}{Damai Dai}, \bibinfo{person}{Daya Guo}, {et~al\mbox{.}}} \bibinfo{year}{2024}\natexlab{c}.
\newblock \showarticletitle{Deepseek-v2: A strong, economical, and efficient mixture-of-experts language model}.
\newblock \bibinfo{journal}{\emph{arXiv preprint arXiv:2405.04434}} (\bibinfo{year}{2024}).
\newblock


\bibitem[Liu et~al\mbox{.}(2024d)]%
        {liu2024tuning}
\bibfield{author}{\bibinfo{person}{Alisa Liu}, \bibinfo{person}{Xiaochuang Han}, \bibinfo{person}{Yizhong Wang}, \bibinfo{person}{Yulia Tsvetkov}, \bibinfo{person}{Yejin Choi}, {and} \bibinfo{person}{Noah~A Smith}.} \bibinfo{year}{2024}\natexlab{d}.
\newblock \showarticletitle{Tuning language models by proxy}.
\newblock \bibinfo{journal}{\emph{arXiv preprint arXiv:2401.08565}} (\bibinfo{year}{2024}).
\newblock


\bibitem[Liu et~al\mbox{.}(2021)]%
        {liu2021towards}
\bibfield{author}{\bibinfo{person}{Jiashuo Liu}, \bibinfo{person}{Zheyan Shen}, \bibinfo{person}{Yue He}, \bibinfo{person}{Xingxuan Zhang}, \bibinfo{person}{Renzhe Xu}, \bibinfo{person}{Han Yu}, {and} \bibinfo{person}{Peng Cui}.} \bibinfo{year}{2021}\natexlab{}.
\newblock \showarticletitle{Towards out-of-distribution generalization: A survey}.
\newblock \bibinfo{journal}{\emph{arXiv preprint arXiv:2108.13624}} (\bibinfo{year}{2021}).
\newblock


\bibitem[Liu et~al\mbox{.}(2024a)]%
        {liu2024once}
\bibfield{author}{\bibinfo{person}{Qijiong Liu}, \bibinfo{person}{Nuo Chen}, \bibinfo{person}{Tetsuya Sakai}, {and} \bibinfo{person}{Xiao-Ming Wu}.} \bibinfo{year}{2024}\natexlab{a}.
\newblock \showarticletitle{Once: Boosting content-based recommendation with both open-and closed-source large language models}. In \bibinfo{booktitle}{\emph{Proceedings of the 17th ACM International Conference on Web Search and Data Mining}}. \bibinfo{pages}{452--461}.
\newblock


\bibitem[Liu et~al\mbox{.}(2024g)]%
        {liu2024can}
\bibfield{author}{\bibinfo{person}{Suqing Liu}, \bibinfo{person}{Zezhu Yu}, \bibinfo{person}{Feiran Huang}, \bibinfo{person}{Yousef Bulbulia}, \bibinfo{person}{Andreas Bergen}, {and} \bibinfo{person}{Michael Liut}.} \bibinfo{year}{2024}\natexlab{g}.
\newblock \showarticletitle{Can Small Language Models With Retrieval-Augmented Generation Replace Large Language Models When Learning Computer Science?}
\newblock In \bibinfo{booktitle}{\emph{Proceedings of the 2024 on Innovation and Technology in Computer Science Education V. 1}}. \bibinfo{pages}{388--393}.
\newblock


\bibitem[Liu et~al\mbox{.}(2023c)]%
        {liu2023makes}
\bibfield{author}{\bibinfo{person}{Wei Liu}, \bibinfo{person}{Weihao Zeng}, \bibinfo{person}{Keqing He}, \bibinfo{person}{Yong Jiang}, {and} \bibinfo{person}{Junxian He}.} \bibinfo{year}{2023}\natexlab{c}.
\newblock \showarticletitle{What makes good data for alignment? a comprehensive study of automatic data selection in instruction tuning}.
\newblock \bibinfo{journal}{\emph{arXiv preprint arXiv:2312.15685}} (\bibinfo{year}{2023}).
\newblock


\bibitem[Liu et~al\mbox{.}(2024e)]%
        {liu2024let}
\bibfield{author}{\bibinfo{person}{Yinpeng Liu}, \bibinfo{person}{Jiawei Liu}, \bibinfo{person}{Xiang Shi}, \bibinfo{person}{Qikai Cheng}, {and} \bibinfo{person}{Wei Lu}.} \bibinfo{year}{2024}\natexlab{e}.
\newblock \showarticletitle{Let's Learn Step by Step: Enhancing In-Context Learning Ability with Curriculum Learning}.
\newblock \bibinfo{journal}{\emph{arXiv preprint arXiv:2402.10738}} (\bibinfo{year}{2024}).
\newblock


\bibitem[Liu et~al\mbox{.}(2019)]%
        {liu2019roberta}
\bibfield{author}{\bibinfo{person}{Yinhan Liu}, \bibinfo{person}{Myle Ott}, \bibinfo{person}{Naman Goyal}, \bibinfo{person}{Jingfei Du}, \bibinfo{person}{Mandar Joshi}, \bibinfo{person}{Danqi Chen}, \bibinfo{person}{Omer Levy}, \bibinfo{person}{Mike Lewis}, \bibinfo{person}{Luke Zettlemoyer}, {and} \bibinfo{person}{Veselin Stoyanov}.} \bibinfo{year}{2019}\natexlab{}.
\newblock \showarticletitle{Roberta: A robustly optimized bert pretraining approach}.
\newblock \bibinfo{journal}{\emph{arXiv preprint arXiv:1907.11692}} (\bibinfo{year}{2019}).
\newblock


\bibitem[Liu et~al\mbox{.}(2024f)]%
        {liu2024ra}
\bibfield{author}{\bibinfo{person}{Yanming Liu}, \bibinfo{person}{Xinyue Peng}, \bibinfo{person}{Xuhong Zhang}, \bibinfo{person}{Weihao Liu}, \bibinfo{person}{Jianwei Yin}, \bibinfo{person}{Jiannan Cao}, {and} \bibinfo{person}{Tianyu Du}.} \bibinfo{year}{2024}\natexlab{f}.
\newblock \showarticletitle{RA-ISF: Learning to Answer and Understand from Retrieval Augmentation via Iterative Self-Feedback}.
\newblock \bibinfo{journal}{\emph{arXiv preprint arXiv:2403.06840}} (\bibinfo{year}{2024}).
\newblock


\bibitem[Liu et~al\mbox{.}(2024b)]%
        {liu2024scissorhands}
\bibfield{author}{\bibinfo{person}{Zichang Liu}, \bibinfo{person}{Aditya Desai}, \bibinfo{person}{Fangshuo Liao}, \bibinfo{person}{Weitao Wang}, \bibinfo{person}{Victor Xie}, \bibinfo{person}{Zhaozhuo Xu}, \bibinfo{person}{Anastasios Kyrillidis}, {and} \bibinfo{person}{Anshumali Shrivastava}.} \bibinfo{year}{2024}\natexlab{b}.
\newblock \showarticletitle{Scissorhands: Exploiting the persistence of importance hypothesis for llm kv cache compression at test time}.
\newblock \bibinfo{journal}{\emph{Advances in Neural Information Processing Systems}}  \bibinfo{volume}{36} (\bibinfo{year}{2024}).
\newblock


\bibitem[Liu et~al\mbox{.}(2023a)]%
        {liu2023llm}
\bibfield{author}{\bibinfo{person}{Zechun Liu}, \bibinfo{person}{Barlas Oguz}, \bibinfo{person}{Changsheng Zhao}, \bibinfo{person}{Ernie Chang}, \bibinfo{person}{Pierre Stock}, \bibinfo{person}{Yashar Mehdad}, \bibinfo{person}{Yangyang Shi}, \bibinfo{person}{Raghuraman Krishnamoorthi}, {and} \bibinfo{person}{Vikas Chandra}.} \bibinfo{year}{2023}\natexlab{a}.
\newblock \showarticletitle{Llm-qat: Data-free quantization aware training for large language models}.
\newblock \bibinfo{journal}{\emph{arXiv preprint arXiv:2305.17888}} (\bibinfo{year}{2023}).
\newblock


\bibitem[Liu et~al\mbox{.}(2023b)]%
        {liu-etal-2023-maximum}
\bibfield{author}{\bibinfo{person}{Zhengxiao Liu}, \bibinfo{person}{Bowen Shen}, \bibinfo{person}{Zheng Lin}, \bibinfo{person}{Fali Wang}, {and} \bibinfo{person}{Weiping Wang}.} \bibinfo{year}{2023}\natexlab{b}.
\newblock \showarticletitle{Maximum Entropy Loss, the Silver Bullet Targeting Backdoor Attacks in Pre-trained Language Models}. In \bibinfo{booktitle}{\emph{Findings of the Association for Computational Linguistics: ACL 2023}}, \bibfield{editor}{\bibinfo{person}{Anna Rogers}, \bibinfo{person}{Jordan Boyd-Graber}, {and} \bibinfo{person}{Naoaki Okazaki}} (Eds.). \bibinfo{publisher}{Association for Computational Linguistics}, \bibinfo{address}{Toronto, Canada}, \bibinfo{pages}{3850--3868}.
\newblock
\urldef\tempurl%
\url{https://doi.org/10.18653/v1/2023.findings-acl.237}
\showDOI{\tempurl}


\bibitem[Liu et~al\mbox{.}(2024h)]%
        {liu2024mobilellm}
\bibfield{author}{\bibinfo{person}{Zechun Liu}, \bibinfo{person}{Changsheng Zhao}, \bibinfo{person}{Forrest Iandola}, \bibinfo{person}{Chen Lai}, \bibinfo{person}{Yuandong Tian}, \bibinfo{person}{Igor Fedorov}, \bibinfo{person}{Yunyang Xiong}, \bibinfo{person}{Ernie Chang}, \bibinfo{person}{Yangyang Shi}, \bibinfo{person}{Raghuraman Krishnamoorthi}, {et~al\mbox{.}}} \bibinfo{year}{2024}\natexlab{h}.
\newblock \showarticletitle{Mobilellm: Optimizing sub-billion parameter language models for on-device use cases}.
\newblock \bibinfo{journal}{\emph{arXiv preprint arXiv:2402.14905}} (\bibinfo{year}{2024}).
\newblock


\bibitem[Long et~al\mbox{.}(2024)]%
        {long2024llms}
\bibfield{author}{\bibinfo{person}{Lin Long}, \bibinfo{person}{Rui Wang}, \bibinfo{person}{Ruixuan Xiao}, \bibinfo{person}{Junbo Zhao}, \bibinfo{person}{Xiao Ding}, \bibinfo{person}{Gang Chen}, {and} \bibinfo{person}{Haobo Wang}.} \bibinfo{year}{2024}\natexlab{}.
\newblock \showarticletitle{On LLMs-Driven Synthetic Data Generation, Curation, and Evaluation: A Survey}. In \bibinfo{booktitle}{\emph{Findings of the Association for Computational Linguistics ACL 2024}}. \bibinfo{pages}{11065--11082}.
\newblock


\bibitem[Longpre et~al\mbox{.}(2023a)]%
        {longpre2023flan}
\bibfield{author}{\bibinfo{person}{Shayne Longpre}, \bibinfo{person}{Le Hou}, \bibinfo{person}{Tu Vu}, \bibinfo{person}{Albert Webson}, \bibinfo{person}{Hyung~Won Chung}, \bibinfo{person}{Yi Tay}, \bibinfo{person}{Denny Zhou}, \bibinfo{person}{Quoc~V Le}, \bibinfo{person}{Barret Zoph}, \bibinfo{person}{Jason Wei}, {et~al\mbox{.}}} \bibinfo{year}{2023}\natexlab{a}.
\newblock \showarticletitle{The flan collection: Designing data and methods for effective instruction tuning}. In \bibinfo{booktitle}{\emph{International Conference on Machine Learning}}. PMLR, \bibinfo{pages}{22631--22648}.
\newblock


\bibitem[Longpre et~al\mbox{.}(2023b)]%
        {longpre2023data}
\bibfield{author}{\bibinfo{person}{Shayne Longpre}, \bibinfo{person}{Robert Mahari}, \bibinfo{person}{Anthony Chen}, \bibinfo{person}{Naana Obeng-Marnu}, \bibinfo{person}{Damien Sileo}, \bibinfo{person}{William Brannon}, \bibinfo{person}{Niklas Muennighoff}, \bibinfo{person}{Nathan Khazam}, \bibinfo{person}{Jad Kabbara}, \bibinfo{person}{Kartik Perisetla}, {et~al\mbox{.}}} \bibinfo{year}{2023}\natexlab{b}.
\newblock \showarticletitle{The data provenance initiative: A large scale audit of dataset licensing \& attribution in ai}.
\newblock \bibinfo{journal}{\emph{arXiv preprint arXiv:2310.16787}} (\bibinfo{year}{2023}).
\newblock


\bibitem[Lozhkov et~al\mbox{.}(2024)]%
        {lozhkov2024starcoder}
\bibfield{author}{\bibinfo{person}{Anton Lozhkov}, \bibinfo{person}{Raymond Li}, \bibinfo{person}{Loubna~Ben Allal}, \bibinfo{person}{Federico Cassano}, \bibinfo{person}{Joel Lamy-Poirier}, \bibinfo{person}{Nouamane Tazi}, \bibinfo{person}{Ao Tang}, \bibinfo{person}{Dmytro Pykhtar}, \bibinfo{person}{Jiawei Liu}, \bibinfo{person}{Yuxiang Wei}, {et~al\mbox{.}}} \bibinfo{year}{2024}\natexlab{}.
\newblock \showarticletitle{Starcoder 2 and the stack v2: The next generation}.
\newblock \bibinfo{journal}{\emph{arXiv preprint arXiv:2402.19173}} (\bibinfo{year}{2024}).
\newblock


\bibitem[Lu et~al\mbox{.}(2020)]%
        {lu2020twinbert}
\bibfield{author}{\bibinfo{person}{Wenhao Lu}, \bibinfo{person}{Jian Jiao}, {and} \bibinfo{person}{Ruofei Zhang}.} \bibinfo{year}{2020}\natexlab{}.
\newblock \showarticletitle{Twinbert: Distilling knowledge to twin-structured compressed bert models for large-scale retrieval}. In \bibinfo{booktitle}{\emph{Proceedings of the 29th ACM International Conference on Information \& Knowledge Management}}. \bibinfo{pages}{2645--2652}.
\newblock


\bibitem[Lu et~al\mbox{.}(2024)]%
        {lu2024small}
\bibfield{author}{\bibinfo{person}{Zhenyan Lu}, \bibinfo{person}{Xiang Li}, \bibinfo{person}{Dongqi Cai}, \bibinfo{person}{Rongjie Yi}, \bibinfo{person}{Fangming Liu}, \bibinfo{person}{Xiwen Zhang}, \bibinfo{person}{Nicholas~D Lane}, {and} \bibinfo{person}{Mengwei Xu}.} \bibinfo{year}{2024}\natexlab{}.
\newblock \showarticletitle{Small language models: Survey, measurements, and insights}.
\newblock \bibinfo{journal}{\emph{arXiv preprint arXiv:2409.15790}} (\bibinfo{year}{2024}).
\newblock


\bibitem[Luo et~al\mbox{.}(2022)]%
        {luo2022biogpt}
\bibfield{author}{\bibinfo{person}{Renqian Luo}, \bibinfo{person}{Liai Sun}, \bibinfo{person}{Yingce Xia}, \bibinfo{person}{Tao Qin}, \bibinfo{person}{Sheng Zhang}, \bibinfo{person}{Hoifung Poon}, {and} \bibinfo{person}{Tie-Yan Liu}.} \bibinfo{year}{2022}\natexlab{}.
\newblock \showarticletitle{BioGPT: generative pre-trained transformer for biomedical text generation and mining}.
\newblock \bibinfo{journal}{\emph{Briefings in bioinformatics}} \bibinfo{volume}{23}, \bibinfo{number}{6} (\bibinfo{year}{2022}), \bibinfo{pages}{bbac409}.
\newblock


\bibitem[Ma et~al\mbox{.}(2024)]%
        {ma2024era}
\bibfield{author}{\bibinfo{person}{Shuming Ma}, \bibinfo{person}{Hongyu Wang}, \bibinfo{person}{Lingxiao Ma}, \bibinfo{person}{Lei Wang}, \bibinfo{person}{Wenhui Wang}, \bibinfo{person}{Shaohan Huang}, \bibinfo{person}{Li Dong}, \bibinfo{person}{Ruiping Wang}, \bibinfo{person}{Jilong Xue}, {and} \bibinfo{person}{Furu Wei}.} \bibinfo{year}{2024}\natexlab{}.
\newblock \showarticletitle{The era of 1-bit llms: All large language models are in 1.58 bits}.
\newblock \bibinfo{journal}{\emph{arXiv preprint arXiv:2402.17764}} (\bibinfo{year}{2024}).
\newblock


\bibitem[Ma et~al\mbox{.}(2023c)]%
        {ma2023llm}
\bibfield{author}{\bibinfo{person}{Xinyin Ma}, \bibinfo{person}{Gongfan Fang}, {and} \bibinfo{person}{Xinchao Wang}.} \bibinfo{year}{2023}\natexlab{c}.
\newblock \showarticletitle{Llm-pruner: On the structural pruning of large language models}.
\newblock \bibinfo{journal}{\emph{Advances in neural information processing systems}}  \bibinfo{volume}{36} (\bibinfo{year}{2023}), \bibinfo{pages}{21702--21720}.
\newblock


\bibitem[Ma et~al\mbox{.}(2023d)]%
        {ma2023query}
\bibfield{author}{\bibinfo{person}{Xinbei Ma}, \bibinfo{person}{Yeyun Gong}, \bibinfo{person}{Pengcheng He}, \bibinfo{person}{Hai Zhao}, {and} \bibinfo{person}{Nan Duan}.} \bibinfo{year}{2023}\natexlab{d}.
\newblock \showarticletitle{Query Rewriting in Retrieval-Augmented Large Language Models}. In \bibinfo{booktitle}{\emph{Proceedings of the 2023 Conference on Empirical Methods in Natural Language Processing}}. \bibinfo{pages}{5303--5315}.
\newblock


\bibitem[Ma et~al\mbox{.}(2023a)]%
        {ma2023large}
\bibfield{author}{\bibinfo{person}{Yubo Ma}, \bibinfo{person}{Yixin Cao}, \bibinfo{person}{YongChing Hong}, {and} \bibinfo{person}{Aixin Sun}.} \bibinfo{year}{2023}\natexlab{a}.
\newblock \showarticletitle{Large language model is not a good few-shot information extractor, but a good reranker for hard samples!}
\newblock \bibinfo{journal}{\emph{arXiv preprint arXiv:2303.08559}} (\bibinfo{year}{2023}).
\newblock


\bibitem[Ma et~al\mbox{.}(2023b)]%
        {ma2023sci}
\bibfield{author}{\bibinfo{person}{Yuhan Ma}, \bibinfo{person}{Chenyou Fan}, {and} \bibinfo{person}{Haiqi Jiang}.} \bibinfo{year}{2023}\natexlab{b}.
\newblock \showarticletitle{Sci-cot: Leveraging large language models for enhanced knowledge distillation in small models for scientific qa}. In \bibinfo{booktitle}{\emph{2023 9th International Conference on Computer and Communications (ICCC)}}. IEEE, \bibinfo{pages}{2394--2398}.
\newblock


\bibitem[MA et~al\mbox{.}(2024)]%
        {ma2024at}
\bibfield{author}{\bibinfo{person}{YINGWEI MA}, \bibinfo{person}{Yue Liu}, \bibinfo{person}{Yue Yu}, \bibinfo{person}{Yuanliang Zhang}, \bibinfo{person}{Yu Jiang}, \bibinfo{person}{Changjian Wang}, {and} \bibinfo{person}{Shanshan Li}.} \bibinfo{year}{2024}\natexlab{}.
\newblock \showarticletitle{At Which Training Stage Does Code Data Help {LLM}s Reasoning?}. In \bibinfo{booktitle}{\emph{The Twelfth International Conference on Learning Representations}}.
\newblock
\urldef\tempurl%
\url{https://openreview.net/forum?id=KIPJKST4gw}
\showURL{%
\tempurl}


\bibitem[Magnusson et~al\mbox{.}(2023)]%
        {magnusson2023paloma}
\bibfield{author}{\bibinfo{person}{Ian Magnusson}, \bibinfo{person}{Akshita Bhagia}, \bibinfo{person}{Valentin Hofmann}, \bibinfo{person}{Luca Soldaini}, \bibinfo{person}{Ananya~Harsh Jha}, \bibinfo{person}{Oyvind Tafjord}, \bibinfo{person}{Dustin Schwenk}, \bibinfo{person}{Evan~Pete Walsh}, \bibinfo{person}{Yanai Elazar}, \bibinfo{person}{Kyle Lo}, {et~al\mbox{.}}} \bibinfo{year}{2023}\natexlab{}.
\newblock \showarticletitle{Paloma: A benchmark for evaluating language model fit}.
\newblock \bibinfo{journal}{\emph{arXiv preprint arXiv:2312.10523}} (\bibinfo{year}{2023}).
\newblock


\bibitem[Mahan et~al\mbox{.}({[n.\,d.]})]%
        {StableBelugaModels}
\bibfield{author}{\bibinfo{person}{Dakota Mahan}, \bibinfo{person}{Ryan Carlow}, \bibinfo{person}{Louis Castricato}, \bibinfo{person}{Nathan Cooper}, {and} \bibinfo{person}{Christian Laforte}.} \bibinfo{year}{[n.\,d.]}\natexlab{}.
\newblock \bibinfo{title}{Stable Beluga models}.
\newblock
\newblock
\urldef\tempurl%
\url{[https://huggingface.co/stabilityai/StableBeluga2](https://huggingface.co/stabilityai/StableBeluga2)}
\showURL{%
\tempurl}


\bibitem[Malinovskii et~al\mbox{.}(2024)]%
        {malinovskii2024pv}
\bibfield{author}{\bibinfo{person}{Vladimir Malinovskii}, \bibinfo{person}{Denis Mazur}, \bibinfo{person}{Ivan Ilin}, \bibinfo{person}{Denis Kuznedelev}, \bibinfo{person}{Konstantin Burlachenko}, \bibinfo{person}{Kai Yi}, \bibinfo{person}{Dan Alistarh}, {and} \bibinfo{person}{Peter Richtarik}.} \bibinfo{year}{2024}\natexlab{}.
\newblock \showarticletitle{PV-Tuning: Beyond Straight-Through Estimation for Extreme LLM Compression}.
\newblock \bibinfo{journal}{\emph{arXiv preprint arXiv:2405.14852}} (\bibinfo{year}{2024}).
\newblock


\bibitem[Manakul et~al\mbox{.}(2023)]%
        {manakul-etal-2023-selfcheckgpt}
\bibfield{author}{\bibinfo{person}{Potsawee Manakul}, \bibinfo{person}{Adian Liusie}, {and} \bibinfo{person}{Mark Gales}.} \bibinfo{year}{2023}\natexlab{}.
\newblock \showarticletitle{{S}elf{C}heck{GPT}: Zero-Resource Black-Box Hallucination Detection for Generative Large Language Models}. In \bibinfo{booktitle}{\emph{Proceedings of the 2023 Conference on Empirical Methods in Natural Language Processing}}, \bibfield{editor}{\bibinfo{person}{Houda Bouamor}, \bibinfo{person}{Juan Pino}, {and} \bibinfo{person}{Kalika Bali}} (Eds.). \bibinfo{publisher}{Association for Computational Linguistics}, \bibinfo{address}{Singapore}, \bibinfo{pages}{9004--9017}.
\newblock
\urldef\tempurl%
\url{https://doi.org/10.18653/v1/2023.emnlp-main.557}
\showDOI{\tempurl}


\bibitem[Mehta et~al\mbox{.}(2024)]%
        {mehta2024openelm}
\bibfield{author}{\bibinfo{person}{Sachin Mehta}, \bibinfo{person}{Mohammad~Hossein Sekhavat}, \bibinfo{person}{Qingqing Cao}, \bibinfo{person}{Maxwell Horton}, \bibinfo{person}{Yanzi Jin}, \bibinfo{person}{Chenfan Sun}, \bibinfo{person}{Seyed~Iman Mirzadeh}, \bibinfo{person}{Mahyar Najibi}, \bibinfo{person}{Dmitry Belenko}, \bibinfo{person}{Peter Zatloukal}, {et~al\mbox{.}}} \bibinfo{year}{2024}\natexlab{}.
\newblock \showarticletitle{Openelm: An efficient language model family with open training and inference framework}. In \bibinfo{booktitle}{\emph{Workshop on Efficient Systems for Foundation Models II@ ICML2024}}.
\newblock


\bibitem[Mekala et~al\mbox{.}(2024)]%
        {mekala2024smaller}
\bibfield{author}{\bibinfo{person}{Dheeraj Mekala}, \bibinfo{person}{Alex Nguyen}, {and} \bibinfo{person}{Jingbo Shang}.} \bibinfo{year}{2024}\natexlab{}.
\newblock \showarticletitle{Smaller language models are capable of selecting instruction-tuning training data for larger language models}.
\newblock \bibinfo{journal}{\emph{arXiv preprint arXiv:2402.10430}} (\bibinfo{year}{2024}).
\newblock


\bibitem[Men et~al\mbox{.}(2024)]%
        {men2024shortgpt}
\bibfield{author}{\bibinfo{person}{Xin Men}, \bibinfo{person}{Mingyu Xu}, \bibinfo{person}{Qingyu Zhang}, \bibinfo{person}{Bingning Wang}, \bibinfo{person}{Hongyu Lin}, \bibinfo{person}{Yaojie Lu}, \bibinfo{person}{Xianpei Han}, {and} \bibinfo{person}{Weipeng Chen}.} \bibinfo{year}{2024}\natexlab{}.
\newblock \showarticletitle{Shortgpt: Layers in large language models are more redundant than you expect}.
\newblock \bibinfo{journal}{\emph{arXiv preprint arXiv:2403.03853}} (\bibinfo{year}{2024}).
\newblock


\bibitem[Min et~al\mbox{.}(2023)]%
        {min-etal-2023-factscore}
\bibfield{author}{\bibinfo{person}{Sewon Min}, \bibinfo{person}{Kalpesh Krishna}, \bibinfo{person}{Xinxi Lyu}, \bibinfo{person}{Mike Lewis}, \bibinfo{person}{Wen-tau Yih}, \bibinfo{person}{Pang Koh}, \bibinfo{person}{Mohit Iyyer}, \bibinfo{person}{Luke Zettlemoyer}, {and} \bibinfo{person}{Hannaneh Hajishirzi}.} \bibinfo{year}{2023}\natexlab{}.
\newblock \showarticletitle{{FA}ct{S}core: Fine-grained Atomic Evaluation of Factual Precision in Long Form Text Generation}. In \bibinfo{booktitle}{\emph{Proceedings of the 2023 Conference on Empirical Methods in Natural Language Processing}}, \bibfield{editor}{\bibinfo{person}{Houda Bouamor}, \bibinfo{person}{Juan Pino}, {and} \bibinfo{person}{Kalika Bali}} (Eds.). \bibinfo{publisher}{Association for Computational Linguistics}, \bibinfo{address}{Singapore}, \bibinfo{pages}{12076--12100}.
\newblock
\urldef\tempurl%
\url{https://doi.org/10.18653/v1/2023.emnlp-main.741}
\showDOI{\tempurl}


\bibitem[Min-su(2024)]%
        {gominsu2024bible}
\bibfield{author}{\bibinfo{person}{Go Min-su}.} \bibinfo{year}{2024}\natexlab{}.
\newblock \bibinfo{title}{Deep Learning Bible - 8. Large Language Models}.
\newblock \bibinfo{howpublished}{WikiDocs}.
\newblock
\urldef\tempurl%
\url{https://wikidocs.net/237419}
\showURL{%
\tempurl}


\bibitem[Mitchell et~al\mbox{.}(2024)]%
        {mitchell2024an}
\bibfield{author}{\bibinfo{person}{Eric Mitchell}, \bibinfo{person}{Rafael Rafailov}, \bibinfo{person}{Archit Sharma}, \bibinfo{person}{Chelsea Finn}, {and} \bibinfo{person}{Christopher~D Manning}.} \bibinfo{year}{2024}\natexlab{}.
\newblock \showarticletitle{An Emulator for Fine-tuning Large Language Models using Small Language Models}. In \bibinfo{booktitle}{\emph{The Twelfth International Conference on Learning Representations}}.
\newblock
\urldef\tempurl%
\url{https://openreview.net/forum?id=Eo7kv0sllr}
\showURL{%
\tempurl}


\bibitem[Mitra et~al\mbox{.}(2023)]%
        {mitra2023orca}
\bibfield{author}{\bibinfo{person}{Arindam Mitra}, \bibinfo{person}{Luciano Del~Corro}, \bibinfo{person}{Shweti Mahajan}, \bibinfo{person}{Andres Codas}, \bibinfo{person}{Clarisse Simoes}, \bibinfo{person}{Sahaj Agarwal}, \bibinfo{person}{Xuxi Chen}, \bibinfo{person}{Anastasia Razdaibiedina}, \bibinfo{person}{Erik Jones}, \bibinfo{person}{Kriti Aggarwal}, {et~al\mbox{.}}} \bibinfo{year}{2023}\natexlab{}.
\newblock \showarticletitle{Orca 2: Teaching small language models how to reason}.
\newblock \bibinfo{journal}{\emph{arXiv preprint arXiv:2311.11045}} (\bibinfo{year}{2023}).
\newblock


\bibitem[Mo et~al\mbox{.}(2024)]%
        {mo2023trustworthy}
\bibfield{author}{\bibinfo{person}{Lingbo Mo}, \bibinfo{person}{Boshi Wang}, \bibinfo{person}{Muhao Chen}, {and} \bibinfo{person}{Huan Sun}.} \bibinfo{year}{2024}\natexlab{}.
\newblock \showarticletitle{How Trustworthy are Open-Source {LLM}s? An Assessment under Malicious Demonstrations Shows their Vulnerabilities}. In \bibinfo{booktitle}{\emph{Proceedings of the 2024 Conference of the North American Chapter of the Association for Computational Linguistics: Human Language Technologies (Volume 1: Long Papers)}}. \bibinfo{pages}{2775--2792}.
\newblock


\bibitem[Morris et~al\mbox{.}(2024)]%
        {morris2024language}
\bibfield{author}{\bibinfo{person}{John~Xavier Morris}, \bibinfo{person}{Wenting Zhao}, \bibinfo{person}{Justin~T Chiu}, \bibinfo{person}{Vitaly Shmatikov}, {and} \bibinfo{person}{Alexander~M Rush}.} \bibinfo{year}{2024}\natexlab{}.
\newblock \showarticletitle{Language Model Inversion}. In \bibinfo{booktitle}{\emph{The Twelfth International Conference on Learning Representations}}.
\newblock
\urldef\tempurl%
\url{https://openreview.net/forum?id=t9dWHpGkPj}
\showURL{%
\tempurl}


\bibitem[Mozes et~al\mbox{.}(2023)]%
        {mozes2023use}
\bibfield{author}{\bibinfo{person}{Maximilian Mozes}, \bibinfo{person}{Xuanli He}, \bibinfo{person}{Bennett Kleinberg}, {and} \bibinfo{person}{Lewis~D Griffin}.} \bibinfo{year}{2023}\natexlab{}.
\newblock \showarticletitle{Use of llms for illicit purposes: Threats, prevention measures, and vulnerabilities}.
\newblock \bibinfo{journal}{\emph{arXiv preprint arXiv:2308.12833}} (\bibinfo{year}{2023}).
\newblock


\bibitem[Mukherjee et~al\mbox{.}(2023)]%
        {mukherjee2023orca}
\bibfield{author}{\bibinfo{person}{Subhabrata Mukherjee}, \bibinfo{person}{Arindam Mitra}, \bibinfo{person}{Ganesh Jawahar}, \bibinfo{person}{Sahaj Agarwal}, \bibinfo{person}{Hamid Palangi}, {and} \bibinfo{person}{Ahmed Awadallah}.} \bibinfo{year}{2023}\natexlab{}.
\newblock \showarticletitle{Orca: Progressive learning from complex explanation traces of gpt-4}.
\newblock \bibinfo{journal}{\emph{arXiv preprint arXiv:2306.02707}} (\bibinfo{year}{2023}).
\newblock


\bibitem[Muralidharan et~al\mbox{.}(2024)]%
        {muralidharan2024compact}
\bibfield{author}{\bibinfo{person}{Saurav Muralidharan}, \bibinfo{person}{Sharath~Turuvekere Sreenivas}, \bibinfo{person}{Raviraj Joshi}, \bibinfo{person}{Marcin Chochowski}, \bibinfo{person}{Mostofa Patwary}, \bibinfo{person}{Mohammad Shoeybi}, \bibinfo{person}{Bryan Catanzaro}, \bibinfo{person}{Jan Kautz}, {and} \bibinfo{person}{Pavlo Molchanov}.} \bibinfo{year}{2024}\natexlab{}.
\newblock \showarticletitle{Compact language models via pruning and knowledge distillation}.
\newblock \bibinfo{journal}{\emph{arXiv preprint arXiv:2407.14679}} (\bibinfo{year}{2024}).
\newblock


\bibitem[Murthy et~al\mbox{.}(2024)]%
        {murthy2024mobileaibenchbenchmarkingllmslmms}
\bibfield{author}{\bibinfo{person}{Rithesh Murthy}, \bibinfo{person}{Liangwei Yang}, \bibinfo{person}{Juntao Tan}, \bibinfo{person}{Tulika~Manoj Awalgaonkar}, \bibinfo{person}{Yilun Zhou}, \bibinfo{person}{Shelby Heinecke}, \bibinfo{person}{Sachin Desai}, \bibinfo{person}{Jason Wu}, \bibinfo{person}{Ran Xu}, \bibinfo{person}{Sarah Tan}, \bibinfo{person}{Jianguo Zhang}, \bibinfo{person}{Zhiwei Liu}, \bibinfo{person}{Shirley Kokane}, \bibinfo{person}{Zuxin Liu}, \bibinfo{person}{Ming Zhu}, \bibinfo{person}{Huan Wang}, \bibinfo{person}{Caiming Xiong}, {and} \bibinfo{person}{Silvio Savarese}.} \bibinfo{year}{2024}\natexlab{}.
\newblock \bibinfo{title}{MobileAIBench: Benchmarking LLMs and LMMs for On-Device Use Cases}.
\newblock
\newblock
\showeprint[arxiv]{2406.10290}~[cs.CL]
\urldef\tempurl%
\url{https://arxiv.org/abs/2406.10290}
\showURL{%
\tempurl}


\bibitem[Nakka et~al\mbox{.}(2024)]%
        {nakka2024device}
\bibfield{author}{\bibinfo{person}{Kalyan Nakka}, \bibinfo{person}{Jimmy Dani}, {and} \bibinfo{person}{Nitesh Saxena}.} \bibinfo{year}{2024}\natexlab{}.
\newblock \showarticletitle{Is On-Device AI Broken and Exploitable? Assessing the Trust and Ethics in Small Language Models}.
\newblock \bibinfo{journal}{\emph{arXiv preprint arXiv:2406.05364}} (\bibinfo{year}{2024}).
\newblock


\bibitem[Nam et~al\mbox{.}(2024)]%
        {nam2024using}
\bibfield{author}{\bibinfo{person}{Daye Nam}, \bibinfo{person}{Andrew Macvean}, \bibinfo{person}{Vincent Hellendoorn}, \bibinfo{person}{Bogdan Vasilescu}, {and} \bibinfo{person}{Brad Myers}.} \bibinfo{year}{2024}\natexlab{}.
\newblock \showarticletitle{Using an llm to help with code understanding}. In \bibinfo{booktitle}{\emph{Proceedings of the IEEE/ACM 46th International Conference on Software Engineering}}. \bibinfo{pages}{1--13}.
\newblock


\bibitem[Nawrot et~al\mbox{.}(2024)]%
        {nawrot2024dynamic}
\bibfield{author}{\bibinfo{person}{Piotr Nawrot}, \bibinfo{person}{Adrian {\L}a{\'n}cucki}, \bibinfo{person}{Marcin Chochowski}, \bibinfo{person}{David Tarjan}, {and} \bibinfo{person}{Edoardo Ponti}.} \bibinfo{year}{2024}\natexlab{}.
\newblock \showarticletitle{Dynamic Memory Compression: Retrofitting {LLM}s for Accelerated Inference}. In \bibinfo{booktitle}{\emph{Forty-first International Conference on Machine Learning}}.
\newblock
\urldef\tempurl%
\url{https://openreview.net/forum?id=tDRYrAkOB7}
\showURL{%
\tempurl}


\bibitem[Nguyen et~al\mbox{.}(2024)]%
        {nguyen2024culturax}
\bibfield{author}{\bibinfo{person}{Thuat Nguyen}, \bibinfo{person}{Chien Van~Nguyen}, \bibinfo{person}{Viet~Dac Lai}, \bibinfo{person}{Hieu Man}, \bibinfo{person}{Nghia~Trung Ngo}, \bibinfo{person}{Franck Dernoncourt}, \bibinfo{person}{Ryan~A Rossi}, {and} \bibinfo{person}{Thien~Huu Nguyen}.} \bibinfo{year}{2024}\natexlab{}.
\newblock \showarticletitle{CulturaX: A Cleaned, Enormous, and Multilingual Dataset for Large Language Models in 167 Languages}. In \bibinfo{booktitle}{\emph{Proceedings of the 2024 Joint International Conference on Computational Linguistics, Language Resources and Evaluation (LREC-COLING 2024)}}. \bibinfo{pages}{4226--4237}.
\newblock


\bibitem[Nguyen et~al\mbox{.}(2023)]%
        {nguyen2023astrollama}
\bibfield{author}{\bibinfo{person}{Tuan~Dung Nguyen}, \bibinfo{person}{Yuan-Sen Ting}, \bibinfo{person}{Ioana Ciuca}, \bibinfo{person}{Charles O’Neill}, \bibinfo{person}{Ze-Chang Sun}, \bibinfo{person}{Maja Jab{\l}o{\'n}ska}, \bibinfo{person}{Sandor Kruk}, \bibinfo{person}{Ernest Perkowski}, \bibinfo{person}{Jack Miller}, \bibinfo{person}{Jason Jason~Jingsh Li}, {et~al\mbox{.}}} \bibinfo{year}{2023}\natexlab{}.
\newblock \showarticletitle{AstroLLaMA: Towards Specialized Foundation Models in Astronomy}. In \bibinfo{booktitle}{\emph{Proceedings of the Second Workshop on Information Extraction from Scientific Publications}}. \bibinfo{pages}{49--55}.
\newblock


\bibitem[Ni et~al\mbox{.}(2022)]%
        {ni-etal-2022-large}
\bibfield{author}{\bibinfo{person}{Jianmo Ni}, \bibinfo{person}{Chen Qu}, \bibinfo{person}{Jing Lu}, \bibinfo{person}{Zhuyun Dai}, \bibinfo{person}{Gustavo Hernandez~Abrego}, \bibinfo{person}{Ji Ma}, \bibinfo{person}{Vincent Zhao}, \bibinfo{person}{Yi Luan}, \bibinfo{person}{Keith Hall}, \bibinfo{person}{Ming-Wei Chang}, {and} \bibinfo{person}{Yinfei Yang}.} \bibinfo{year}{2022}\natexlab{}.
\newblock \showarticletitle{Large Dual Encoders Are Generalizable Retrievers}. In \bibinfo{booktitle}{\emph{Proceedings of the 2022 Conference on Empirical Methods in Natural Language Processing}}. \bibinfo{publisher}{Association for Computational Linguistics}, \bibinfo{pages}{9844--9855}.
\newblock
\urldef\tempurl%
\url{https://doi.org/10.18653/v1/2022.emnlp-main.669}
\showDOI{\tempurl}


\bibitem[Nogueira and Cho(2019)]%
        {nogueira2019passage}
\bibfield{author}{\bibinfo{person}{Rodrigo Nogueira} {and} \bibinfo{person}{Kyunghyun Cho}.} \bibinfo{year}{2019}\natexlab{}.
\newblock \showarticletitle{Passage Re-ranking with BERT}.
\newblock \bibinfo{journal}{\emph{arXiv preprint arXiv:1901.04085}} (\bibinfo{year}{2019}).
\newblock


\bibitem[Noorian(2024)]%
        {noorian2024bert}
\bibfield{author}{\bibinfo{person}{A Noorian}.} \bibinfo{year}{2024}\natexlab{}.
\newblock \showarticletitle{A BERT-based sequential POI recommender system in social media}.
\newblock \bibinfo{journal}{\emph{Computer Standards \& Interfaces}}  \bibinfo{volume}{87} (\bibinfo{year}{2024}), \bibinfo{pages}{103766}.
\newblock


\bibitem[OpenAI(2024a)]%
        {GPT-4o-mini}
\bibfield{author}{\bibinfo{person}{OpenAI}.} \bibinfo{year}{2024}\natexlab{a}.
\newblock \bibinfo{booktitle}{\emph{GPT-4o mini: advancing cost-efficient intelligence}}.
\newblock
\urldef\tempurl%
\url{https://openai.com/index/gpt-4o-mini-advancing-cost-efficient-intelligence/}
\showURL{%
\tempurl}
\newblock
\shownote{Accessed: 2024-7-18}.


\bibitem[OpenAI(2024b)]%
        {gpt4o}
\bibfield{author}{\bibinfo{person}{OpenAI}.} \bibinfo{year}{2024}\natexlab{b}.
\newblock \bibinfo{booktitle}{\emph{Hello GPT-4o}}.
\newblock
\urldef\tempurl%
\url{https://openai.com/index/hello-gpt-4o/}
\showURL{%
\tempurl}
\newblock
\shownote{Accessed: 2024-5-13}.


\bibitem[Ouyang et~al\mbox{.}(2022)]%
        {ouyang2022training}
\bibfield{author}{\bibinfo{person}{Long Ouyang}, \bibinfo{person}{Jeffrey Wu}, \bibinfo{person}{Xu Jiang}, \bibinfo{person}{Diogo Almeida}, \bibinfo{person}{Carroll Wainwright}, \bibinfo{person}{Pamela Mishkin}, \bibinfo{person}{Chong Zhang}, \bibinfo{person}{Sandhini Agarwal}, \bibinfo{person}{Katarina Slama}, \bibinfo{person}{Alex Ray}, {et~al\mbox{.}}} \bibinfo{year}{2022}\natexlab{}.
\newblock \showarticletitle{Training language models to follow instructions with human feedback}.
\newblock \bibinfo{journal}{\emph{Advances in neural information processing systems}}  \bibinfo{volume}{35} (\bibinfo{year}{2022}), \bibinfo{pages}{27730--27744}.
\newblock


\bibitem[Padmanabhan et~al\mbox{.}(2024)]%
        {padmanabhan2024propagating}
\bibfield{author}{\bibinfo{person}{Shankar Padmanabhan}, \bibinfo{person}{Yasumasa Onoe}, \bibinfo{person}{Michael Zhang}, \bibinfo{person}{Greg Durrett}, {and} \bibinfo{person}{Eunsol Choi}.} \bibinfo{year}{2024}\natexlab{}.
\newblock \showarticletitle{Propagating knowledge updates to lms through distillation}.
\newblock \bibinfo{journal}{\emph{Advances in Neural Information Processing Systems}}  \bibinfo{volume}{36} (\bibinfo{year}{2024}).
\newblock


\bibitem[Parmar et~al\mbox{.}(2024)]%
        {parmar2024nemotron}
\bibfield{author}{\bibinfo{person}{Jupinder Parmar}, \bibinfo{person}{Shrimai Prabhumoye}, \bibinfo{person}{Joseph Jennings}, \bibinfo{person}{Mostofa Patwary}, \bibinfo{person}{Sandeep Subramanian}, \bibinfo{person}{Dan Su}, \bibinfo{person}{Chen Zhu}, \bibinfo{person}{Deepak Narayanan}, \bibinfo{person}{Aastha Jhunjhunwala}, \bibinfo{person}{Ayush Dattagupta}, {et~al\mbox{.}}} \bibinfo{year}{2024}\natexlab{}.
\newblock \showarticletitle{Nemotron-4 15B Technical Report}.
\newblock \bibinfo{journal}{\emph{arXiv preprint arXiv:2402.16819}} (\bibinfo{year}{2024}).
\newblock


\bibitem[Paster et~al\mbox{.}(2024)]%
        {paster2024openwebmath}
\bibfield{author}{\bibinfo{person}{Keiran Paster}, \bibinfo{person}{Marco~Dos Santos}, \bibinfo{person}{Zhangir Azerbayev}, {and} \bibinfo{person}{Jimmy Ba}.} \bibinfo{year}{2024}\natexlab{}.
\newblock \showarticletitle{OpenWebMath: An Open Dataset of High-Quality Mathematical Web Text}. In \bibinfo{booktitle}{\emph{The Twelfth International Conference on Learning Representations}}.
\newblock
\urldef\tempurl%
\url{https://openreview.net/forum?id=jKHmjlpViu}
\showURL{%
\tempurl}


\bibitem[Penedo et~al\mbox{.}(2024)]%
        {penedo2024finewebdatasetsdecantingweb}
\bibfield{author}{\bibinfo{person}{Guilherme Penedo}, \bibinfo{person}{Hynek Kydlíček}, \bibinfo{person}{Loubna~Ben allal}, \bibinfo{person}{Anton Lozhkov}, \bibinfo{person}{Margaret Mitchell}, \bibinfo{person}{Colin Raffel}, \bibinfo{person}{Leandro~Von Werra}, {and} \bibinfo{person}{Thomas Wolf}.} \bibinfo{year}{2024}\natexlab{}.
\newblock \bibinfo{title}{The FineWeb Datasets: Decanting the Web for the Finest Text Data at Scale}.
\newblock
\newblock
\showeprint[arxiv]{2406.17557}~[cs.CL]


\bibitem[Penedo et~al\mbox{.}(2023)]%
        {penedo2023refinedweb}
\bibfield{author}{\bibinfo{person}{Guilherme Penedo}, \bibinfo{person}{Quentin Malartic}, \bibinfo{person}{Daniel Hesslow}, \bibinfo{person}{Ruxandra Cojocaru}, \bibinfo{person}{Alessandro Cappelli}, \bibinfo{person}{Hamza Alobeidli}, \bibinfo{person}{Baptiste Pannier}, \bibinfo{person}{Ebtesam Almazrouei}, {and} \bibinfo{person}{Julien Launay}.} \bibinfo{year}{2023}\natexlab{}.
\newblock \showarticletitle{The RefinedWeb dataset for Falcon LLM: outperforming curated corpora with web data, and web data only}.
\newblock \bibinfo{journal}{\emph{arXiv preprint arXiv:2306.01116}} (\bibinfo{year}{2023}).
\newblock


\bibitem[Peng et~al\mbox{.}(2023a)]%
        {peng-etal-2023-rwkv}
\bibfield{author}{\bibinfo{person}{Bo Peng}, \bibinfo{person}{Eric Alcaide}, \bibinfo{person}{Quentin Anthony}, \bibinfo{person}{Alon Albalak}, \bibinfo{person}{Samuel Arcadinho}, \bibinfo{person}{Stella Biderman}, \bibinfo{person}{Huanqi Cao}, \bibinfo{person}{Xin Cheng}, \bibinfo{person}{Michael Chung}, \bibinfo{person}{Leon Derczynski}, \bibinfo{person}{Xingjian Du}, \bibinfo{person}{Matteo Grella}, \bibinfo{person}{Kranthi Gv}, \bibinfo{person}{Xuzheng He}, \bibinfo{person}{Haowen Hou}, \bibinfo{person}{Przemyslaw Kazienko}, \bibinfo{person}{Jan Kocon}, \bibinfo{person}{Jiaming Kong}, \bibinfo{person}{Bart{\l}omiej Koptyra}, \bibinfo{person}{Hayden Lau}, \bibinfo{person}{Jiaju Lin}, \bibinfo{person}{Krishna Sri~Ipsit Mantri}, \bibinfo{person}{Ferdinand Mom}, \bibinfo{person}{Atsushi Saito}, \bibinfo{person}{Guangyu Song}, \bibinfo{person}{Xiangru Tang}, \bibinfo{person}{Johan Wind}, \bibinfo{person}{Stanis{\l}aw Wo{\'z}niak}, \bibinfo{person}{Zhenyuan Zhang}, \bibinfo{person}{Qinghua Zhou},
  \bibinfo{person}{Jian Zhu}, {and} \bibinfo{person}{Rui-Jie Zhu}.} \bibinfo{year}{2023}\natexlab{a}.
\newblock \showarticletitle{{RWKV}: Reinventing {RNN}s for the Transformer Era}. In \bibinfo{booktitle}{\emph{Findings of the Association for Computational Linguistics: EMNLP 2023}}, \bibfield{editor}{\bibinfo{person}{Houda Bouamor}, \bibinfo{person}{Juan Pino}, {and} \bibinfo{person}{Kalika Bali}} (Eds.). \bibinfo{publisher}{Association for Computational Linguistics}, \bibinfo{address}{Singapore}, \bibinfo{pages}{14048--14077}.
\newblock
\urldef\tempurl%
\url{https://doi.org/10.18653/v1/2023.findings-emnlp.936}
\showDOI{\tempurl}


\bibitem[Peng et~al\mbox{.}(2023b)]%
        {peng2023instruction}
\bibfield{author}{\bibinfo{person}{Baolin Peng}, \bibinfo{person}{Chunyuan Li}, \bibinfo{person}{Pengcheng He}, \bibinfo{person}{Michel Galley}, {and} \bibinfo{person}{Jianfeng Gao}.} \bibinfo{year}{2023}\natexlab{b}.
\newblock \showarticletitle{Instruction tuning with gpt-4}.
\newblock \bibinfo{journal}{\emph{arXiv preprint arXiv:2304.03277}} (\bibinfo{year}{2023}).
\newblock


\bibitem[Peng et~al\mbox{.}(2023c)]%
        {peng2023soft}
\bibfield{author}{\bibinfo{person}{Zhiyuan Peng}, \bibinfo{person}{Xuyang Wu}, \bibinfo{person}{Qifan Wang}, {and} \bibinfo{person}{Yi Fang}.} \bibinfo{year}{2023}\natexlab{c}.
\newblock \showarticletitle{Soft prompt tuning for augmenting dense retrieval with large language models}.
\newblock \bibinfo{journal}{\emph{arXiv preprint arXiv:2307.08303}} (\bibinfo{year}{2023}).
\newblock


\bibitem[Perez et~al\mbox{.}(2023)]%
        {perez-etal-2023-discovering}
\bibfield{author}{\bibinfo{person}{Ethan Perez}, \bibinfo{person}{Sam Ringer}, \bibinfo{person}{Kamile Lukosiute}, \bibinfo{person}{Karina Nguyen}, \bibinfo{person}{Edwin Chen}, \bibinfo{person}{Scott Heiner}, \bibinfo{person}{Craig Pettit}, \bibinfo{person}{Catherine Olsson}, \bibinfo{person}{Sandipan Kundu}, \bibinfo{person}{Saurav Kadavath}, \bibinfo{person}{Andy Jones}, \bibinfo{person}{Anna Chen}, \bibinfo{person}{Benjamin Mann}, \bibinfo{person}{Brian Israel}, \bibinfo{person}{Bryan Seethor}, \bibinfo{person}{Cameron McKinnon}, \bibinfo{person}{Christopher Olah}, \bibinfo{person}{Da Yan}, \bibinfo{person}{Daniela Amodei}, \bibinfo{person}{Dario Amodei}, \bibinfo{person}{Dawn Drain}, \bibinfo{person}{Dustin Li}, \bibinfo{person}{Eli Tran-Johnson}, \bibinfo{person}{Guro Khundadze}, \bibinfo{person}{Jackson Kernion}, \bibinfo{person}{James Landis}, \bibinfo{person}{Jamie Kerr}, \bibinfo{person}{Jared Mueller}, \bibinfo{person}{Jeeyoon Hyun}, \bibinfo{person}{Joshua Landau}, \bibinfo{person}{Kamal Ndousse},
  \bibinfo{person}{Landon Goldberg}, \bibinfo{person}{Liane Lovitt}, \bibinfo{person}{Martin Lucas}, \bibinfo{person}{Michael Sellitto}, \bibinfo{person}{Miranda Zhang}, \bibinfo{person}{Neerav Kingsland}, \bibinfo{person}{Nelson Elhage}, \bibinfo{person}{Nicholas Joseph}, \bibinfo{person}{Noemi Mercado}, \bibinfo{person}{Nova DasSarma}, \bibinfo{person}{Oliver Rausch}, \bibinfo{person}{Robin Larson}, \bibinfo{person}{Sam McCandlish}, \bibinfo{person}{Scott Johnston}, \bibinfo{person}{Shauna Kravec}, \bibinfo{person}{Sheer El~Showk}, \bibinfo{person}{Tamera Lanham}, \bibinfo{person}{Timothy Telleen-Lawton}, \bibinfo{person}{Tom Brown}, \bibinfo{person}{Tom Henighan}, \bibinfo{person}{Tristan Hume}, \bibinfo{person}{Yuntao Bai}, \bibinfo{person}{Zac Hatfield-Dodds}, \bibinfo{person}{Jack Clark}, \bibinfo{person}{Samuel~R. Bowman}, \bibinfo{person}{Amanda Askell}, \bibinfo{person}{Roger Grosse}, \bibinfo{person}{Danny Hernandez}, \bibinfo{person}{Deep Ganguli}, \bibinfo{person}{Evan Hubinger},
  \bibinfo{person}{Nicholas Schiefer}, {and} \bibinfo{person}{Jared Kaplan}.} \bibinfo{year}{2023}\natexlab{}.
\newblock \showarticletitle{Discovering Language Model Behaviors with Model-Written Evaluations}. In \bibinfo{booktitle}{\emph{Findings of the Association for Computational Linguistics: ACL 2023}}. \bibinfo{pages}{13387--13434}.
\newblock


\bibitem[Pfeiffer et~al\mbox{.}(2024)]%
        {pfeiffer2024h2o}
\bibfield{author}{\bibinfo{person}{Pascal Pfeiffer}, \bibinfo{person}{Philipp Singer}, \bibinfo{person}{Yauhen Babakhin}, \bibinfo{person}{Gabor Fodor}, \bibinfo{person}{Nischay Dhankhar}, {and} \bibinfo{person}{Sri~Satish Ambati}.} \bibinfo{year}{2024}\natexlab{}.
\newblock \showarticletitle{H2O-Danube3 Technical Report}.
\newblock \bibinfo{journal}{\emph{arXiv preprint arXiv:2407.09276}} (\bibinfo{year}{2024}).
\newblock


\bibitem[Phogat et~al\mbox{.}(2024)]%
        {phogat2024finetuningsmallerlanguagemodels}
\bibfield{author}{\bibinfo{person}{Karmvir~Singh Phogat}, \bibinfo{person}{Sai~Akhil Puranam}, \bibinfo{person}{Sridhar Dasaratha}, \bibinfo{person}{Chetan Harsha}, {and} \bibinfo{person}{Shashishekar Ramakrishna}.} \bibinfo{year}{2024}\natexlab{}.
\newblock \showarticletitle{Fine-tuning Smaller Language Models for Question Answering over Financial Documents}. In \bibinfo{booktitle}{\emph{Findings of the Association for Computational Linguistics: EMNLP 2024}}, \bibfield{editor}{\bibinfo{person}{Yaser Al-Onaizan}, \bibinfo{person}{Mohit Bansal}, {and} \bibinfo{person}{Yun-Nung Chen}} (Eds.). \bibinfo{publisher}{Association for Computational Linguistics}, \bibinfo{address}{Miami, Florida, USA}, \bibinfo{pages}{10528--10548}.
\newblock
\urldef\tempurl%
\url{https://doi.org/10.18653/v1/2024.findings-emnlp.617}
\showDOI{\tempurl}


\bibitem[Pope et~al\mbox{.}(2023)]%
        {pope2023efficiently}
\bibfield{author}{\bibinfo{person}{Reiner Pope}, \bibinfo{person}{Sholto Douglas}, \bibinfo{person}{Aakanksha Chowdhery}, \bibinfo{person}{Jacob Devlin}, \bibinfo{person}{James Bradbury}, \bibinfo{person}{Jonathan Heek}, \bibinfo{person}{Kefan Xiao}, \bibinfo{person}{Shivani Agrawal}, {and} \bibinfo{person}{Jeff Dean}.} \bibinfo{year}{2023}\natexlab{}.
\newblock \showarticletitle{Efficiently scaling transformer inference}.
\newblock \bibinfo{journal}{\emph{Proceedings of Machine Learning and Systems}}  \bibinfo{volume}{5} (\bibinfo{year}{2023}), \bibinfo{pages}{606--624}.
\newblock


\bibitem[Press et~al\mbox{.}(2022)]%
        {press2022train}
\bibfield{author}{\bibinfo{person}{Ofir Press}, \bibinfo{person}{Noah Smith}, {and} \bibinfo{person}{Mike Lewis}.} \bibinfo{year}{2022}\natexlab{}.
\newblock \showarticletitle{Train Short, Test Long: Attention with Linear Biases Enables Input Length Extrapolation}. In \bibinfo{booktitle}{\emph{International Conference on Learning Representations}}.
\newblock
\urldef\tempurl%
\url{https://openreview.net/forum?id=R8sQPpGCv0}
\showURL{%
\tempurl}


\bibitem[Qin et~al\mbox{.}(2023)]%
        {qin2023enabling}
\bibfield{author}{\bibinfo{person}{Ruiyang Qin}, \bibinfo{person}{Jun Xia}, \bibinfo{person}{Zhenge Jia}, \bibinfo{person}{Meng Jiang}, \bibinfo{person}{Ahmed Abbasi}, \bibinfo{person}{Peipei Zhou}, \bibinfo{person}{Jingtong Hu}, {and} \bibinfo{person}{Yiyu Shi}.} \bibinfo{year}{2023}\natexlab{}.
\newblock \showarticletitle{Enabling on-device large language model personalization with self-supervised data selection and synthesis}.
\newblock \bibinfo{journal}{\emph{arXiv preprint arXiv:2311.12275}} (\bibinfo{year}{2023}).
\newblock


\bibitem[Qin et~al\mbox{.}({[n.\,d.]})]%
        {qintoolllm}
\bibfield{author}{\bibinfo{person}{Yujia Qin}, \bibinfo{person}{Shihao Liang}, \bibinfo{person}{Yining Ye}, \bibinfo{person}{Kunlun Zhu}, \bibinfo{person}{Lan Yan}, \bibinfo{person}{Yaxi Lu}, \bibinfo{person}{Yankai Lin}, \bibinfo{person}{Xin Cong}, \bibinfo{person}{Xiangru Tang}, \bibinfo{person}{Bill Qian}, {et~al\mbox{.}}} \bibinfo{year}{[n.\,d.]}\natexlab{}.
\newblock \showarticletitle{ToolLLM: Facilitating Large Language Models to Master 16000+ Real-world APIs}. In \bibinfo{booktitle}{\emph{The Twelfth International Conference on Learning Representations}}.
\newblock


\bibitem[Qu et~al\mbox{.}(2024a)]%
        {qu2024survey}
\bibfield{author}{\bibinfo{person}{Haohao Qu}, \bibinfo{person}{Liangbo Ning}, \bibinfo{person}{Rui An}, \bibinfo{person}{Wenqi Fan}, \bibinfo{person}{Tyler Derr}, \bibinfo{person}{Hui Liu}, \bibinfo{person}{Xin Xu}, {and} \bibinfo{person}{Qing Li}.} \bibinfo{year}{2024}\natexlab{a}.
\newblock \showarticletitle{A survey of mamba}.
\newblock \bibinfo{journal}{\emph{arXiv preprint arXiv:2408.01129}} (\bibinfo{year}{2024}).
\newblock


\bibitem[Qu et~al\mbox{.}(2024b)]%
        {qu2024ssd4rec}
\bibfield{author}{\bibinfo{person}{Haohao Qu}, \bibinfo{person}{Yifeng Zhang}, \bibinfo{person}{Liangbo Ning}, \bibinfo{person}{Wenqi Fan}, {and} \bibinfo{person}{Qing Li}.} \bibinfo{year}{2024}\natexlab{b}.
\newblock \showarticletitle{Ssd4rec: a structured state space duality model for efficient sequential recommendation}.
\newblock \bibinfo{journal}{\emph{arXiv preprint arXiv:2409.01192}} (\bibinfo{year}{2024}).
\newblock


\bibitem[Radford et~al\mbox{.}(2019)]%
        {radford2019language}
\bibfield{author}{\bibinfo{person}{Alec Radford}, \bibinfo{person}{Jeffrey Wu}, \bibinfo{person}{Rewon Child}, \bibinfo{person}{David Luan}, \bibinfo{person}{Dario Amodei}, \bibinfo{person}{Ilya Sutskever}, {et~al\mbox{.}}} \bibinfo{year}{2019}\natexlab{}.
\newblock \showarticletitle{Language models are unsupervised multitask learners}.
\newblock \bibinfo{journal}{\emph{OpenAI blog}} \bibinfo{volume}{1}, \bibinfo{number}{8} (\bibinfo{year}{2019}), \bibinfo{pages}{9}.
\newblock


\bibitem[Rafailov et~al\mbox{.}(2024)]%
        {rafailov2024direct}
\bibfield{author}{\bibinfo{person}{Rafael Rafailov}, \bibinfo{person}{Archit Sharma}, \bibinfo{person}{Eric Mitchell}, \bibinfo{person}{Christopher~D Manning}, \bibinfo{person}{Stefano Ermon}, {and} \bibinfo{person}{Chelsea Finn}.} \bibinfo{year}{2024}\natexlab{}.
\newblock \showarticletitle{Direct preference optimization: Your language model is secretly a reward model}.
\newblock \bibinfo{journal}{\emph{Advances in Neural Information Processing Systems}}  \bibinfo{volume}{36} (\bibinfo{year}{2024}).
\newblock


\bibitem[Raffel et~al\mbox{.}(2020)]%
        {raffel2020exploring}
\bibfield{author}{\bibinfo{person}{Colin Raffel}, \bibinfo{person}{Noam Shazeer}, \bibinfo{person}{Adam Roberts}, \bibinfo{person}{Katherine Lee}, \bibinfo{person}{Sharan Narang}, \bibinfo{person}{Michael Matena}, \bibinfo{person}{Yanqi Zhou}, \bibinfo{person}{Wei Li}, {and} \bibinfo{person}{Peter~J Liu}.} \bibinfo{year}{2020}\natexlab{}.
\newblock \showarticletitle{Exploring the limits of transfer learning with a unified text-to-text transformer}.
\newblock \bibinfo{journal}{\emph{Journal of machine learning research}} \bibinfo{volume}{21}, \bibinfo{number}{140} (\bibinfo{year}{2020}), \bibinfo{pages}{1--67}.
\newblock


\bibitem[Rahman et~al\mbox{.}(2023)]%
        {rahman2023quantizedtransformerlanguagemodel}
\bibfield{author}{\bibinfo{person}{Mohammad Wali~Ur Rahman}, \bibinfo{person}{Murad~Mehrab Abrar}, \bibinfo{person}{Hunter~Gibbons Copening}, \bibinfo{person}{Salim Hariri}, \bibinfo{person}{Sicong Shao}, \bibinfo{person}{Pratik Satam}, {and} \bibinfo{person}{Soheil Salehi}.} \bibinfo{year}{2023}\natexlab{}.
\newblock \bibinfo{title}{Quantized Transformer Language Model Implementations on Edge Devices}.
\newblock
\newblock
\showeprint[arxiv]{2310.03971}~[cs.CL]
\urldef\tempurl%
\url{https://arxiv.org/abs/2310.03971}
\showURL{%
\tempurl}


\bibitem[Rajbhandari et~al\mbox{.}(2020)]%
        {rajbhandari2020zero}
\bibfield{author}{\bibinfo{person}{Samyam Rajbhandari}, \bibinfo{person}{Jeff Rasley}, \bibinfo{person}{Olatunji Ruwase}, {and} \bibinfo{person}{Yuxiong He}.} \bibinfo{year}{2020}\natexlab{}.
\newblock \showarticletitle{Zero: Memory optimizations toward training trillion parameter models}. In \bibinfo{booktitle}{\emph{SC20: International Conference for High Performance Computing, Networking, Storage and Analysis}}. IEEE, \bibinfo{pages}{1--16}.
\newblock


\bibitem[Ram et~al\mbox{.}(2023)]%
        {ram2023context}
\bibfield{author}{\bibinfo{person}{Ori Ram}, \bibinfo{person}{Yoav Levine}, \bibinfo{person}{Itay Dalmedigos}, \bibinfo{person}{Dor Muhlgay}, \bibinfo{person}{Amnon Shashua}, \bibinfo{person}{Kevin Leyton-Brown}, {and} \bibinfo{person}{Yoav Shoham}.} \bibinfo{year}{2023}\natexlab{}.
\newblock \showarticletitle{In-Context Retrieval-Augmented Language Models}.
\newblock \bibinfo{journal}{\emph{Transactions of the Association for Computational Linguistics}}  \bibinfo{volume}{11} (\bibinfo{year}{2023}), \bibinfo{pages}{1316--1331}.
\newblock


\bibitem[Ramesh et~al\mbox{.}(2023)]%
        {ramesh-etal-2023-comparative}
\bibfield{author}{\bibinfo{person}{Krithika Ramesh}, \bibinfo{person}{Arnav Chavan}, \bibinfo{person}{Shrey Pandit}, {and} \bibinfo{person}{Sunayana Sitaram}.} \bibinfo{year}{2023}\natexlab{}.
\newblock \showarticletitle{A Comparative Study on the Impact of Model Compression Techniques on Fairness in Language Models}. In \bibinfo{booktitle}{\emph{Proceedings of the 61st Annual Meeting of the Association for Computational Linguistics (Volume 1: Long Papers)}}. \bibinfo{publisher}{Association for Computational Linguistics}, \bibinfo{pages}{15762--15782}.
\newblock
\urldef\tempurl%
\url{https://aclanthology.org/2023.acl-long.878}
\showURL{%
\tempurl}


\bibitem[Rashid et~al\mbox{.}(2008)]%
        {rashid2008learning}
\bibfield{author}{\bibinfo{person}{Al~Mamunur Rashid}, \bibinfo{person}{George Karypis}, {and} \bibinfo{person}{John Riedl}.} \bibinfo{year}{2008}\natexlab{}.
\newblock \showarticletitle{Learning preferences of new users in recommender systems: an information theoretic approach}.
\newblock \bibinfo{journal}{\emph{Acm Sigkdd Explorations Newsletter}} \bibinfo{volume}{10}, \bibinfo{number}{2} (\bibinfo{year}{2008}), \bibinfo{pages}{90--100}.
\newblock


\bibitem[Rawles et~al\mbox{.}(2024)]%
        {rawles2024androidinthewild}
\bibfield{author}{\bibinfo{person}{Christopher Rawles}, \bibinfo{person}{Alice Li}, \bibinfo{person}{Daniel Rodriguez}, \bibinfo{person}{Oriana Riva}, {and} \bibinfo{person}{Timothy Lillicrap}.} \bibinfo{year}{2024}\natexlab{}.
\newblock \showarticletitle{Androidinthewild: A large-scale dataset for android device control}.
\newblock \bibinfo{journal}{\emph{Advances in Neural Information Processing Systems}}  \bibinfo{volume}{36} (\bibinfo{year}{2024}).
\newblock


\bibitem[Robertson et~al\mbox{.}(2009)]%
        {robertson2009probabilistic}
\bibfield{author}{\bibinfo{person}{Stephen Robertson}, \bibinfo{person}{Hugo Zaragoza}, {et~al\mbox{.}}} \bibinfo{year}{2009}\natexlab{}.
\newblock \showarticletitle{The probabilistic relevance framework: BM25 and beyond}.
\newblock \bibinfo{journal}{\emph{Foundations and Trends{\textregistered} in Information Retrieval}} \bibinfo{volume}{3}, \bibinfo{number}{4} (\bibinfo{year}{2009}), \bibinfo{pages}{333--389}.
\newblock


\bibitem[Roziere et~al\mbox{.}(2023)]%
        {roziere2023code}
\bibfield{author}{\bibinfo{person}{Baptiste Roziere}, \bibinfo{person}{Jonas Gehring}, \bibinfo{person}{Fabian Gloeckle}, \bibinfo{person}{Sten Sootla}, \bibinfo{person}{Itai Gat}, \bibinfo{person}{Xiaoqing~Ellen Tan}, \bibinfo{person}{Yossi Adi}, \bibinfo{person}{Jingyu Liu}, \bibinfo{person}{Tal Remez}, \bibinfo{person}{J{\'e}r{\'e}my Rapin}, {et~al\mbox{.}}} \bibinfo{year}{2023}\natexlab{}.
\newblock \showarticletitle{Code llama: Open foundation models for code}.
\newblock \bibinfo{journal}{\emph{arXiv preprint arXiv:2308.12950}} (\bibinfo{year}{2023}).
\newblock


\bibitem[Sadowski and Levin(2007)]%
        {sadowski2007simhash}
\bibfield{author}{\bibinfo{person}{Caitlin Sadowski} {and} \bibinfo{person}{Greg Levin}.} \bibinfo{year}{2007}\natexlab{}.
\newblock \bibinfo{title}{Simhash: Hash-based similarity detection}.
\newblock
\newblock


\bibitem[Sakaguchi et~al\mbox{.}(2021)]%
        {sakaguchi2021winogrande}
\bibfield{author}{\bibinfo{person}{Keisuke Sakaguchi}, \bibinfo{person}{Ronan~Le Bras}, \bibinfo{person}{Chandra Bhagavatula}, {and} \bibinfo{person}{Yejin Choi}.} \bibinfo{year}{2021}\natexlab{}.
\newblock \showarticletitle{Winogrande: An adversarial winograd schema challenge at scale}.
\newblock \bibinfo{journal}{\emph{Commun. ACM}} \bibinfo{volume}{64}, \bibinfo{number}{9} (\bibinfo{year}{2021}), \bibinfo{pages}{99--106}.
\newblock


\bibitem[Schick et~al\mbox{.}(2024)]%
        {schick2024toolformer}
\bibfield{author}{\bibinfo{person}{Timo Schick}, \bibinfo{person}{Jane Dwivedi-Yu}, \bibinfo{person}{Roberto Dess{\`\i}}, \bibinfo{person}{Roberta Raileanu}, \bibinfo{person}{Maria Lomeli}, \bibinfo{person}{Eric Hambro}, \bibinfo{person}{Luke Zettlemoyer}, \bibinfo{person}{Nicola Cancedda}, {and} \bibinfo{person}{Thomas Scialom}.} \bibinfo{year}{2024}\natexlab{}.
\newblock \showarticletitle{Toolformer: Language models can teach themselves to use tools}.
\newblock \bibinfo{journal}{\emph{Advances in Neural Information Processing Systems}}  \bibinfo{volume}{36} (\bibinfo{year}{2024}).
\newblock


\bibitem[Sennrich et~al\mbox{.}(2023)]%
        {sennrich2023mitigating}
\bibfield{author}{\bibinfo{person}{Rico Sennrich}, \bibinfo{person}{Jannis Vamvas}, {and} \bibinfo{person}{Alireza Mohammadshahi}.} \bibinfo{year}{2023}\natexlab{}.
\newblock \showarticletitle{Mitigating Hallucinations and Off-target Machine Translation with Source-Contrastive and Language-Contrastive Decoding}.
\newblock \bibinfo{journal}{\emph{arXiv preprint arXiv:2309.07098}} (\bibinfo{year}{2023}).
\newblock


\bibitem[Sha and Zhang(2024)]%
        {sha2024prompt}
\bibfield{author}{\bibinfo{person}{Zeyang Sha} {and} \bibinfo{person}{Yang Zhang}.} \bibinfo{year}{2024}\natexlab{}.
\newblock \showarticletitle{Prompt stealing attacks against large language models}.
\newblock \bibinfo{journal}{\emph{arXiv preprint arXiv:2402.12959}} (\bibinfo{year}{2024}).
\newblock


\bibitem[Shang et~al\mbox{.}(2024)]%
        {shang2024defint}
\bibfield{author}{\bibinfo{person}{Yu Shang}, \bibinfo{person}{Yu Li}, \bibinfo{person}{Fengli Xu}, {and} \bibinfo{person}{Yong Li}.} \bibinfo{year}{2024}\natexlab{}.
\newblock \showarticletitle{Synergy-of-Thoughts: Eliciting Efficient Reasoning in Hybrid Language Models}.
\newblock \bibinfo{journal}{\emph{arXiv preprint arXiv:2402.02563}} (\bibinfo{year}{2024}).
\newblock


\bibitem[Shang et~al\mbox{.}(2023)]%
        {shang2023pbllmpartiallybinarizedlarge}
\bibfield{author}{\bibinfo{person}{Yuzhang Shang}, \bibinfo{person}{Zhihang Yuan}, \bibinfo{person}{Qiang Wu}, {and} \bibinfo{person}{Zhen Dong}.} \bibinfo{year}{2023}\natexlab{}.
\newblock \bibinfo{title}{PB-LLM: Partially Binarized Large Language Models}.
\newblock
\newblock
\showeprint[arxiv]{2310.00034}~[cs.LG]
\urldef\tempurl%
\url{https://arxiv.org/abs/2310.00034}
\showURL{%
\tempurl}


\bibitem[Shao et~al\mbox{.}(2024)]%
        {shao2024one}
\bibfield{author}{\bibinfo{person}{Hang Shao}, \bibinfo{person}{Bei Liu}, {and} \bibinfo{person}{Yanmin Qian}.} \bibinfo{year}{2024}\natexlab{}.
\newblock \showarticletitle{One-shot sensitivity-aware mixed sparsity pruning for large language models}. In \bibinfo{booktitle}{\emph{ICASSP 2024-2024 IEEE International Conference on Acoustics, Speech and Signal Processing (ICASSP)}}. IEEE, \bibinfo{pages}{11296--11300}.
\newblock


\bibitem[Sharma et~al\mbox{.}(2023)]%
        {sharma2023towards}
\bibfield{author}{\bibinfo{person}{Mrinank Sharma}, \bibinfo{person}{Meg Tong}, \bibinfo{person}{Tomasz Korbak}, \bibinfo{person}{David Duvenaud}, \bibinfo{person}{Amanda Askell}, \bibinfo{person}{Samuel~R Bowman}, \bibinfo{person}{Newton Cheng}, \bibinfo{person}{Esin Durmus}, \bibinfo{person}{Zac Hatfield-Dodds}, \bibinfo{person}{Scott~R Johnston}, {et~al\mbox{.}}} \bibinfo{year}{2023}\natexlab{}.
\newblock \showarticletitle{Towards understanding sycophancy in language models}.
\newblock \bibinfo{journal}{\emph{arXiv preprint arXiv:2310.13548}} (\bibinfo{year}{2023}).
\newblock


\bibitem[Shayegani et~al\mbox{.}(2023)]%
        {shayegani2023survey}
\bibfield{author}{\bibinfo{person}{Erfan Shayegani}, \bibinfo{person}{Md~Abdullah~Al Mamun}, \bibinfo{person}{Yu Fu}, \bibinfo{person}{Pedram Zaree}, \bibinfo{person}{Yue Dong}, {and} \bibinfo{person}{Nael Abu-Ghazaleh}.} \bibinfo{year}{2023}\natexlab{}.
\newblock \showarticletitle{Survey of vulnerabilities in large language models revealed by adversarial attacks}.
\newblock \bibinfo{journal}{\emph{arXiv preprint arXiv:2310.10844}} (\bibinfo{year}{2023}).
\newblock


\bibitem[Shazeer(2019)]%
        {shazeer2019fast}
\bibfield{author}{\bibinfo{person}{Noam Shazeer}.} \bibinfo{year}{2019}\natexlab{}.
\newblock \showarticletitle{Fast transformer decoding: One write-head is all you need}.
\newblock \bibinfo{journal}{\emph{arXiv preprint arXiv:1911.02150}} (\bibinfo{year}{2019}).
\newblock


\bibitem[Shazeer(2020)]%
        {swiglu}
\bibfield{author}{\bibinfo{person}{Noam Shazeer}.} \bibinfo{year}{2020}\natexlab{}.
\newblock \showarticletitle{Glu variants improve transformer}.
\newblock \bibinfo{journal}{\emph{arXiv preprint arXiv:2002.05202}} (\bibinfo{year}{2020}).
\newblock


\bibitem[Shen et~al\mbox{.}(2022)]%
        {shen-etal-2022-cost}
\bibfield{author}{\bibinfo{person}{Bowen Shen}, \bibinfo{person}{Zheng Lin}, \bibinfo{person}{Yuanxin Liu}, \bibinfo{person}{Zhengxiao Liu}, \bibinfo{person}{Lei Wang}, {and} \bibinfo{person}{Weiping Wang}.} \bibinfo{year}{2022}\natexlab{}.
\newblock \showarticletitle{{COST}-{EFF}: Collaborative Optimization of Spatial and Temporal Efficiency with Slenderized Multi-exit Language Models}. In \bibinfo{booktitle}{\emph{Proceedings of the 2022 Conference on Empirical Methods in Natural Language Processing}}, \bibfield{editor}{\bibinfo{person}{Yoav Goldberg}, \bibinfo{person}{Zornitsa Kozareva}, {and} \bibinfo{person}{Yue Zhang}} (Eds.). \bibinfo{publisher}{Association for Computational Linguistics}, \bibinfo{address}{Abu Dhabi, United Arab Emirates}, \bibinfo{pages}{1719--1730}.
\newblock
\urldef\tempurl%
\url{https://doi.org/10.18653/v1/2022.emnlp-main.112}
\showDOI{\tempurl}


\bibitem[Shen et~al\mbox{.}(2024b)]%
        {shen-etal-2024-pruning}
\bibfield{author}{\bibinfo{person}{Bowen Shen}, \bibinfo{person}{Zheng Lin}, \bibinfo{person}{Daren Zha}, \bibinfo{person}{Wei Liu}, \bibinfo{person}{Jian Luan}, \bibinfo{person}{Bin Wang}, {and} \bibinfo{person}{Weiping Wang}.} \bibinfo{year}{2024}\natexlab{b}.
\newblock \showarticletitle{Pruning Large Language Models to Intra-module Low-rank Architecture with Transitional Activations}. In \bibinfo{booktitle}{\emph{Findings of the Association for Computational Linguistics: ACL 2024}}, \bibfield{editor}{\bibinfo{person}{Lun-Wei Ku}, \bibinfo{person}{Andre Martins}, {and} \bibinfo{person}{Vivek Srikumar}} (Eds.). \bibinfo{publisher}{Association for Computational Linguistics}, \bibinfo{address}{Bangkok, Thailand}, \bibinfo{pages}{9781--9793}.
\newblock
\urldef\tempurl%
\url{https://doi.org/10.18653/v1/2024.findings-acl.582}
\showDOI{\tempurl}


\bibitem[Shen et~al\mbox{.}(2024a)]%
        {shen-etal-2024-small}
\bibfield{author}{\bibinfo{person}{Weizhou Shen}, \bibinfo{person}{Chenliang Li}, \bibinfo{person}{Hongzhan Chen}, \bibinfo{person}{Ming Yan}, \bibinfo{person}{Xiaojun Quan}, \bibinfo{person}{Hehong Chen}, \bibinfo{person}{Ji Zhang}, {and} \bibinfo{person}{Fei Huang}.} \bibinfo{year}{2024}\natexlab{a}.
\newblock \showarticletitle{Small {LLM}s Are Weak Tool Learners: A Multi-{LLM} Agent}. In \bibinfo{booktitle}{\emph{Proceedings of the 2024 Conference on Empirical Methods in Natural Language Processing}}. \bibinfo{publisher}{Association for Computational Linguistics}, \bibinfo{address}{Miami, Florida, USA}, \bibinfo{pages}{16658--16680}.
\newblock
\urldef\tempurl%
\url{https://doi.org/10.18653/v1/2024.emnlp-main.929}
\showDOI{\tempurl}


\bibitem[Sheng et~al\mbox{.}(2023)]%
        {sheng2023flexgen}
\bibfield{author}{\bibinfo{person}{Ying Sheng}, \bibinfo{person}{Lianmin Zheng}, \bibinfo{person}{Binhang Yuan}, \bibinfo{person}{Zhuohan Li}, \bibinfo{person}{Max Ryabinin}, \bibinfo{person}{Beidi Chen}, \bibinfo{person}{Percy Liang}, \bibinfo{person}{Christopher R{\'e}}, \bibinfo{person}{Ion Stoica}, {and} \bibinfo{person}{Ce Zhang}.} \bibinfo{year}{2023}\natexlab{}.
\newblock \showarticletitle{Flexgen: High-throughput generative inference of large language models with a single gpu}. In \bibinfo{booktitle}{\emph{International Conference on Machine Learning}}. PMLR, \bibinfo{pages}{31094--31116}.
\newblock


\bibitem[Shi et~al\mbox{.}(2024)]%
        {shi2024large}
\bibfield{author}{\bibinfo{person}{Wentao Shi}, \bibinfo{person}{Xiangnan He}, \bibinfo{person}{Yang Zhang}, \bibinfo{person}{Chongming Gao}, \bibinfo{person}{Xinyue Li}, \bibinfo{person}{Jizhi Zhang}, \bibinfo{person}{Qifan Wang}, {and} \bibinfo{person}{Fuli Feng}.} \bibinfo{year}{2024}\natexlab{}.
\newblock \showarticletitle{Large language models are learnable planners for long-term recommendation}. In \bibinfo{booktitle}{\emph{Proceedings of the 47th International ACM SIGIR Conference on Research and Development in Information Retrieval}}. \bibinfo{pages}{1893--1903}.
\newblock


\bibitem[Shoeybi et~al\mbox{.}(2019)]%
        {shoeybi2019megatron}
\bibfield{author}{\bibinfo{person}{Mohammad Shoeybi}, \bibinfo{person}{Mostofa Patwary}, \bibinfo{person}{Raul Puri}, \bibinfo{person}{Patrick LeGresley}, \bibinfo{person}{Jared Casper}, {and} \bibinfo{person}{Bryan Catanzaro}.} \bibinfo{year}{2019}\natexlab{}.
\newblock \showarticletitle{Megatron-lm: Training multi-billion parameter language models using model parallelism}.
\newblock \bibinfo{journal}{\emph{arXiv preprint arXiv:1909.08053}} (\bibinfo{year}{2019}).
\newblock


\bibitem[Shojaee et~al\mbox{.}(2024)]%
        {shojaee2024llm}
\bibfield{author}{\bibinfo{person}{Parshin Shojaee}, \bibinfo{person}{Kazem Meidani}, \bibinfo{person}{Shashank Gupta}, \bibinfo{person}{Amir~Barati Farimani}, {and} \bibinfo{person}{Chandan~K Reddy}.} \bibinfo{year}{2024}\natexlab{}.
\newblock \showarticletitle{Llm-sr: Scientific equation discovery via programming with large language models}.
\newblock \bibinfo{journal}{\emph{arXiv preprint arXiv:2404.18400}} (\bibinfo{year}{2024}).
\newblock


\bibitem[Shu et~al\mbox{.}(2022)]%
        {shu2022test}
\bibfield{author}{\bibinfo{person}{Manli Shu}, \bibinfo{person}{Weili Nie}, \bibinfo{person}{De-An Huang}, \bibinfo{person}{Zhiding Yu}, \bibinfo{person}{Tom Goldstein}, \bibinfo{person}{Anima Anandkumar}, {and} \bibinfo{person}{Chaowei Xiao}.} \bibinfo{year}{2022}\natexlab{}.
\newblock \showarticletitle{Test-time prompt tuning for zero-shot generalization in vision-language models}.
\newblock \bibinfo{journal}{\emph{Advances in Neural Information Processing Systems}}  \bibinfo{volume}{35} (\bibinfo{year}{2022}), \bibinfo{pages}{14274--14289}.
\newblock


\bibitem[Singhal et~al\mbox{.}(2023)]%
        {singhal2023large}
\bibfield{author}{\bibinfo{person}{Karan Singhal}, \bibinfo{person}{Shekoofeh Azizi}, \bibinfo{person}{Tao Tu}, \bibinfo{person}{S~Sara Mahdavi}, \bibinfo{person}{Jason Wei}, \bibinfo{person}{Hyung~Won Chung}, \bibinfo{person}{Nathan Scales}, \bibinfo{person}{Ajay Tanwani}, \bibinfo{person}{Heather Cole-Lewis}, \bibinfo{person}{Stephen Pfohl}, {et~al\mbox{.}}} \bibinfo{year}{2023}\natexlab{}.
\newblock \showarticletitle{Large language models encode clinical knowledge}.
\newblock \bibinfo{journal}{\emph{Nature}} \bibinfo{volume}{620}, \bibinfo{number}{7972} (\bibinfo{year}{2023}), \bibinfo{pages}{172--180}.
\newblock


\bibitem[Soboleva et~al\mbox{.}(2023)]%
        {cerebras2023slimpajama}
\bibfield{author}{\bibinfo{person}{Daria Soboleva}, \bibinfo{person}{Faisal Al-Khateeb}, \bibinfo{person}{Robert Myers}, \bibinfo{person}{Jacob~R Steeves}, \bibinfo{person}{Joel Hestness}, {and} \bibinfo{person}{Nolan Dey}.} \bibinfo{year}{2023}\natexlab{}.
\newblock \bibinfo{title}{{SlimPajama: A 627B token cleaned and deduplicated version of RedPajama}}.
\newblock \bibinfo{howpublished}{\url{https://cerebras.ai/blog/slimpajama-a-627b-token-cleaned-and-deduplicated-version-of-redpajama}}.
\newblock
\urldef\tempurl%
\url{https://huggingface.co/datasets/cerebras/SlimPajama-627B}
\showURL{%
\tempurl}


\bibitem[Soldaini et~al\mbox{.}(2024)]%
        {dolma}
\bibfield{author}{\bibinfo{person}{Luca Soldaini}, \bibinfo{person}{Rodney Kinney}, \bibinfo{person}{Akshita Bhagia}, \bibinfo{person}{Dustin Schwenk}, \bibinfo{person}{David Atkinson}, \bibinfo{person}{Russell Authur}, \bibinfo{person}{Ben Bogin}, \bibinfo{person}{Khyathi Chandu}, \bibinfo{person}{Jennifer Dumas}, \bibinfo{person}{Yanai Elazar}, \bibinfo{person}{Valentin Hofmann}, \bibinfo{person}{Ananya~Harsh Jha}, \bibinfo{person}{Sachin Kumar}, \bibinfo{person}{Li Lucy}, \bibinfo{person}{Xinxi Lyu}, \bibinfo{person}{Nathan Lambert}, \bibinfo{person}{Ian Magnusson}, \bibinfo{person}{Jacob Morrison}, \bibinfo{person}{Niklas Muennighoff}, \bibinfo{person}{Aakanksha Naik}, \bibinfo{person}{Crystal Nam}, \bibinfo{person}{Matthew~E. Peters}, \bibinfo{person}{Abhilasha Ravichander}, \bibinfo{person}{Kyle Richardson}, \bibinfo{person}{Zejiang Shen}, \bibinfo{person}{Emma Strubell}, \bibinfo{person}{Nishant Subramani}, \bibinfo{person}{Oyvind Tafjord}, \bibinfo{person}{Pete Walsh}, \bibinfo{person}{Luke
  Zettlemoyer}, \bibinfo{person}{Noah~A. Smith}, \bibinfo{person}{Hannaneh Hajishirzi}, \bibinfo{person}{Iz Beltagy}, \bibinfo{person}{Dirk Groeneveld}, \bibinfo{person}{Jesse Dodge}, {and} \bibinfo{person}{Kyle Lo}.} \bibinfo{year}{2024}\natexlab{}.
\newblock \showarticletitle{{Dolma: an Open Corpus of Three Trillion Tokens for Language Model Pretraining Research}}.
\newblock \bibinfo{journal}{\emph{arXiv preprint}} (\bibinfo{year}{2024}).
\newblock


\bibitem[Song et~al\mbox{.}(2021)]%
        {song2021fast}
\bibfield{author}{\bibinfo{person}{Xinying Song}, \bibinfo{person}{Alex Salcianu}, \bibinfo{person}{Yang Song}, \bibinfo{person}{Dave Dopson}, {and} \bibinfo{person}{Denny Zhou}.} \bibinfo{year}{2021}\natexlab{}.
\newblock \showarticletitle{Fast WordPiece Tokenization}. In \bibinfo{booktitle}{\emph{Proceedings of the 2021 Conference on Empirical Methods in Natural Language Processing}}. \bibinfo{pages}{2089--2103}.
\newblock


\bibitem[Spatharioti et~al\mbox{.}(2023)]%
        {spatharioti2023comparing}
\bibfield{author}{\bibinfo{person}{Sofia~Eleni Spatharioti}, \bibinfo{person}{David~M Rothschild}, \bibinfo{person}{Daniel~G Goldstein}, {and} \bibinfo{person}{Jake~M Hofman}.} \bibinfo{year}{2023}\natexlab{}.
\newblock \showarticletitle{Comparing traditional and llm-based search for consumer choice: A randomized experiment}.
\newblock \bibinfo{journal}{\emph{arXiv preprint arXiv:2307.03744}} (\bibinfo{year}{2023}).
\newblock


\bibitem[Su et~al\mbox{.}(2024)]%
        {su2024roformer}
\bibfield{author}{\bibinfo{person}{Jianlin Su}, \bibinfo{person}{Murtadha Ahmed}, \bibinfo{person}{Yu Lu}, \bibinfo{person}{Shengfeng Pan}, \bibinfo{person}{Wen Bo}, {and} \bibinfo{person}{Yunfeng Liu}.} \bibinfo{year}{2024}\natexlab{}.
\newblock \showarticletitle{Roformer: Enhanced transformer with rotary position embedding}.
\newblock \bibinfo{journal}{\emph{Neurocomputing}}  \bibinfo{volume}{568} (\bibinfo{year}{2024}), \bibinfo{pages}{127063}.
\newblock


\bibitem[Sun et~al\mbox{.}(2024a)]%
        {sun2024scieval}
\bibfield{author}{\bibinfo{person}{Liangtai Sun}, \bibinfo{person}{Yang Han}, \bibinfo{person}{Zihan Zhao}, \bibinfo{person}{Da Ma}, \bibinfo{person}{Zhennan Shen}, \bibinfo{person}{Baocai Chen}, \bibinfo{person}{Lu Chen}, {and} \bibinfo{person}{Kai Yu}.} \bibinfo{year}{2024}\natexlab{a}.
\newblock \showarticletitle{Scieval: A multi-level large language model evaluation benchmark for scientific research}. In \bibinfo{booktitle}{\emph{Proceedings of the AAAI Conference on Artificial Intelligence}}, Vol.~\bibinfo{volume}{38}. \bibinfo{pages}{19053--19061}.
\newblock


\bibitem[Sun et~al\mbox{.}(2024b)]%
        {sun2024trustllm}
\bibfield{author}{\bibinfo{person}{Lichao Sun}, \bibinfo{person}{Yue Huang}, \bibinfo{person}{Haoran Wang}, \bibinfo{person}{Siyuan Wu}, \bibinfo{person}{Qihui Zhang}, \bibinfo{person}{Chujie Gao}, \bibinfo{person}{Yixin Huang}, \bibinfo{person}{Wenhan Lyu}, \bibinfo{person}{Yixuan Zhang}, \bibinfo{person}{Xiner Li}, {et~al\mbox{.}}} \bibinfo{year}{2024}\natexlab{b}.
\newblock \showarticletitle{Trustllm: Trustworthiness in large language models}.
\newblock \bibinfo{journal}{\emph{arXiv preprint arXiv:2401.05561}} (\bibinfo{year}{2024}).
\newblock


\bibitem[Sun et~al\mbox{.}(2024c)]%
        {sun2024a}
\bibfield{author}{\bibinfo{person}{Mingjie Sun}, \bibinfo{person}{Zhuang Liu}, \bibinfo{person}{Anna Bair}, {and} \bibinfo{person}{J~Zico Kolter}.} \bibinfo{year}{2024}\natexlab{c}.
\newblock \showarticletitle{A Simple and Effective Pruning Approach for Large Language Models}. In \bibinfo{booktitle}{\emph{Proceedings of the Twelfth International Conference on Learning Representations, ICLR}}.
\newblock


\bibitem[Sun et~al\mbox{.}(2020)]%
        {sun2020mobilebert}
\bibfield{author}{\bibinfo{person}{Zhiqing Sun}, \bibinfo{person}{Hongkun Yu}, \bibinfo{person}{Xiaodan Song}, \bibinfo{person}{Renjie Liu}, \bibinfo{person}{Yiming Yang}, {and} \bibinfo{person}{Denny Zhou}.} \bibinfo{year}{2020}\natexlab{}.
\newblock \showarticletitle{Mobilebert: a compact task-agnostic bert for resource-limited devices}.
\newblock \bibinfo{journal}{\emph{arXiv preprint arXiv:2004.02984}} (\bibinfo{year}{2020}).
\newblock


\bibitem[Swayamdipta et~al\mbox{.}(2020)]%
        {swayamdipta2020dataset}
\bibfield{author}{\bibinfo{person}{Swabha Swayamdipta}, \bibinfo{person}{Roy Schwartz}, \bibinfo{person}{Nicholas Lourie}, \bibinfo{person}{Yizhong Wang}, \bibinfo{person}{Hannaneh Hajishirzi}, \bibinfo{person}{Noah~A Smith}, {and} \bibinfo{person}{Yejin Choi}.} \bibinfo{year}{2020}\natexlab{}.
\newblock \showarticletitle{Dataset Cartography: Mapping and Diagnosing Datasets with Training Dynamics}. In \bibinfo{booktitle}{\emph{Proceedings of the 2020 Conference on Empirical Methods in Natural Language Processing (EMNLP)}}. \bibinfo{pages}{9275--9293}.
\newblock


\bibitem[Talmor and Berant(2018)]%
        {talmor-berant-2018-web}
\bibfield{author}{\bibinfo{person}{Alon Talmor} {and} \bibinfo{person}{Jonathan Berant}.} \bibinfo{year}{2018}\natexlab{}.
\newblock \showarticletitle{The Web as a Knowledge-Base for Answering Complex Questions}. In \bibinfo{booktitle}{\emph{Proceedings of the 2018 Conference of the North {A}merican Chapter of the Association for Computational Linguistics: Human Language Technologies, Volume 1 (Long Papers)}}, \bibfield{editor}{\bibinfo{person}{Marilyn Walker}, \bibinfo{person}{Heng Ji}, {and} \bibinfo{person}{Amanda Stent}} (Eds.). \bibinfo{publisher}{Association for Computational Linguistics}, \bibinfo{address}{New Orleans, Louisiana}, \bibinfo{pages}{641--651}.
\newblock
\urldef\tempurl%
\url{https://doi.org/10.18653/v1/N18-1059}
\showDOI{\tempurl}


\bibitem[Tan et~al\mbox{.}(2024)]%
        {tan2024small}
\bibfield{author}{\bibinfo{person}{Jiejun Tan}, \bibinfo{person}{Zhicheng Dou}, \bibinfo{person}{Yutao Zhu}, \bibinfo{person}{Peidong Guo}, \bibinfo{person}{Kun Fang}, {and} \bibinfo{person}{Ji-Rong Wen}.} \bibinfo{year}{2024}\natexlab{}.
\newblock \showarticletitle{Small Models, Big Insights: Leveraging Slim Proxy Models To Decide When and What to Retrieve for LLMs}.
\newblock \bibinfo{journal}{\emph{arXiv preprint arXiv:2402.12052}} (\bibinfo{year}{2024}).
\newblock


\bibitem[Tang et~al\mbox{.}(2023)]%
        {tang2023toolalpaca}
\bibfield{author}{\bibinfo{person}{Qiaoyu Tang}, \bibinfo{person}{Ziliang Deng}, \bibinfo{person}{Hongyu Lin}, \bibinfo{person}{Xianpei Han}, \bibinfo{person}{Qiao Liang}, \bibinfo{person}{Boxi Cao}, {and} \bibinfo{person}{Le Sun}.} \bibinfo{year}{2023}\natexlab{}.
\newblock \showarticletitle{Toolalpaca: Generalized tool learning for language models with 3000 simulated cases}.
\newblock \bibinfo{journal}{\emph{arXiv preprint arXiv:2306.05301}} (\bibinfo{year}{2023}).
\newblock


\bibitem[Tang et~al\mbox{.}(2024b)]%
        {tang2024small}
\bibfield{author}{\bibinfo{person}{Xuemei Tang}, \bibinfo{person}{Jun Wang}, {and} \bibinfo{person}{Qi Su}.} \bibinfo{year}{2024}\natexlab{b}.
\newblock \showarticletitle{Small Language Model Is a Good Guide for Large Language Model in Chinese Entity Relation Extraction}.
\newblock \bibinfo{journal}{\emph{arXiv preprint arXiv:2402.14373}} (\bibinfo{year}{2024}).
\newblock


\bibitem[Tang et~al\mbox{.}(2024a)]%
        {tang2024rethinking}
\bibfield{author}{\bibinfo{person}{Yehui Tang}, \bibinfo{person}{Fangcheng Liu}, \bibinfo{person}{Yunsheng Ni}, \bibinfo{person}{Yuchuan Tian}, \bibinfo{person}{Zheyuan Bai}, \bibinfo{person}{Yi-Qi Hu}, \bibinfo{person}{Sichao Liu}, \bibinfo{person}{Shangling Jui}, \bibinfo{person}{Kai Han}, {and} \bibinfo{person}{Yunhe Wang}.} \bibinfo{year}{2024}\natexlab{a}.
\newblock \showarticletitle{Rethinking optimization and architecture for tiny language models}.
\newblock \bibinfo{journal}{\emph{arXiv preprint arXiv:2402.02791}} (\bibinfo{year}{2024}).
\newblock


\bibitem[Taori et~al\mbox{.}(2023)]%
        {alpaca}
\bibfield{author}{\bibinfo{person}{Rohan Taori}, \bibinfo{person}{Ishaan Gulrajani}, \bibinfo{person}{Tianyi Zhang}, \bibinfo{person}{Yann Dubois}, \bibinfo{person}{Xuechen Li}, \bibinfo{person}{Carlos Guestrin}, \bibinfo{person}{Percy Liang}, {and} \bibinfo{person}{Tatsunori~B. Hashimoto}.} \bibinfo{year}{2023}\natexlab{}.
\newblock \bibinfo{title}{Stanford Alpaca: An Instruction-following LLaMA model}.
\newblock \bibinfo{howpublished}{\url{https://github.com/tatsu-lab/stanford_alpaca}}.
\newblock


\bibitem[Taylor et~al\mbox{.}(2022)]%
        {taylor2022galactica}
\bibfield{author}{\bibinfo{person}{Ross Taylor}, \bibinfo{person}{Marcin Kardas}, \bibinfo{person}{Guillem Cucurull}, \bibinfo{person}{Thomas Scialom}, \bibinfo{person}{Anthony Hartshorn}, \bibinfo{person}{Elvis Saravia}, \bibinfo{person}{Andrew Poulton}, \bibinfo{person}{Viktor Kerkez}, {and} \bibinfo{person}{Robert Stojnic}.} \bibinfo{year}{2022}\natexlab{}.
\newblock \showarticletitle{Galactica: A large language model for science}.
\newblock \bibinfo{journal}{\emph{arXiv preprint arXiv:2211.09085}} (\bibinfo{year}{2022}).
\newblock


\bibitem[Team(2024a)]%
        {team2024codegemma}
\bibfield{author}{\bibinfo{person}{CodeGemma Team}.} \bibinfo{year}{2024}\natexlab{a}.
\newblock \showarticletitle{Codegemma: Open code models based on gemma}.
\newblock \bibinfo{journal}{\emph{arXiv preprint arXiv:2406.11409}} (\bibinfo{year}{2024}).
\newblock


\bibitem[Team et~al\mbox{.}(2024a)]%
        {team2024gemma}
\bibfield{author}{\bibinfo{person}{Gemma Team}, \bibinfo{person}{Thomas Mesnard}, \bibinfo{person}{Cassidy Hardin}, \bibinfo{person}{Robert Dadashi}, \bibinfo{person}{Surya Bhupatiraju}, \bibinfo{person}{Shreya Pathak}, \bibinfo{person}{Laurent Sifre}, \bibinfo{person}{Morgane Rivi{\`e}re}, \bibinfo{person}{Mihir~Sanjay Kale}, \bibinfo{person}{Juliette Love}, {et~al\mbox{.}}} \bibinfo{year}{2024}\natexlab{a}.
\newblock \showarticletitle{Gemma: Open models based on gemini research and technology}.
\newblock \bibinfo{journal}{\emph{arXiv preprint arXiv:2403.08295}} (\bibinfo{year}{2024}).
\newblock


\bibitem[Team et~al\mbox{.}(2024b)]%
        {team2024gemma2}
\bibfield{author}{\bibinfo{person}{Gemma Team}, \bibinfo{person}{Morgane Riviere}, \bibinfo{person}{Shreya Pathak}, \bibinfo{person}{Pier~Giuseppe Sessa}, \bibinfo{person}{Cassidy Hardin}, \bibinfo{person}{Surya Bhupatiraju}, \bibinfo{person}{L{\'e}onard Hussenot}, \bibinfo{person}{Thomas Mesnard}, \bibinfo{person}{Bobak Shahriari}, \bibinfo{person}{Alexandre Ram{\'e}}, {et~al\mbox{.}}} \bibinfo{year}{2024}\natexlab{b}.
\newblock \showarticletitle{Gemma 2: Improving open language models at a practical size}.
\newblock \bibinfo{journal}{\emph{arXiv preprint arXiv:2408.00118}} (\bibinfo{year}{2024}).
\newblock


\bibitem[Team(2024b)]%
        {fox1}
\bibfield{author}{\bibinfo{person}{TensorOpera Team}.} \bibinfo{year}{2024}\natexlab{b}.
\newblock \bibinfo{booktitle}{\emph{TensorOpera Unveils Fox Foundation Model: A Pioneering Small Language Model (SLM) for Cloud and Edge}}.
\newblock
\urldef\tempurl%
\url{https://blog.tensoropera.ai/tensoropera-unveils-fox-foundation-model-a-pioneering-open-source-slm-leading-the-way-against-tech-giants/}
\showURL{%
\tempurl}
\newblock
\shownote{Accessed: 2024-6-13}.


\bibitem[Teerapittayanon et~al\mbox{.}(2016)]%
        {teerapittayanon2016branchynet}
\bibfield{author}{\bibinfo{person}{Surat Teerapittayanon}, \bibinfo{person}{Bradley McDanel}, {and} \bibinfo{person}{Hsiang-Tsung Kung}.} \bibinfo{year}{2016}\natexlab{}.
\newblock \showarticletitle{Branchynet: Fast inference via early exiting from deep neural networks}. In \bibinfo{booktitle}{\emph{2016 23rd international conference on pattern recognition (ICPR)}}. IEEE, \bibinfo{pages}{2464--2469}.
\newblock


\bibitem[Thawakar et~al\mbox{.}(2024)]%
        {thawakar2024mobillama}
\bibfield{author}{\bibinfo{person}{Omkar Thawakar}, \bibinfo{person}{Ashmal Vayani}, \bibinfo{person}{Salman Khan}, \bibinfo{person}{Hisham Cholakal}, \bibinfo{person}{Rao~M Anwer}, \bibinfo{person}{Michael Felsberg}, \bibinfo{person}{Tim Baldwin}, \bibinfo{person}{Eric~P Xing}, {and} \bibinfo{person}{Fahad~Shahbaz Khan}.} \bibinfo{year}{2024}\natexlab{}.
\newblock \showarticletitle{Mobillama: Towards accurate and lightweight fully transparent gpt}.
\newblock \bibinfo{journal}{\emph{arXiv preprint arXiv:2402.16840}} (\bibinfo{year}{2024}).
\newblock


\bibitem[Tian et~al\mbox{.}(2024)]%
        {tian2024toward}
\bibfield{author}{\bibinfo{person}{Ye Tian}, \bibinfo{person}{Baolin Peng}, \bibinfo{person}{Linfeng Song}, \bibinfo{person}{Lifeng Jin}, \bibinfo{person}{Dian Yu}, \bibinfo{person}{Haitao Mi}, {and} \bibinfo{person}{Dong Yu}.} \bibinfo{year}{2024}\natexlab{}.
\newblock \showarticletitle{Toward Self-Improvement of LLMs via Imagination, Searching, and Criticizing}.
\newblock \bibinfo{journal}{\emph{arXiv preprint arXivko2024distillm:2404.12253}} (\bibinfo{year}{2024}).
\newblock


\bibitem[Touvron et~al\mbox{.}(2023a)]%
        {touvron2023llama}
\bibfield{author}{\bibinfo{person}{Hugo Touvron}, \bibinfo{person}{Thibaut Lavril}, \bibinfo{person}{Gautier Izacard}, \bibinfo{person}{Xavier Martinet}, \bibinfo{person}{Marie-Anne Lachaux}, \bibinfo{person}{Timoth{\'e}e Lacroix}, \bibinfo{person}{Baptiste Rozi{\`e}re}, \bibinfo{person}{Naman Goyal}, \bibinfo{person}{Eric Hambro}, \bibinfo{person}{Faisal Azhar}, {et~al\mbox{.}}} \bibinfo{year}{2023}\natexlab{a}.
\newblock \showarticletitle{Llama: Open and efficient foundation language models}.
\newblock \bibinfo{journal}{\emph{arXiv preprint arXiv:2302.13971}} (\bibinfo{year}{2023}).
\newblock


\bibitem[Touvron et~al\mbox{.}(2023b)]%
        {touvron2023llama2}
\bibfield{author}{\bibinfo{person}{Hugo Touvron}, \bibinfo{person}{Louis Martin}, \bibinfo{person}{Kevin Stone}, \bibinfo{person}{Peter Albert}, \bibinfo{person}{Amjad Almahairi}, \bibinfo{person}{Yasmine Babaei}, \bibinfo{person}{Nikolay Bashlykov}, \bibinfo{person}{Soumya Batra}, \bibinfo{person}{Prajjwal Bhargava}, \bibinfo{person}{Shruti Bhosale}, {et~al\mbox{.}}} \bibinfo{year}{2023}\natexlab{b}.
\newblock \showarticletitle{Llama 2: Open foundation and fine-tuned chat models}.
\newblock \bibinfo{journal}{\emph{arXiv preprint arXiv:2307.09288}} (\bibinfo{year}{2023}).
\newblock


\bibitem[Tow et~al\mbox{.}({[n.\,d.]})]%
        {StableLM-3B-4E1T}
\bibfield{author}{\bibinfo{person}{Jonathan Tow}, \bibinfo{person}{Marco Bellagente}, \bibinfo{person}{Dakota Mahan}, {and} \bibinfo{person}{Carlos Riquelme}.} \bibinfo{year}{[n.\,d.]}\natexlab{}.
\newblock \bibinfo{title}{StableLM 3B 4E1T}.
\newblock
\newblock
\urldef\tempurl%
\url{[https://huggingface.co/stabilityai/stablelm-3b-4e1t](https://huggingface.co/stabilityai/stablelm-3b-4e1t)}
\showURL{%
\tempurl}


\bibitem[Trinh and Le(2018)]%
        {trinh2018simple}
\bibfield{author}{\bibinfo{person}{Trieu~H Trinh} {and} \bibinfo{person}{Quoc~V Le}.} \bibinfo{year}{2018}\natexlab{}.
\newblock \showarticletitle{A simple method for commonsense reasoning}.
\newblock \bibinfo{journal}{\emph{arXiv preprint arXiv:1806.02847}} (\bibinfo{year}{2018}).
\newblock


\bibitem[Trufinescu(2024)]%
        {Trufinescu2024}
\bibfield{author}{\bibinfo{person}{Adina Trufinescu}.} \bibinfo{year}{2024}\natexlab{}.
\newblock \bibinfo{title}{Discover the New Multi-Lingual High-Quality Phi-3.5 SLMs}.
\newblock \bibinfo{howpublished}{\url{https://techcommunity.microsoft.com/t5/ai-azure-ai-services-blog/discover-the-new-multi-lingual-high-quality-phi-3-5-slms/ba-p/4225280}}.
\newblock


\bibitem[Ulmer et~al\mbox{.}(2024)]%
        {ulmer2024calibrating}
\bibfield{author}{\bibinfo{person}{Dennis Ulmer}, \bibinfo{person}{Martin Gubri}, \bibinfo{person}{Hwaran Lee}, \bibinfo{person}{Sangdoo Yun}, {and} \bibinfo{person}{Seong~Joon Oh}.} \bibinfo{year}{2024}\natexlab{}.
\newblock \showarticletitle{Calibrating Large Language Models Using Their Generations Only}.
\newblock \bibinfo{journal}{\emph{arXiv preprint arXiv:2403.05973}} (\bibinfo{year}{2024}).
\newblock


\bibitem[Van Der~Linden(2022)]%
        {van2022misinformation}
\bibfield{author}{\bibinfo{person}{Sander Van Der~Linden}.} \bibinfo{year}{2022}\natexlab{}.
\newblock \showarticletitle{Misinformation: susceptibility, spread, and interventions to immunize the public}.
\newblock \bibinfo{journal}{\emph{Nature medicine}} \bibinfo{volume}{28}, \bibinfo{number}{3} (\bibinfo{year}{2022}), \bibinfo{pages}{460--467}.
\newblock


\bibitem[Van~Nguyen et~al\mbox{.}(2024)]%
        {van2024survey}
\bibfield{author}{\bibinfo{person}{Chien Van~Nguyen}, \bibinfo{person}{Xuan Shen}, \bibinfo{person}{Ryan Aponte}, \bibinfo{person}{Yu Xia}, \bibinfo{person}{Samyadeep Basu}, \bibinfo{person}{Zhengmian Hu}, \bibinfo{person}{Jian Chen}, \bibinfo{person}{Mihir Parmar}, \bibinfo{person}{Sasidhar Kunapuli}, \bibinfo{person}{Joe Barrow}, {et~al\mbox{.}}} \bibinfo{year}{2024}\natexlab{}.
\newblock \showarticletitle{A Survey of Small Language Models}.
\newblock \bibinfo{journal}{\emph{arXiv preprint arXiv:2410.20011}} (\bibinfo{year}{2024}).
\newblock


\bibitem[Vaswani(2017)]%
        {vaswani2017attention}
\bibfield{author}{\bibinfo{person}{A Vaswani}.} \bibinfo{year}{2017}\natexlab{}.
\newblock \showarticletitle{Attention is all you need}.
\newblock \bibinfo{journal}{\emph{Advances in Neural Information Processing Systems}} (\bibinfo{year}{2017}).
\newblock


\bibitem[Veksler(2023)]%
        {veksler2023test}
\bibfield{author}{\bibinfo{person}{Olga Veksler}.} \bibinfo{year}{2023}\natexlab{}.
\newblock \showarticletitle{Test time adaptation with regularized loss for weakly supervised salient object detection}. In \bibinfo{booktitle}{\emph{Proceedings of the IEEE/CVF Conference on Computer Vision and Pattern Recognition}}. \bibinfo{pages}{7360--7369}.
\newblock


\bibitem[Voigt and Von~dem Bussche(2017)]%
        {voigt2017eu}
\bibfield{author}{\bibinfo{person}{Paul Voigt} {and} \bibinfo{person}{Axel Von~dem Bussche}.} \bibinfo{year}{2017}\natexlab{}.
\newblock \showarticletitle{The eu general data protection regulation (gdpr)}.
\newblock \bibinfo{journal}{\emph{A Practical Guide, 1st Ed., Cham: Springer International Publishing}} \bibinfo{volume}{10}, \bibinfo{number}{3152676} (\bibinfo{year}{2017}), \bibinfo{pages}{10--5555}.
\newblock


\bibitem[Wan et~al\mbox{.}(2023)]%
        {wan2023multi}
\bibfield{author}{\bibinfo{person}{Yuxian Wan}, \bibinfo{person}{Wenlin Zhang}, {and} \bibinfo{person}{Zhen Li}.} \bibinfo{year}{2023}\natexlab{}.
\newblock \showarticletitle{Multi-Task Feature Self-Distillation for Semi-Supervised Machine Translation}. In \bibinfo{booktitle}{\emph{International Conference on Neural Information Processing}}. Springer, \bibinfo{pages}{238--254}.
\newblock


\bibitem[Wang et~al\mbox{.}(2018)]%
        {wang2018glue}
\bibfield{author}{\bibinfo{person}{Alex Wang}, \bibinfo{person}{Amanpreet Singh}, \bibinfo{person}{Julian Michael}, \bibinfo{person}{Felix Hill}, \bibinfo{person}{Omer Levy}, {and} \bibinfo{person}{Samuel Bowman}.} \bibinfo{year}{2018}\natexlab{}.
\newblock \showarticletitle{GLUE: A Multi-Task Benchmark and Analysis Platform for Natural Language Understanding}. In \bibinfo{booktitle}{\emph{Proceedings of the 2018 EMNLP Workshop BlackboxNLP: Analyzing and Interpreting Neural Networks for NLP}}. \bibinfo{pages}{353--355}.
\newblock


\bibitem[Wang et~al\mbox{.}(2023a)]%
        {WangCPXKZXXDSTA23}
\bibfield{author}{\bibinfo{person}{Boxin Wang}, \bibinfo{person}{Weixin Chen}, \bibinfo{person}{Hengzhi Pei}, \bibinfo{person}{Chulin Xie}, \bibinfo{person}{Mintong Kang}, \bibinfo{person}{Chenhui Zhang}, \bibinfo{person}{Chejian Xu}, \bibinfo{person}{Zidi Xiong}, \bibinfo{person}{Ritik Dutta}, \bibinfo{person}{Rylan Schaeffer}, \bibinfo{person}{Sang~T. Truong}, \bibinfo{person}{Simran Arora}, \bibinfo{person}{Mantas Mazeika}, \bibinfo{person}{Dan Hendrycks}, \bibinfo{person}{Zinan Lin}, \bibinfo{person}{Yu Cheng}, \bibinfo{person}{Sanmi Koyejo}, \bibinfo{person}{Dawn Song}, {and} \bibinfo{person}{Bo Li}.} \bibinfo{year}{2023}\natexlab{a}.
\newblock \showarticletitle{DecodingTrust: {A} Comprehensive Assessment of Trustworthiness in {GPT} Models}. In \bibinfo{booktitle}{\emph{Proceedings of the Annual Conference on Neural Information Processing Systems}}.
\newblock


\bibitem[Wang et~al\mbox{.}(2021b)]%
        {wang2021adversarial}
\bibfield{author}{\bibinfo{person}{Boxin Wang}, \bibinfo{person}{Chejian Xu}, \bibinfo{person}{Shuohang Wang}, \bibinfo{person}{Zhe Gan}, \bibinfo{person}{Yu Cheng}, \bibinfo{person}{Jianfeng Gao}, \bibinfo{person}{Ahmed~Hassan Awadallah}, {and} \bibinfo{person}{Bo Li}.} \bibinfo{year}{2021}\natexlab{b}.
\newblock \showarticletitle{Adversarial glue: A multi-task benchmark for robustness evaluation of language models}.
\newblock \bibinfo{journal}{\emph{arXiv preprint arXiv:2111.02840}} (\bibinfo{year}{2021}).
\newblock


\bibitem[Wang et~al\mbox{.}(2024a)]%
        {wang2024infuserki}
\bibfield{author}{\bibinfo{person}{Fali Wang}, \bibinfo{person}{Runxue Bao}, \bibinfo{person}{Suhang Wang}, \bibinfo{person}{Wenchao Yu}, \bibinfo{person}{Yanchi Liu}, \bibinfo{person}{Wei Cheng}, {and} \bibinfo{person}{Haifeng Chen}.} \bibinfo{year}{2024}\natexlab{a}.
\newblock \showarticletitle{InfuserKI: Enhancing Large Language Models with Knowledge Graphs via Infuser-Guided Knowledge Integration}.
\newblock \bibinfo{journal}{\emph{arXiv preprint arXiv:2402.11441}} (\bibinfo{year}{2024}).
\newblock


\bibitem[Wang et~al\mbox{.}(2021a)]%
        {wang2021macrobert}
\bibfield{author}{\bibinfo{person}{Fali Wang}, \bibinfo{person}{Zheng Lin}, \bibinfo{person}{Zhengxiao Liu}, \bibinfo{person}{Mingyu Zheng}, \bibinfo{person}{Lei Wang}, {and} \bibinfo{person}{Daren Zha}.} \bibinfo{year}{2021}\natexlab{a}.
\newblock \showarticletitle{Macrobert: Maximizing certified region of bert to adversarial word substitutions}. In \bibinfo{booktitle}{\emph{Database Systems for Advanced Applications: 26th International Conference, DASFAA 2021, Taipei, Taiwan, April 11--14, 2021, Proceedings, Part II 26}}. Springer, \bibinfo{pages}{253--261}.
\newblock


\bibitem[Wang et~al\mbox{.}(2023b)]%
        {openllms23}
\bibfield{author}{\bibinfo{person}{Guan Wang}, \bibinfo{person}{Sijie Cheng}, \bibinfo{person}{Qiying Yu}, {and} \bibinfo{person}{Changling Liu}.} \bibinfo{year}{2023}\natexlab{b}.
\newblock \bibinfo{booktitle}{\emph{OpenLLMs: Less is More for Open-source Models}}.
\newblock
\urldef\tempurl%
\url{https://doi.org/10.5281/zenodo.8105775}
\showDOI{\tempurl}


\bibitem[Wang et~al\mbox{.}(2024d)]%
        {wang2024openchat}
\bibfield{author}{\bibinfo{person}{Guan Wang}, \bibinfo{person}{Sijie Cheng}, \bibinfo{person}{Xianyuan Zhan}, \bibinfo{person}{Xiangang Li}, \bibinfo{person}{Sen Song}, {and} \bibinfo{person}{Yang Liu}.} \bibinfo{year}{2024}\natexlab{d}.
\newblock \showarticletitle{OpenChat: Advancing Open-source Language Models with Mixed-Quality Data}. In \bibinfo{booktitle}{\emph{The Twelfth International Conference on Learning Representations}}.
\newblock
\urldef\tempurl%
\url{https://openreview.net/forum?id=AOJyfhWYHf}
\showURL{%
\tempurl}


\bibitem[Wang et~al\mbox{.}(2023e)]%
        {wang2023bitnet}
\bibfield{author}{\bibinfo{person}{Hongyu Wang}, \bibinfo{person}{Shuming Ma}, \bibinfo{person}{Li Dong}, \bibinfo{person}{Shaohan Huang}, \bibinfo{person}{Huaijie Wang}, \bibinfo{person}{Lingxiao Ma}, \bibinfo{person}{Fan Yang}, \bibinfo{person}{Ruiping Wang}, \bibinfo{person}{Yi Wu}, {and} \bibinfo{person}{Furu Wei}.} \bibinfo{year}{2023}\natexlab{e}.
\newblock \showarticletitle{Bitnet: Scaling 1-bit transformers for large language models}.
\newblock \bibinfo{journal}{\emph{arXiv preprint arXiv:2310.11453}} (\bibinfo{year}{2023}).
\newblock


\bibitem[Wang et~al\mbox{.}({[n.\,d.]})]%
        {wang2023robustness}
\bibfield{author}{\bibinfo{person}{Jindong Wang}, \bibinfo{person}{HU Xixu}, \bibinfo{person}{Wenxin Hou}, \bibinfo{person}{Hao Chen}, \bibinfo{person}{Runkai Zheng}, \bibinfo{person}{Yidong Wang}, \bibinfo{person}{Linyi Yang}, \bibinfo{person}{Wei Ye}, \bibinfo{person}{Haojun Huang}, \bibinfo{person}{Xiubo Geng}, {et~al\mbox{.}}} \bibinfo{year}{[n.\,d.]}\natexlab{}.
\newblock \showarticletitle{On the Robustness of ChatGPT: An Adversarial and Out-of-distribution Perspective}. In \bibinfo{booktitle}{\emph{ICLR 2023 Workshop on Trustworthy and Reliable Large-Scale Machine Learning Models}}.
\newblock


\bibitem[Wang et~al\mbox{.}(2024h)]%
        {wang2023improving}
\bibfield{author}{\bibinfo{person}{Liang Wang}, \bibinfo{person}{Nan Yang}, \bibinfo{person}{Xiaolong Huang}, \bibinfo{person}{Linjun Yang}, \bibinfo{person}{Rangan Majumder}, {and} \bibinfo{person}{Furu Wei}.} \bibinfo{year}{2024}\natexlab{h}.
\newblock \showarticletitle{Improving Text Embeddings with Large Language Models}. In \bibinfo{booktitle}{\emph{Proceedings of the 62nd Annual Meeting of the Association for Computational Linguistics (Volume 1: Long Papers)}}. \bibinfo{publisher}{Association for Computational Linguistics}, \bibinfo{pages}{11897--11916}.
\newblock
\urldef\tempurl%
\url{https://doi.org/10.18653/v1/2024.acl-long.642}
\showDOI{\tempurl}


\bibitem[Wang et~al\mbox{.}(2023f)]%
        {wang2023scott}
\bibfield{author}{\bibinfo{person}{Peifeng Wang}, \bibinfo{person}{Zhengyang Wang}, \bibinfo{person}{Zheng Li}, \bibinfo{person}{Yifan Gao}, \bibinfo{person}{Bing Yin}, {and} \bibinfo{person}{Xiang Ren}.} \bibinfo{year}{2023}\natexlab{f}.
\newblock \showarticletitle{SCOTT: Self-Consistent Chain-of-Thought Distillation}. In \bibinfo{booktitle}{\emph{Proceedings of the 61st Annual Meeting of the Association for Computational Linguistics (Volume 1: Long Papers)}}. \bibinfo{pages}{5546--5558}.
\newblock


\bibitem[Wang et~al\mbox{.}(2024c)]%
        {wang2024model}
\bibfield{author}{\bibinfo{person}{Wenxiao Wang}, \bibinfo{person}{Wei Chen}, \bibinfo{person}{Yicong Luo}, \bibinfo{person}{Yongliu Long}, \bibinfo{person}{Zhengkai Lin}, \bibinfo{person}{Liye Zhang}, \bibinfo{person}{Binbin Lin}, \bibinfo{person}{Deng Cai}, {and} \bibinfo{person}{Xiaofei He}.} \bibinfo{year}{2024}\natexlab{c}.
\newblock \showarticletitle{Model compression and efficient inference for large language models: A survey}.
\newblock \bibinfo{journal}{\emph{arXiv preprint arXiv:2402.09748}} (\bibinfo{year}{2024}).
\newblock


\bibitem[Wang et~al\mbox{.}(2020)]%
        {wang2020minilmdeepselfattentiondistillation}
\bibfield{author}{\bibinfo{person}{Wenhui Wang}, \bibinfo{person}{Furu Wei}, \bibinfo{person}{Li Dong}, \bibinfo{person}{Hangbo Bao}, \bibinfo{person}{Nan Yang}, {and} \bibinfo{person}{Ming Zhou}.} \bibinfo{year}{2020}\natexlab{}.
\newblock \bibinfo{title}{MiniLM: Deep Self-Attention Distillation for Task-Agnostic Compression of Pre-Trained Transformers}.
\newblock
\newblock
\showeprint[arxiv]{2002.10957}~[cs.CL]
\urldef\tempurl%
\url{https://arxiv.org/abs/2002.10957}
\showURL{%
\tempurl}


\bibitem[Wang et~al\mbox{.}(2024e)]%
        {wang2024scibench}
\bibfield{author}{\bibinfo{person}{Xiaoxuan Wang}, \bibinfo{person}{Ziniu Hu}, \bibinfo{person}{Pan Lu}, \bibinfo{person}{Yanqiao Zhu}, \bibinfo{person}{Jieyu Zhang}, \bibinfo{person}{Satyen Subramaniam}, \bibinfo{person}{Arjun~R Loomba}, \bibinfo{person}{Shichang Zhang}, \bibinfo{person}{Yizhou Sun}, {and} \bibinfo{person}{Wei Wang}.} \bibinfo{year}{2024}\natexlab{e}.
\newblock \showarticletitle{SciBench: Evaluating College-Level Scientific Problem-Solving Abilities of Large Language Models}. In \bibinfo{booktitle}{\emph{Forty-first International Conference on Machine Learning}}.
\newblock
\urldef\tempurl%
\url{https://openreview.net/forum?id=bq1JEgioLr}
\showURL{%
\tempurl}


\bibitem[Wang et~al\mbox{.}(2024f)]%
        {wang-etal-2024-answer}
\bibfield{author}{\bibinfo{person}{Yuxia Wang}, \bibinfo{person}{Haonan Li}, \bibinfo{person}{Xudong Han}, \bibinfo{person}{Preslav Nakov}, {and} \bibinfo{person}{Timothy Baldwin}.} \bibinfo{year}{2024}\natexlab{f}.
\newblock \showarticletitle{Do-Not-Answer: Evaluating Safeguards in {LLM}s}. In \bibinfo{booktitle}{\emph{Findings of the Association for Computational Linguistics: EACL 2024}}. \bibinfo{publisher}{Association for Computational Linguistics}, \bibinfo{pages}{896--911}.
\newblock
\urldef\tempurl%
\url{https://aclanthology.org/2024.findings-eacl.61}
\showURL{%
\tempurl}


\bibitem[Wang et~al\mbox{.}(2023c)]%
        {wang2023self}
\bibfield{author}{\bibinfo{person}{Yile Wang}, \bibinfo{person}{Peng Li}, \bibinfo{person}{Maosong Sun}, {and} \bibinfo{person}{Yang Liu}.} \bibinfo{year}{2023}\natexlab{c}.
\newblock \showarticletitle{Self-Knowledge Guided Retrieval Augmentation for Large Language Models}. In \bibinfo{booktitle}{\emph{Findings of the Association for Computational Linguistics: EMNLP 2023}}. \bibinfo{pages}{10303--10315}.
\newblock


\bibitem[Wang et~al\mbox{.}(2023d)]%
        {wang2023augmenting}
\bibfield{author}{\bibinfo{person}{Yubo Wang}, \bibinfo{person}{Xueguang Ma}, {and} \bibinfo{person}{Wenhu Chen}.} \bibinfo{year}{2023}\natexlab{d}.
\newblock \showarticletitle{Augmenting black-box llms with medical textbooks for clinical question answering}.
\newblock \bibinfo{journal}{\emph{arXiv preprint arXiv:2309.02233}} (\bibinfo{year}{2023}).
\newblock


\bibitem[Wang et~al\mbox{.}(2022a)]%
        {wang-etal-2022-super}
\bibfield{author}{\bibinfo{person}{Yizhong Wang}, \bibinfo{person}{Swaroop Mishra}, \bibinfo{person}{Pegah Alipoormolabashi}, \bibinfo{person}{Yeganeh Kordi}, \bibinfo{person}{Amirreza Mirzaei}, \bibinfo{person}{Atharva Naik}, \bibinfo{person}{Arjun Ashok}, \bibinfo{person}{Arut~Selvan Dhanasekaran}, \bibinfo{person}{Anjana Arunkumar}, \bibinfo{person}{David Stap}, \bibinfo{person}{Eshaan Pathak}, \bibinfo{person}{Giannis Karamanolakis}, \bibinfo{person}{Haizhi Lai}, \bibinfo{person}{Ishan Purohit}, \bibinfo{person}{Ishani Mondal}, \bibinfo{person}{Jacob Anderson}, \bibinfo{person}{Kirby Kuznia}, \bibinfo{person}{Krima Doshi}, \bibinfo{person}{Kuntal~Kumar Pal}, \bibinfo{person}{Maitreya Patel}, \bibinfo{person}{Mehrad Moradshahi}, \bibinfo{person}{Mihir Parmar}, \bibinfo{person}{Mirali Purohit}, \bibinfo{person}{Neeraj Varshney}, \bibinfo{person}{Phani~Rohitha Kaza}, \bibinfo{person}{Pulkit Verma}, \bibinfo{person}{Ravsehaj~Singh Puri}, \bibinfo{person}{Rushang Karia}, \bibinfo{person}{Savan Doshi},
  \bibinfo{person}{Shailaja~Keyur Sampat}, \bibinfo{person}{Siddhartha Mishra}, \bibinfo{person}{Sujan Reddy~A}, \bibinfo{person}{Sumanta Patro}, \bibinfo{person}{Tanay Dixit}, {and} \bibinfo{person}{Xudong Shen}.} \bibinfo{year}{2022}\natexlab{a}.
\newblock \showarticletitle{Super-{N}atural{I}nstructions: Generalization via Declarative Instructions on 1600+ {NLP} Tasks}. In \bibinfo{booktitle}{\emph{Proceedings of the 2022 Conference on Empirical Methods in Natural Language Processing}}. \bibinfo{publisher}{Association for Computational Linguistics}, \bibinfo{pages}{5085--5109}.
\newblock
\urldef\tempurl%
\url{https://doi.org/10.18653/v1/2022.emnlp-main.340}
\showDOI{\tempurl}


\bibitem[Wang et~al\mbox{.}(2022b)]%
        {10.1145/3534678.3539073}
\bibfield{author}{\bibinfo{person}{Yuyan Wang}, \bibinfo{person}{Mohit Sharma}, \bibinfo{person}{Can Xu}, \bibinfo{person}{Sriraj Badam}, \bibinfo{person}{Qian Sun}, \bibinfo{person}{Lee Richardson}, \bibinfo{person}{Lisa Chung}, \bibinfo{person}{Ed~H. Chi}, {and} \bibinfo{person}{Minmin Chen}.} \bibinfo{year}{2022}\natexlab{b}.
\newblock \showarticletitle{Surrogate for Long-Term User Experience in Recommender Systems}. In \bibinfo{booktitle}{\emph{Proceedings of the 28th ACM SIGKDD Conference on Knowledge Discovery and Data Mining}}. \bibinfo{publisher}{Association for Computing Machinery}, \bibinfo{pages}{4100–4109}.
\newblock
\urldef\tempurl%
\url{https://doi.org/10.1145/3534678.3539073}
\showDOI{\tempurl}


\bibitem[Wang et~al\mbox{.}(2024g)]%
        {wang2024can}
\bibfield{author}{\bibinfo{person}{Yuling Wang}, \bibinfo{person}{Changxin Tian}, \bibinfo{person}{Binbin Hu}, \bibinfo{person}{Yanhua Yu}, \bibinfo{person}{Ziqi Liu}, \bibinfo{person}{Zhiqiang Zhang}, \bibinfo{person}{Jun Zhou}, \bibinfo{person}{Liang Pang}, {and} \bibinfo{person}{Xiao Wang}.} \bibinfo{year}{2024}\natexlab{g}.
\newblock \showarticletitle{Can Small Language Models be Good Reasoners for Sequential Recommendation?}. In \bibinfo{booktitle}{\emph{Proceedings of the ACM on Web Conference 2024}}. \bibinfo{pages}{3876--3887}.
\newblock


\bibitem[Wang and Zhao(2024)]%
        {wang2024rupbench}
\bibfield{author}{\bibinfo{person}{Yuqing Wang} {and} \bibinfo{person}{Yun Zhao}.} \bibinfo{year}{2024}\natexlab{}.
\newblock \showarticletitle{RUPBench: Benchmarking Reasoning Under Perturbations for Robustness Evaluation in Large Language Models}.
\newblock \bibinfo{journal}{\emph{arXiv preprint arXiv:2406.11020}} (\bibinfo{year}{2024}).
\newblock


\bibitem[Wang et~al\mbox{.}(2024b)]%
        {wang-etal-2024-unlocking}
\bibfield{author}{\bibinfo{person}{Zhepeng Wang}, \bibinfo{person}{Runxue Bao}, \bibinfo{person}{Yawen Wu}, \bibinfo{person}{Jackson Taylor}, \bibinfo{person}{Cao Xiao}, \bibinfo{person}{Feng Zheng}, \bibinfo{person}{Weiwen Jiang}, \bibinfo{person}{Shangqian Gao}, {and} \bibinfo{person}{Yanfu Zhang}.} \bibinfo{year}{2024}\natexlab{b}.
\newblock \showarticletitle{Unlocking Memorization in Large Language Models with Dynamic Soft Prompting}. In \bibinfo{booktitle}{\emph{Proceedings of the 2024 Conference on Empirical Methods in Natural Language Processing}}, \bibfield{editor}{\bibinfo{person}{Yaser Al-Onaizan}, \bibinfo{person}{Mohit Bansal}, {and} \bibinfo{person}{Yun-Nung Chen}} (Eds.). \bibinfo{publisher}{Association for Computational Linguistics}, \bibinfo{address}{Miami, Florida, USA}, \bibinfo{pages}{9782--9796}.
\newblock
\urldef\tempurl%
\url{https://doi.org/10.18653/v1/2024.emnlp-main.546}
\showDOI{\tempurl}


\bibitem[Wei et~al\mbox{.}(2022)]%
        {wei2022finetuned}
\bibfield{author}{\bibinfo{person}{Jason Wei}, \bibinfo{person}{Maarten Bosma}, \bibinfo{person}{Vincent Zhao}, \bibinfo{person}{Kelvin Guu}, \bibinfo{person}{Adams~Wei Yu}, \bibinfo{person}{Brian Lester}, \bibinfo{person}{Nan Du}, \bibinfo{person}{Andrew~M. Dai}, {and} \bibinfo{person}{Quoc~V Le}.} \bibinfo{year}{2022}\natexlab{}.
\newblock \showarticletitle{Finetuned Language Models are Zero-Shot Learners}. In \bibinfo{booktitle}{\emph{International Conference on Learning Representations}}.
\newblock
\urldef\tempurl%
\url{https://openreview.net/forum?id=gEZrGCozdqR}
\showURL{%
\tempurl}


\bibitem[Wei et~al\mbox{.}(2023)]%
        {wei2023magicoder}
\bibfield{author}{\bibinfo{person}{Yuxiang Wei}, \bibinfo{person}{Zhe Wang}, \bibinfo{person}{Jiawei Liu}, \bibinfo{person}{Yifeng Ding}, {and} \bibinfo{person}{Lingming Zhang}.} \bibinfo{year}{2023}\natexlab{}.
\newblock \showarticletitle{Magicoder: Source code is all you need}.
\newblock \bibinfo{journal}{\emph{arXiv preprint arXiv:2312.02120}} (\bibinfo{year}{2023}).
\newblock


\bibitem[Welbl et~al\mbox{.}(2021)]%
        {welbl-etal-2021-challenges-detoxifying}
\bibfield{author}{\bibinfo{person}{Johannes Welbl}, \bibinfo{person}{Amelia Glaese}, \bibinfo{person}{Jonathan Uesato}, \bibinfo{person}{Sumanth Dathathri}, \bibinfo{person}{John Mellor}, \bibinfo{person}{Lisa~Anne Hendricks}, \bibinfo{person}{Kirsty Anderson}, \bibinfo{person}{Pushmeet Kohli}, \bibinfo{person}{Ben Coppin}, {and} \bibinfo{person}{Po-Sen Huang}.} \bibinfo{year}{2021}\natexlab{}.
\newblock \showarticletitle{Challenges in Detoxifying Language Models}. In \bibinfo{booktitle}{\emph{Findings of the Association for Computational Linguistics: EMNLP 2021}}. \bibinfo{publisher}{Association for Computational Linguistics}, \bibinfo{pages}{2447--2469}.
\newblock
\urldef\tempurl%
\url{https://doi.org/10.18653/v1/2021.findings-emnlp.210}
\showDOI{\tempurl}


\bibitem[Welbl et~al\mbox{.}(2017)]%
        {welbl2017crowdsourcing}
\bibfield{author}{\bibinfo{person}{Johannes Welbl}, \bibinfo{person}{Nelson~F Liu}, {and} \bibinfo{person}{Matt Gardner}.} \bibinfo{year}{2017}\natexlab{}.
\newblock \showarticletitle{Crowdsourcing Multiple Choice Science Questions}. In \bibinfo{booktitle}{\emph{Proceedings of the 3rd Workshop on Noisy User-generated Text}}. \bibinfo{pages}{94--106}.
\newblock


\bibitem[Wen et~al\mbox{.}(2024)]%
        {wen2024autodroidllmpoweredtaskautomation}
\bibfield{author}{\bibinfo{person}{Hao Wen}, \bibinfo{person}{Yuanchun Li}, \bibinfo{person}{Guohong Liu}, \bibinfo{person}{Shanhui Zhao}, \bibinfo{person}{Tao Yu}, \bibinfo{person}{Toby Jia-Jun Li}, \bibinfo{person}{Shiqi Jiang}, \bibinfo{person}{Yunhao Liu}, \bibinfo{person}{Yaqin Zhang}, {and} \bibinfo{person}{Yunxin Liu}.} \bibinfo{year}{2024}\natexlab{}.
\newblock \bibinfo{title}{AutoDroid: LLM-powered Task Automation in Android}.
\newblock
\newblock
\showeprint[arxiv]{2308.15272}~[cs.AI]
\urldef\tempurl%
\url{https://arxiv.org/abs/2308.15272}
\showURL{%
\tempurl}


\bibitem[Wettig et~al\mbox{.}(2024)]%
        {wettig2024qurating}
\bibfield{author}{\bibinfo{person}{Alexander Wettig}, \bibinfo{person}{Aatmik Gupta}, \bibinfo{person}{Saumya Malik}, {and} \bibinfo{person}{Danqi Chen}.} \bibinfo{year}{2024}\natexlab{}.
\newblock \showarticletitle{QuRating: Selecting High-Quality Data for Training Language Models}. In \bibinfo{booktitle}{\emph{Forty-first International Conference on Machine Learning}}.
\newblock
\urldef\tempurl%
\url{https://openreview.net/forum?id=GLGYYqPwjy}
\showURL{%
\tempurl}


\bibitem[Williams et~al\mbox{.}(2018)]%
        {williams-etal-2018-broad}
\bibfield{author}{\bibinfo{person}{Adina Williams}, \bibinfo{person}{Nikita Nangia}, {and} \bibinfo{person}{Samuel Bowman}.} \bibinfo{year}{2018}\natexlab{}.
\newblock \showarticletitle{A Broad-Coverage Challenge Corpus for Sentence Understanding through Inference}. In \bibinfo{booktitle}{\emph{Proceedings of the 2018 Conference of the North {A}merican Chapter of the Association for Computational Linguistics: Human Language Technologies, Volume 1 (Long Papers)}}. \bibinfo{publisher}{Association for Computational Linguistics}, \bibinfo{address}{New Orleans, Louisiana}, \bibinfo{pages}{1112--1122}.
\newblock
\urldef\tempurl%
\url{https://doi.org/10.18653/v1/N18-1101}
\showDOI{\tempurl}


\bibitem[Wu et~al\mbox{.}(2021)]%
        {wu2021empowering}
\bibfield{author}{\bibinfo{person}{Chuhan Wu}, \bibinfo{person}{Fangzhao Wu}, \bibinfo{person}{Tao Qi}, {and} \bibinfo{person}{Yongfeng Huang}.} \bibinfo{year}{2021}\natexlab{}.
\newblock \showarticletitle{Empowering news recommendation with pre-trained language models}. In \bibinfo{booktitle}{\emph{Proceedings of the 44th international ACM SIGIR conference on research and development in information retrieval}}. \bibinfo{pages}{1652--1656}.
\newblock


\bibitem[Wu et~al\mbox{.}(2024c)]%
        {wu-etal-2024-lamini}
\bibfield{author}{\bibinfo{person}{Minghao Wu}, \bibinfo{person}{Abdul Waheed}, \bibinfo{person}{Chiyu Zhang}, \bibinfo{person}{Muhammad Abdul-Mageed}, {and} \bibinfo{person}{Alham~Fikri Aji}.} \bibinfo{year}{2024}\natexlab{c}.
\newblock \showarticletitle{{L}a{M}ini-{LM}: A Diverse Herd of Distilled Models from Large-Scale Instructions}. In \bibinfo{booktitle}{\emph{Proceedings of the 18th Conference of the European Chapter of the Association for Computational Linguistics (Volume 1: Long Papers)}}, \bibfield{editor}{\bibinfo{person}{Yvette Graham} {and} \bibinfo{person}{Matthew Purver}} (Eds.). \bibinfo{publisher}{Association for Computational Linguistics}, \bibinfo{address}{St. Julian{'}s, Malta}, \bibinfo{pages}{944--964}.
\newblock
\urldef\tempurl%
\url{https://aclanthology.org/2024.eacl-long.57}
\showURL{%
\tempurl}


\bibitem[Wu et~al\mbox{.}(2024b)]%
        {wu-etal-2024-weight}
\bibfield{author}{\bibinfo{person}{Taiqiang Wu}, \bibinfo{person}{Cheng Hou}, \bibinfo{person}{Shanshan Lao}, \bibinfo{person}{Jiayi Li}, \bibinfo{person}{Ngai Wong}, \bibinfo{person}{Zhe Zhao}, {and} \bibinfo{person}{Yujiu Yang}.} \bibinfo{year}{2024}\natexlab{b}.
\newblock \showarticletitle{Weight-Inherited Distillation for Task-Agnostic {BERT} Compression}. In \bibinfo{booktitle}{\emph{Findings of the Association for Computational Linguistics: NAACL 2024}}, \bibfield{editor}{\bibinfo{person}{Kevin Duh}, \bibinfo{person}{Helena Gomez}, {and} \bibinfo{person}{Steven Bethard}} (Eds.). \bibinfo{publisher}{Association for Computational Linguistics}, \bibinfo{address}{Mexico City, Mexico}, \bibinfo{pages}{13--28}.
\newblock
\urldef\tempurl%
\url{https://doi.org/10.18653/v1/2024.findings-naacl.2}
\showDOI{\tempurl}


\bibitem[Wu et~al\mbox{.}(2024d)]%
        {wu2024could}
\bibfield{author}{\bibinfo{person}{Xuansheng Wu}, \bibinfo{person}{Huachi Zhou}, \bibinfo{person}{Yucheng Shi}, \bibinfo{person}{Wenlin Yao}, \bibinfo{person}{Xiao Huang}, {and} \bibinfo{person}{Ninghao Liu}.} \bibinfo{year}{2024}\natexlab{d}.
\newblock \showarticletitle{Could Small Language Models Serve as Recommenders? Towards Data-centric Cold-start Recommendation}. In \bibinfo{booktitle}{\emph{Proceedings of the ACM on Web Conference 2024}}. \bibinfo{pages}{3566--3575}.
\newblock


\bibitem[Wu et~al\mbox{.}(2024a)]%
        {wu2024divide}
\bibfield{author}{\bibinfo{person}{Zhuofeng Wu}, \bibinfo{person}{He Bai}, \bibinfo{person}{Aonan Zhang}, \bibinfo{person}{Jiatao Gu}, \bibinfo{person}{VG Vydiswaran}, \bibinfo{person}{Navdeep Jaitly}, {and} \bibinfo{person}{Yizhe Zhang}.} \bibinfo{year}{2024}\natexlab{a}.
\newblock \showarticletitle{Divide-or-Conquer? Which Part Should You Distill Your LLM?}
\newblock \bibinfo{journal}{\emph{arXiv preprint arXiv:2402.15000}} (\bibinfo{year}{2024}).
\newblock


\bibitem[Xi et~al\mbox{.}(2024)]%
        {xi2024learning}
\bibfield{author}{\bibinfo{person}{Nuwa Xi}, \bibinfo{person}{Yuhan Chen}, \bibinfo{person}{Sendong Zhao}, \bibinfo{person}{Haochun Wang}, \bibinfo{person}{Bing Qin}, {and} \bibinfo{person}{Ting Liu}.} \bibinfo{year}{2024}\natexlab{}.
\newblock \showarticletitle{AS-ES Learning: Towards Efficient CoT Learning in Small Models}.
\newblock \bibinfo{journal}{\emph{arXiv preprint arXiv:2403.01969}} (\bibinfo{year}{2024}).
\newblock


\bibitem[Xia et~al\mbox{.}(2024)]%
        {xia2024sheared}
\bibfield{author}{\bibinfo{person}{Mengzhou Xia}, \bibinfo{person}{Tianyu Gao}, \bibinfo{person}{Zhiyuan Zeng}, {and} \bibinfo{person}{Danqi Chen}.} \bibinfo{year}{2024}\natexlab{}.
\newblock \showarticletitle{Sheared {LL}a{MA}: Accelerating Language Model Pre-training via Structured Pruning}. In \bibinfo{booktitle}{\emph{The Twelfth International Conference on Learning Representations}}.
\newblock
\urldef\tempurl%
\url{https://openreview.net/forum?id=09iOdaeOzp}
\showURL{%
\tempurl}


\bibitem[Xiao et~al\mbox{.}(2023)]%
        {xiao2023smoothquant}
\bibfield{author}{\bibinfo{person}{Guangxuan Xiao}, \bibinfo{person}{Ji Lin}, \bibinfo{person}{Mickael Seznec}, \bibinfo{person}{Hao Wu}, \bibinfo{person}{Julien Demouth}, {and} \bibinfo{person}{Song Han}.} \bibinfo{year}{2023}\natexlab{}.
\newblock \showarticletitle{Smoothquant: Accurate and efficient post-training quantization for large language models}. In \bibinfo{booktitle}{\emph{International Conference on Machine Learning}}. PMLR, \bibinfo{pages}{38087--38099}.
\newblock


\bibitem[Xie et~al\mbox{.}(2024a)]%
        {xie2024sorry}
\bibfield{author}{\bibinfo{person}{Tinghao Xie}, \bibinfo{person}{Xiangyu Qi}, \bibinfo{person}{Yi Zeng}, \bibinfo{person}{Yangsibo Huang}, \bibinfo{person}{Udari~Madhushani Sehwag}, \bibinfo{person}{Kaixuan Huang}, \bibinfo{person}{Luxi He}, \bibinfo{person}{Boyi Wei}, \bibinfo{person}{Dacheng Li}, \bibinfo{person}{Ying Sheng}, {et~al\mbox{.}}} \bibinfo{year}{2024}\natexlab{a}.
\newblock \showarticletitle{Sorry-bench: Systematically evaluating large language model safety refusal behaviors}.
\newblock \bibinfo{journal}{\emph{arXiv preprint arXiv:2406.14598}} (\bibinfo{year}{2024}).
\newblock


\bibitem[Xie et~al\mbox{.}(2023)]%
        {xie2023darwin}
\bibfield{author}{\bibinfo{person}{Tong Xie}, \bibinfo{person}{Yuwei Wan}, \bibinfo{person}{Wei Huang}, \bibinfo{person}{Zhenyu Yin}, \bibinfo{person}{Yixuan Liu}, \bibinfo{person}{Shaozhou Wang}, \bibinfo{person}{Qingyuan Linghu}, \bibinfo{person}{Chunyu Kit}, \bibinfo{person}{Clara Grazian}, \bibinfo{person}{Wenjie Zhang}, {et~al\mbox{.}}} \bibinfo{year}{2023}\natexlab{}.
\newblock \showarticletitle{Darwin series: Domain specific large language models for natural science}.
\newblock \bibinfo{journal}{\emph{arXiv preprint arXiv:2308.13565}} (\bibinfo{year}{2023}).
\newblock


\bibitem[Xie et~al\mbox{.}(2024c)]%
        {xie2024droidcall}
\bibfield{author}{\bibinfo{person}{Weikai Xie}, \bibinfo{person}{Li Zhang}, \bibinfo{person}{Shihe Wang}, \bibinfo{person}{Rongjie Yi}, {and} \bibinfo{person}{Mengwei Xu}.} \bibinfo{year}{2024}\natexlab{c}.
\newblock \showarticletitle{DroidCall: A Dataset for LLM-powered Android Intent Invocation}.
\newblock \bibinfo{journal}{\emph{arXiv preprint arXiv:2412.00402}} (\bibinfo{year}{2024}).
\newblock


\bibitem[Xie et~al\mbox{.}(2024b)]%
        {xie2024online}
\bibfield{author}{\bibinfo{person}{Xuan Xie}, \bibinfo{person}{Jiayang Song}, \bibinfo{person}{Zhehua Zhou}, \bibinfo{person}{Yuheng Huang}, \bibinfo{person}{Da Song}, {and} \bibinfo{person}{Lei Ma}.} \bibinfo{year}{2024}\natexlab{b}.
\newblock \showarticletitle{Online Safety Analysis for LLMs: a Benchmark, an Assessment, and a Path Forward}.
\newblock \bibinfo{journal}{\emph{arXiv preprint arXiv:2404.08517}} (\bibinfo{year}{2024}).
\newblock


\bibitem[Xin et~al\mbox{.}(2021)]%
        {xin-etal-2021-berxit}
\bibfield{author}{\bibinfo{person}{Ji Xin}, \bibinfo{person}{Raphael Tang}, \bibinfo{person}{Yaoliang Yu}, {and} \bibinfo{person}{Jimmy Lin}.} \bibinfo{year}{2021}\natexlab{}.
\newblock \showarticletitle{{BER}xi{T}: Early Exiting for {BERT} with Better Fine-Tuning and Extension to Regression}. In \bibinfo{booktitle}{\emph{Proceedings of the 16th Conference of the European Chapter of the Association for Computational Linguistics: Main Volume}}, \bibfield{editor}{\bibinfo{person}{Paola Merlo}, \bibinfo{person}{Jorg Tiedemann}, {and} \bibinfo{person}{Reut Tsarfaty}} (Eds.). \bibinfo{publisher}{Association for Computational Linguistics}, \bibinfo{address}{Online}, \bibinfo{pages}{91--104}.
\newblock
\urldef\tempurl%
\url{https://doi.org/10.18653/v1/2021.eacl-main.8}
\showDOI{\tempurl}


\bibitem[Xu et~al\mbox{.}(2023b)]%
        {xu2023wizardlm}
\bibfield{author}{\bibinfo{person}{Can Xu}, \bibinfo{person}{Qingfeng Sun}, \bibinfo{person}{Kai Zheng}, \bibinfo{person}{Xiubo Geng}, \bibinfo{person}{Pu Zhao}, \bibinfo{person}{Jiazhan Feng}, \bibinfo{person}{Chongyang Tao}, {and} \bibinfo{person}{Daxin Jiang}.} \bibinfo{year}{2023}\natexlab{b}.
\newblock \showarticletitle{Wizardlm: Empowering large language models to follow complex instructions}.
\newblock \bibinfo{journal}{\emph{arXiv preprint arXiv:2304.12244}} (\bibinfo{year}{2023}).
\newblock


\bibitem[Xu et~al\mbox{.}(2023c)]%
        {xu2023small}
\bibfield{author}{\bibinfo{person}{Canwen Xu}, \bibinfo{person}{Yichong Xu}, \bibinfo{person}{Shuohang Wang}, \bibinfo{person}{Yang Liu}, \bibinfo{person}{Chenguang Zhu}, {and} \bibinfo{person}{Julian McAuley}.} \bibinfo{year}{2023}\natexlab{c}.
\newblock \showarticletitle{Small models are valuable plug-ins for large language models}.
\newblock \bibinfo{journal}{\emph{arXiv preprint arXiv:2305.08848}} (\bibinfo{year}{2023}).
\newblock


\bibitem[Xu et~al\mbox{.}(2023d)]%
        {xu2023llmcadfastscalableondevice}
\bibfield{author}{\bibinfo{person}{Daliang Xu}, \bibinfo{person}{Wangsong Yin}, \bibinfo{person}{Xin Jin}, \bibinfo{person}{Ying Zhang}, \bibinfo{person}{Shiyun Wei}, \bibinfo{person}{Mengwei Xu}, {and} \bibinfo{person}{Xuanzhe Liu}.} \bibinfo{year}{2023}\natexlab{d}.
\newblock \bibinfo{title}{LLMCad: Fast and Scalable On-device Large Language Model Inference}.
\newblock
\newblock
\showeprint[arxiv]{2309.04255}~[cs.NI]
\urldef\tempurl%
\url{https://arxiv.org/abs/2309.04255}
\showURL{%
\tempurl}


\bibitem[Xu et~al\mbox{.}(2024d)]%
        {xu2024empowering}
\bibfield{author}{\bibinfo{person}{Daliang Xu}, \bibinfo{person}{Hao Zhang}, \bibinfo{person}{Liming Yang}, \bibinfo{person}{Ruiqi Liu}, \bibinfo{person}{Gang Huang}, \bibinfo{person}{Mengwei Xu}, {and} \bibinfo{person}{Xuanzhe Liu}.} \bibinfo{year}{2024}\natexlab{d}.
\newblock \showarticletitle{Empowering 1000 tokens/second on-device llm prefilling with mllm-npu}.
\newblock \bibinfo{journal}{\emph{arXiv preprint arXiv:2407.05858}} (\bibinfo{year}{2024}).
\newblock


\bibitem[Xu et~al\mbox{.}(2015)]%
        {xu-etal-2015-short}
\bibfield{author}{\bibinfo{person}{Jiaming Xu}, \bibinfo{person}{Peng Wang}, \bibinfo{person}{Guanhua Tian}, \bibinfo{person}{Bo Xu}, \bibinfo{person}{Jun Zhao}, \bibinfo{person}{Fangyuan Wang}, {and} \bibinfo{person}{Hongwei Hao}.} \bibinfo{year}{2015}\natexlab{}.
\newblock \showarticletitle{Short Text Clustering via Convolutional Neural Networks}. In \bibinfo{booktitle}{\emph{Proceedings of the 1st Workshop on Vector Space Modeling for Natural Language Processing}}, \bibfield{editor}{\bibinfo{person}{Phil Blunsom}, \bibinfo{person}{Shay Cohen}, \bibinfo{person}{Paramveer Dhillon}, {and} \bibinfo{person}{Percy Liang}} (Eds.). \bibinfo{publisher}{Association for Computational Linguistics}, \bibinfo{address}{Denver, Colorado}, \bibinfo{pages}{62--69}.
\newblock
\urldef\tempurl%
\url{https://doi.org/10.3115/v1/W15-1509}
\showDOI{\tempurl}


\bibitem[Xu et~al\mbox{.}(2024a)]%
        {xu2024large}
\bibfield{author}{\bibinfo{person}{Minrui Xu}, \bibinfo{person}{Niyato Dusit}, \bibinfo{person}{Jiawen Kang}, \bibinfo{person}{Zehui Xiong}, \bibinfo{person}{Shiwen Mao}, \bibinfo{person}{Zhu Han}, \bibinfo{person}{Dong~In Kim}, {and} \bibinfo{person}{Khaled~B Letaief}.} \bibinfo{year}{2024}\natexlab{a}.
\newblock \showarticletitle{When large language model agents meet 6G networks: Perception, grounding, and alignment}.
\newblock \bibinfo{journal}{\emph{arXiv preprint arXiv:2401.07764}} (\bibinfo{year}{2024}).
\newblock


\bibitem[Xu et~al\mbox{.}(2024c)]%
        {xu2024survey}
\bibfield{author}{\bibinfo{person}{Mengwei Xu}, \bibinfo{person}{Wangsong Yin}, \bibinfo{person}{Dongqi Cai}, \bibinfo{person}{Rongjie Yi}, \bibinfo{person}{Daliang Xu}, \bibinfo{person}{Qipeng Wang}, \bibinfo{person}{Bingyang Wu}, \bibinfo{person}{Yihao Zhao}, \bibinfo{person}{Chen Yang}, \bibinfo{person}{Shihe Wang}, {et~al\mbox{.}}} \bibinfo{year}{2024}\natexlab{c}.
\newblock \showarticletitle{A survey of resource-efficient llm and multimodal foundation models}.
\newblock \bibinfo{journal}{\emph{arXiv preprint arXiv:2401.08092}} (\bibinfo{year}{2024}).
\newblock


\bibitem[Xu et~al\mbox{.}(2023a)]%
        {xu2023understanding}
\bibfield{author}{\bibinfo{person}{Weijia Xu}, \bibinfo{person}{Sweta Agrawal}, \bibinfo{person}{Eleftheria Briakou}, \bibinfo{person}{Marianna Martindale}, {and} \bibinfo{person}{Marine Carpuat}.} \bibinfo{year}{2023}\natexlab{a}.
\newblock \showarticletitle{Understanding and Detecting Hallucinations in Neural Machine Translation via Model Introspection}.
\newblock \bibinfo{journal}{\emph{Transactions of the Association for Computational Linguistics}}  \bibinfo{volume}{11} (\bibinfo{year}{2023}), \bibinfo{pages}{546--564}.
\newblock


\bibitem[Xu et~al\mbox{.}(2024b)]%
        {xu2024onebit}
\bibfield{author}{\bibinfo{person}{Yuzhuang Xu}, \bibinfo{person}{Xu Han}, \bibinfo{person}{Zonghan Yang}, \bibinfo{person}{Shuo Wang}, \bibinfo{person}{Qingfu Zhu}, \bibinfo{person}{Zhiyuan Liu}, \bibinfo{person}{Weidong Liu}, {and} \bibinfo{person}{Wanxiang Che}.} \bibinfo{year}{2024}\natexlab{b}.
\newblock \showarticletitle{OneBit: Towards Extremely Low-bit Large Language Models}.
\newblock \bibinfo{journal}{\emph{arXiv preprint arXiv:2402.11295}} (\bibinfo{year}{2024}).
\newblock


\bibitem[Yan et~al\mbox{.}(2024)]%
        {yan2024corrective}
\bibfield{author}{\bibinfo{person}{Shi-Qi Yan}, \bibinfo{person}{Jia-Chen Gu}, \bibinfo{person}{Yun Zhu}, {and} \bibinfo{person}{Zhen-Hua Ling}.} \bibinfo{year}{2024}\natexlab{}.
\newblock \showarticletitle{Corrective retrieval augmented generation}.
\newblock \bibinfo{journal}{\emph{arXiv preprint arXiv:2401.15884}} (\bibinfo{year}{2024}).
\newblock


\bibitem[Yang et~al\mbox{.}(2024h)]%
        {yang2024qwen2}
\bibfield{author}{\bibinfo{person}{An Yang}, \bibinfo{person}{Baosong Yang}, \bibinfo{person}{Binyuan Hui}, \bibinfo{person}{Bo Zheng}, \bibinfo{person}{Bowen Yu}, \bibinfo{person}{Chang Zhou}, \bibinfo{person}{Chengpeng Li}, \bibinfo{person}{Chengyuan Li}, \bibinfo{person}{Dayiheng Liu}, \bibinfo{person}{Fei Huang}, {et~al\mbox{.}}} \bibinfo{year}{2024}\natexlab{h}.
\newblock \showarticletitle{Qwen2 technical report}.
\newblock \bibinfo{journal}{\emph{arXiv preprint arXiv:2407.10671}} (\bibinfo{year}{2024}).
\newblock


\bibitem[Yang et~al\mbox{.}(2024c)]%
        {yang2024survey}
\bibfield{author}{\bibinfo{person}{Chuanpeng Yang}, \bibinfo{person}{Wang Lu}, \bibinfo{person}{Yao Zhu}, \bibinfo{person}{Yidong Wang}, \bibinfo{person}{Qian Chen}, \bibinfo{person}{Chenlong Gao}, \bibinfo{person}{Bingjie Yan}, {and} \bibinfo{person}{Yiqiang Chen}.} \bibinfo{year}{2024}\natexlab{c}.
\newblock \showarticletitle{Survey on Knowledge Distillation for Large Language Models: Methods, Evaluation, and Application}.
\newblock \bibinfo{journal}{\emph{arXiv preprint arXiv:2407.01885}} (\bibinfo{year}{2024}).
\newblock


\bibitem[Yang et~al\mbox{.}(2024f)]%
        {yang-etal-2024-moe}
\bibfield{author}{\bibinfo{person}{Cheng Yang}, \bibinfo{person}{Yang Sui}, \bibinfo{person}{Jinqi Xiao}, \bibinfo{person}{Lingyi Huang}, \bibinfo{person}{Yu Gong}, \bibinfo{person}{Yuanlin Duan}, \bibinfo{person}{Wenqi Jia}, \bibinfo{person}{Miao Yin}, \bibinfo{person}{Yu Cheng}, {and} \bibinfo{person}{Bo Yuan}.} \bibinfo{year}{2024}\natexlab{f}.
\newblock \showarticletitle{{M}o{E}-I$^2$: Compressing Mixture of Experts Models through Inter-Expert Pruning and Intra-Expert Low-Rank Decomposition}. In \bibinfo{booktitle}{\emph{Findings of the Association for Computational Linguistics: EMNLP 2024}}, \bibfield{editor}{\bibinfo{person}{Yaser Al-Onaizan}, \bibinfo{person}{Mohit Bansal}, {and} \bibinfo{person}{Yun-Nung Chen}} (Eds.). \bibinfo{publisher}{Association for Computational Linguistics}, \bibinfo{address}{Miami, Florida, USA}, \bibinfo{pages}{10456--10466}.
\newblock
\urldef\tempurl%
\url{https://doi.org/10.18653/v1/2024.findings-emnlp.612}
\showDOI{\tempurl}


\bibitem[Yang et~al\mbox{.}(2024b)]%
        {yang2024rlcd}
\bibfield{author}{\bibinfo{person}{Kevin Yang}, \bibinfo{person}{Dan Klein}, \bibinfo{person}{Asli Celikyilmaz}, \bibinfo{person}{Nanyun Peng}, {and} \bibinfo{person}{Yuandong Tian}.} \bibinfo{year}{2024}\natexlab{b}.
\newblock \showarticletitle{{RLCD}: Reinforcement Learning from Contrastive Distillation for {LM} Alignment}. In \bibinfo{booktitle}{\emph{The Twelfth International Conference on Learning Representations}}.
\newblock
\urldef\tempurl%
\url{https://openreview.net/forum?id=v3XXtxWKi6}
\showURL{%
\tempurl}


\bibitem[Yang et~al\mbox{.}(2024i)]%
        {yang2024mentallama}
\bibfield{author}{\bibinfo{person}{Kailai Yang}, \bibinfo{person}{Tianlin Zhang}, \bibinfo{person}{Ziyan Kuang}, \bibinfo{person}{Qianqian Xie}, \bibinfo{person}{Jimin Huang}, {and} \bibinfo{person}{Sophia Ananiadou}.} \bibinfo{year}{2024}\natexlab{i}.
\newblock \showarticletitle{MentaLLaMA: interpretable mental health analysis on social media with large language models}. In \bibinfo{booktitle}{\emph{Proceedings of the ACM on Web Conference 2024}}. \bibinfo{pages}{4489--4500}.
\newblock


\bibitem[Yang et~al\mbox{.}(2023b)]%
        {yang-etal-2023-glue}
\bibfield{author}{\bibinfo{person}{Linyi Yang}, \bibinfo{person}{Shuibai Zhang}, \bibinfo{person}{Libo Qin}, \bibinfo{person}{Yafu Li}, \bibinfo{person}{Yidong Wang}, \bibinfo{person}{Hanmeng Liu}, \bibinfo{person}{Jindong Wang}, \bibinfo{person}{Xing Xie}, {and} \bibinfo{person}{Yue Zhang}.} \bibinfo{year}{2023}\natexlab{b}.
\newblock \showarticletitle{{GLUE}-{X}: Evaluating Natural Language Understanding Models from an Out-of-Distribution Generalization Perspective}. In \bibinfo{booktitle}{\emph{Findings of the Association for Computational Linguistics: ACL 2023}}. \bibinfo{publisher}{Association for Computational Linguistics}, \bibinfo{address}{Toronto, Canada}, \bibinfo{pages}{12731--12750}.
\newblock
\urldef\tempurl%
\url{https://doi.org/10.18653/v1/2023.findings-acl.806}
\showDOI{\tempurl}


\bibitem[Yang et~al\mbox{.}(2024j)]%
        {yang2024supervised}
\bibfield{author}{\bibinfo{person}{Linyi Yang}, \bibinfo{person}{Shuibai Zhang}, \bibinfo{person}{Zhuohao Yu}, \bibinfo{person}{Guangsheng Bao}, \bibinfo{person}{Yidong Wang}, \bibinfo{person}{Jindong Wang}, \bibinfo{person}{Ruochen Xu}, \bibinfo{person}{Wei Ye}, \bibinfo{person}{Xing Xie}, \bibinfo{person}{Weizhu Chen}, {and} \bibinfo{person}{Yue Zhang}.} \bibinfo{year}{2024}\natexlab{j}.
\newblock \showarticletitle{Supervised Knowledge Makes Large Language Models Better In-context Learners}. In \bibinfo{booktitle}{\emph{The Twelfth International Conference on Learning Representations}}.
\newblock
\urldef\tempurl%
\url{https://openreview.net/forum?id=bAMPOUF227}
\showURL{%
\tempurl}


\bibitem[Yang et~al\mbox{.}(2024g)]%
        {yang2024llm}
\bibfield{author}{\bibinfo{person}{Runming Yang}, \bibinfo{person}{Taiqiang Wu}, \bibinfo{person}{Jiahao Wang}, \bibinfo{person}{Pengfei Hu}, \bibinfo{person}{Ngai Wong}, {and} \bibinfo{person}{Yujiu Yang}.} \bibinfo{year}{2024}\natexlab{g}.
\newblock \showarticletitle{LLM-Neo: Parameter Efficient Knowledge Distillation for Large Language Models}.
\newblock \bibinfo{journal}{\emph{arXiv preprint arXiv:2411.06839}} (\bibinfo{year}{2024}).
\newblock


\bibitem[Yang et~al\mbox{.}(2024a)]%
        {yang2024laco}
\bibfield{author}{\bibinfo{person}{Yifei Yang}, \bibinfo{person}{Zouying Cao}, {and} \bibinfo{person}{Hai Zhao}.} \bibinfo{year}{2024}\natexlab{a}.
\newblock \showarticletitle{Laco: Large language model pruning via layer collapse}.
\newblock \bibinfo{journal}{\emph{arXiv preprint arXiv:2402.11187}} (\bibinfo{year}{2024}).
\newblock


\bibitem[Yang et~al\mbox{.}(2024d)]%
        {yang2024smalltolarge}
\bibfield{author}{\bibinfo{person}{Yu Yang}, \bibinfo{person}{Siddhartha Mishra}, \bibinfo{person}{Jeffrey~N Chiang}, {and} \bibinfo{person}{Baharan Mirzasoleiman}.} \bibinfo{year}{2024}\natexlab{d}.
\newblock \showarticletitle{SmallToLarge (S2L): Scalable Data Selection for Fine-tuning Large Language Models by Summarizing Training Trajectories of Small Models}. In \bibinfo{booktitle}{\emph{The Thirty-eighth Annual Conference on Neural Information Processing Systems}}.
\newblock
\urldef\tempurl%
\url{https://openreview.net/forum?id=K9IGlMQpif}
\showURL{%
\tempurl}


\bibitem[Yang et~al\mbox{.}(2023a)]%
        {yang2023mindllm}
\bibfield{author}{\bibinfo{person}{Yizhe Yang}, \bibinfo{person}{Huashan Sun}, \bibinfo{person}{Jiawei Li}, \bibinfo{person}{Runheng Liu}, \bibinfo{person}{Yinghao Li}, \bibinfo{person}{Yuhang Liu}, \bibinfo{person}{Heyan Huang}, {and} \bibinfo{person}{Yang Gao}.} \bibinfo{year}{2023}\natexlab{a}.
\newblock \showarticletitle{Mindllm: Pre-training lightweight large language model from scratch, evaluations and domain applications}.
\newblock \bibinfo{journal}{\emph{arXiv preprint arXiv:2310.15777}} (\bibinfo{year}{2023}).
\newblock


\bibitem[Yang et~al\mbox{.}(2024e)]%
        {yang2024enhancing}
\bibfield{author}{\bibinfo{person}{Zhou Yang}, \bibinfo{person}{Zhaochun Ren}, \bibinfo{person}{Wang Yufeng}, \bibinfo{person}{Shizhong Peng}, \bibinfo{person}{Haizhou Sun}, \bibinfo{person}{Xiaofei Zhu}, {and} \bibinfo{person}{Xiangwen Liao}.} \bibinfo{year}{2024}\natexlab{e}.
\newblock \showarticletitle{Enhancing Empathetic Response Generation by Augmenting LLMs with Small-scale Empathetic Models}.
\newblock \bibinfo{journal}{\emph{arXiv preprint arXiv:2402.11801}} (\bibinfo{year}{2024}).
\newblock


\bibitem[Yao et~al\mbox{.}(2021)]%
        {yao2021adapt}
\bibfield{author}{\bibinfo{person}{Yunzhi Yao}, \bibinfo{person}{Shaohan Huang}, \bibinfo{person}{Wenhui Wang}, \bibinfo{person}{Li Dong}, {and} \bibinfo{person}{Furu Wei}.} \bibinfo{year}{2021}\natexlab{}.
\newblock \showarticletitle{Adapt-and-distill: Developing small, fast and effective pretrained language models for domains}.
\newblock \bibinfo{journal}{\emph{arXiv preprint arXiv:2106.13474}} (\bibinfo{year}{2021}), \bibinfo{pages}{460--470}.
\newblock


\bibitem[Yazan et~al\mbox{.}(2024)]%
        {yazan2024impact}
\bibfield{author}{\bibinfo{person}{Mert Yazan}, \bibinfo{person}{Suzan Verberne}, {and} \bibinfo{person}{Frederik Situmeang}.} \bibinfo{year}{2024}\natexlab{}.
\newblock \showarticletitle{The Impact of Quantization on Retrieval-Augmented Generation: An Analysis of Small LLMs}.
\newblock \bibinfo{journal}{\emph{arXiv preprint arXiv:2406.10251}} (\bibinfo{year}{2024}).
\newblock


\bibitem[Yi et~al\mbox{.}(2023)]%
        {yi2023edgemoefastondeviceinference}
\bibfield{author}{\bibinfo{person}{Rongjie Yi}, \bibinfo{person}{Liwei Guo}, \bibinfo{person}{Shiyun Wei}, \bibinfo{person}{Ao Zhou}, \bibinfo{person}{Shangguang Wang}, {and} \bibinfo{person}{Mengwei Xu}.} \bibinfo{year}{2023}\natexlab{}.
\newblock \bibinfo{title}{EdgeMoE: Fast On-Device Inference of MoE-based Large Language Models}.
\newblock
\newblock
\showeprint[arxiv]{2308.14352}~[cs.LG]
\urldef\tempurl%
\url{https://arxiv.org/abs/2308.14352}
\showURL{%
\tempurl}


\bibitem[Yi et~al\mbox{.}(2024)]%
        {yi2024phonelm}
\bibfield{author}{\bibinfo{person}{Rongjie Yi}, \bibinfo{person}{Xiang Li}, \bibinfo{person}{Weikai Xie}, \bibinfo{person}{Zhenyan Lu}, \bibinfo{person}{Chenghua Wang}, \bibinfo{person}{Ao Zhou}, \bibinfo{person}{Shangguang Wang}, \bibinfo{person}{Xiwen Zhang}, {and} \bibinfo{person}{Mengwei Xu}.} \bibinfo{year}{2024}\natexlab{}.
\newblock \showarticletitle{PhoneLM: an Efficient and Capable Small Language Model Family through Principled Pre-training}.
\newblock \bibinfo{journal}{\emph{arXiv preprint arXiv:2411.05046}} (\bibinfo{year}{2024}).
\newblock


\bibitem[Yih et~al\mbox{.}(2016)]%
        {yih-etal-2016-value}
\bibfield{author}{\bibinfo{person}{Wen-tau Yih}, \bibinfo{person}{Matthew Richardson}, \bibinfo{person}{Chris Meek}, \bibinfo{person}{Ming-Wei Chang}, {and} \bibinfo{person}{Jina Suh}.} \bibinfo{year}{2016}\natexlab{}.
\newblock \showarticletitle{The Value of Semantic Parse Labeling for Knowledge Base Question Answering}. In \bibinfo{booktitle}{\emph{Proceedings of the 54th Annual Meeting of the Association for Computational Linguistics (Volume 2: Short Papers)}}, \bibfield{editor}{\bibinfo{person}{Katrin Erk} {and} \bibinfo{person}{Noah~A. Smith}} (Eds.). \bibinfo{publisher}{Association for Computational Linguistics}, \bibinfo{address}{Berlin, Germany}, \bibinfo{pages}{201--206}.
\newblock
\urldef\tempurl%
\url{https://doi.org/10.18653/v1/P16-2033}
\showDOI{\tempurl}


\bibitem[Yin et~al\mbox{.}(2024)]%
        {yin2024llmservicemobiledevices}
\bibfield{author}{\bibinfo{person}{Wangsong Yin}, \bibinfo{person}{Mengwei Xu}, \bibinfo{person}{Yuanchun Li}, {and} \bibinfo{person}{Xuanzhe Liu}.} \bibinfo{year}{2024}\natexlab{}.
\newblock \bibinfo{title}{LLM as a System Service on Mobile Devices}.
\newblock
\newblock
\showeprint[arxiv]{2403.11805}~[cs.OS]
\urldef\tempurl%
\url{https://arxiv.org/abs/2403.11805}
\showURL{%
\tempurl}


\bibitem[Yu et~al\mbox{.}(2024a)]%
        {yu2024metamath}
\bibfield{author}{\bibinfo{person}{Longhui Yu}, \bibinfo{person}{Weisen Jiang}, \bibinfo{person}{Han Shi}, \bibinfo{person}{Jincheng YU}, \bibinfo{person}{Zhengying Liu}, \bibinfo{person}{Yu Zhang}, \bibinfo{person}{James Kwok}, \bibinfo{person}{Zhenguo Li}, \bibinfo{person}{Adrian Weller}, {and} \bibinfo{person}{Weiyang Liu}.} \bibinfo{year}{2024}\natexlab{a}.
\newblock \showarticletitle{MetaMath: Bootstrap Your Own Mathematical Questions for Large Language Models}. In \bibinfo{booktitle}{\emph{The Twelfth International Conference on Learning Representations}}.
\newblock
\urldef\tempurl%
\url{https://openreview.net/forum?id=N8N0hgNDRt}
\showURL{%
\tempurl}


\bibitem[Yu et~al\mbox{.}(2024b)]%
        {yu2024rankrag}
\bibfield{author}{\bibinfo{person}{Yue Yu}, \bibinfo{person}{Wei Ping}, \bibinfo{person}{Zihan Liu}, \bibinfo{person}{Boxin Wang}, \bibinfo{person}{Jiaxuan You}, \bibinfo{person}{Chao Zhang}, \bibinfo{person}{Mohammad Shoeybi}, {and} \bibinfo{person}{Bryan Catanzaro}.} \bibinfo{year}{2024}\natexlab{b}.
\newblock \showarticletitle{Rankrag: Unifying context ranking with retrieval-augmented generation in llms}.
\newblock \bibinfo{journal}{\emph{arXiv preprint arXiv:2407.02485}} (\bibinfo{year}{2024}).
\newblock


\bibitem[Yu et~al\mbox{.}(2024c)]%
        {yu2024edge}
\bibfield{author}{\bibinfo{person}{Zhongzhi Yu}, \bibinfo{person}{Zheng Wang}, \bibinfo{person}{Yuhan Li}, \bibinfo{person}{Haoran You}, \bibinfo{person}{Ruijie Gao}, \bibinfo{person}{Xiaoya Zhou}, \bibinfo{person}{Sreenidhi~Reedy Bommu}, \bibinfo{person}{Yang~Katie Zhao}, {and} \bibinfo{person}{Yingyan~Celine Lin}.} \bibinfo{year}{2024}\natexlab{c}.
\newblock \showarticletitle{EDGE-LLM: Enabling Efficient Large Language Model Adaptation on Edge Devices via Layerwise Unified Compression and Adaptive Layer Tuning and Voting}.
\newblock \bibinfo{journal}{\emph{arXiv preprint arXiv:2406.15758}} (\bibinfo{year}{2024}).
\newblock


\bibitem[Yuan et~al\mbox{.}(2024c)]%
        {yuan2024mobile}
\bibfield{author}{\bibinfo{person}{Jinliang Yuan}, \bibinfo{person}{Chen Yang}, \bibinfo{person}{Dongqi Cai}, \bibinfo{person}{Shihe Wang}, \bibinfo{person}{Xin Yuan}, \bibinfo{person}{Zeling Zhang}, \bibinfo{person}{Xiang Li}, \bibinfo{person}{Dingge Zhang}, \bibinfo{person}{Hanzi Mei}, \bibinfo{person}{Xianqing Jia}, {et~al\mbox{.}}} \bibinfo{year}{2024}\natexlab{c}.
\newblock \showarticletitle{Mobile Foundation Model as Firmware}. In \bibinfo{booktitle}{\emph{Proceedings of the 30th Annual International Conference on Mobile Computing and Networking}}. \bibinfo{pages}{279--295}.
\newblock


\bibitem[Yuan et~al\mbox{.}(2021)]%
        {yuan2021wudaocorpora}
\bibfield{author}{\bibinfo{person}{Sha Yuan}, \bibinfo{person}{Hanyu Zhao}, \bibinfo{person}{Zhengxiao Du}, \bibinfo{person}{Ming Ding}, \bibinfo{person}{Xiao Liu}, \bibinfo{person}{Yukuo Cen}, \bibinfo{person}{Xu Zou}, \bibinfo{person}{Zhilin Yang}, {and} \bibinfo{person}{Jie Tang}.} \bibinfo{year}{2021}\natexlab{}.
\newblock \showarticletitle{Wudaocorpora: A super large-scale chinese corpora for pre-training language models}.
\newblock \bibinfo{journal}{\emph{AI Open}}  \bibinfo{volume}{2} (\bibinfo{year}{2021}), \bibinfo{pages}{65--68}.
\newblock


\bibitem[Yuan et~al\mbox{.}(2024b)]%
        {yuan2024s}
\bibfield{author}{\bibinfo{person}{Xiaohan Yuan}, \bibinfo{person}{Jinfeng Li}, \bibinfo{person}{Dongxia Wang}, \bibinfo{person}{Yuefeng Chen}, \bibinfo{person}{Xiaofeng Mao}, \bibinfo{person}{Longtao Huang}, \bibinfo{person}{Hui Xue}, \bibinfo{person}{Wenhai Wang}, \bibinfo{person}{Kui Ren}, {and} \bibinfo{person}{Jingyi Wang}.} \bibinfo{year}{2024}\natexlab{b}.
\newblock \showarticletitle{S-Eval: Automatic and Adaptive Test Generation for Benchmarking Safety Evaluation of Large Language Models}.
\newblock \bibinfo{journal}{\emph{arXiv preprint arXiv:2405.14191}} (\bibinfo{year}{2024}).
\newblock


\bibitem[Yuan et~al\mbox{.}(2024a)]%
        {yuan2024gpt}
\bibfield{author}{\bibinfo{person}{Youliang Yuan}, \bibinfo{person}{Wenxiang Jiao}, \bibinfo{person}{Wenxuan Wang}, \bibinfo{person}{Jen tse Huang}, \bibinfo{person}{Pinjia He}, \bibinfo{person}{Shuming Shi}, {and} \bibinfo{person}{Zhaopeng Tu}.} \bibinfo{year}{2024}\natexlab{a}.
\newblock \showarticletitle{{GPT}-4 Is Too Smart To Be Safe: Stealthy Chat with {LLM}s via Cipher}. In \bibinfo{booktitle}{\emph{The Twelfth International Conference on Learning Representations}}.
\newblock
\urldef\tempurl%
\url{https://openreview.net/forum?id=MbfAK4s61A}
\showURL{%
\tempurl}


\bibitem[Yue et~al\mbox{.}(2023)]%
        {yue2023disc}
\bibfield{author}{\bibinfo{person}{Shengbin Yue}, \bibinfo{person}{Wei Chen}, \bibinfo{person}{Siyuan Wang}, \bibinfo{person}{Bingxuan Li}, \bibinfo{person}{Chenchen Shen}, \bibinfo{person}{Shujun Liu}, \bibinfo{person}{Yuxuan Zhou}, \bibinfo{person}{Yao Xiao}, \bibinfo{person}{Song Yun}, \bibinfo{person}{Wei Lin}, {et~al\mbox{.}}} \bibinfo{year}{2023}\natexlab{}.
\newblock \showarticletitle{Disc-lawllm: Fine-tuning large language models for intelligent legal services}.
\newblock \bibinfo{journal}{\emph{arXiv preprint arXiv:2309.11325}} (\bibinfo{year}{2023}).
\newblock


\bibitem[Yue et~al\mbox{.}({[n.\,d.]})]%
        {yue2309mammoth}
\bibfield{author}{\bibinfo{person}{Xiang Yue}, \bibinfo{person}{Xingwei Qu}, \bibinfo{person}{Ge Zhang}, \bibinfo{person}{Yao Fu}, \bibinfo{person}{Wenhao Huang}, \bibinfo{person}{Huan Sun}, \bibinfo{person}{Yu Su}, {and} \bibinfo{person}{Wenhu Chen}.} \bibinfo{year}{[n.\,d.]}\natexlab{}.
\newblock \showarticletitle{Mammoth: Building math generalist models through hybrid instruction tuning, 2023}.
\newblock \bibinfo{journal}{\emph{URL https://arxiv. org/abs/2309.05653}} (\bibinfo{year}{[n.\,d.]}).
\newblock


\bibitem[Zellers et~al\mbox{.}(2019a)]%
        {zellers2019hellaswag}
\bibfield{author}{\bibinfo{person}{Rowan Zellers}, \bibinfo{person}{Ari Holtzman}, \bibinfo{person}{Yonatan Bisk}, \bibinfo{person}{Ali Farhadi}, {and} \bibinfo{person}{Yejin Choi}.} \bibinfo{year}{2019}\natexlab{a}.
\newblock \showarticletitle{HellaSwag: Can a Machine Really Finish Your Sentence?}. In \bibinfo{booktitle}{\emph{Proceedings of the 57th Annual Meeting of the Association for Computational Linguistics}}. \bibinfo{pages}{4791--4800}.
\newblock


\bibitem[Zellers et~al\mbox{.}(2019b)]%
        {zellers2019defending}
\bibfield{author}{\bibinfo{person}{Rowan Zellers}, \bibinfo{person}{Ari Holtzman}, \bibinfo{person}{Hannah Rashkin}, \bibinfo{person}{Yonatan Bisk}, \bibinfo{person}{Ali Farhadi}, \bibinfo{person}{Franziska Roesner}, {and} \bibinfo{person}{Yejin Choi}.} \bibinfo{year}{2019}\natexlab{b}.
\newblock \showarticletitle{Defending against neural fake news}.
\newblock \bibinfo{journal}{\emph{Advances in neural information processing systems}}  \bibinfo{volume}{32} (\bibinfo{year}{2019}).
\newblock


\bibitem[Zhang and Sennrich(2019)]%
        {zhang2019root}
\bibfield{author}{\bibinfo{person}{Biao Zhang} {and} \bibinfo{person}{Rico Sennrich}.} \bibinfo{year}{2019}\natexlab{}.
\newblock \showarticletitle{Root mean square layer normalization}.
\newblock \bibinfo{journal}{\emph{Advances in Neural Information Processing Systems}}  \bibinfo{volume}{32} (\bibinfo{year}{2019}).
\newblock


\bibitem[Zhang et~al\mbox{.}(2024c)]%
        {zhang2024lqer}
\bibfield{author}{\bibinfo{person}{Cheng Zhang}, \bibinfo{person}{Jianyi Cheng}, \bibinfo{person}{George~A Constantinides}, {and} \bibinfo{person}{Yiren Zhao}.} \bibinfo{year}{2024}\natexlab{c}.
\newblock \showarticletitle{LQER: Low-Rank Quantization Error Reconstruction for LLMs}.
\newblock \bibinfo{journal}{\emph{arXiv preprint arXiv:2402.02446}} (\bibinfo{year}{2024}).
\newblock


\bibitem[Zhang et~al\mbox{.}(2024f)]%
        {zhang2024extracting}
\bibfield{author}{\bibinfo{person}{Collin Zhang}, \bibinfo{person}{John~X Morris}, {and} \bibinfo{person}{Vitaly Shmatikov}.} \bibinfo{year}{2024}\natexlab{f}.
\newblock \showarticletitle{Extracting Prompts by Inverting LLM Outputs}.
\newblock \bibinfo{journal}{\emph{arXiv preprint arXiv:2405.15012}} (\bibinfo{year}{2024}).
\newblock


\bibitem[Zhang et~al\mbox{.}(2023b)]%
        {zhang2023towards}
\bibfield{author}{\bibinfo{person}{Chen Zhang}, \bibinfo{person}{Dawei Song}, \bibinfo{person}{Zheyu Ye}, {and} \bibinfo{person}{Yan Gao}.} \bibinfo{year}{2023}\natexlab{b}.
\newblock \showarticletitle{Towards the law of capacity gap in distilling language models}.
\newblock \bibinfo{journal}{\emph{arXiv preprint arXiv:2311.07052}} (\bibinfo{year}{2023}).
\newblock


\bibitem[Zhang et~al\mbox{.}(2024d)]%
        {zhang2024sciglm}
\bibfield{author}{\bibinfo{person}{Dan Zhang}, \bibinfo{person}{Ziniu Hu}, \bibinfo{person}{Sining Zhoubian}, \bibinfo{person}{Zhengxiao Du}, \bibinfo{person}{Kaiyu Yang}, \bibinfo{person}{Zihan Wang}, \bibinfo{person}{Yisong Yue}, \bibinfo{person}{Yuxiao Dong}, {and} \bibinfo{person}{Jie Tang}.} \bibinfo{year}{2024}\natexlab{d}.
\newblock \showarticletitle{Sciglm: Training scientific language models with self-reflective instruction annotation and tuning}.
\newblock \bibinfo{journal}{\emph{arXiv preprint arXiv:2401.07950}} (\bibinfo{year}{2024}).
\newblock


\bibitem[Zhang et~al\mbox{.}(2024e)]%
        {zhang2024chemllm}
\bibfield{author}{\bibinfo{person}{Di Zhang}, \bibinfo{person}{Wei Liu}, \bibinfo{person}{Qian Tan}, \bibinfo{person}{Jingdan Chen}, \bibinfo{person}{Hang Yan}, \bibinfo{person}{Yuliang Yan}, \bibinfo{person}{Jiatong Li}, \bibinfo{person}{Weiran Huang}, \bibinfo{person}{Xiangyu Yue}, \bibinfo{person}{Dongzhan Zhou}, {et~al\mbox{.}}} \bibinfo{year}{2024}\natexlab{e}.
\newblock \showarticletitle{Chemllm: A chemical large language model}.
\newblock \bibinfo{journal}{\emph{arXiv preprint arXiv:2402.06852}} (\bibinfo{year}{2024}).
\newblock


\bibitem[Zhang et~al\mbox{.}(2024h)]%
        {zhang2024cogenesis}
\bibfield{author}{\bibinfo{person}{Kaiyan Zhang}, \bibinfo{person}{Jianyu Wang}, \bibinfo{person}{Ermo Hua}, \bibinfo{person}{Biqing Qi}, \bibinfo{person}{Ning Ding}, {and} \bibinfo{person}{Bowen Zhou}.} \bibinfo{year}{2024}\natexlab{h}.
\newblock \showarticletitle{Cogenesis: A framework collaborating large and small language models for secure context-aware instruction following}.
\newblock \bibinfo{journal}{\emph{arXiv preprint arXiv:2403.03129}} (\bibinfo{year}{2024}).
\newblock


\bibitem[Zhang et~al\mbox{.}(2023a)]%
        {zhang2023loraprune}
\bibfield{author}{\bibinfo{person}{Mingyang Zhang}, \bibinfo{person}{Hao Chen}, \bibinfo{person}{Chunhua Shen}, \bibinfo{person}{Zhen Yang}, \bibinfo{person}{Linlin Ou}, \bibinfo{person}{Xinyi Yu}, {and} \bibinfo{person}{Bohan Zhuang}.} \bibinfo{year}{2023}\natexlab{a}.
\newblock \showarticletitle{Loraprune: Pruning meets low-rank parameter-efficient fine-tuning}.
\newblock \bibinfo{journal}{\emph{arXiv preprint arXiv:2305.18403}} (\bibinfo{year}{2023}).
\newblock


\bibitem[Zhang et~al\mbox{.}(2024j)]%
        {zhang2024tinyllamaopensourcesmalllanguage}
\bibfield{author}{\bibinfo{person}{Peiyuan Zhang}, \bibinfo{person}{Guangtao Zeng}, \bibinfo{person}{Tianduo Wang}, {and} \bibinfo{person}{Wei Lu}.} \bibinfo{year}{2024}\natexlab{j}.
\newblock \bibinfo{title}{TinyLlama: An Open-Source Small Language Model}.
\newblock
\newblock
\showeprint[arxiv]{2401.02385}~[cs.CL]
\urldef\tempurl%
\url{https://arxiv.org/abs/2401.02385}
\showURL{%
\tempurl}


\bibitem[Zhang et~al\mbox{.}(2022)]%
        {zhang2022opt}
\bibfield{author}{\bibinfo{person}{Susan Zhang}, \bibinfo{person}{Stephen Roller}, \bibinfo{person}{Naman Goyal}, \bibinfo{person}{Mikel Artetxe}, \bibinfo{person}{Moya Chen}, \bibinfo{person}{Shuohui Chen}, \bibinfo{person}{Christopher Dewan}, \bibinfo{person}{Mona Diab}, \bibinfo{person}{Xian Li}, \bibinfo{person}{Xi~Victoria Lin}, {et~al\mbox{.}}} \bibinfo{year}{2022}\natexlab{}.
\newblock \showarticletitle{Opt: Open pre-trained transformer language models}.
\newblock \bibinfo{journal}{\emph{arXiv preprint arXiv:2205.01068}} (\bibinfo{year}{2022}).
\newblock


\bibitem[Zhang et~al\mbox{.}(2019)]%
        {zhang2019cold}
\bibfield{author}{\bibinfo{person}{Xinran Zhang}, \bibinfo{person}{Xin Yuan}, \bibinfo{person}{Yunwei Li}, {and} \bibinfo{person}{Yanru Zhang}.} \bibinfo{year}{2019}\natexlab{}.
\newblock \showarticletitle{Cold-Start representation learning: A recommendation approach with bert4Movie and movie2Vec}. In \bibinfo{booktitle}{\emph{Proceedings of the 27th ACM International Conference on Multimedia}}. \bibinfo{pages}{2612--2616}.
\newblock


\bibitem[Zhang et~al\mbox{.}(2024a)]%
        {zhang2024plug}
\bibfield{author}{\bibinfo{person}{Yingtao Zhang}, \bibinfo{person}{Haoli Bai}, \bibinfo{person}{Haokun Lin}, \bibinfo{person}{Jialin Zhao}, \bibinfo{person}{Lu Hou}, {and} \bibinfo{person}{Carlo~Vittorio Cannistraci}.} \bibinfo{year}{2024}\natexlab{a}.
\newblock \showarticletitle{Plug-and-play: An efficient post-training pruning method for large language models}. In \bibinfo{booktitle}{\emph{The Twelfth International Conference on Learning Representations}}.
\newblock


\bibitem[Zhang et~al\mbox{.}(2024b)]%
        {zhang2024effectivepromptextractionlanguage}
\bibfield{author}{\bibinfo{person}{Yiming Zhang}, \bibinfo{person}{Nicholas Carlini}, {and} \bibinfo{person}{Daphne Ippolito}.} \bibinfo{year}{2024}\natexlab{b}.
\newblock \bibinfo{title}{Effective Prompt Extraction from Language Models}.
\newblock
\newblock
\showeprint[arxiv]{2307.06865}~[cs.CL]
\urldef\tempurl%
\url{https://arxiv.org/abs/2307.06865}
\showURL{%
\tempurl}


\bibitem[Zhang et~al\mbox{.}(2024g)]%
        {zhang2024h2o}
\bibfield{author}{\bibinfo{person}{Zhenyu Zhang}, \bibinfo{person}{Ying Sheng}, \bibinfo{person}{Tianyi Zhou}, \bibinfo{person}{Tianlong Chen}, \bibinfo{person}{Lianmin Zheng}, \bibinfo{person}{Ruisi Cai}, \bibinfo{person}{Zhao Song}, \bibinfo{person}{Yuandong Tian}, \bibinfo{person}{Christopher R{\'e}}, \bibinfo{person}{Clark Barrett}, {et~al\mbox{.}}} \bibinfo{year}{2024}\natexlab{g}.
\newblock \showarticletitle{H2o: Heavy-hitter oracle for efficient generative inference of large language models}.
\newblock \bibinfo{journal}{\emph{Advances in Neural Information Processing Systems}}  \bibinfo{volume}{36} (\bibinfo{year}{2024}).
\newblock


\bibitem[Zhang et~al\mbox{.}(2024i)]%
        {zhang2024does}
\bibfield{author}{\bibinfo{person}{Zhiwei Zhang}, \bibinfo{person}{Fali Wang}, \bibinfo{person}{Xiaomin Li}, \bibinfo{person}{Zongyu Wu}, \bibinfo{person}{Xianfeng Tang}, \bibinfo{person}{Hui Liu}, \bibinfo{person}{Qi He}, \bibinfo{person}{Wenpeng Yin}, {and} \bibinfo{person}{Suhang Wang}.} \bibinfo{year}{2024}\natexlab{i}.
\newblock \showarticletitle{Does your LLM truly unlearn? An embarrassingly simple approach to recover unlearned knowledge}.
\newblock \bibinfo{journal}{\emph{arXiv preprint arXiv:2410.16454}} (\bibinfo{year}{2024}).
\newblock


\bibitem[Zhao et~al\mbox{.}(2024a)]%
        {zhaoapt}
\bibfield{author}{\bibinfo{person}{Bowen Zhao}, \bibinfo{person}{Hannaneh Hajishirzi}, {and} \bibinfo{person}{Qingqing Cao}.} \bibinfo{year}{2024}\natexlab{a}.
\newblock \showarticletitle{APT: Adaptive Pruning and Tuning Pretrained Language Models for Efficient Training and Inference}. In \bibinfo{booktitle}{\emph{Forty-first International Conference on Machine Learning}}.
\newblock


\bibitem[Zhao et~al\mbox{.}(2023a)]%
        {zhao2023lingualinkeddistributedlargelanguage}
\bibfield{author}{\bibinfo{person}{Junchen Zhao}, \bibinfo{person}{Yurun Song}, \bibinfo{person}{Simeng Liu}, \bibinfo{person}{Ian~G. Harris}, {and} \bibinfo{person}{Sangeetha~Abdu Jyothi}.} \bibinfo{year}{2023}\natexlab{a}.
\newblock \bibinfo{title}{LinguaLinked: A Distributed Large Language Model Inference System for Mobile Devices}.
\newblock
\newblock
\showeprint[arxiv]{2312.00388}~[cs.LG]
\urldef\tempurl%
\url{https://arxiv.org/abs/2312.00388}
\showURL{%
\tempurl}


\bibitem[Zhao et~al\mbox{.}(2024c)]%
        {zhao2024llm}
\bibfield{author}{\bibinfo{person}{Juntao Zhao}, \bibinfo{person}{Borui Wan}, \bibinfo{person}{Yanghua Peng}, \bibinfo{person}{Haibin Lin}, {and} \bibinfo{person}{Chuan Wu}.} \bibinfo{year}{2024}\natexlab{c}.
\newblock \showarticletitle{LLM-PQ: Serving LLM on Heterogeneous Clusters with Phase-Aware Partition and Adaptive Quantization}.
\newblock \bibinfo{journal}{\emph{arXiv preprint arXiv:2403.01136}} (\bibinfo{year}{2024}).
\newblock


\bibitem[Zhao et~al\mbox{.}(2024d)]%
        {zhao2024slide}
\bibfield{author}{\bibinfo{person}{Kun Zhao}, \bibinfo{person}{Bohao Yang}, \bibinfo{person}{Chen Tang}, \bibinfo{person}{Chenghua Lin}, {and} \bibinfo{person}{Liang Zhan}.} \bibinfo{year}{2024}\natexlab{d}.
\newblock \showarticletitle{SLIDE: A Framework Integrating Small and Large Language Models for Open-Domain Dialogues Evaluation}.
\newblock \bibinfo{journal}{\emph{arXiv preprint arXiv:2405.15924}} (\bibinfo{year}{2024}).
\newblock


\bibitem[Zhao et~al\mbox{.}(2023b)]%
        {zhao2023automatic}
\bibfield{author}{\bibinfo{person}{Theodore Zhao}, \bibinfo{person}{Mu Wei}, \bibinfo{person}{J~Samuel Preston}, {and} \bibinfo{person}{Hoifung Poon}.} \bibinfo{year}{2023}\natexlab{b}.
\newblock \showarticletitle{Automatic calibration and error correction for large language models via pareto optimal self-supervision}.
\newblock \bibinfo{journal}{\emph{arXiv preprint arXiv:2306.16564}} (\bibinfo{year}{2023}).
\newblock


\bibitem[Zhao et~al\mbox{.}(2024b)]%
        {zhao2024merinoentropydrivendesigngenerative}
\bibfield{author}{\bibinfo{person}{Youpeng Zhao}, \bibinfo{person}{Ming Lin}, \bibinfo{person}{Huadong Tang}, \bibinfo{person}{Qiang Wu}, {and} \bibinfo{person}{Jun Wang}.} \bibinfo{year}{2024}\natexlab{b}.
\newblock \bibinfo{title}{Merino: Entropy-driven Design for Generative Language Models on IoT Devices}.
\newblock
\newblock
\showeprint[arxiv]{2403.07921}~[cs.LG]
\urldef\tempurl%
\url{https://arxiv.org/abs/2403.07921}
\showURL{%
\tempurl}


\bibitem[Zhao et~al\mbox{.}(2022)]%
        {zhao2022pmc}
\bibfield{author}{\bibinfo{person}{Zhengyun Zhao}, \bibinfo{person}{Qiao Jin}, \bibinfo{person}{Fangyuan Chen}, \bibinfo{person}{Tuorui Peng}, {and} \bibinfo{person}{Sheng Yu}.} \bibinfo{year}{2022}\natexlab{}.
\newblock \showarticletitle{Pmc-patients: A large-scale dataset of patient summaries and relations for benchmarking retrieval-based clinical decision support systems}.
\newblock \bibinfo{journal}{\emph{arXiv preprint arXiv:2202.13876}} (\bibinfo{year}{2022}).
\newblock


\bibitem[Zhou et~al\mbox{.}(2024)]%
        {zhou2024weaktostrong}
\bibfield{author}{\bibinfo{person}{Zhanhui Zhou}, \bibinfo{person}{Zhixuan Liu}, \bibinfo{person}{Jie Liu}, \bibinfo{person}{Zhichen Dong}, \bibinfo{person}{Chao Yang}, {and} \bibinfo{person}{Yu Qiao}.} \bibinfo{year}{2024}\natexlab{}.
\newblock \showarticletitle{Weak-to-Strong Search: Align Large Language Models via Searching over Small Language Models}. In \bibinfo{booktitle}{\emph{The Thirty-eighth Annual Conference on Neural Information Processing Systems}}.
\newblock
\urldef\tempurl%
\url{https://openreview.net/forum?id=dOJ6CqWDf1}
\showURL{%
\tempurl}


\bibitem[Zhu et~al\mbox{.}(2021)]%
        {zhu2021tat}
\bibfield{author}{\bibinfo{person}{Fengbin Zhu}, \bibinfo{person}{Wenqiang Lei}, \bibinfo{person}{Youcheng Huang}, \bibinfo{person}{Chao Wang}, \bibinfo{person}{Shuo Zhang}, \bibinfo{person}{Jiancheng Lv}, \bibinfo{person}{Fuli Feng}, {and} \bibinfo{person}{Tat-Seng Chua}.} \bibinfo{year}{2021}\natexlab{}.
\newblock \showarticletitle{TAT-QA: A Question Answering Benchmark on a Hybrid of Tabular and Textual Content in Finance}. In \bibinfo{booktitle}{\emph{Proceedings of the 59th Annual Meeting of the Association for Computational Linguistics and the 11th International Joint Conference on Natural Language Processing (Volume 1: Long Papers)}}. \bibinfo{pages}{3277--3287}.
\newblock


\bibitem[Zhu et~al\mbox{.}(2024)]%
        {zhu2024lifelong}
\bibfield{author}{\bibinfo{person}{Jiachen Zhu}, \bibinfo{person}{Jianghao Lin}, \bibinfo{person}{Xinyi Dai}, \bibinfo{person}{Bo Chen}, \bibinfo{person}{Rong Shan}, \bibinfo{person}{Jieming Zhu}, \bibinfo{person}{Ruiming Tang}, \bibinfo{person}{Yong Yu}, {and} \bibinfo{person}{Weinan Zhang}.} \bibinfo{year}{2024}\natexlab{}.
\newblock \showarticletitle{Lifelong Personalized Low-Rank Adaptation of Large Language Models for Recommendation}.
\newblock \bibinfo{journal}{\emph{arXiv preprint arXiv:2408.03533}} (\bibinfo{year}{2024}).
\newblock


\bibitem[Zhu et~al\mbox{.}(2023c)]%
        {zhu2023promptbench}
\bibfield{author}{\bibinfo{person}{Kaijie Zhu}, \bibinfo{person}{Jindong Wang}, \bibinfo{person}{Jiaheng Zhou}, \bibinfo{person}{Zichen Wang}, \bibinfo{person}{Hao Chen}, \bibinfo{person}{Yidong Wang}, \bibinfo{person}{Linyi Yang}, \bibinfo{person}{Wei Ye}, \bibinfo{person}{Yue Zhang}, \bibinfo{person}{Neil~Zhenqiang Gong}, {et~al\mbox{.}}} \bibinfo{year}{2023}\natexlab{c}.
\newblock \showarticletitle{Promptbench: Towards evaluating the robustness of large language models on adversarial prompts}.
\newblock \bibinfo{journal}{\emph{arXiv preprint arXiv:2306.04528}} (\bibinfo{year}{2023}).
\newblock


\bibitem[Zhu et~al\mbox{.}(2023a)]%
        {zhu2023survey}
\bibfield{author}{\bibinfo{person}{Xunyu Zhu}, \bibinfo{person}{Jian Li}, \bibinfo{person}{Yong Liu}, \bibinfo{person}{Can Ma}, {and} \bibinfo{person}{Weiping Wang}.} \bibinfo{year}{2023}\natexlab{a}.
\newblock \showarticletitle{A survey on model compression for large language models}.
\newblock \bibinfo{journal}{\emph{arXiv preprint arXiv:2308.07633}} (\bibinfo{year}{2023}).
\newblock


\bibitem[Zhu et~al\mbox{.}(2023b)]%
        {zhu2023ondeviceagenttextrewriting}
\bibfield{author}{\bibinfo{person}{Yun Zhu}, \bibinfo{person}{Yinxiao Liu}, \bibinfo{person}{Felix Stahlberg}, \bibinfo{person}{Shankar Kumar}, \bibinfo{person}{Yu hui Chen}, \bibinfo{person}{Liangchen Luo}, \bibinfo{person}{Lei Shu}, \bibinfo{person}{Renjie Liu}, \bibinfo{person}{Jindong Chen}, {and} \bibinfo{person}{Lei Meng}.} \bibinfo{year}{2023}\natexlab{b}.
\newblock \bibinfo{title}{Towards an On-device Agent for Text Rewriting}.
\newblock
\newblock
\showeprint[arxiv]{2308.11807}~[cs.CL]
\urldef\tempurl%
\url{https://arxiv.org/abs/2308.11807}
\showURL{%
\tempurl}


\bibitem[Zhuang and Kim(2021)]%
        {zhuang2021bert}
\bibfield{author}{\bibinfo{person}{Yuanyuan Zhuang} {and} \bibinfo{person}{Jaekyeong Kim}.} \bibinfo{year}{2021}\natexlab{}.
\newblock \showarticletitle{A bert-based multi-criteria recommender system for hotel promotion management}.
\newblock \bibinfo{journal}{\emph{Sustainability}} \bibinfo{volume}{13}, \bibinfo{number}{14} (\bibinfo{year}{2021}), \bibinfo{pages}{8039}.
\newblock


\bibitem[Zhuo et~al\mbox{.}(2023)]%
        {zhuo2023red}
\bibfield{author}{\bibinfo{person}{Terry~Yue Zhuo}, \bibinfo{person}{Yujin Huang}, \bibinfo{person}{Chunyang Chen}, {and} \bibinfo{person}{Zhenchang Xing}.} \bibinfo{year}{2023}\natexlab{}.
\newblock \showarticletitle{Red teaming chatgpt via jailbreaking: Bias, robustness, reliability and toxicity}.
\newblock \bibinfo{journal}{\emph{arXiv preprint arXiv:2301.12867}} (\bibinfo{year}{2023}).
\newblock


\bibitem[Zou et~al\mbox{.}(2023)]%
        {zou2023universal}
\bibfield{author}{\bibinfo{person}{Andy Zou}, \bibinfo{person}{Zifan Wang}, \bibinfo{person}{Nicholas Carlini}, \bibinfo{person}{Milad Nasr}, \bibinfo{person}{J~Zico Kolter}, {and} \bibinfo{person}{Matt Fredrikson}.} \bibinfo{year}{2023}\natexlab{}.
\newblock \showarticletitle{Universal and transferable adversarial attacks on aligned language models}.
\newblock \bibinfo{journal}{\emph{arXiv preprint arXiv:2307.15043}} (\bibinfo{year}{2023}).
\newblock


\bibitem[Zou et~al\mbox{.}(2022)]%
        {zou2022pre}
\bibfield{author}{\bibinfo{person}{Lixin Zou}, \bibinfo{person}{Weixue Lu}, \bibinfo{person}{Yiding Liu}, \bibinfo{person}{Hengyi Cai}, \bibinfo{person}{Xiaokai Chu}, \bibinfo{person}{Dehong Ma}, \bibinfo{person}{Daiting Shi}, \bibinfo{person}{Yu Sun}, \bibinfo{person}{Zhicong Cheng}, \bibinfo{person}{Simiu Gu}, {et~al\mbox{.}}} \bibinfo{year}{2022}\natexlab{}.
\newblock \showarticletitle{Pre-trained language model-based retrieval and ranking for web search}.
\newblock \bibinfo{journal}{\emph{ACM Transactions on the Web}} \bibinfo{volume}{17}, \bibinfo{number}{1} (\bibinfo{year}{2022}), \bibinfo{pages}{1--36}.
\newblock


\bibitem[Zuo et~al\mbox{.}(2024)]%
        {zuo2024falcon}
\bibfield{author}{\bibinfo{person}{Jingwei Zuo}, \bibinfo{person}{Maksim Velikanov}, \bibinfo{person}{Dhia~Eddine Rhaiem}, \bibinfo{person}{Ilyas Chahed}, \bibinfo{person}{Younes Belkada}, \bibinfo{person}{Guillaume Kunsch}, {and} \bibinfo{person}{Hakim Hacid}.} \bibinfo{year}{2024}\natexlab{}.
\newblock \showarticletitle{Falcon mamba: The first competitive attention-free 7b language model}.
\newblock \bibinfo{journal}{\emph{arXiv preprint arXiv:2410.05355}} (\bibinfo{year}{2024}).
\newblock


\end{thebibliography}
